\documentclass{article} %

\PassOptionsToPackage{table}{xcolor}

\usepackage[T1]{fontenc}
\usepackage{iclr2027_conference,times}

\usepackage{amsmath,amsfonts,bm}

\def\eqref#1{equation~\ref{#1}}

\def\1{\bm{1}}

\DeclareMathAlphabet{\mathsfit}{\encodingdefault}{\sfdefault}{m}{sl}
\SetMathAlphabet{\mathsfit}{bold}{\encodingdefault}{\sfdefault}{bx}{n}

\usepackage{hyperref}
\usepackage{xurl} %
\usepackage{amsmath,amssymb}
\usepackage{graphicx}
\usepackage{xcolor}          %
\usepackage{booktabs}
\usepackage{multirow}        %
\usepackage{array}           %
\usepackage{caption}         %
\usepackage{wrapfig}
\usepackage{placeins}        %
\usepackage[capitalize,noabbrev]{cleveref}


\newcommand{\stdrow}[8]{%
  {\footnotesize\textcolor{black!55}{$\pm$ std (3 seeds)}}%
  & \textcolor{black!55}{#1} & \textcolor{black!55}{#2} & \textcolor{black!55}{#3}%
  & \textcolor{black!55}{#4} & \textcolor{black!55}{#5} & \textcolor{black!55}{#6}%
  & \textcolor{black!55}{#7} & \textcolor{black!55}{#8}}

\title{When Text-to-Image Helps Editing:\\
The Effects of Conditioning During Denoising}

\author{Lidia Troeshestova \\
HSE University \& Yandex
\And
Alexander Ustyuzhanin \\
Yandex
\And
Sergey Kastryulin \\
Yandex
}
\iclrfinalcopy
\hypersetup{
pdftitle={When Text-to-Image Helps Editing: The Effects of Conditioning During Denoising},
pdfauthor={Lidia Troeshestova, Alexander Ustyuzhanin, Sergey Kastryulin},
pdfsubject={Computer Vision and Pattern Recognition}
}
\begin{document}

\maketitle
\lhead{Preprint}
\suppressfloats[t] %

\begin{abstract}
Unified models are trained for both instruction-based image editing and text-to-image (T2I) generation, but standard editing pipelines keep source-image conditioning throughout denoising. We ask whether editing can benefit from T2I, and study how the effects of conditioning vary across edits and denoising stages. In pure editing, source attention declines for some edits over the sampling trajectory. This observation led us to task switching, which lets the model draw on its T2I capabilities. Across three unified editors and four benchmarks, switching to the T2I task for bounded intervals improves edit quality, while mean perceptual preservation remains close to pure editing across all three models. Unified editors therefore benefit from using both conditioning modes they are trained for, and the timing of the switch sets the balance between quality and preservation.
\end{abstract}

\begingroup
\renewcommand{\thefootnote}{}
\footnotetext{An earlier version of this work appeared as ``DuET: Dual Expert Trajectories for Diffusion Image Editing''~\citep{troeshestova2026duet} (arXiv:2606.13303).}
\endgroup

\section{Introduction}
\label{sec:intro}

\begin{figure}[!htbp]
	\centering
\includegraphics[width=\linewidth]{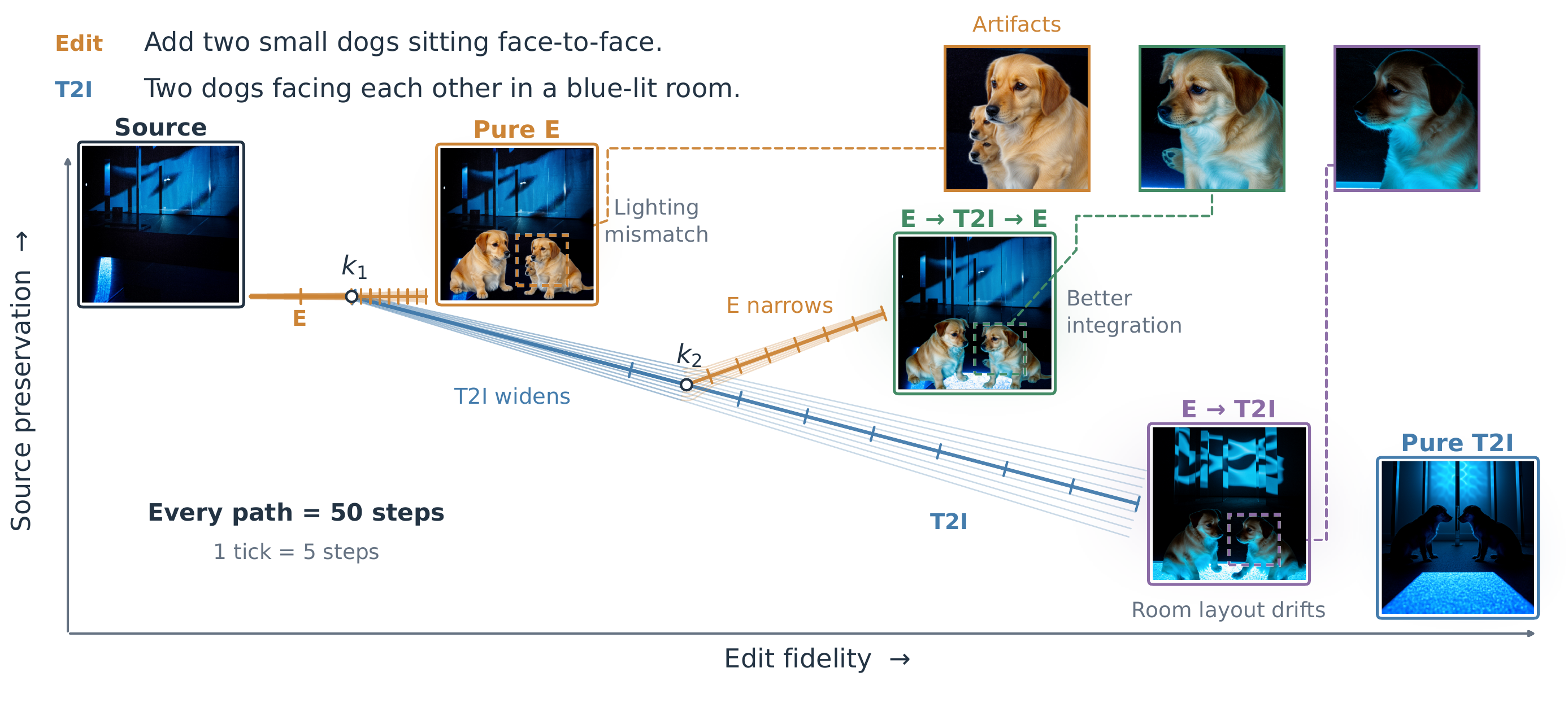}
\caption{\textbf{Switching conditioning helps added objects fit the scene.} Adding two dogs to a blue-lit room with pure editing produces artifacts and mismatched lighting (orange close-up). Switching to caption-only T2I at $k_1$ improves integration but changes the room. The $\mathrm{E}\to\mathrm{T2I}\to\mathrm{E}$ schedule returns to editing at $k_2$ to retain more of the source.}
	\label{fig:intuition}
\end{figure}

Unified multimodal models support text-to-image (T2I) generation and instruction-based image editing with the same architecture and weights~\citep{flux2025, deng2025emerging}. T2I generation synthesizes an image from a caption describing the desired scene~\citep{rombach2022high}, while editing transforms a source image according to a natural-language instruction~\citep{brooks2023instructpix2pix}. Within a unified model, the two tasks offer a synergy between synthesizing a coherent target scene and retaining relevant content from an existing image.

Standard pipelines nevertheless use these capabilities separately. In diffusion editors such as FLUX2-Klein~\citep{flux2025}, editing conditions on the source image and instruction throughout denoising, while T2I generation uses a caption alone. Source-image conditioning therefore stays fixed over the course of an edit, and the model's dual capability goes unused within a single trajectory.

Persistent source-image anchoring preserves structure, but it can also suppress intended changes~\citep{hertz2022prompt, zhou2025scale}. When the target scene diverges substantially from the input, editors can under-execute the instruction or integrate the change poorly, with visible seams and mismatched lighting. This raises a question: does every edit benefit from source-image conditioning at every stage of denoising?

To answer it, we examine source-image attention in pure-editing trajectories. With source conditioning available throughout denoising, some edits progressively attend less to it, while localized changes retain strong source attention through the final steps (\cref{sec:intuition}). This led us to test whether temporarily removing source conditioning can improve edit fidelity and naturalness.

For a fixed interval, we remove source-image conditioning and replace the editing instruction with a target caption generated by a vision-language model (VLM). We then restore the original editing inputs. The model weights and number of denoiser forward passes remain unchanged. \Cref{fig:intuition} shows an example: switching to T2I helps integrate the added objects into the scene, and returning to editing preserves more of the original room.

\paragraph{Contributions.}
\begin{itemize}
	\item We find that the strength of source-image attention depends on the type of edit and varies over the course of denoising. This suggests that for some classes of edits, source conditioning is redundant at certain stages.
	\item We identify a conditioning regime that consistently improves edit quality across three models and four benchmarks while keeping preservation virtually the same.
	\item We vary the degrees of freedom of this regime, the conditioning inputs, the interval placement and duration, and the guidance formulation, and characterize how each affects edit quality and preservation.
\end{itemize}

\section{Related Work}
\label{sec:related}

\paragraph{Attention and source conditioning.}
Prompt-to-Prompt reuses cross-attention maps over text tokens from a source generation and limits their injection in time to permit structural changes~\citep{hertz2022prompt}. SaaS identifies source-image dominance over neglected instructions in OmniGen and adaptively scales instruction attention. In its examples, jointly replacing image and text inputs late in denoising has little effect~\citep{zhou2025scale}. GRAG modulates the relative influence of source-image and instruction tokens to control editing intensity~\citep{zhang2025grouprelativeattentionguidance}. We ask the same question about source conditioning, but instead of rescaling attention we change the conditioning inputs themselves and measure the outcome, using attention maps only as a diagnostic.

\paragraph{Temporal conditioning.}
eDiff-I observes that reliance on text conditioning decreases during sampling and trains separate denoisers for different noise levels~\citep{balaji2022ediff}. Restricting classifier-free guidance to a limited interval also improves generation~\citep{kynkaanniemi2024applying}. PCI intervenes on the text prompt during T2I sampling to find when a concept can still be inserted. The resulting temporal profiles differ across concepts and models and carry over to editing~\citep{gorgun2026temporal}. We extend this line of work from varying denoisers, guidance, or text prompts across denoising stages to switching between image-conditioned editing and caption-only generation within a unified model.

\paragraph{Combining generation and editing.}
HiDream-E1's released pipeline provides a direct precedent for an editing-to-generation handoff: it continues the same latent with a target caption, removes source-image conditioning, and finishes the remaining steps with the HiDream-I1 generator weights~\citep{hidream2025e1}. For refinement of generated images, RvR argues that pixel-level preservation restricts the space of possible changes, and trains a regeneration model conditioned on target prompts and semantic image tokens without source VAE tokens~\citep{guo2026refinement}. Building on these pipelines, we study how source-image conditioning, the text paired with it, and interval timing affect edit quality and source preservation across unified editors. Conditioning controls and timing sweeps use the models' native editing and T2I modes within one denoising trajectory under unchanged weights.

\section{Source conditioning along editing trajectories}
\label{sec:method}

A unified model is trained for two tasks but runs one at a time. Every editing request has a source image, and the pipeline conditions on it at every step regardless of what the request asks for. Editing requests, however, differ widely in nature and in difficulty: enhancing the lighting adjusts an intact scene, whereas replacing a background asks the model to synthesize most of the image from scratch. It would be surprising if one conditioning policy served both equally well. This section examines how an editor actually uses the source image across such edits and across denoising steps.

\subsection{Source attention varies across edits and timesteps}
\label{sec:intuition}

\begin{figure}[!htb]
	\centering
	\includegraphics[width=0.85\linewidth]{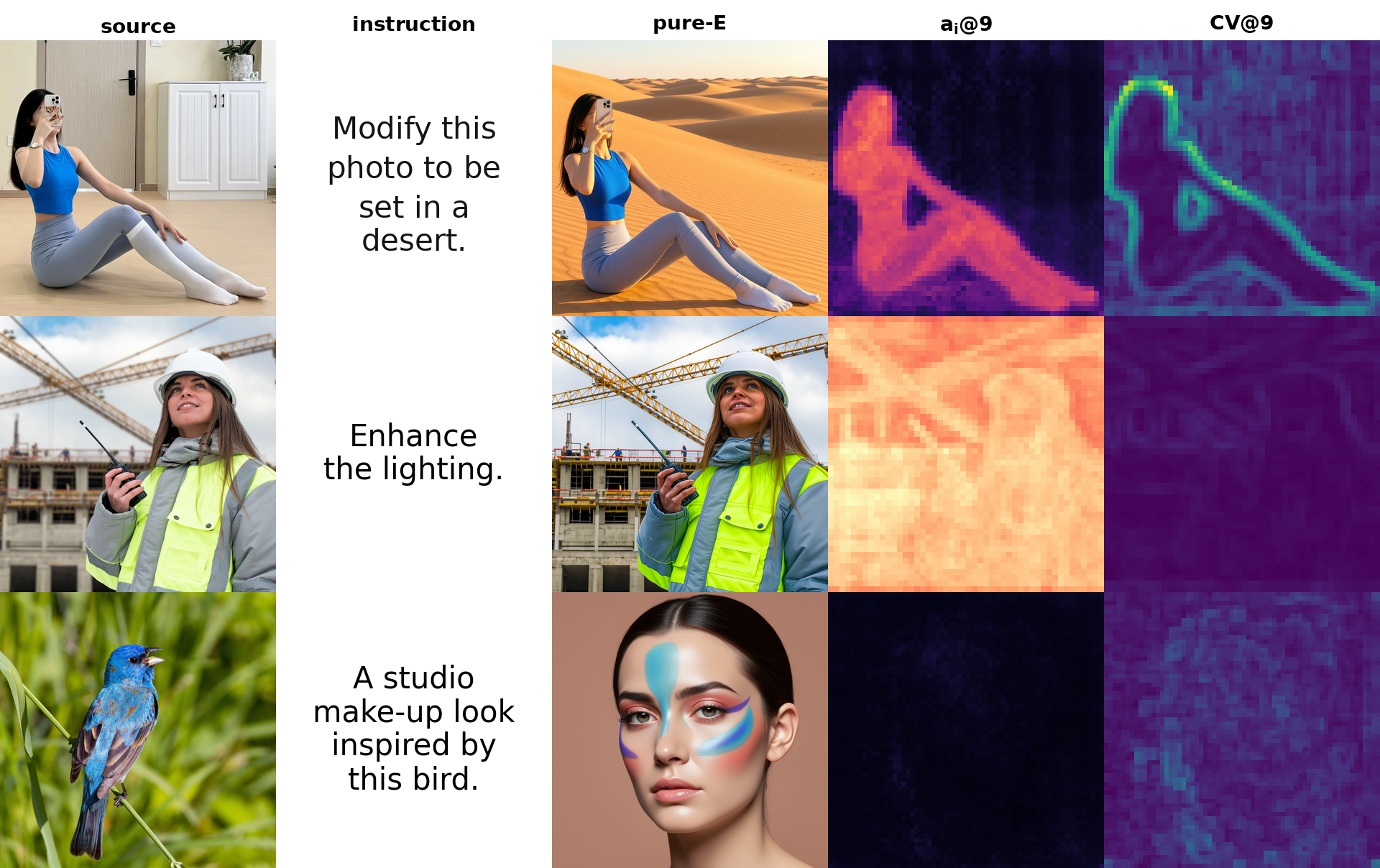}
	\caption{\textbf{Source attention varies across edits (FLUX2-Klein~4B, step 9).} Maps average over heads and blocks. $a_i$@9 shows per-patch source attention; CV@9 shows the coefficient of variation in local $3\times3$ neighborhoods.}
	\label{fig:attention}
\end{figure}

If the source image matters more for some edits than for others, the most direct place to look is the model's attention to it. We therefore measure how much each output patch attends to the source-image tokens during ordinary editing, and how that changes as denoising proceeds. We examine source attention along $50$-step pure-editing trajectories on FLUX2-Klein~4B. For each output latent patch $i$, we sum attention over source-image tokens and average across heads and transformer blocks, obtaining $a_i$. At step $k=9$ (\cref{fig:attention}), the three edits attend to the source very differently: only on the kept foreground for the background replacement, strongly everywhere for the lighting change, and weakly everywhere for the makeup look. How much the model uses the source depends on what the edit needs from it.

Attention also changes over time (\cref{fig:disagreement}a): source-attention mass generally decreases in the examined trajectories, whereas some local edits retain high attention through the final steps. The source image and instruction are identical at every step, so this variation reflects how the model uses fixed inputs. Where attention to the source has faded, conditioning on it constrains the model without contributing much. Since T2I generation is free of that constraint, we hypothesize that it can produce a better scene, and the question becomes where in the trajectory a T2I phase would disturb the edit the least.

\subsection{Comparing editing and T2I predictions}
\label{sec:disagreement}

\begin{figure}[!htb]
	\centering
	\includegraphics[width=\linewidth]{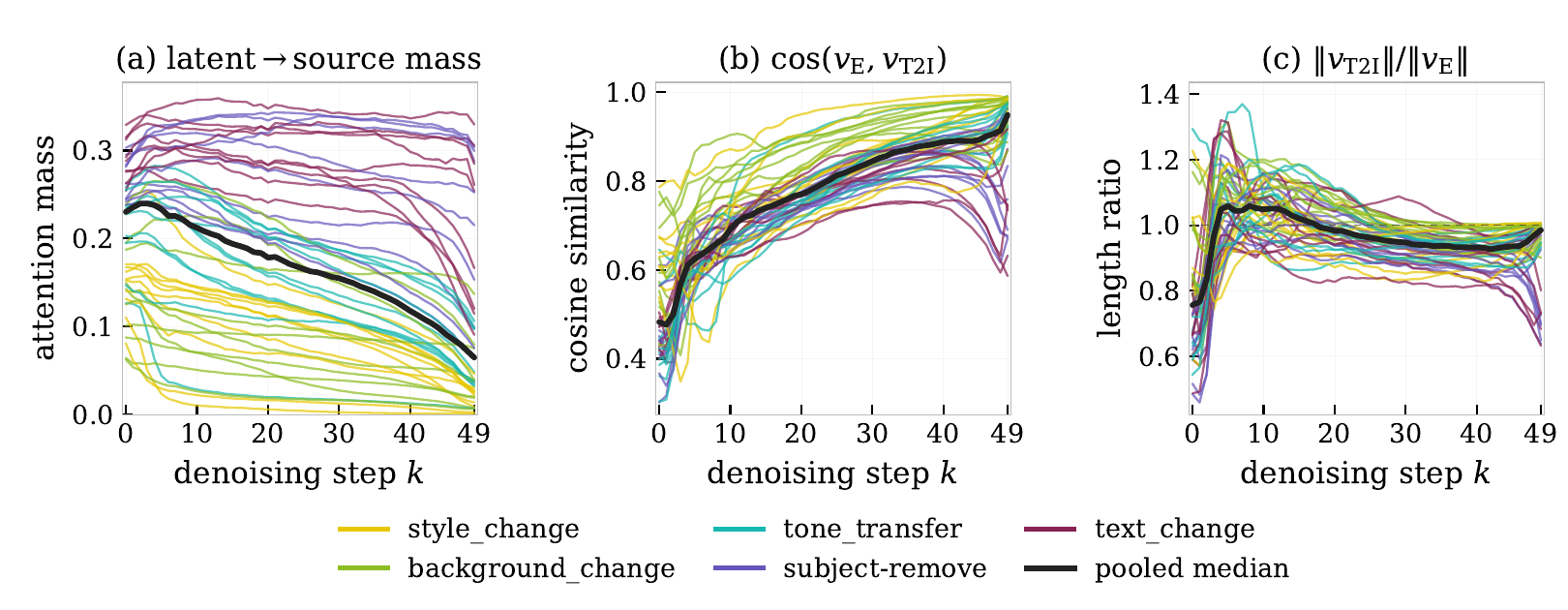}
	\caption{\textbf{Conditioning disagreement and source attention along pure editing (FLUX2-Klein~4B).} \emph{(a)} Source-attention mass generally decreases, while local edits retain high attention late in sampling. \emph{(b)} Prediction cosine similarity, initially low (median $0.48$). \emph{(c)} T2I/edit prediction norm ratio. Both are evaluated at identical latent states.}
	\label{fig:disagreement}
\end{figure}

The two conditioning modes propose different updates to the same latent. Where they agree, switching between them changes little; where they diverge, a switch pulls the trajectory off the edit. To find the low-disturbance region, we compute both predictions at every step of an ordinary edit. At each latent state of a pure-editing trajectory, we compare predictions conditioned on the source image and instruction ($v_{\mathrm{E}}$) with predictions conditioned on a target caption ($v_{\mathrm{T2I}}$). Their directional agreement and norm ratio (\cref{fig:disagreement}b,c) are highly variable early, comparatively stable in the middle, and diverge again late for local edits. This suggests that the effect of switching depends on its timing, so we test both one-way switches and bounded intervals placed in the stable middle region. The comparison is a diagnostic: it reads both predictions at each step without altering the sampled trajectory, whereas the switching schedules below evaluate one mode per step.

\subsection{Conditioning interventions and controls}
\label{sec:fixed-duet}

\begin{wraptable}[10]{R}{0.48\linewidth}
\centering
\captionsetup{skip=4pt}
\caption{Conditioning inputs.}
\label{tab:cond-modes}
\small
\setlength{\tabcolsep}{4pt}
\begin{tabular}{@{}lcc@{}}
\toprule
Mode / control & Image & Text \\
\midrule
$\mathrm{E}$ & $x_{\mathrm{src}}$ & $p_{\mathrm{edit}}$ \\
$\mathrm{E}^{*}$ & $x_{\mathrm{src}}$ & $p_{\mathrm{edit}}^{*}$ \\
$\mathrm{T2I}$ & $\varnothing$ & $c_{\mathrm{tgt}}$ \\
$\mathrm{T2I}$ w/ editing prompt & $\varnothing$ & $p_{\mathrm{edit}}$ \\
$\mathrm{E}$ w/ caption & $x_{\mathrm{src}}$ & $c_{\mathrm{tgt}}$ \\
\bottomrule
\end{tabular}
\end{wraptable}

The diagnostics above say where a switch is likely to be safe. To test whether it helps, we intervene directly: for part of the trajectory we hand the latent to the model's T2I mode and let it continue with a caption of the target scene, then return to editing. Edit conditioning $\mathrm{E}$ uses a source image $x_{\mathrm{src}}$ and instruction $p_{\mathrm{edit}}$; T2I uses a target caption $c_{\mathrm{tgt}}$ generated beforehand by a VLM from those inputs. A one-way switch $\mathrm{E}\to\mathrm{T2I}$ retains caption-only conditioning from step $k$ onward. A bounded schedule $\mathrm{E}\to\mathrm{T2I}\to\mathrm{E}$ uses it on $[k_1,k_2)$ and edit conditioning elsewhere. Both continue denoising the same latent. Only one mode is active per step, so FLUX2-Klein~4B/9B still run the baseline's two denoiser passes per step. The only overhead is one VLM call for the caption (timings in \cref{app:timing}).

Switching changes both image and text inputs. To separate their effects, we test the original instruction without the source image ($\varnothing$) and the target caption with the source image (\cref{tab:cond-modes}). The $\mathrm{E}^{*}$ control supplies a VLM-improved instruction $p_{\mathrm{edit}}^{*}$, at the cost of the same single VLM call as caption generation. Outside the interval, the controls use ordinary edit conditioning. An additional control uses the target caption throughout and removes the source image only during the interval (\cref{tab:appx-gedit-qwen}). Together, the schedules (when conditioning changes) and the controls (what it changes to) let us attribute any gain to the removal of the source image, the change of text, or their timing.

\section{Effects of conditioning and interval timing}
\label{sec:exps}

To test whether the effect is a property of unified editors rather than of one model, we evaluate three that differ in size, architecture, and training: FLUX2-Klein 4B and 9B~\citep{flux2025}, two sizes of one diffusion-transformer family, and BAGEL~\citep{deng2025emerging}, a separately developed model. The benchmarks cover complementary aspects of editing quality. GEdit-Bench~\citep{liu2025step1x} uses real user instructions across task types and scores semantic consistency and perceptual quality separately. ImgEdit~\citep{ye2026imgedit} adds hard categories such as extraction, composition, and actions. MagicBrush~\citep{zhang2023magicbrush} supplies a reference caption for every target, which allows a judge-free alignment score. GIE-Bench~\citep{qian2025gie} scores functional correctness and preservation separately, so quality gains can be priced in preservation.

\paragraph{Setup.}
All models run with unchanged weights and $50$ steps ($k=0,\ldots,49$); target captions and improved instructions come from Gemini-2.5-pro by default~\citep{comanici2025gemini} (Qwen3.5 captioners are compared in \cref{app:timing}). GEdit is judged by GPT-4.1~\citep{openai2025gpt41} on a $0$--$10$ scale, ImgEdit by GPT-4o~\citep{hurst2024gpt} on $1$--$5$, and GIE-Bench correctness by GPT-4o in $\%$; MagicBrush reports CLIP-T cosine similarity to the reference caption, and GIE-Bench preservation is SSIM outside the edit mask and whole-image Unmasked-CLIP similarity to the source. Per GEdit sample, semantic consistency (SC) is the minimum of fidelity and preservation, perceptual quality (PQ) the minimum of naturalness and artifact-free appearance, and overall $O=\sqrt{\mathrm{SC}\cdot\mathrm{PQ}}$; scores are averaged within task types, then across types. The GEdit conditioning matrices and \cref{fig:t2i-vs-i2c,fig:ablation} use Qwen3.5~\citep{qwen2026qwen35}; the extended 4B GIE-Bench sweep uses Gemini-3-Flash~\citep{google2025gemini3flash}.

\paragraph{Variability.}
Baseline standard deviations come from three seeds with one judge pass each, so they mix generation and judge variability. They describe the baseline only, and the uncertainty of the switching gains is assessed separately. Paired quality tests on 4B use two-sided Wilcoxon signed-rank tests and exact McNemar tests for binary correctness on matched per-edit records. Preservation comparisons are described in \cref{app:preservation}.

\subsection{Task switching improves editing quality}
\label{sec:fixed-results}
\label{sec:main-results}

\begin{figure}[!htbp]
	\centering
	\includegraphics[width=\linewidth]{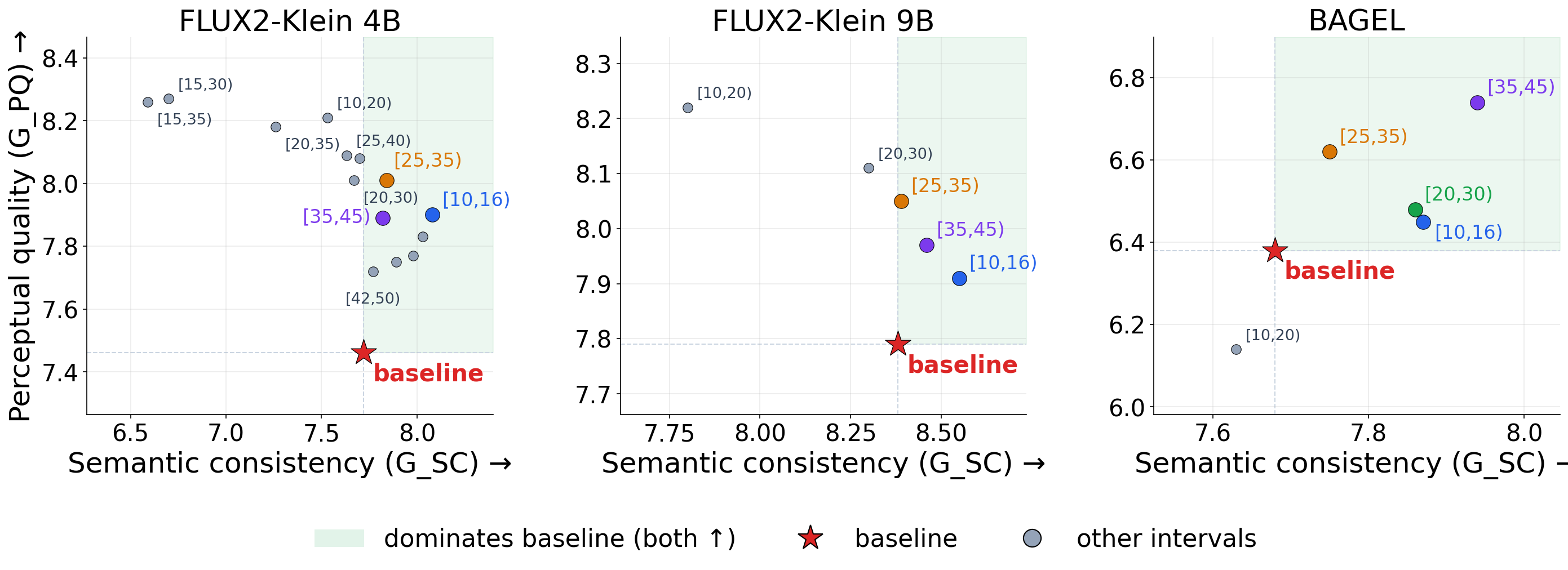}
	\caption{\textbf{A fixed T2I interval can improve both GEdit semantic consistency and perceptual quality on each model.} GEdit scores use the GPT-4.1 judge. Points are $\mathrm{E}\to\mathrm{T2I}\to\mathrm{E}$ schedules, labeled by $[k_1,k_2)$. The shaded region indicates improvement over pure editing on both metrics.}
	\label{fig:existence}
\end{figure}

Trading preservation for quality is easy: any editor can be pushed toward the target by loosening its anchor to the source. Improving quality without giving up preservation is the hard part, and this is what a T2I interval achieves. Our sweeps over interval length and position identify $[10,16)$ as a strong fixed interval: applied to every edit, it raises all six quality scores on all three models (\cref{tab:main}), while preservation as judged by GPT-4.1 stays within a fraction of a point of pure editing. The pixel-level measures, SSIM and Unmasked-CLIP, do drop, and \cref{sec:preservation} examines how much of that drop is perceptible. The interval is not unique: several placements improve GEdit semantic consistency and perceptual quality at once (\cref{fig:existence}), and which one is best depends on the model and the metric.

We feature $[10,16)$ because it was consistently strong in the FLUX2-Klein~4B sweeps across all four benchmarks. On 9B and BAGEL we tested a smaller set of intervals, and $[10,16)$ again ranked among the strongest (\cref{fig:existence,tab:appx-imgedit}), so we report it for all three models and use it for the selective variant. The interval was not re-tuned per model or benchmark. Full GIE-Bench, MagicBrush, and GEdit results are reported in \cref{tab:appx-giebench-fc,tab:appx-magicbrush,tab:appx-gedit-gpt}.

\begin{table}[!htbp]
	\centering
	\caption{\textbf{A fixed T2I interval improves quality across models and benchmarks.} Fixed and selective switching use the six-step interval $[10,16)$. The first six scores measure edit quality. Higher is better throughout. Green/red intensity shows the magnitude of increases/decreases from each model's pure-editing baseline. Standard deviations describe baseline variability across three seeds. GPT-4.1 preservation is scored on a $0$--$10$ scale.}
	\label{tab:main}

	{\footnotesize
	\setlength{\tabcolsep}{3pt}
	\resizebox{\linewidth}{!}{%
	\begin{tabular}{llcccccccc}
	\toprule
	& & \multicolumn{3}{c}{GEdit} & ImgEdit & MagicBrush & \multicolumn{3}{c}{GIE-Bench} \\
	\cmidrule(lr){3-5} \cmidrule(lr){6-6} \cmidrule(lr){7-7} \cmidrule(lr){8-10}
	& & & & & & & Func.\ Corr. & \multicolumn{2}{c}{Preservation} \\
	\cmidrule(lr){8-8} \cmidrule(lr){9-10}
	Model & Sampling & SC & PQ & O & overall & CLIP-T & Overall & SSIM & GPT-4.1 \\
	\midrule
	FLUX2-Klein 4B & \stdrow{0.05}{0.03}{0.04}{0.04}{0.001}{0.35}{0.001}{0.05} \\
	 & Pure editing & 7.72 & 7.46 & 7.11 & 3.74 & 0.317 & 73.40 & 0.805 & 7.65 \\
	 & Fixed & \cellcolor[RGB]{209,242,216}8.08 & \cellcolor[RGB]{224,246,228}7.90 & \cellcolor[RGB]{198,239,206}7.67 & \cellcolor[RGB]{223,246,227}3.99 & \cellcolor[RGB]{219,245,224}0.322 & \cellcolor[RGB]{222,246,226}76.46 & \cellcolor[RGB]{255,221,226}0.781 & \cellcolor[RGB]{255,227,230}7.57 \\
	 & Selective$^{*}$ & \cellcolor[RGB]{218,245,223}8.01 & \cellcolor[RGB]{240,251,242}7.67 & \cellcolor[RGB]{220,245,225}7.45 & \cellcolor[RGB]{237,250,239}3.88 & \cellcolor[RGB]{241,251,243}0.319 & \cellcolor[RGB]{214,244,220}77.11 & \cellcolor[RGB]{255,228,232}0.786 & \cellcolor[RGB]{255,237,240}7.60 \\
	\midrule
	FLUX2-Klein 9B & \stdrow{0.01}{0.01}{0.01}{0.02}{0.001}{0.72}{0.001}{0.03} \\
	 & Pure editing & 8.38 & 7.79 & 7.74 & 4.02 & 0.319 & 77.94 & 0.939 & 8.70 \\
	 & Fixed & \cellcolor[RGB]{233,249,236}8.55 & \cellcolor[RGB]{247,253,248}7.91 & \cellcolor[RGB]{233,249,236}7.96 & \cellcolor[RGB]{228,247,232}4.23 & \cellcolor[RGB]{219,245,224}0.324 & \cellcolor[RGB]{219,245,224}81.28 & \cellcolor[RGB]{255,199,206}0.899 & \cellcolor[RGB]{255,199,206}8.54 \\
	 & Selective$^{*}$ & \cellcolor[RGB]{227,247,231}8.60 & \cellcolor[RGB]{248,253,249}7.89 & \cellcolor[RGB]{234,249,237}7.95 & \cellcolor[RGB]{228,247,232}4.23 & \cellcolor[RGB]{226,247,230}0.323 & \cellcolor[RGB]{225,246,229}80.72 & \cellcolor[RGB]{255,217,222}0.912 & \cellcolor[RGB]{255,224,227}8.61 \\
	\midrule
	BAGEL & \stdrow{0.06}{0.09}{0.07}{0.03}{0.001}{1.25}{0.001}{0.02} \\
	 & Pure editing & 7.68 & 6.38 & 6.64 & 3.32 & 0.320 & 67.16 & 0.960 & 7.90 \\
	 & Fixed & \cellcolor[RGB]{231,248,234}7.87 & \cellcolor[RGB]{250,254,251}6.45 & \cellcolor[RGB]{238,250,240}6.81 & \cellcolor[RGB]{239,251,242}3.44 & \cellcolor[RGB]{226,247,230}0.324 & \cellcolor[RGB]{198,239,206}72.38 & \cellcolor[RGB]{255,244,245}0.952 & \cellcolor[RGB]{255,241,243}7.86 \\
	 & Selective$^{*}$ & \cellcolor[RGB]{232,249,235}7.86 & \cellcolor[RGB]{248,253,249}6.48 & \cellcolor[RGB]{240,251,242}6.79 & \cellcolor[RGB]{245,252,246}3.40 & \cellcolor[RGB]{234,249,237}0.323 & \cellcolor[RGB]{209,242,216}71.36 & \cellcolor[RGB]{255,247,248}0.954 & 7.90 \\
	\bottomrule
	\end{tabular}}}
\par\smallskip
\begin{minipage}{\linewidth}
\footnotesize Fixed applies the interval to every edit. $^{*}$Selective uses the attention rule in \cref{sec:selective-duet}, with thresholds tuned on GEdit only.
\end{minipage}
\end{table}

\Cref{fig:qualitative} shows what the scores summarize: switching executes edits that pure editing leaves undone or half done, and returning to editing is what keeps the rest of the source intact, as the one-way column shows by contrast (more examples in \cref{app:qualitative}).

\begingroup
\captionsetup{skip=4pt}
\begin{figure}[!htb]
\centering

\begingroup
\sffamily
\setlength{\tabcolsep}{0pt}
\renewcommand{\arraystretch}{0}
\setlength{\fboxsep}{0pt}
\newlength{\qinstr}
\newlength{\qcell}
\newlength{\qcelltall}
\setlength{\qinstr}{0.110\linewidth}
\setlength{\qcell}{0.178\linewidth}
\setlength{\qcelltall}{1.472\qcell}
\providecommand{\qimg}[1]{%
  \parbox[c][\qcell][c]{\qcell}{\centering\includegraphics[width=\qcell,height=\qcell,keepaspectratio]{#1}}%
}
\providecommand{\qimgtall}[1]{%
  \parbox[c][\qcelltall][c]{\qcell}{\centering\includegraphics[width=\qcell]{#1}}%
}
\providecommand{\qmissing}{%
  \parbox[c][\qcell][c]{\qcell}{\centering\textcolor{black!10}{\rule{\qcell}{\qcell}}}%
}
\providecommand{\qmissingtall}{%
  \parbox[c][\qcelltall][c]{\qcell}{\centering\textcolor{black!10}{\rule{\qcell}{\qcelltall}}}%
}
\providecommand{\qtextpad}{\vspace{2pt}\vspace{0pt}}
\providecommand{\qtxtinstr}[1]{%
  \parbox[c][\qcell][t]{\dimexpr\qinstr-2pt\relax}{\qtextpad\raggedright #1\par}%
}
\providecommand{\qtxtinstrtall}[1]{%
  \parbox[c][\qcelltall][t]{\dimexpr\qinstr-2pt\relax}{\qtextpad\raggedright #1\par}%
}
\providecommand{\qtxtcap}[1]{%
  \parbox[c][\qcell][t]{\dimexpr\qcell-2pt\relax}{\qtextpad\raggedright #1\par}%
}
\providecommand{\qtxtcaptall}[1]{%
  \parbox[c][\qcelltall][t]{\dimexpr\qcell-2pt\relax}{\qtextpad\raggedright #1\par}%
}
\providecommand{\qhdrinstr}[1]{%
  \parbox[t]{\dimexpr\qinstr-2pt\relax}{\centering\qtextpad #1\par}%
}
\providecommand{\qhdr}[1]{%
  \parbox[t]{\dimexpr\qcell-2pt\relax}{\centering\qtextpad #1\par}%
}

\begin{tabular}{@{}>{\raggedright\arraybackslash}m{\qinstr}*{5}{>{\centering\arraybackslash}m{\qcell}}@{}}
\qhdrinstr{\fontsize{7}{8}\selectfont\bfseries instruction} & \qhdr{\fontsize{7}{8}\selectfont\bfseries source} & \qhdr{\fontsize{7}{8}\selectfont\bfseries pure-E} & \qhdr{\fontsize{7}{8}\selectfont\bfseries T2I caption} & \qhdr{\fontsize{7}{8}\selectfont\bfseries E$\to$T2I $k{=}10$} & \qhdr{\fontsize{7}{8}\selectfont\bfseries E$\to$T2I$\to$E\\$[10,20)$} \\
\noalign{\vskip 1pt}
\qtxtinstr{\fontsize{6}{7}\selectfont Replace the house in the image with a giant tree.} & \qimg{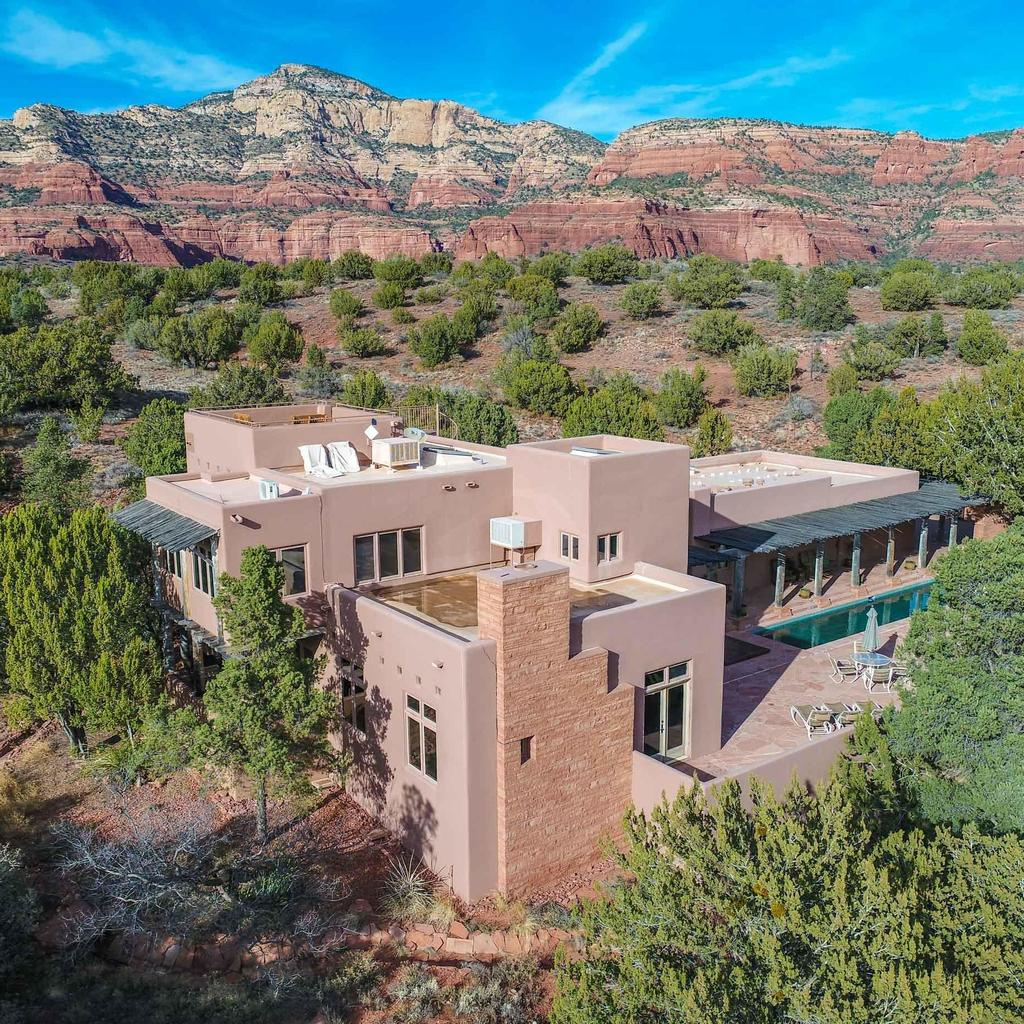} & \qimg{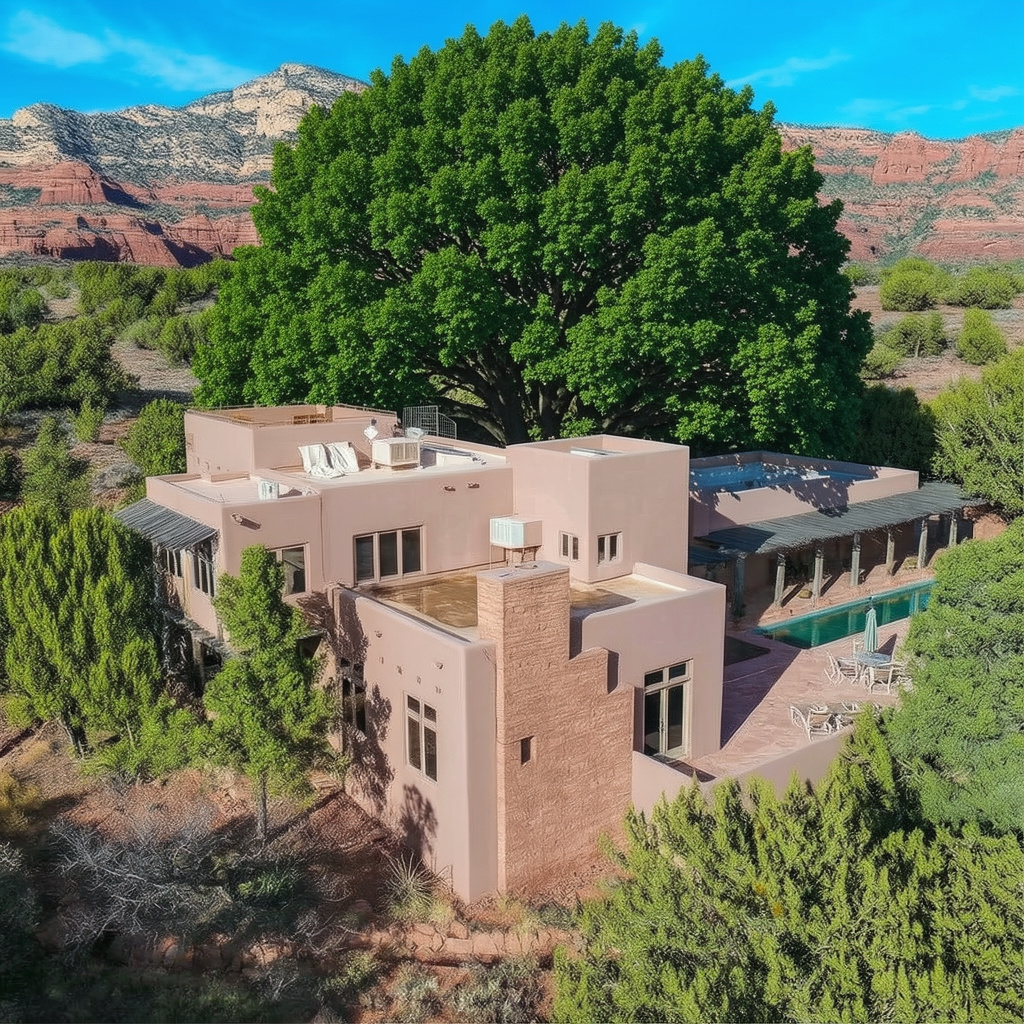} & \qtxtcap{\fontsize{5.2}{5.9}\selectfont An aerial, high-angle drone photograph of a majestic desert landscape where a single, colossal, ancient tree with a massive, thick trunk and a sprawling, dense\,\ldots{}} & \qimg{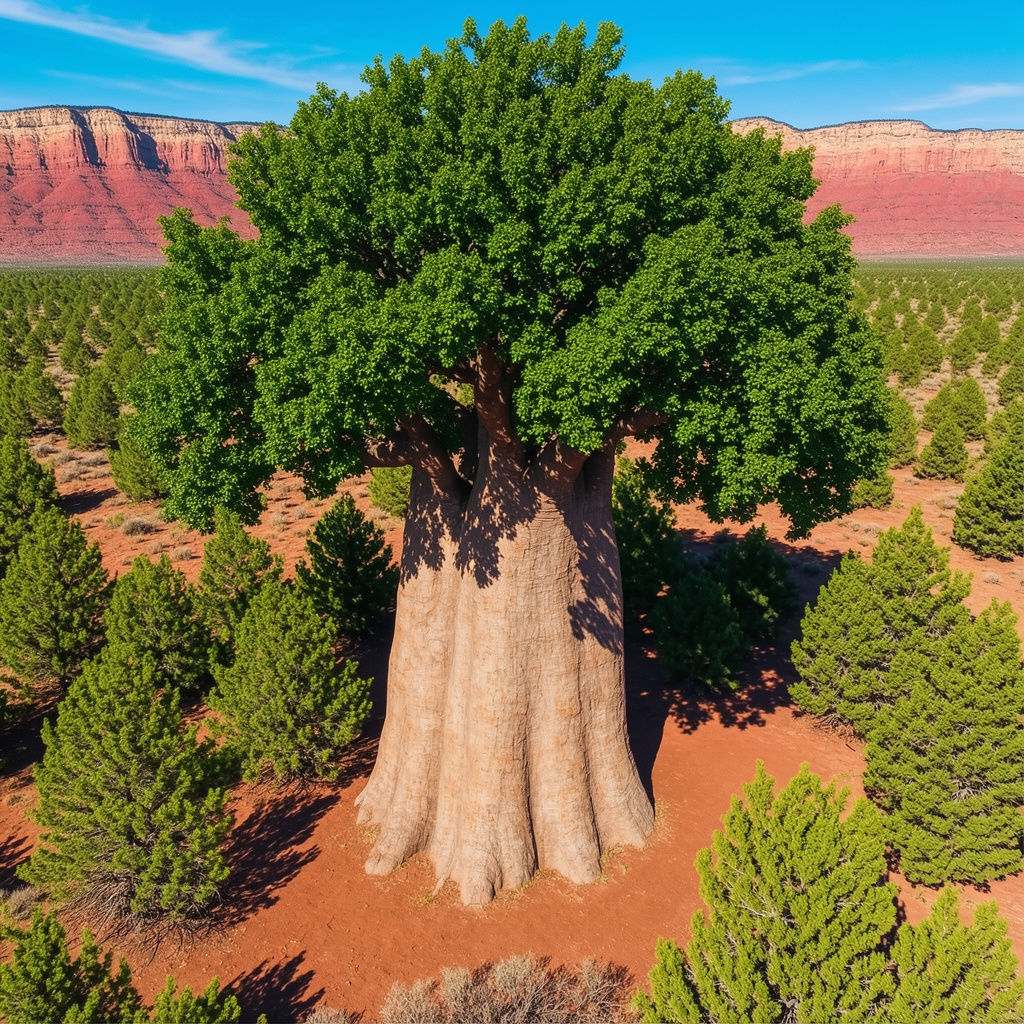} & \qimg{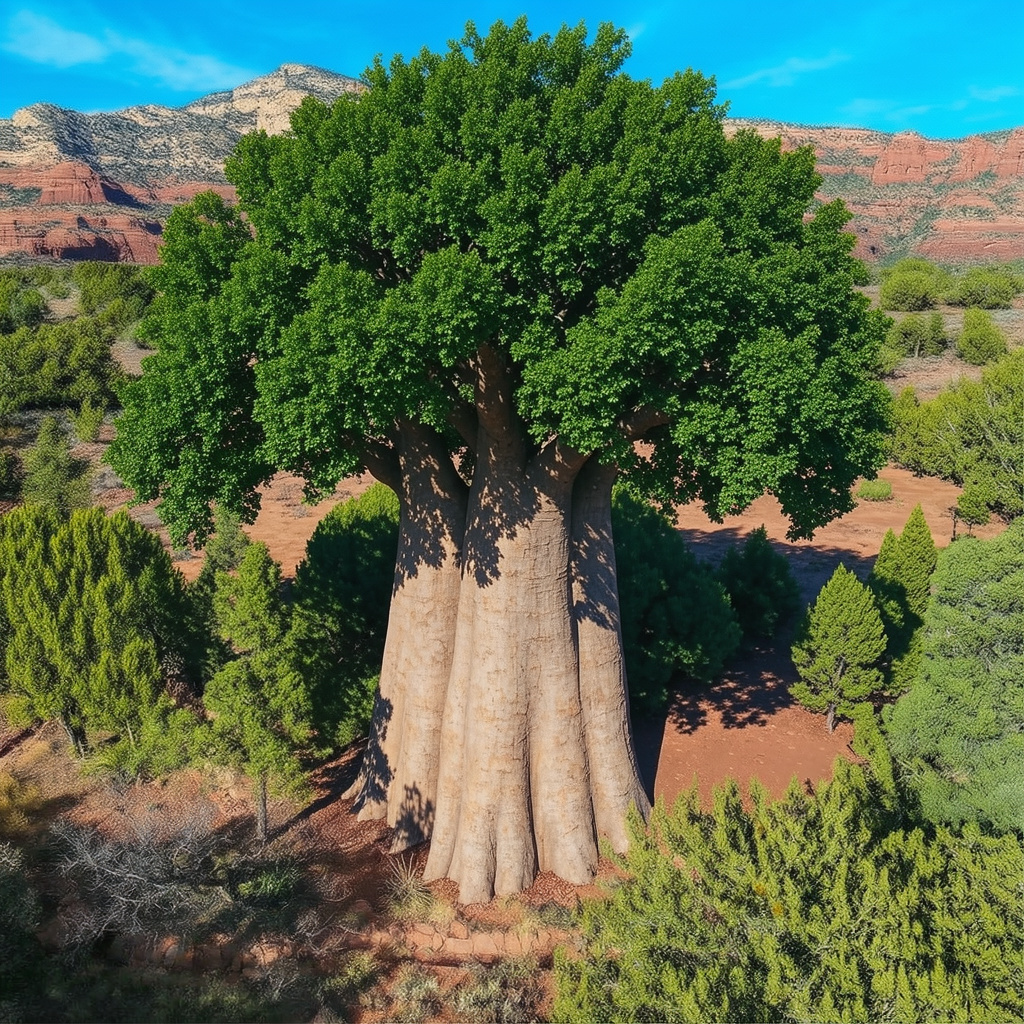} \\
\qtxtinstr{\fontsize{6}{7}\selectfont Replace the blue bird in the image with a red fox.} & \qimg{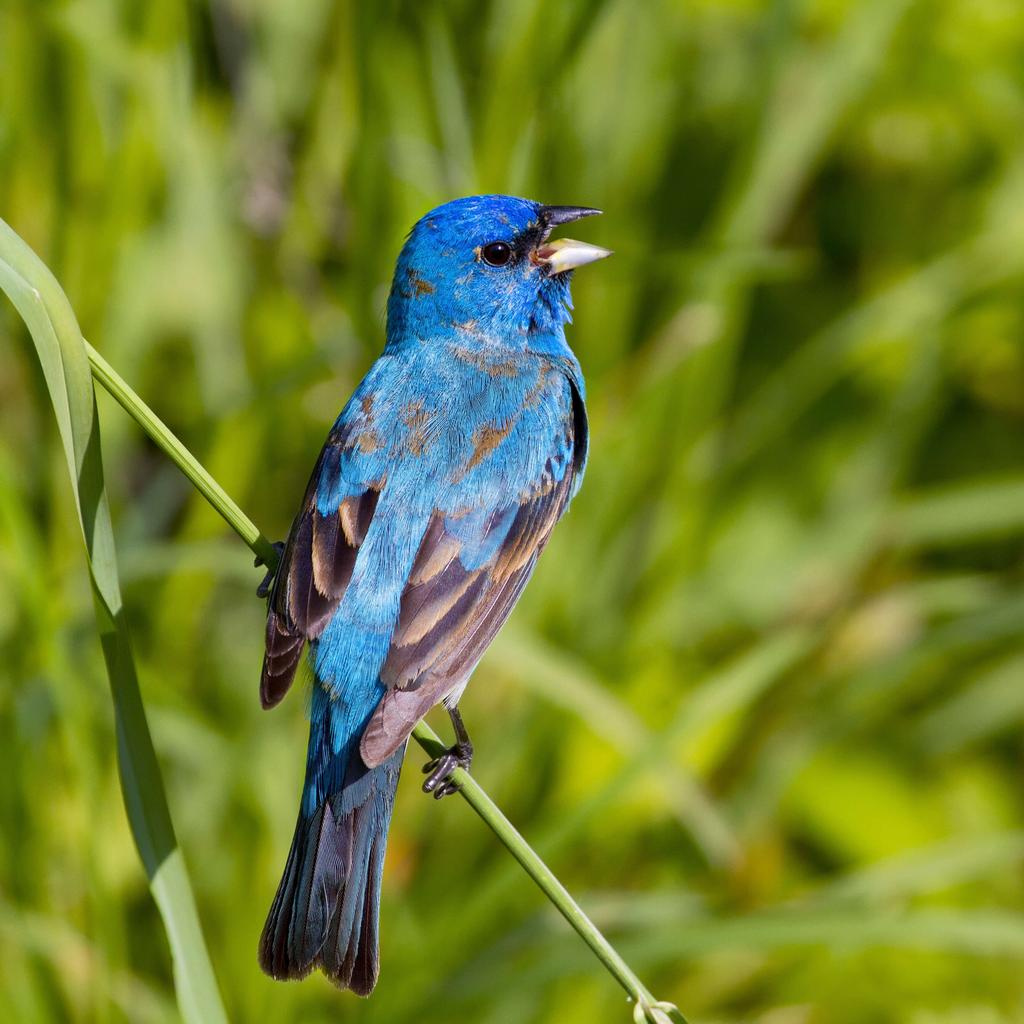} & \qimg{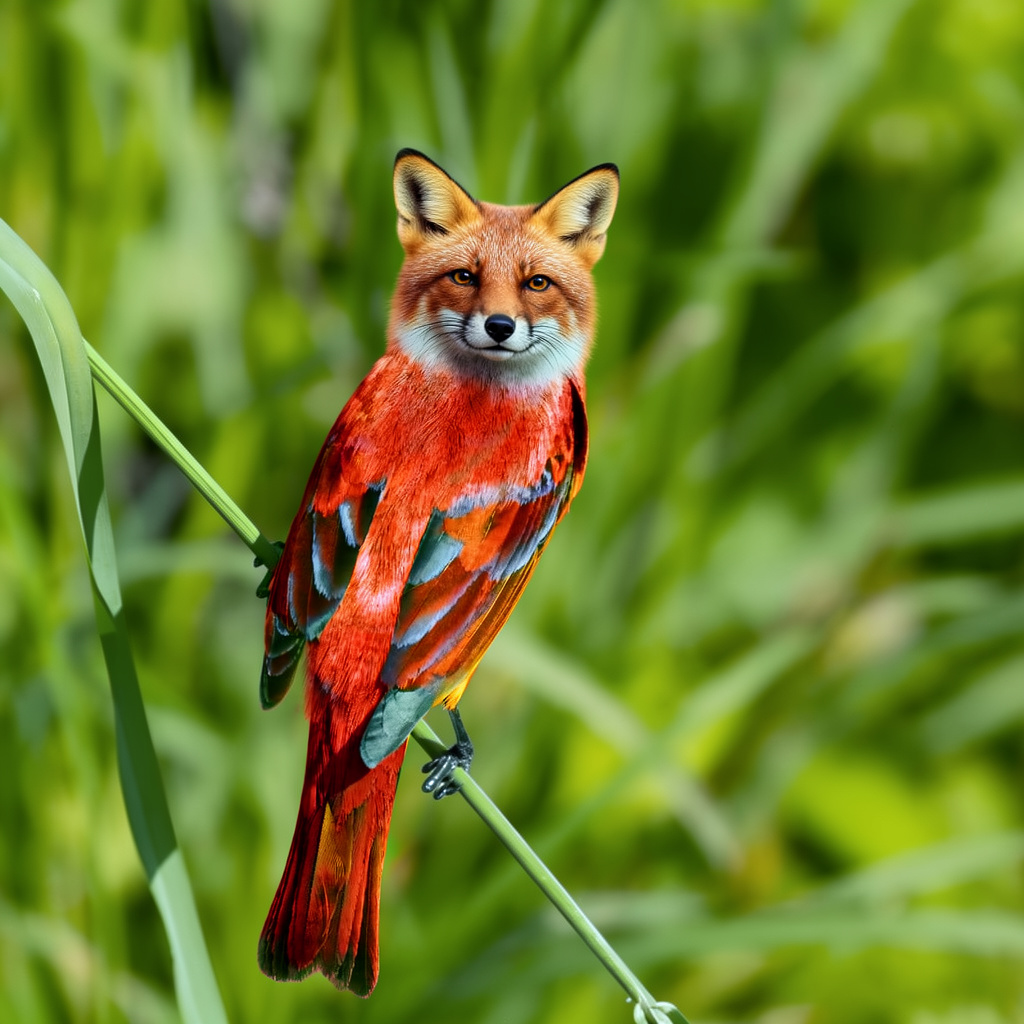} & \qtxtcap{\fontsize{5.2}{5.9}\selectfont A close-up, square-framed wildlife photograph of a vibrant red fox sitting amidst tall, lush green grass. The fox's body is angled away from the camera, but its\,\ldots{}} & \qimg{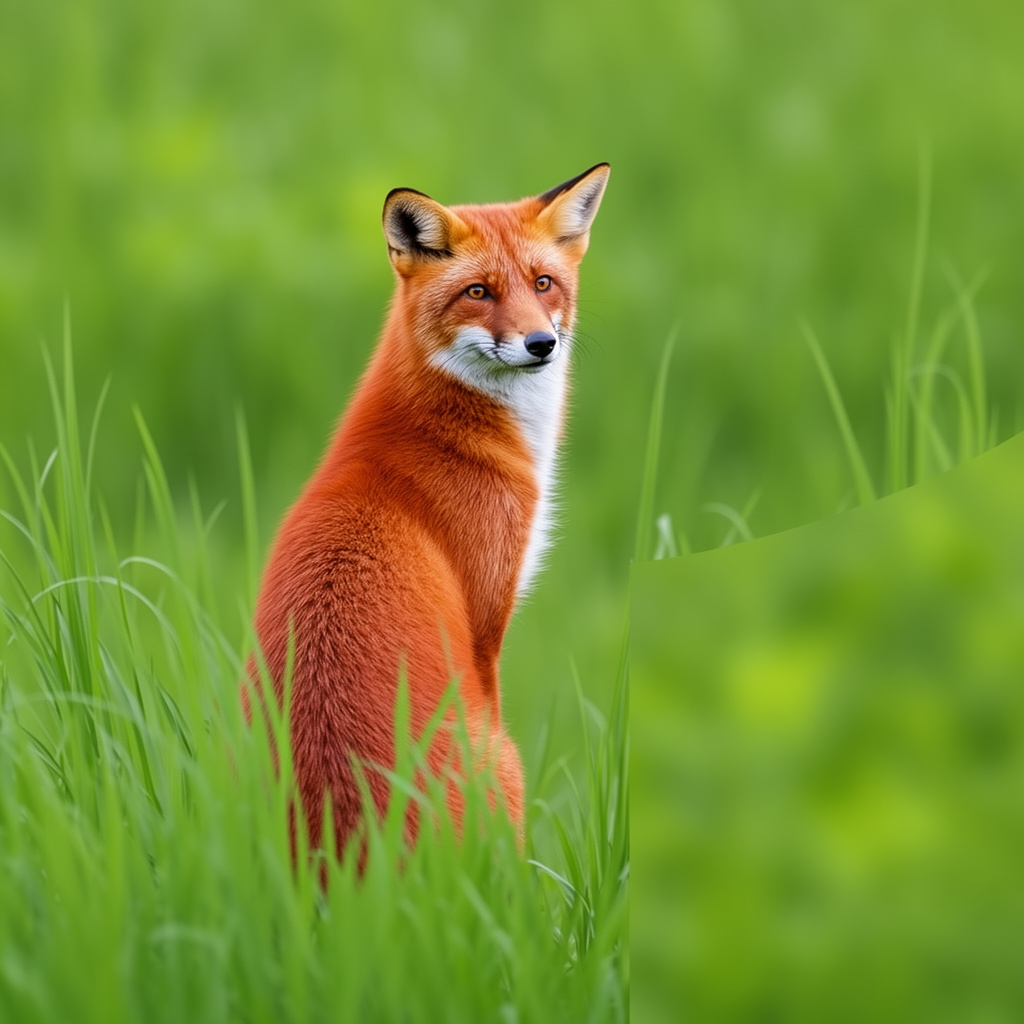} & \qimg{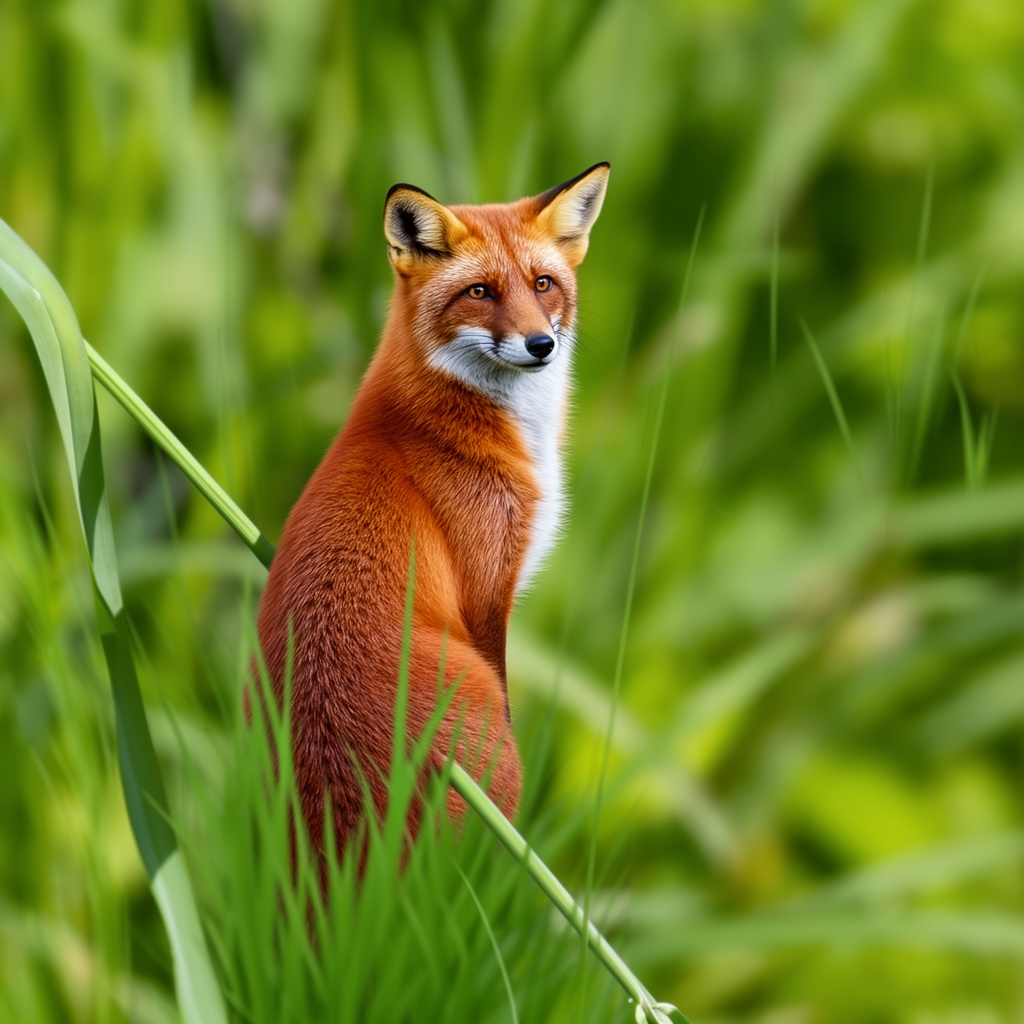} \\
\qtxtinstr{\fontsize{5.4}{6.2}\selectfont Replace the green armchair in the image with a wooden bookshelf.} & \qimg{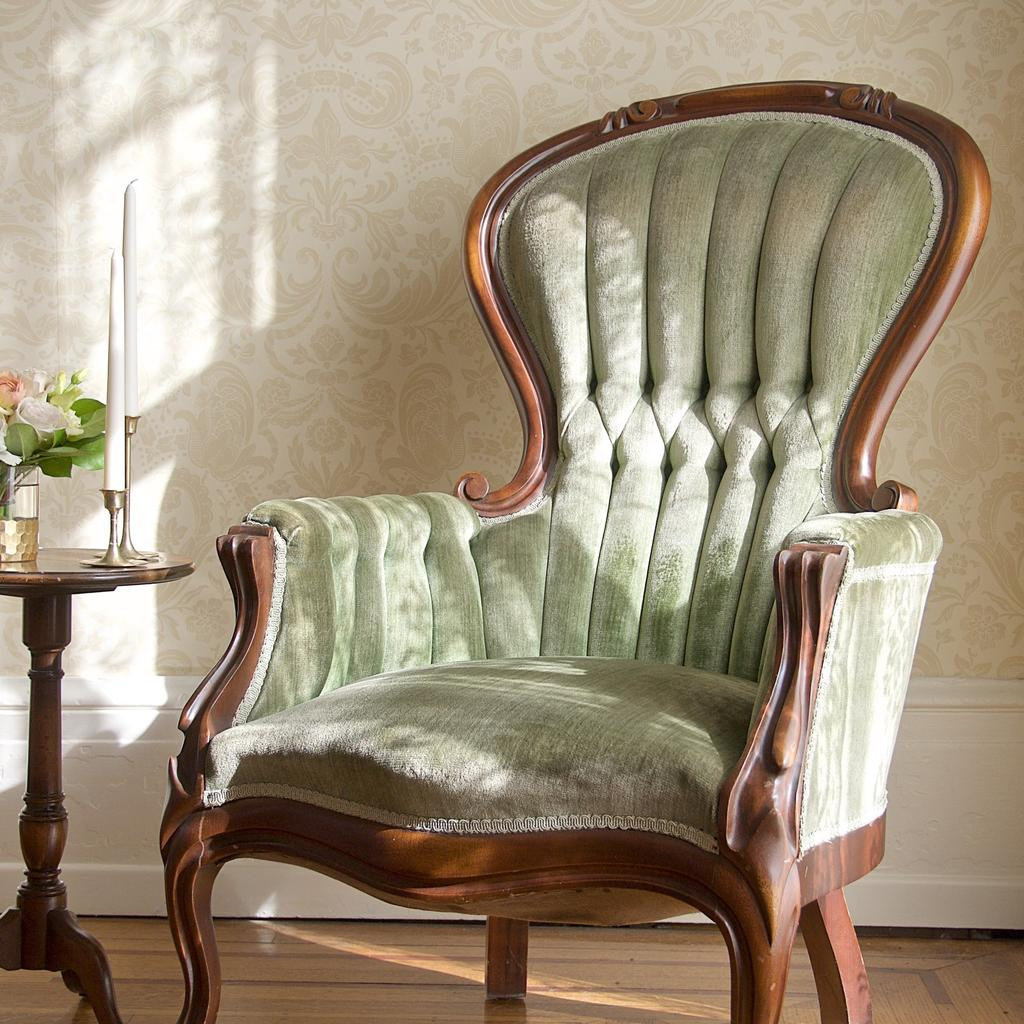} & \qimg{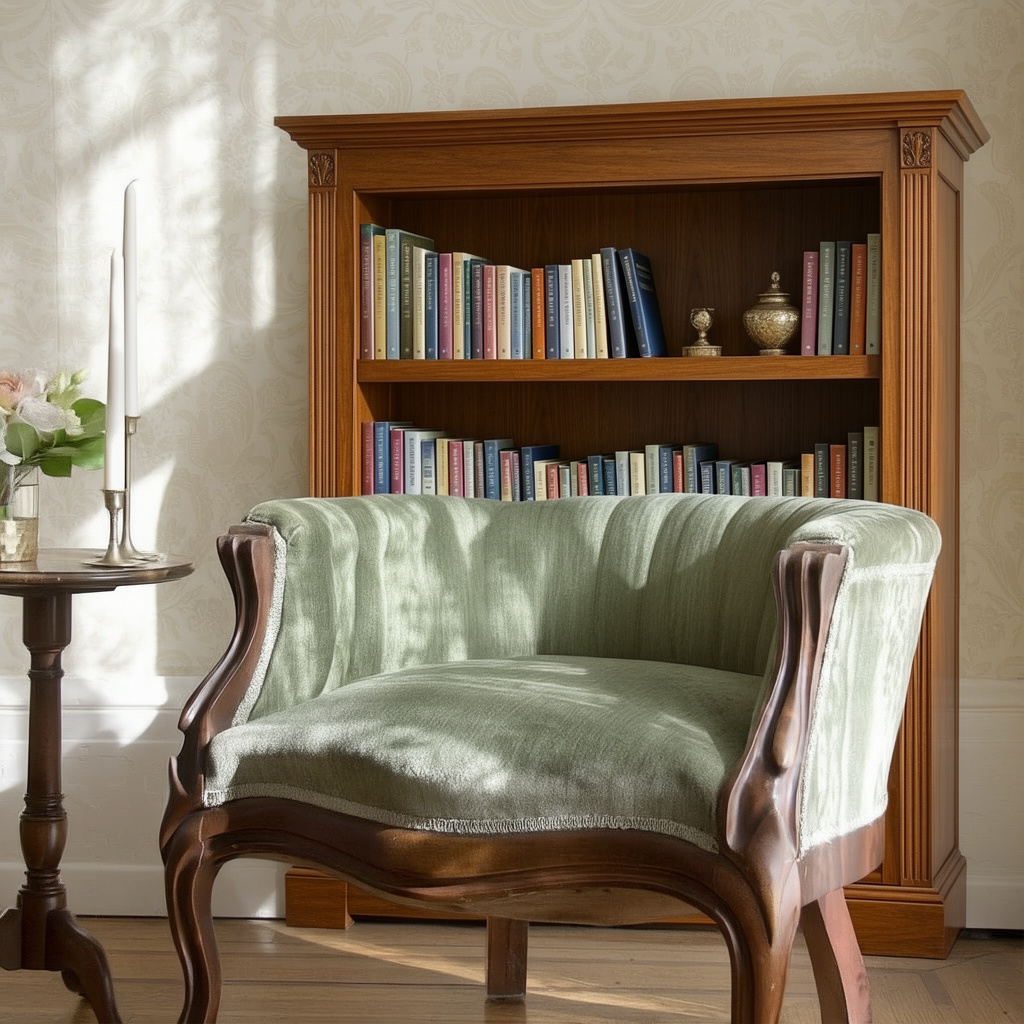} & \qtxtcap{\fontsize{5.2}{5.9}\selectfont A medium shot of a sunlit room corner with an elegant, classic aesthetic, viewed from a slightly right-angled perspective. On the right, a dark wooden bookshelf\,\ldots{}} & \qimg{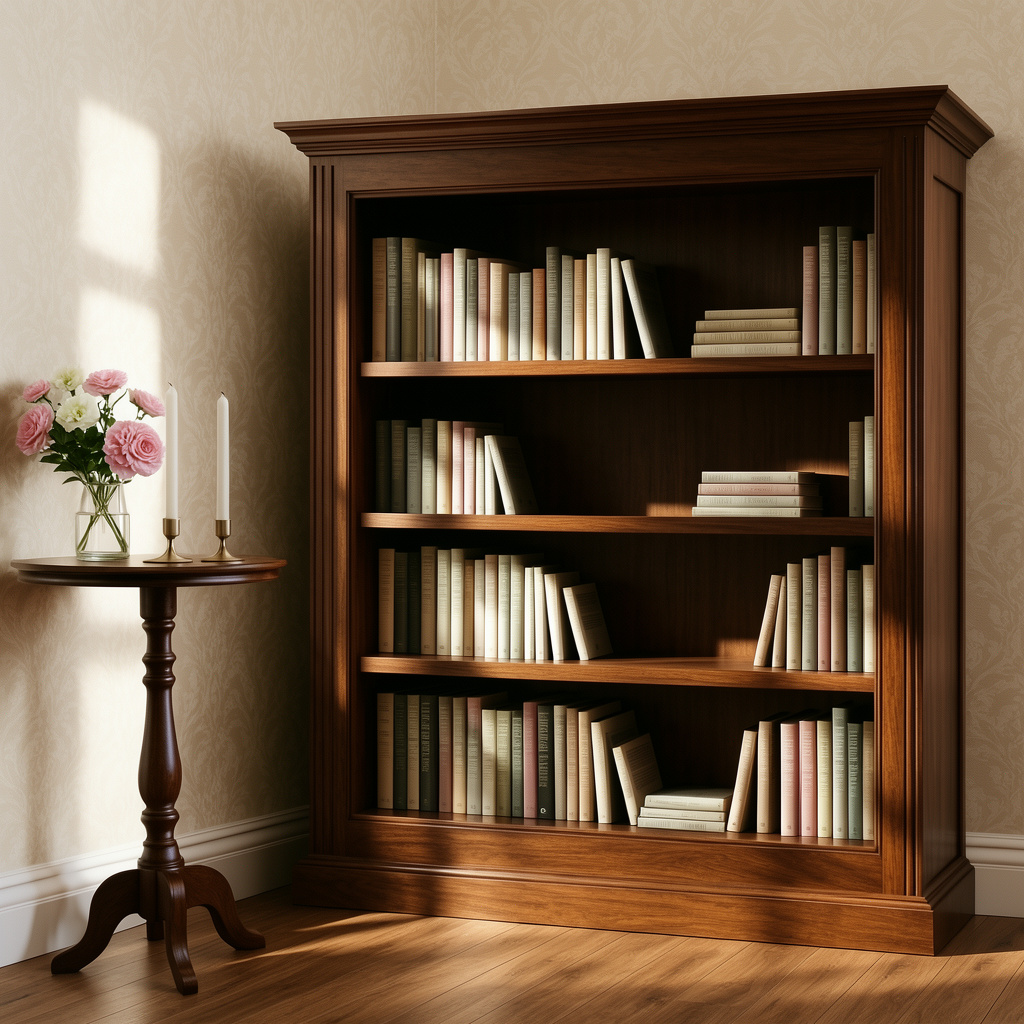} & \qimg{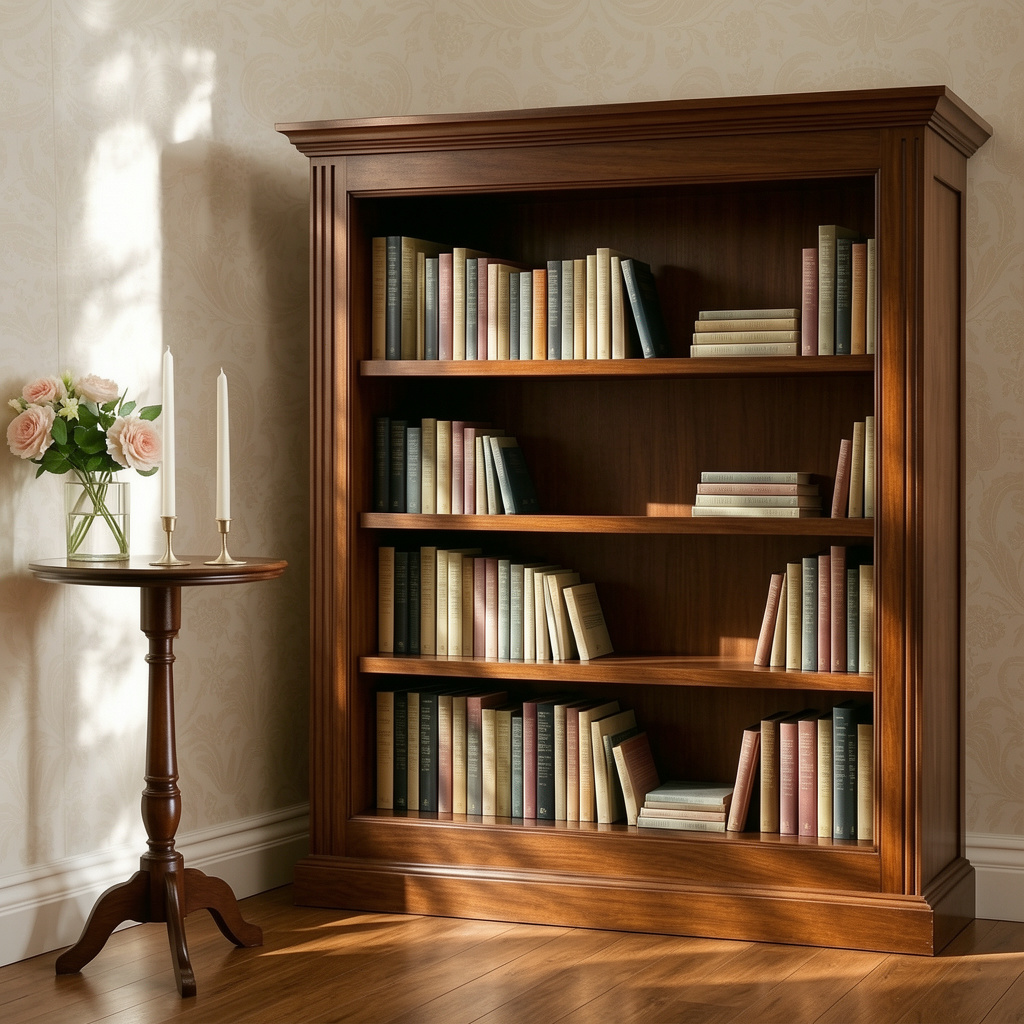} \\
\qtxtinstr{\fontsize{5.4}{6.2}\selectfont Extract the Polaris Ranger XP utility vehicle from the snowy outdoor scene in the image} & \qimg{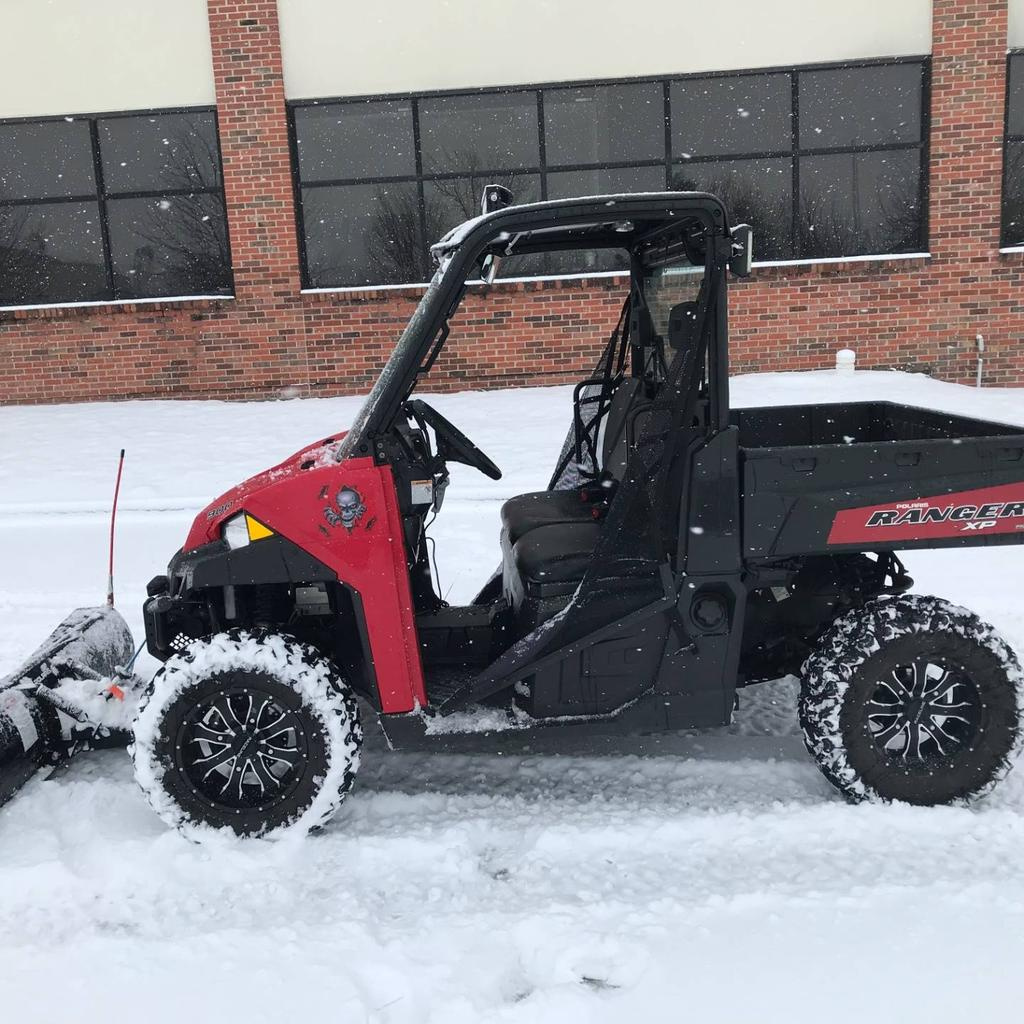} & \qimg{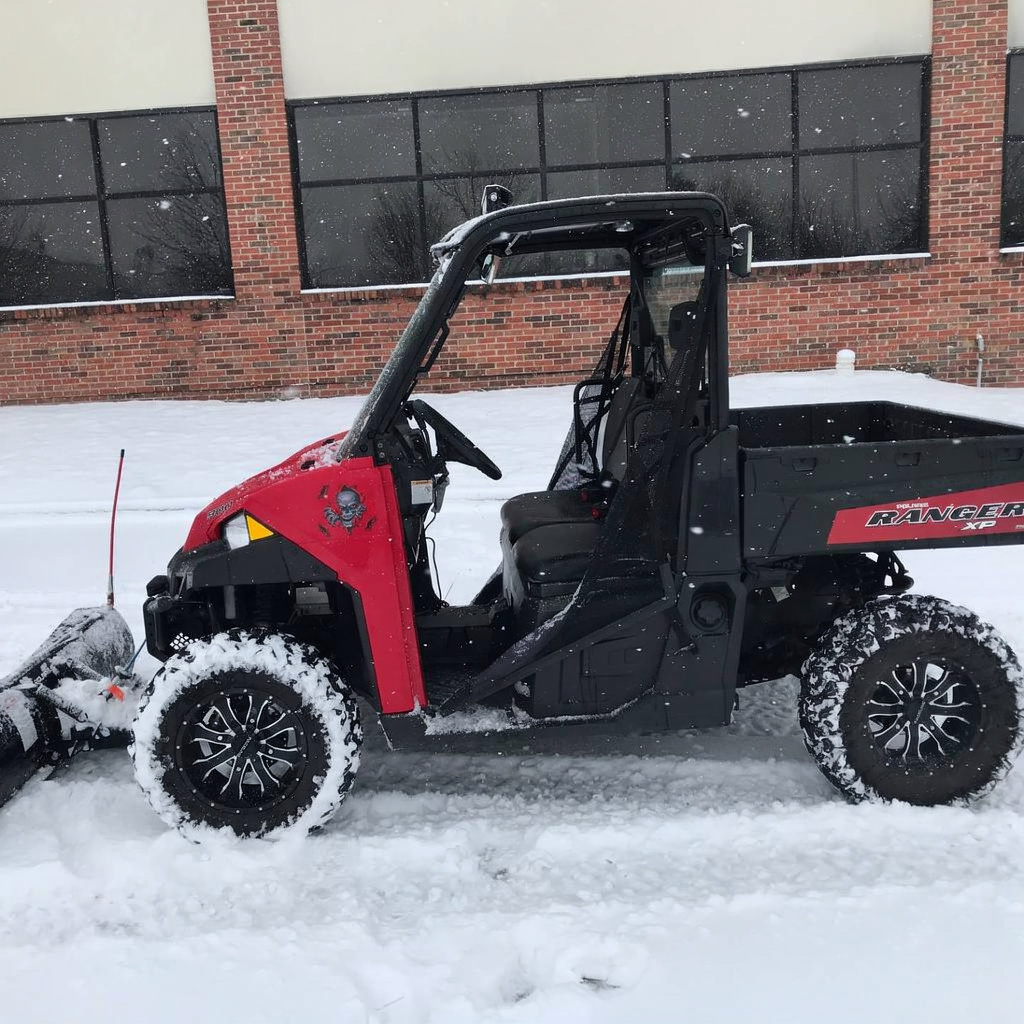} & \qtxtcap{\fontsize{5.2}{5.9}\selectfont A photorealistic shot of a red and black Polaris Ranger XP utility vehicle, isolated against a plain white background. The vehicle is seen from a side profile,\,\ldots{}} & \qimg{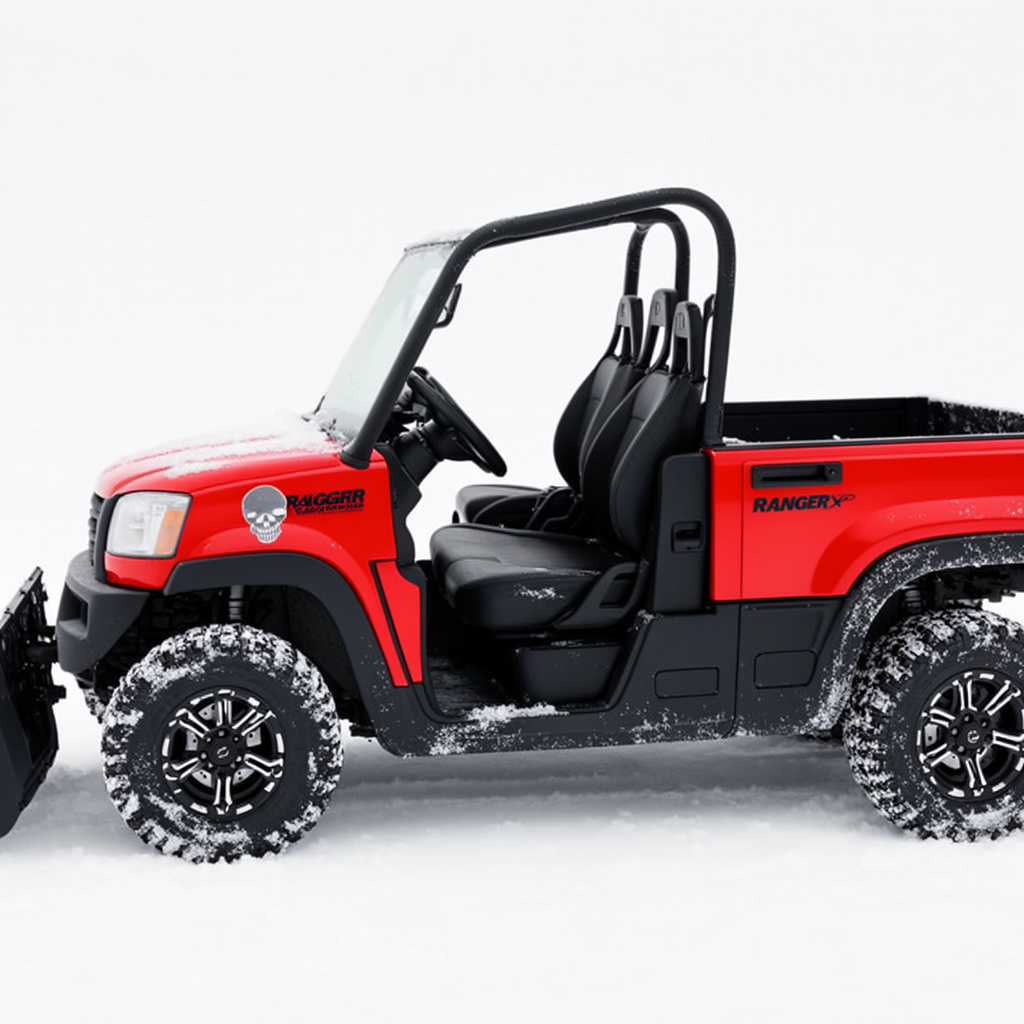} & \qimg{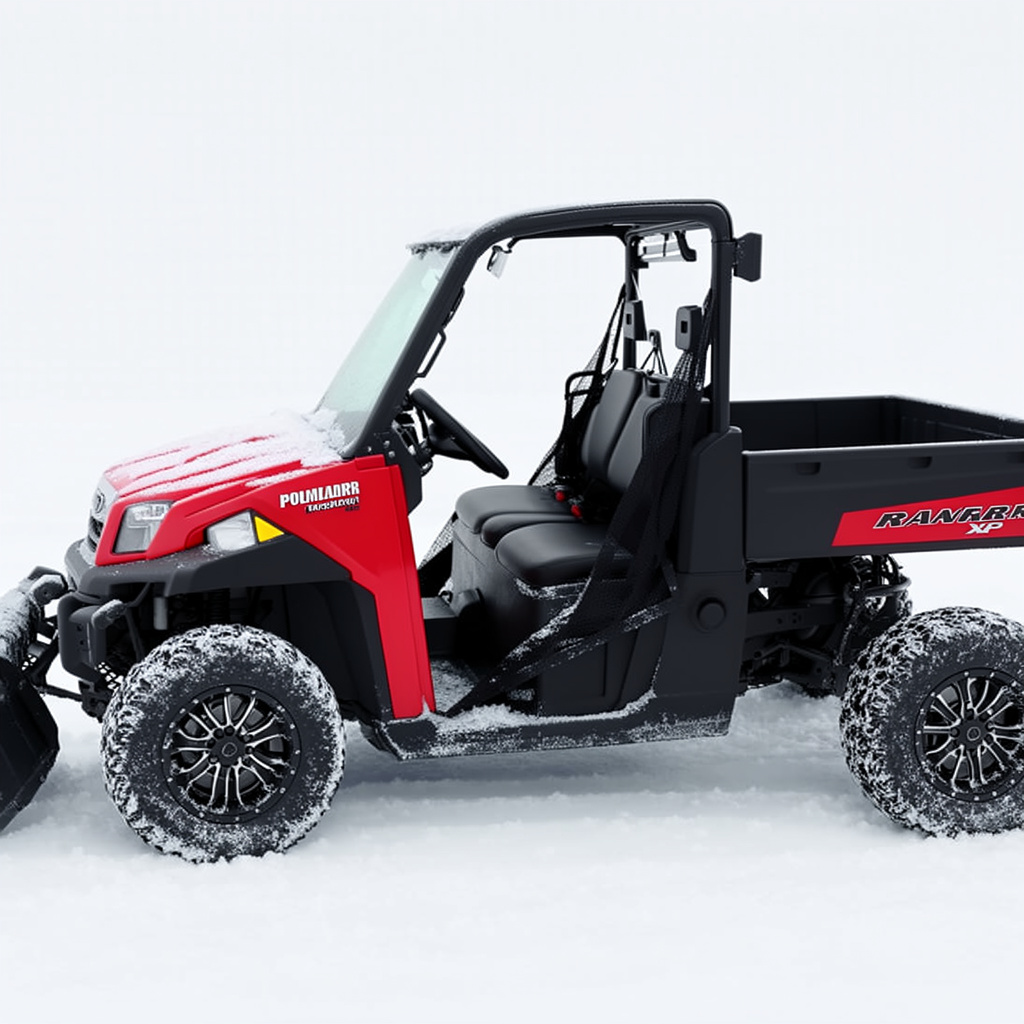} \\
\qtxtinstrtall{\fontsize{5.4}{6.2}\selectfont Change the scene setting to a sunny summer afternoon.} & \qimgtall{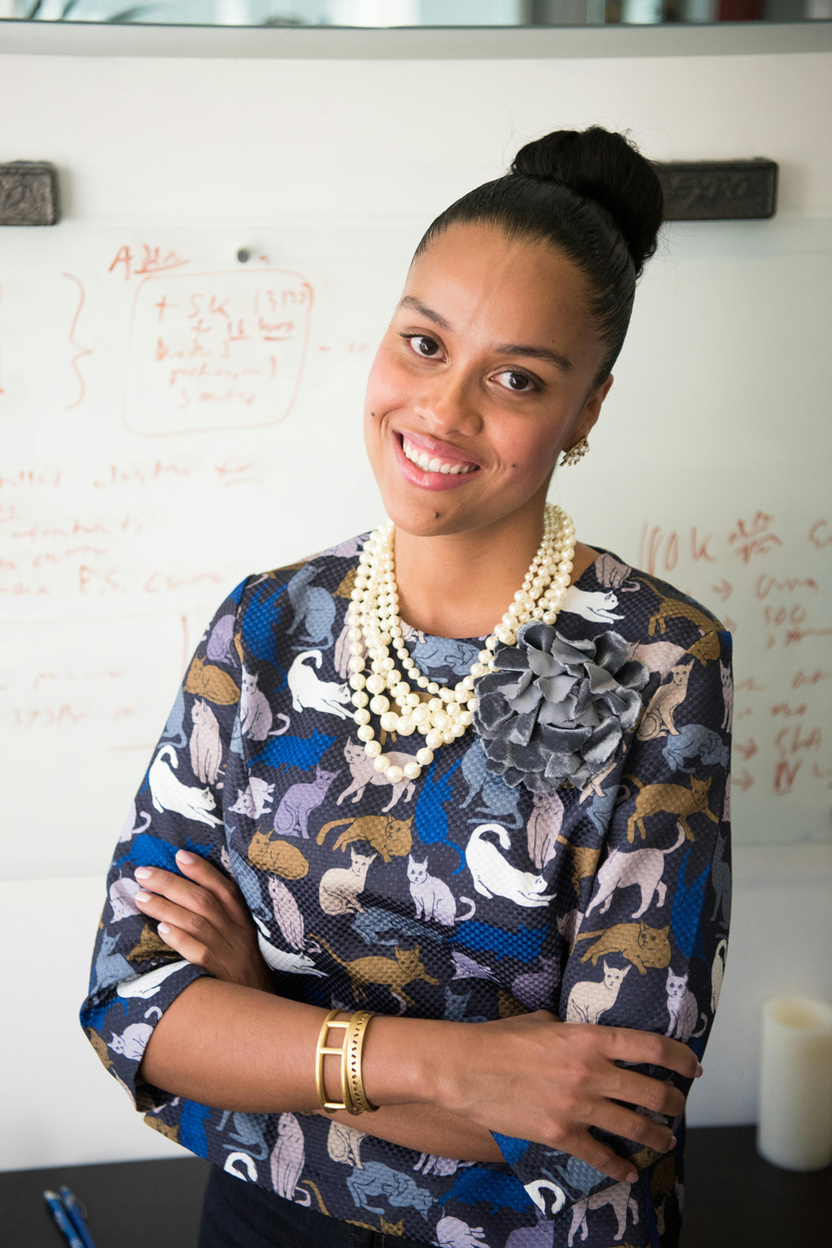} & \qimgtall{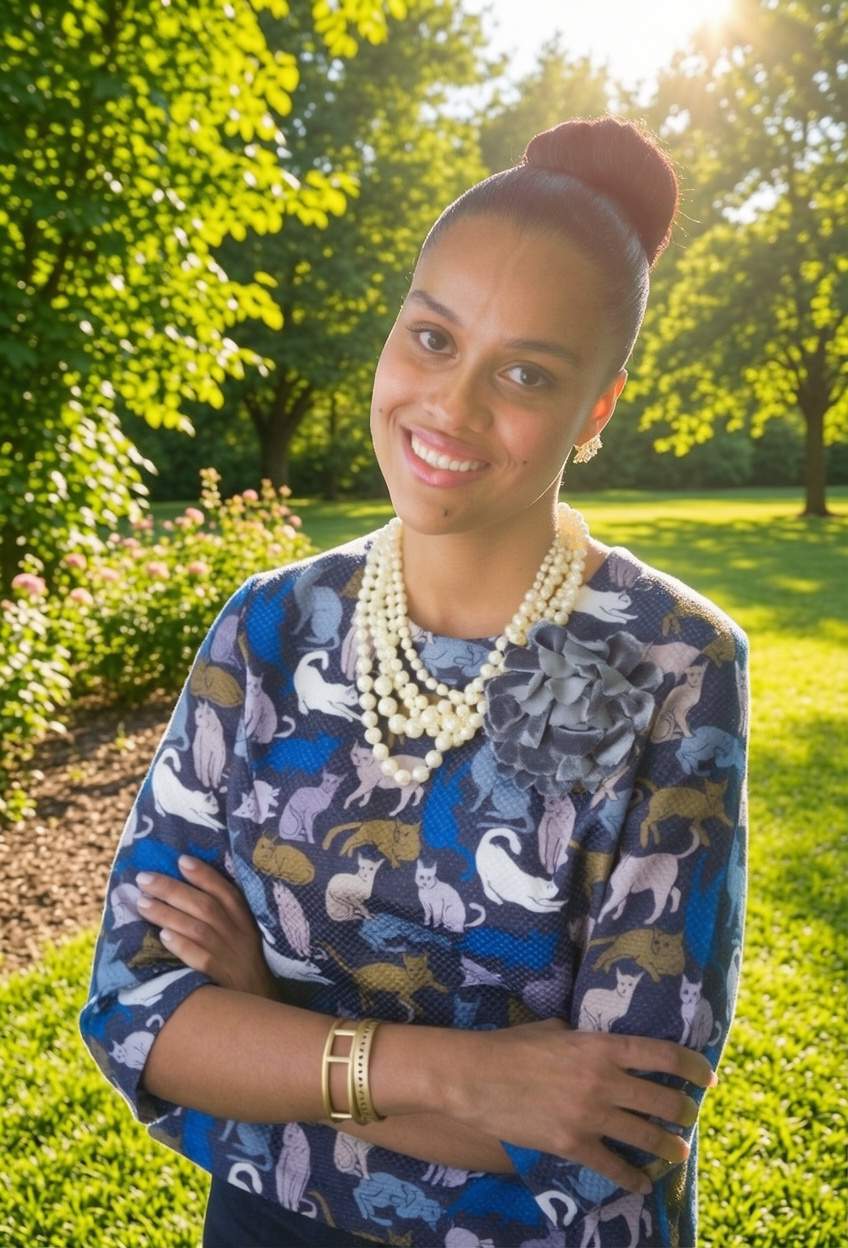} & \qtxtcaptall{\fontsize{5.2}{5.9}\selectfont A photorealistic medium shot of a friendly woman with light brown skin and dark hair pulled up into a neat high bun, standing outdoors on a sunny summer afternoon.\,\ldots{}} & \qimgtall{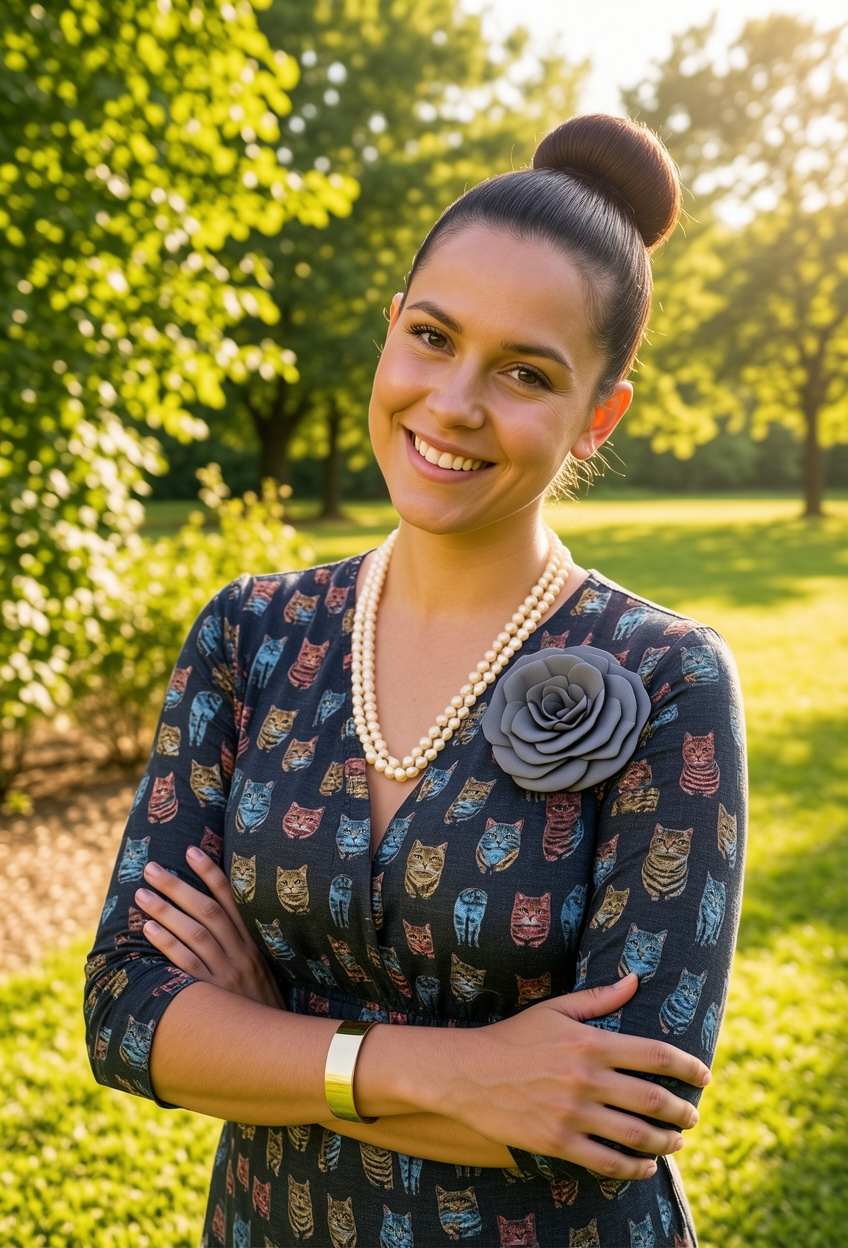} & \qimgtall{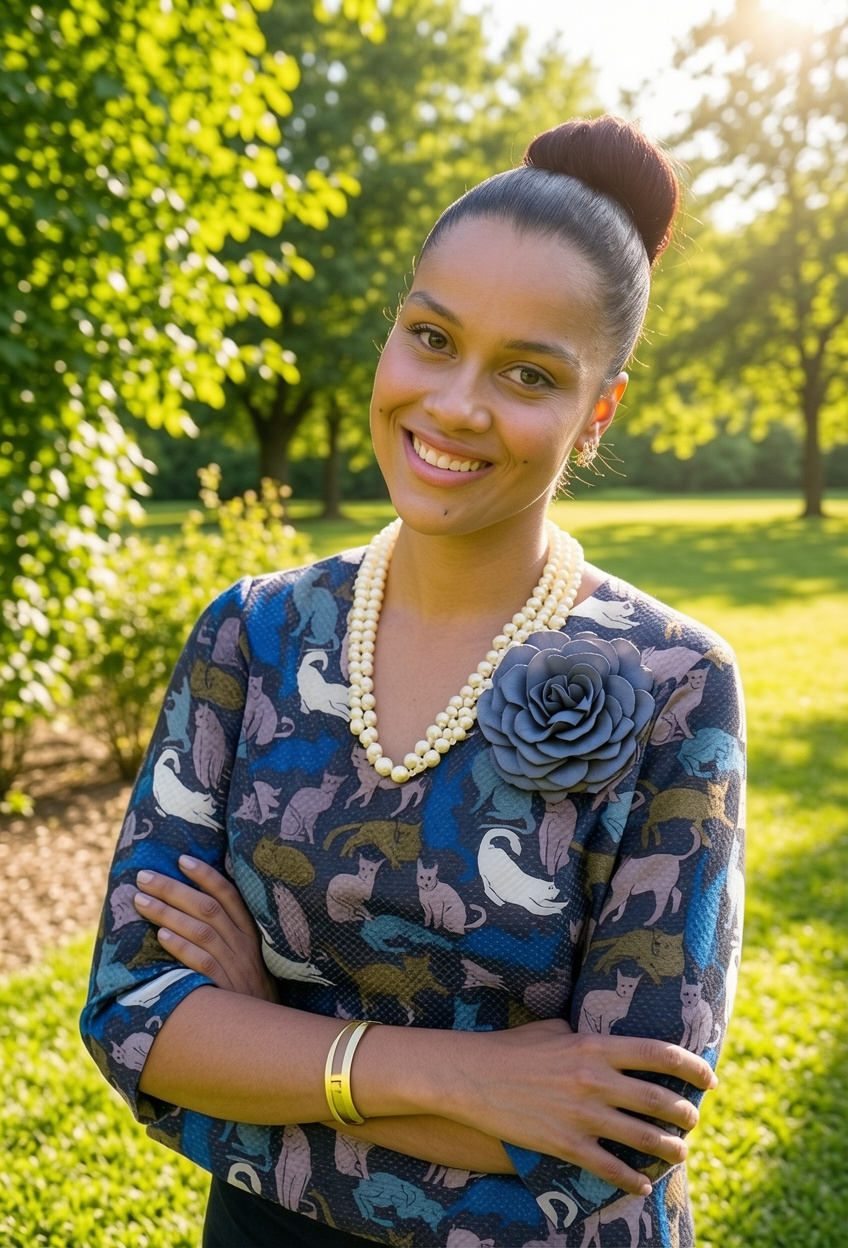} \\
\end{tabular}
\endgroup
\caption{\textbf{Returning to edit conditioning retains more of the source (FLUX2-Klein~4B).} A one-way T2I switch can improve instruction fidelity while changing unrelated content; resuming edit conditioning after the interval limits these changes.}
\label{fig:qualitative}
\end{figure}

\endgroup

\subsection{What accounts for the improvements?}
\label{sec:ablation}
\label{sec:guidance-baselines}

\begin{figure}[!htbp]
	\centering
	\vspace{8pt}
	\captionsetup{skip=4pt}
	\includegraphics[width=0.90\linewidth]{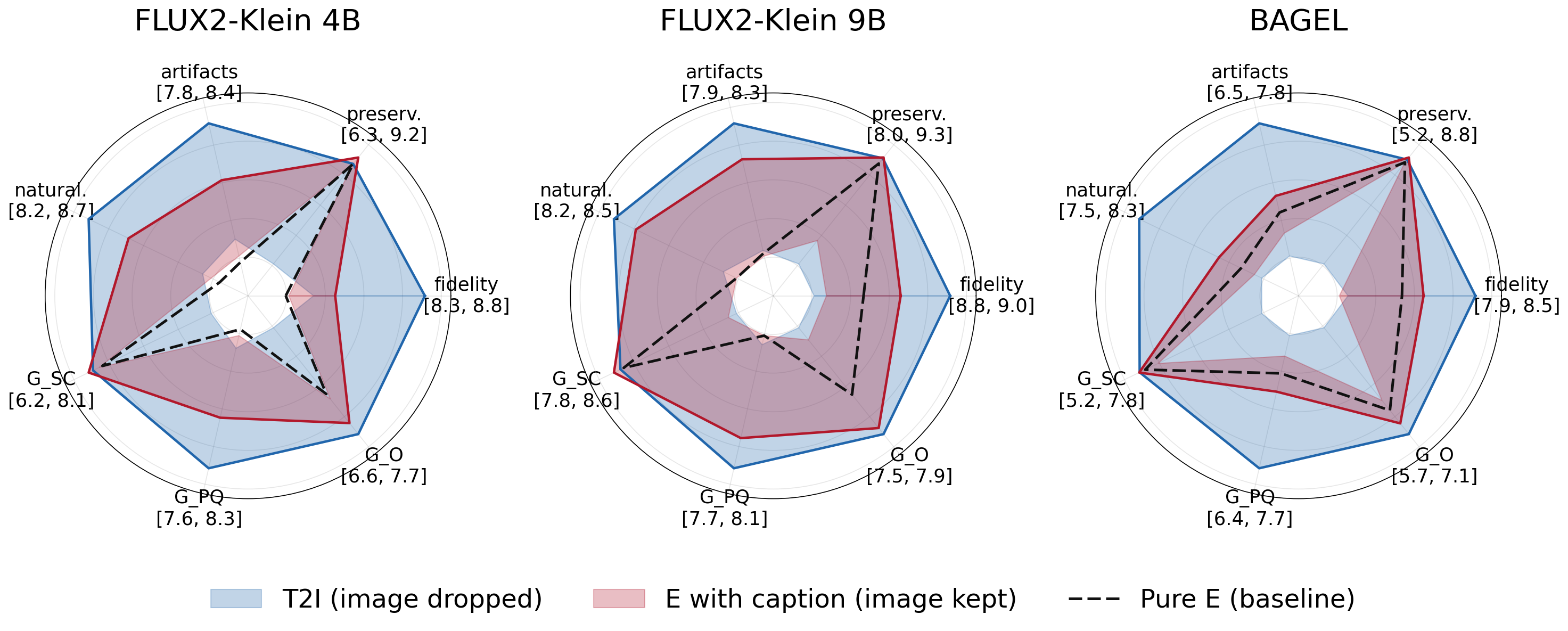}
	\caption{\textbf{Removing the source image allows larger quality gains with the same caption (GEdit, Qwen3.5).} Bands span per-metric extrema across switch timings, with and without a return to editing, so one band edge can combine several configurations. Axes normalized independently.}
	\label{fig:t2i-vs-i2c}
\end{figure}

\paragraph{Image and text conditioning.}
Switching changes two things at once, the image input and the text, and the controls in \cref{tab:cond-modes} pull them apart; \cref{tab:appx-gedit-qwen-summary} summarizes the best scores for each conditioning schedule. Rewording alone does help: a VLM-improved instruction early in the trajectory reaches the highest GEdit overall score among the control winners (\cref{fig:ablation}, left). But the perceptual-quality gain is specific to removing the source image. With the same caption, comparing conditioning with and without the source image (\cref{fig:t2i-vs-i2c}) shows that removing the image allows larger gains in fidelity, naturalness, and artifact scores on every model. The text matters in two different ways. Removing the image while keeping only the editing instruction loses semantic consistency, because nothing tells the model what the scene should become. Pairing the caption with the source image also loses semantic consistency, whether the pair is used for the whole trajectory or only outside the interval; here the caption does describe the target, so the likely cause is that the editor was trained to combine the source image with an instruction, not with a description of the result. The improvement therefore comes from switching the task, source-free generation toward a described target, and not from either change alone.

The strongest single competitor is the improved instruction used for the whole trajectory, pure E$^{*}$, which costs the same one VLM call and the same two denoiser passes per step as switching (\cref{tab:appx-estar-comparison}). It is a strong baseline on the two FLUX models, where it wins GEdit semantic consistency and keeps SSIM higher. But on every model the best T2I interval beats it on GIE-Bench functional correctness, MagicBrush alignment, and GEdit perceptual quality, and the perceptual-quality margin is large. Where E$^{*}$ wins semantic consistency, it does so through the preservation part of that score; fidelity is matched or exceeded by the T2I interval. On BAGEL, the improved instruction alone does not help at all, while the T2I interval improves every quality score. Better prompts make the editor follow the instruction more faithfully; only switching makes the result look better.

\begin{samepage}
\paragraph{Continuous guidance comparisons.}
Continuous guidance provides a natural alternative to switching. Reducing image guidance over the same intervals barely changes GEdit overall and lowers ImgEdit, because image-conditioned predictions still drive the update even at zero guidance. Adding a caption-conditioned branch instead blends edit and T2I velocities at every step. At a quarter editing weight, this improves semantic consistency, correctness, and preservation over the pure $\mathrm{E}\to\mathrm{T2I}\to\mathrm{E}$ switch, but requires four denoiser passes per step rather than two (\cref{app:guidance}). Switching is the endpoint of this family at the cost of pure editing and captures most of the benefit.
\par
\end{samepage}

\subsection{When does switching help?}
\label{sec:timing}

\begin{figure}[!t]
	\centering
	\captionsetup{skip=4pt}
	\includegraphics[width=0.92\linewidth,trim=0 0 0 54bp,clip]{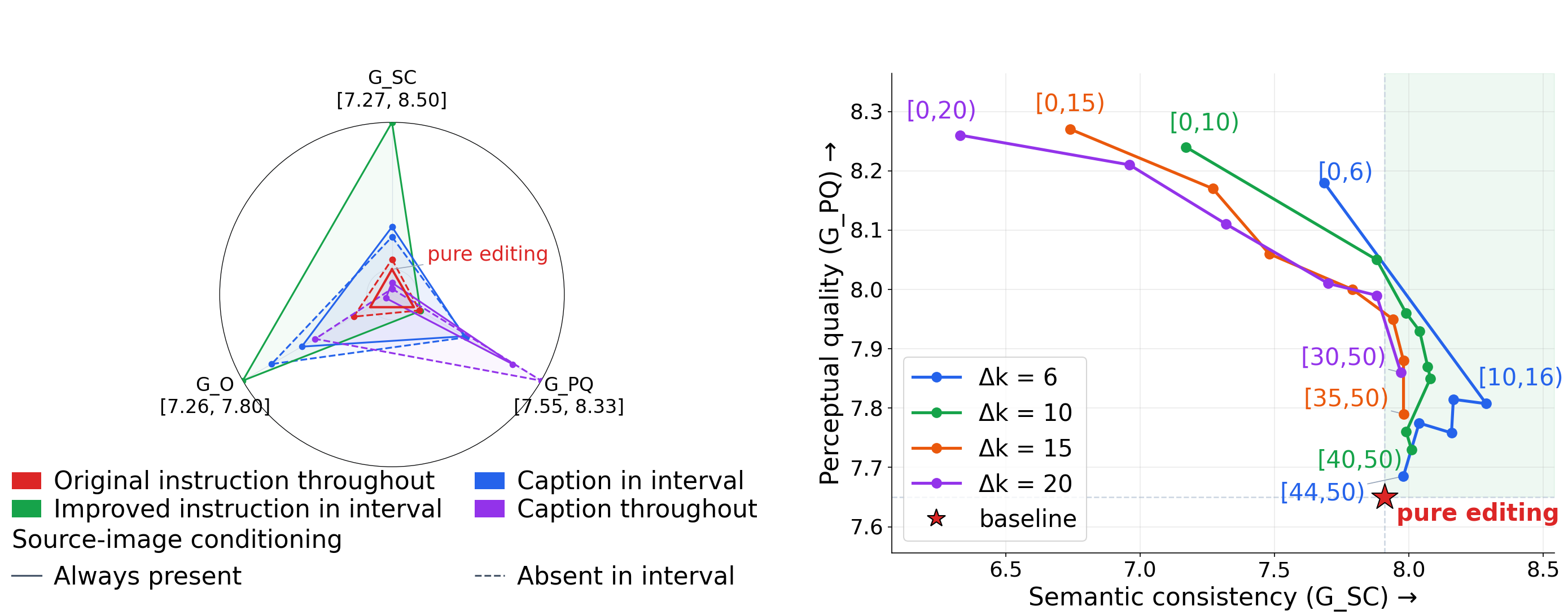}
	\caption{\textbf{Conditioning and timing affect edit quality (FLUX2-Klein~4B, GEdit, Qwen3.5).} \emph{Left:} best-$\mathrm{G\_O}$ configurations per conditioning group over shared placements in \cref{tab:appx-gedit-qwen,tab:appx-gedit-qwen-single}; red: pure editing. Colors denote text schedules; solid/dashed lines indicate source kept/removed in the interval. Axes use independent linear scales above and below pure editing. \emph{Right:} T2I sweeps by duration $\Delta k$; shading marks gains in both metrics.}
	\label{fig:ablation}
\end{figure}

Interval placement and duration are not independent knobs. Across all four durations, the sweeps trace one curve (\cref{fig:ablation}, right): moving an interval earlier or making it longer buys perceptual quality and spends semantic consistency, and every curve returns to the pure-editing point as the interval slides toward the end of the trajectory. What matters is how much of the trajectory is spent in T2I and how early. The semantic-consistency cost comes from its preservation half; because SC takes the per-sample minimum of fidelity and preservation, a fidelity gain is hidden whenever preservation drops further. Only short intervals in the early-middle of the trajectory land in the region that improves both scores, which is why $[10,16)$ is the featured setting. Beyond that region the choice is a trade: an editor that values naturalness should switch earlier or longer, one that values consistency later or shorter.

\subsection{Quality and source preservation}
\label{sec:preservation}
\begin{figure}[!htbp]
	\centering
	\captionsetup{skip=4pt}
	\includegraphics[width=0.9\linewidth]{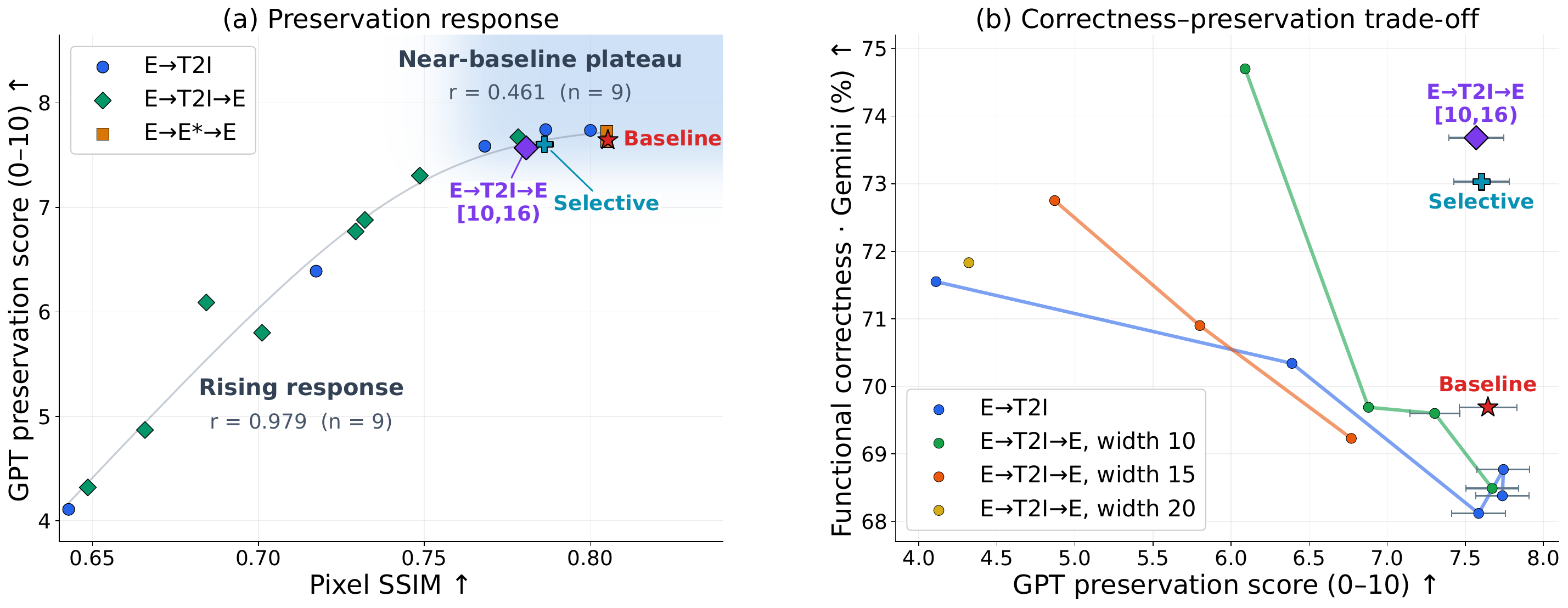}
	\caption{\textbf{Preservation and correctness on GIE-Bench (FLUX2-Klein~4B).} \emph{Left:} mean GPT-4.1 preservation versus SSIM outside the edit mask; shading: near-baseline plateau; faint curve: visual guide. Pearson $r$ is computed separately below and above SSIM $0.76$. \emph{Right:} Gemini-3-Flash correctness versus mean GPT preservation with $95\%$ source-clustered confidence intervals (\cref{app:preservation}).}
	\label{fig:frontier}
\end{figure}

The quality gains are not free, but the two kinds of preservation metric price them very differently. SSIM outside the edit mask and Unmasked-CLIP drop for every switching configuration, and significantly so (\cref{app:preservation}). A judge asked only about unintended changes sees far less. We score preservation with GPT-4.1 using GEdit's preservation criterion with the fidelity part removed, so the judge penalizes only changes that the instruction did not ask for (\cref{app:preservation}). On this measure the fixed interval moves mean preservation by at most $0.16$ points on a ten-point scale across the three models, and none of these shifts is significant. \Cref{fig:frontier} (right) puts the two sides together: the featured intervals gain correctness at judged preservation indistinguishable from pure editing, while the early one-way switches that gain the most correctness are the ones that pay for it. Withholding the interval from part of the edits (\cref{sec:selective}) recovers a portion of the SSIM loss on every model and brings judged preservation closer still to pure editing.

The two metrics disagree because they measure different things. \Cref{fig:frontier} (left) shows how far the pixel drop overstates the cost: across the 4B sweep, judged preservation forms a plateau at the pure-editing level while SSIM varies, and declines only below it. SSIM is known to fall sharply under small translations, rescaling, or rotations that leave an image perceptually unchanged \citep{wang2005translation,ding2020image}, and regenerating a scene from a caption for a few steps is likely to introduce exactly such shifts. SSIM therefore tracks pixel displacement, while the instruction-aware judge is the closer estimate of what a viewer would notice.

\subsection{Selective application as an illustration}
\label{sec:selective-duet}
\label{sec:selective}

\Cref{sec:intuition} showed that source attention reflects how much an edit still depends on the source. Read at the step before the interval, the same signal can decide whether to switch: edits that still depend on the source are more likely to be hurt by removing it. As an illustrative application, we summarize source attention at $k=9$ by its strength and its spatial unevenness:
\begin{equation}
\pi=Q_{0.75}(\{a_i\}),\qquad \mathrm{cv}=\sigma(\{a_i\})/\mu(\{a_i\}).
\label{eq:pi-cv}
\end{equation}
Edits with strong or uneven source attention stay in pure editing; the rest are switched. Both statistics come from the ordinary edit pass at no extra denoiser cost. Thresholds were tuned on GEdit outcomes for 4B and on GEdit probe distributions for 9B and BAGEL, then shared across all benchmarks (\cref{app:hparams}).

Selection does what it was meant to do. It keeps the quality gain over pure editing on every score and every model (\cref{tab:main}), and relative to the fixed interval it moves the operating point toward preservation: SSIM improves significantly on all three models and judged preservation moves closer to pure editing, at the price of somewhat smaller quality gains. The natural question is whether the attention rule is better than choosing edits at random. In a counterfactual replay on GEdit we therefore compare the rule with random routing tuned to spend the same mean preservation (\cref{app:hparams}). The rule wins on 4B and, with thresholds transferred by matching probe distributions alone, again on 9B: at equal preservation it delivers higher overall, perceptual-quality, and fidelity scores. The attention signal identifies edits that can afford the interval.

\section{Discussion and conclusion}
\label{sec:conclusion}
\label{sec:limitations}

Unified editors are trained for two tasks but use one at a time; this study puts the unused mode to work within an edit. An editor's attention to the source image depends on what the edit needs from it and fades over denoising for edits that rebuild the scene, leaving a stretch of the trajectory where source conditioning constrains without contributing. Switching to the model's own T2I mode on that stretch improves edit quality on every model and benchmark we tested, while an instruction-aware judge finds preservation essentially unchanged. The gain is a property of the switch rather than of the prompt: removing the source image is what unlocks perceptual quality, the caption is what keeps the scene on target, and continuous guidance reproduces the effect only at twice the cost. Placement and duration set where on the quality--preservation curve an editor lands, and the same attention signal that motivated the study can decide per edit whether to switch at all.

We use a simple switching schedule to establish that task switching can improve editing quality. More sophisticated switching and conditioning schemes may offer further gains. Our interval sweeps are coarse, so $[10,16)$ is one effective setting among those tested. A caption prepared before sampling cannot react to the trajectory, and its errors produce coherent but unintended changes. Preservation is measured by image similarity and a GPT judge, and how these track human judgments is untested.

\FloatBarrier
\subsection*{AI use statement}
AI assistants helped implement switching conditions, perform statistical tests of preservation, prepare figures, search the literature, polish author-written text, and identify inconsistencies in reported numerical results. The authors verified citations, experimental code, and numerical outputs and take responsibility for the final text, claims, and artifacts.

\subsection*{Reproducibility statement}
\Cref{sec:method,sec:selective-duet} specify the conditioning schedules and attention statistics, while \cref{sec:fixed-results,app:hparams} describe interval selection and model-specific thresholds. The experimental setup and evaluation metrics are described in \cref{sec:exps}, with extended results and guidance formulas in \cref{app:results,app:guidance}. \Cref{app:timing} reports captioner comparisons and timings, and \cref{app:preservation} describes the preservation judge protocol. Generation code and example notebooks for all three models are available at \url{https://github.com/pbnc5r6/conditioning-switching}.

\bibliography{references}
\bibliographystyle{iclr2027_conference}

\clearpage
\appendix

\clearpage
\section{Additional qualitative examples}
\label{app:qualitative}
We extend \cref{fig:qualitative} with 15 GEdit examples, grouped by the observed failure of pure editing: missed edits, incorrect edits, unnatural results, and incomplete edits. The column layout follows \cref{fig:qualitative}: edit instruction, source, pure editing, target caption, a one-way E$\to$T2I switch at step 10, and E$\to$T2I$\to$E on $[10,16)$. All examples use FLUX2-Klein~4B and 50 denoising steps.

\begin{figure}[!htb]
\centering

\begingroup
\sffamily
\setlength{\tabcolsep}{0pt}
\renewcommand{\arraystretch}{0}
\setlength{\fboxsep}{0pt}
\newlength{\geditmissedinstr}
\newlength{\geditmissedcell}
\newlength{\geditmissedcelltall}
\setlength{\geditmissedinstr}{0.110\linewidth}
\setlength{\geditmissedcell}{0.178\linewidth}
\setlength{\geditmissedcelltall}{1.000\geditmissedcell}
\providecommand{\geditmissedimg}[1]{%
  \parbox[c][\geditmissedcell][c]{\geditmissedcell}{\centering\includegraphics[width=\geditmissedcell,height=\geditmissedcell,keepaspectratio]{#1}}%
}
\providecommand{\geditmissedimgtall}[1]{%
  \parbox[c][\geditmissedcelltall][c]{\geditmissedcell}{\centering\includegraphics[width=\geditmissedcell]{#1}}%
}
\providecommand{\geditmissedmissing}{%
  \parbox[c][\geditmissedcell][c]{\geditmissedcell}{\centering\textcolor{black!10}{\rule{\geditmissedcell}{\geditmissedcell}}}%
}
\providecommand{\geditmissedmissingtall}{%
  \parbox[c][\geditmissedcelltall][c]{\geditmissedcell}{\centering\textcolor{black!10}{\rule{\geditmissedcell}{\geditmissedcelltall}}}%
}
\providecommand{\geditmissedtextpad}{\vspace{2pt}\vspace{0pt}}
\providecommand{\geditmissedtxtinstr}[1]{%
  \parbox[c][\geditmissedcell][t]{\dimexpr\geditmissedinstr-2pt\relax}{\geditmissedtextpad\raggedright #1\par}%
}
\providecommand{\geditmissedtxtinstrtall}[1]{%
  \parbox[c][\geditmissedcelltall][t]{\dimexpr\geditmissedinstr-2pt\relax}{\geditmissedtextpad\raggedright #1\par}%
}
\providecommand{\geditmissedtxtcap}[1]{%
  \parbox[c][\geditmissedcell][t]{\dimexpr\geditmissedcell-2pt\relax}{\geditmissedtextpad\raggedright #1\par}%
}
\providecommand{\geditmissedtxtcaptall}[1]{%
  \parbox[c][\geditmissedcelltall][t]{\dimexpr\geditmissedcell-2pt\relax}{\geditmissedtextpad\raggedright #1\par}%
}
\providecommand{\geditmissedhdrinstr}[1]{%
  \parbox[t]{\dimexpr\geditmissedinstr-2pt\relax}{\centering\geditmissedtextpad #1\par}%
}
\providecommand{\geditmissedhdr}[1]{%
  \parbox[t]{\dimexpr\geditmissedcell-2pt\relax}{\centering\geditmissedtextpad #1\par}%
}

\begin{tabular}{@{}>{\raggedright\arraybackslash}m{\geditmissedinstr}*{5}{>{\centering\arraybackslash}m{\geditmissedcell}}@{}}
\geditmissedhdrinstr{\fontsize{7}{8}\selectfont\bfseries instruction} & \geditmissedhdr{\fontsize{7}{8}\selectfont\bfseries source} & \geditmissedhdr{\fontsize{7}{8}\selectfont\bfseries pure-E} & \geditmissedhdr{\fontsize{7}{8}\selectfont\bfseries T2I caption} & \geditmissedhdr{\fontsize{7}{8}\selectfont\bfseries E$\to$T2I $k{=}10$} & \geditmissedhdr{\fontsize{7}{8}\selectfont\bfseries E$\to$T2I$\to$E\\$[10,16)$} \\
\noalign{\vskip 1pt}
\begingroup\setlength{\geditmissedcelltall}{0.849359\geditmissedcell}\geditmissedtxtinstrtall{\fontsize{6}{7}\selectfont remove the freight train}\endgroup & \begingroup\setlength{\geditmissedcelltall}{0.849359\geditmissedcell}\parbox[c][\geditmissedcelltall][c]{\geditmissedcell}{\includegraphics[width=\geditmissedcell,height=\geditmissedcelltall,trim={116.800000bp 0.000000bp 116.800000bp 0.000000bp},clip]{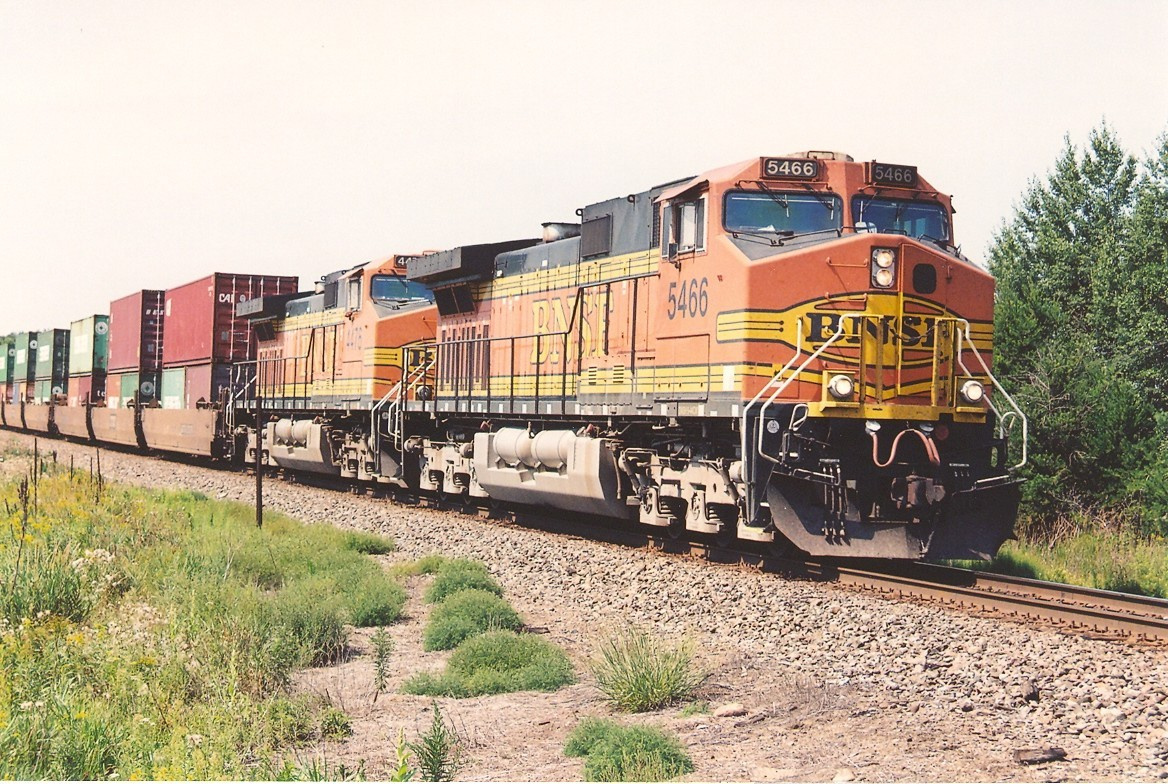}}\endgroup & \begingroup\setlength{\geditmissedcelltall}{0.849359\geditmissedcell}\parbox[c][\geditmissedcelltall][c]{\geditmissedcell}{\includegraphics[width=\geditmissedcell,height=\geditmissedcelltall,trim={124.800000bp 0.000000bp 124.800000bp 0.000000bp},clip]{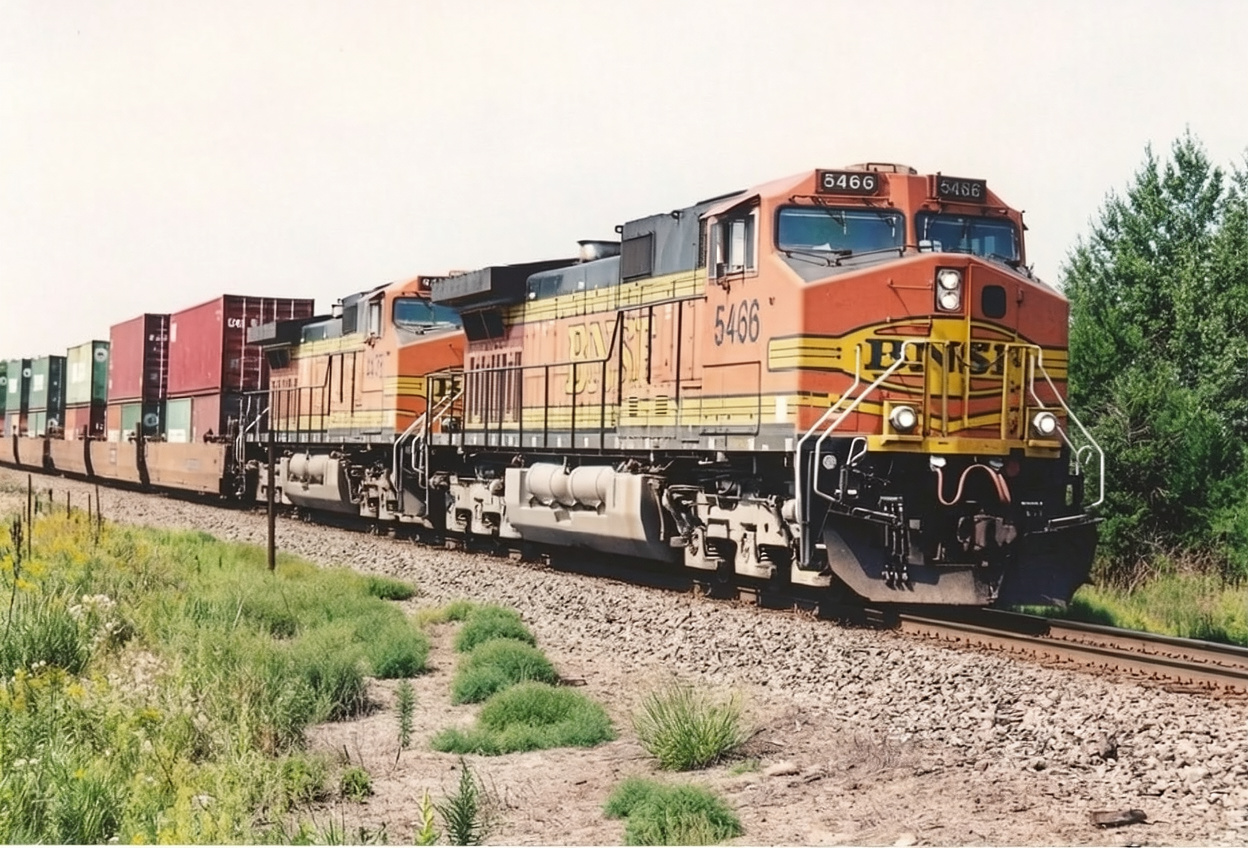}}\endgroup & \begingroup\setlength{\geditmissedcelltall}{0.849359\geditmissedcell}\geditmissedtxtcaptall{\fontsize{5.2}{5.9}\selectfont A low-angle, realistic photograph of empty railroad tracks curving through a rural landscape on a bright, hazy day. The steel rails, set on a raised bed of gray\,\ldots{}}\endgroup & \begingroup\setlength{\geditmissedcelltall}{0.849359\geditmissedcell}\parbox[c][\geditmissedcelltall][c]{\geditmissedcell}{\includegraphics[width=\geditmissedcell,height=\geditmissedcelltall,trim={124.800000bp 0.000000bp 124.800000bp 0.000000bp},clip]{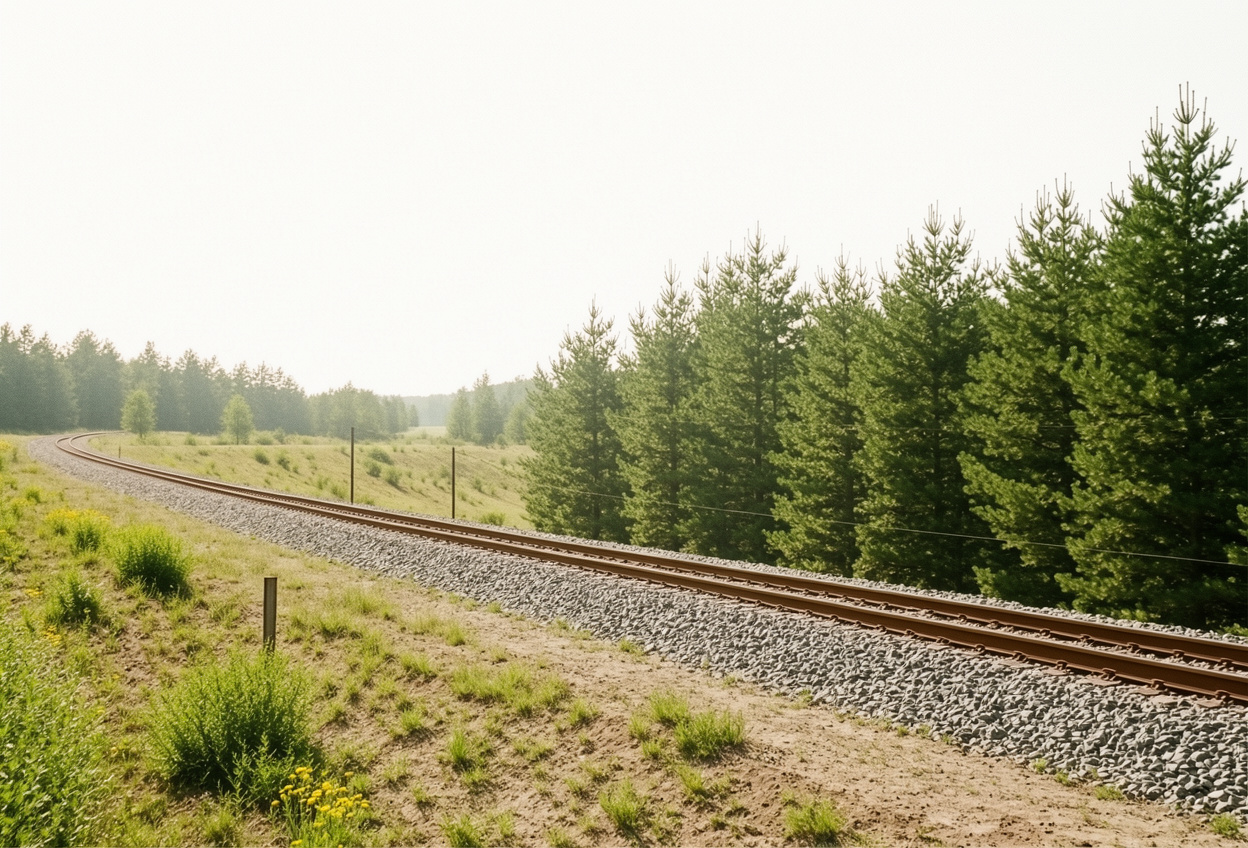}}\endgroup & \begingroup\setlength{\geditmissedcelltall}{0.849359\geditmissedcell}\parbox[c][\geditmissedcelltall][c]{\geditmissedcell}{\includegraphics[width=\geditmissedcell,height=\geditmissedcelltall,trim={124.800000bp 0.000000bp 124.800000bp 0.000000bp},clip]{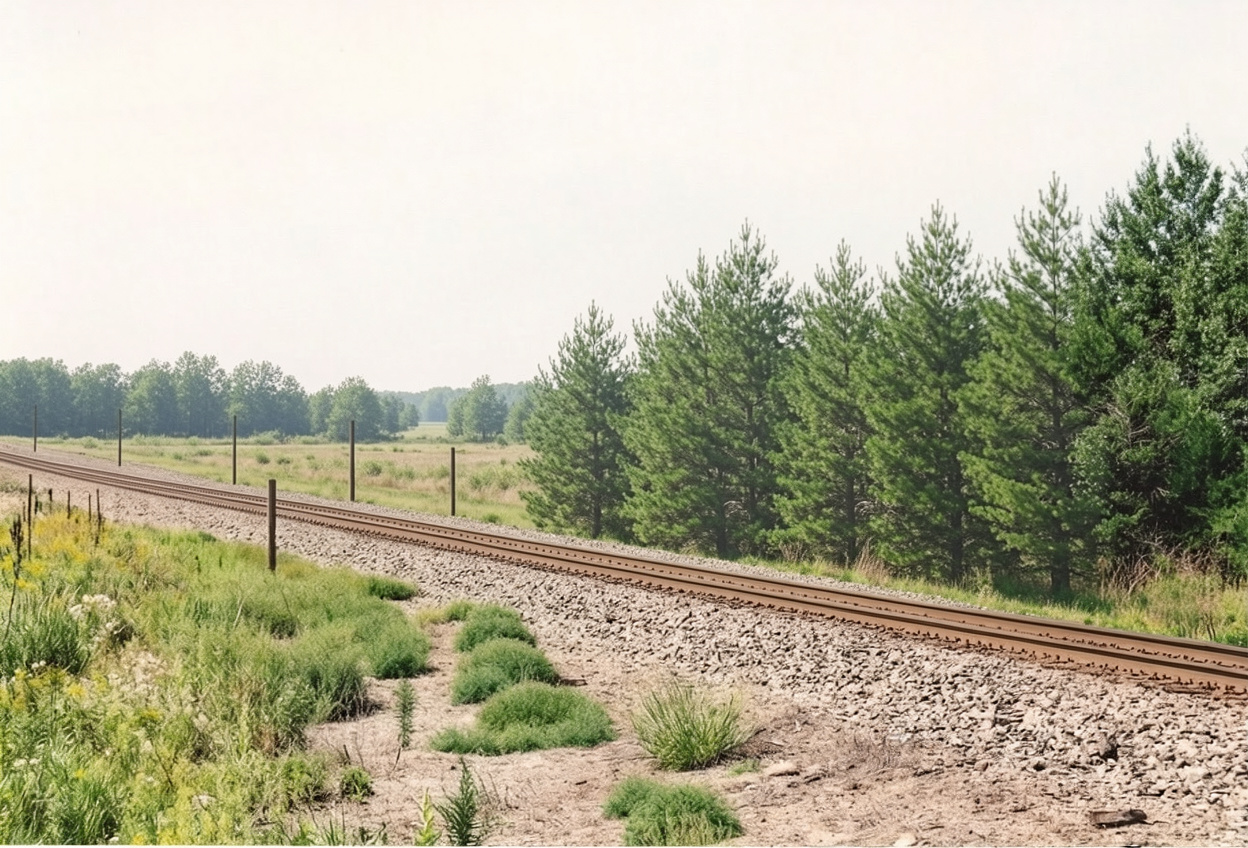}}\endgroup \\
\begingroup\setlength{\geditmissedcelltall}{1.000000\geditmissedcell}\geditmissedtxtinstrtall{\fontsize{6}{7}\selectfont Turn the baby's balloon into an ice cream cone.}\endgroup & \begingroup\setlength{\geditmissedcelltall}{1.000000\geditmissedcell}\parbox[c][\geditmissedcelltall][c]{\geditmissedcell}{\includegraphics[width=\geditmissedcell,height=\geditmissedcelltall,trim={0.000000bp 0.000000bp 0.000000bp 0.000000bp},clip]{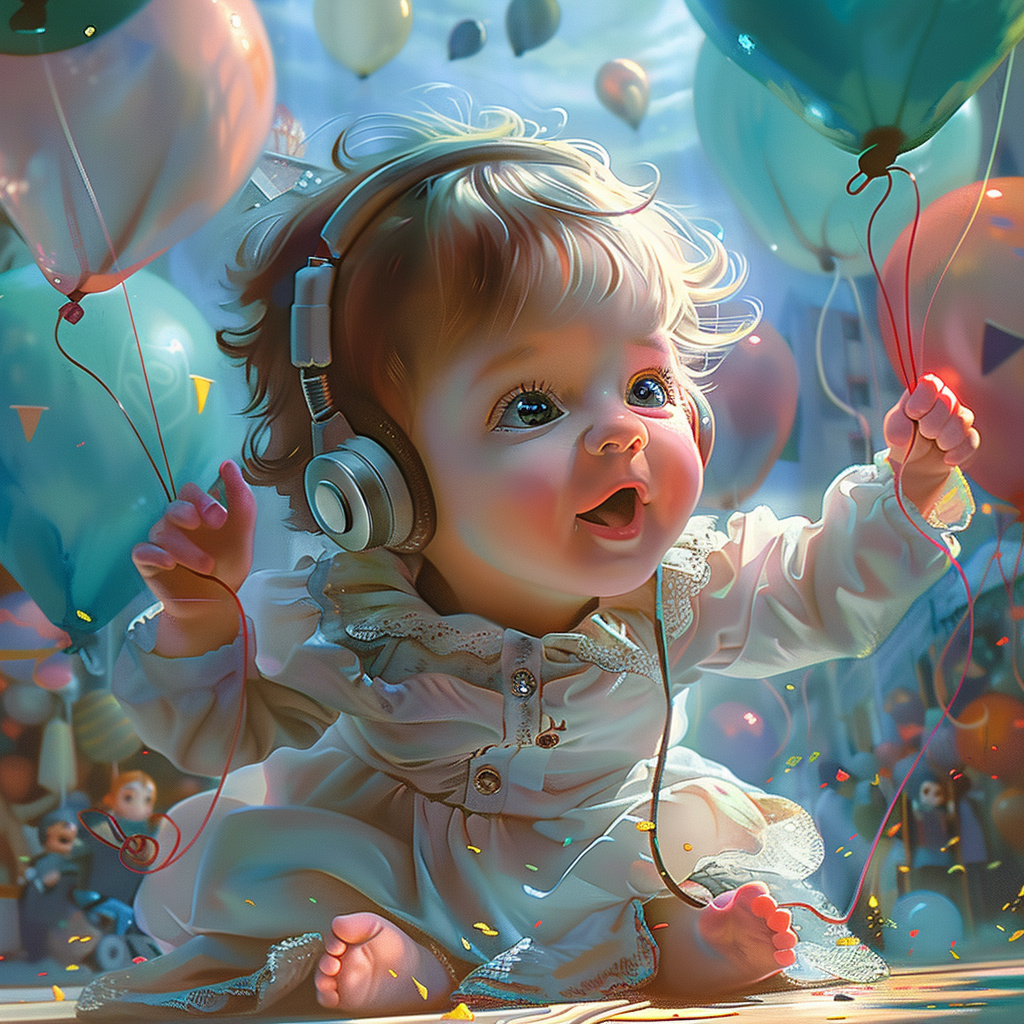}}\endgroup & \begingroup\setlength{\geditmissedcelltall}{1.000000\geditmissedcell}\parbox[c][\geditmissedcelltall][c]{\geditmissedcell}{\includegraphics[width=\geditmissedcell,height=\geditmissedcelltall,trim={0.000000bp 0.000000bp 0.000000bp 0.000000bp},clip]{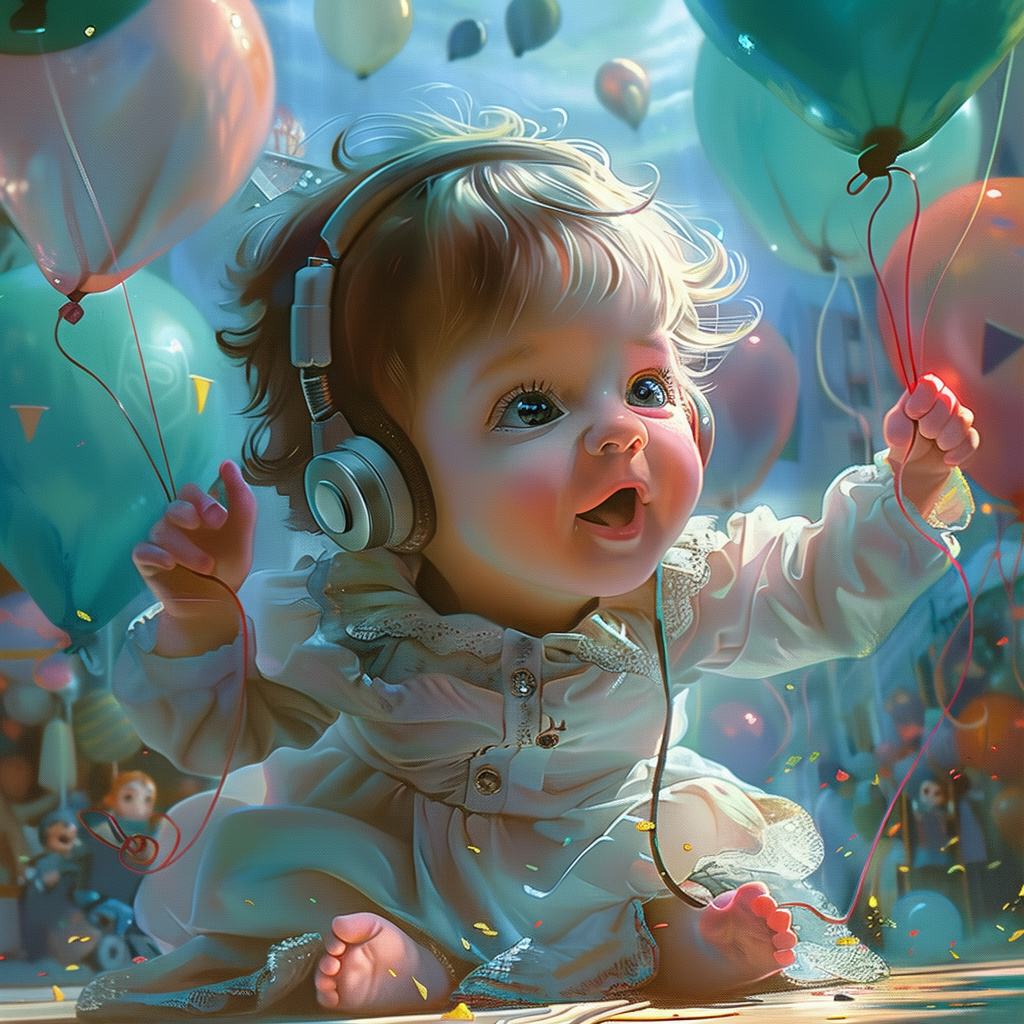}}\endgroup & \begingroup\setlength{\geditmissedcelltall}{1.000000\geditmissedcell}\geditmissedtxtcaptall{\fontsize{5.2}{5.9}\selectfont A hyperrealistic digital painting of an adorable baby girl with wide, curious eyes, rosy cheeks, and wispy blonde hair, wearing large retro headphones. She sits\,\ldots{}}\endgroup & \begingroup\setlength{\geditmissedcelltall}{1.000000\geditmissedcell}\parbox[c][\geditmissedcelltall][c]{\geditmissedcell}{\includegraphics[width=\geditmissedcell,height=\geditmissedcelltall,trim={0.000000bp 0.000000bp 0.000000bp 0.000000bp},clip]{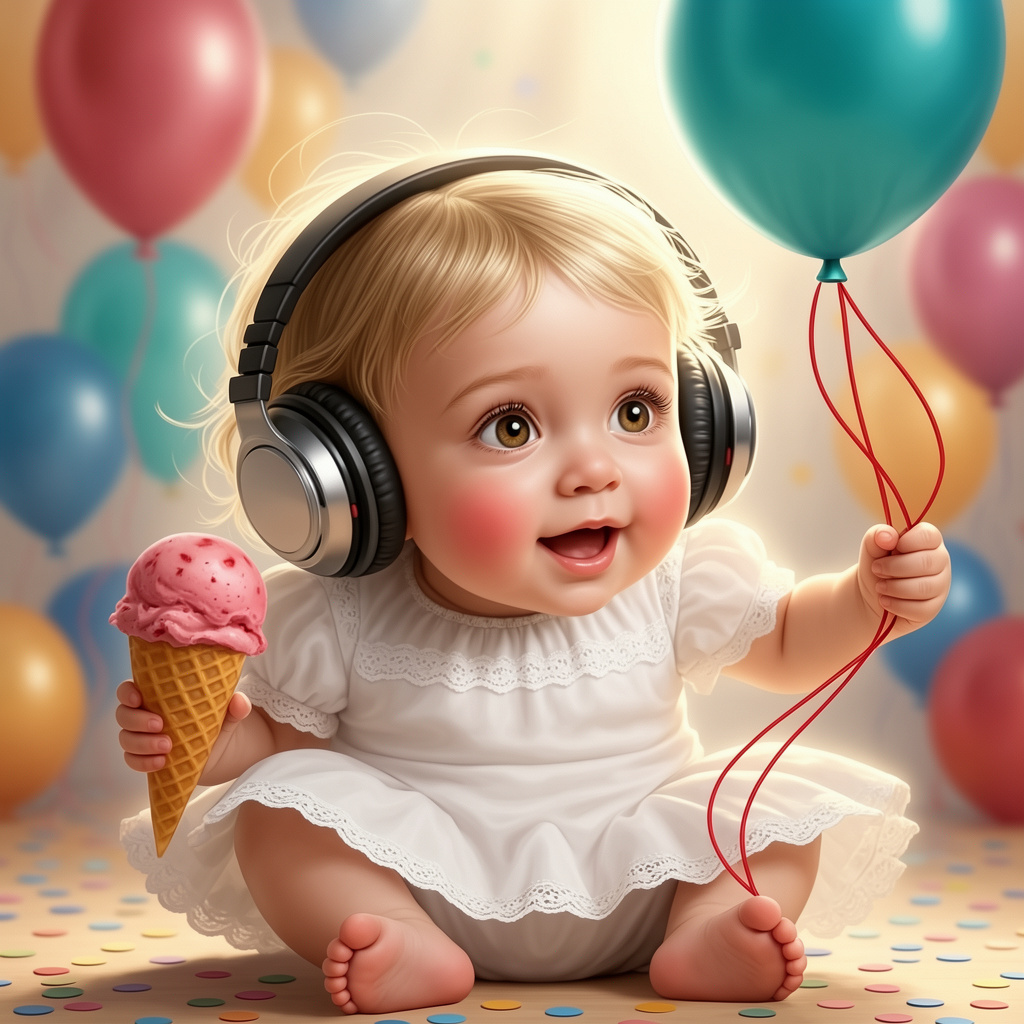}}\endgroup & \begingroup\setlength{\geditmissedcelltall}{1.000000\geditmissedcell}\parbox[c][\geditmissedcelltall][c]{\geditmissedcell}{\includegraphics[width=\geditmissedcell,height=\geditmissedcelltall,trim={0.000000bp 0.000000bp 0.000000bp 0.000000bp},clip]{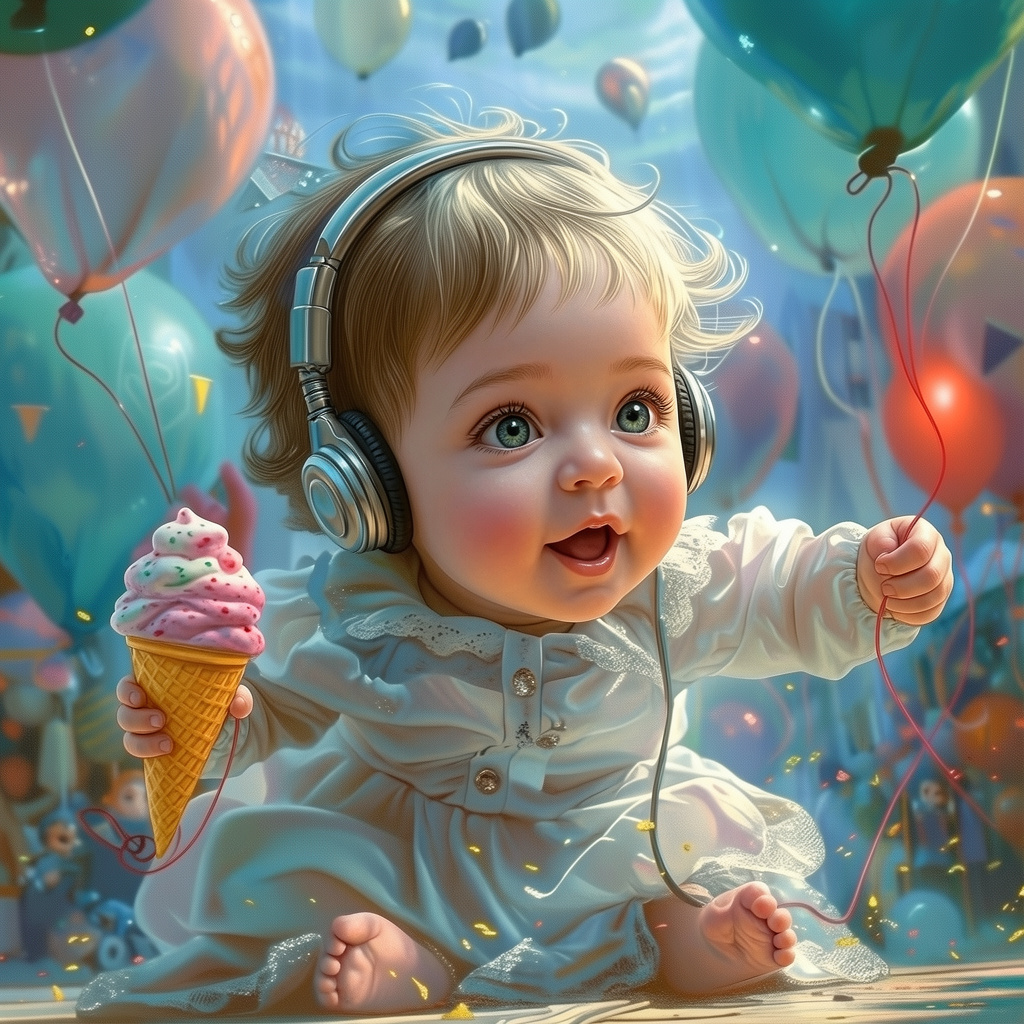}}\endgroup \\
\begingroup\setlength{\geditmissedcelltall}{0.705882\geditmissedcell}\geditmissedtxtinstrtall{\fontsize{6}{7}\selectfont Make her look younger}\endgroup & \begingroup\setlength{\geditmissedcelltall}{0.705882\geditmissedcell}\parbox[c][\geditmissedcelltall][c]{\geditmissedcell}{\includegraphics[width=\geditmissedcell,height=\geditmissedcelltall,trim={200.000000bp 0.000000bp 200.000000bp 0.000000bp},clip]{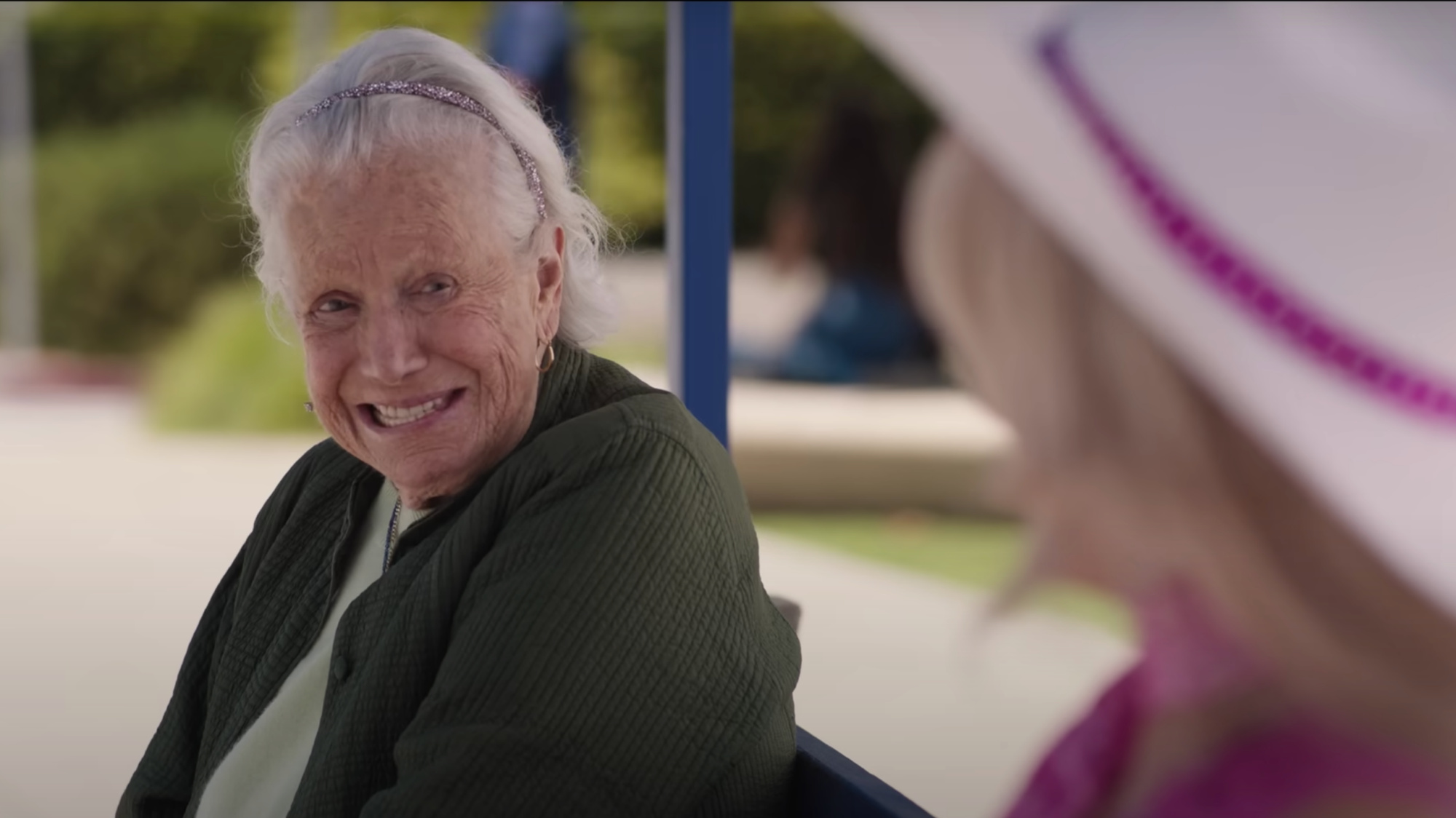}}\endgroup & \begingroup\setlength{\geditmissedcelltall}{0.705882\geditmissedcell}\parbox[c][\geditmissedcelltall][c]{\geditmissedcell}{\includegraphics[width=\geditmissedcell,height=\geditmissedcelltall,trim={136.000000bp 0.000000bp 136.000000bp 0.000000bp},clip]{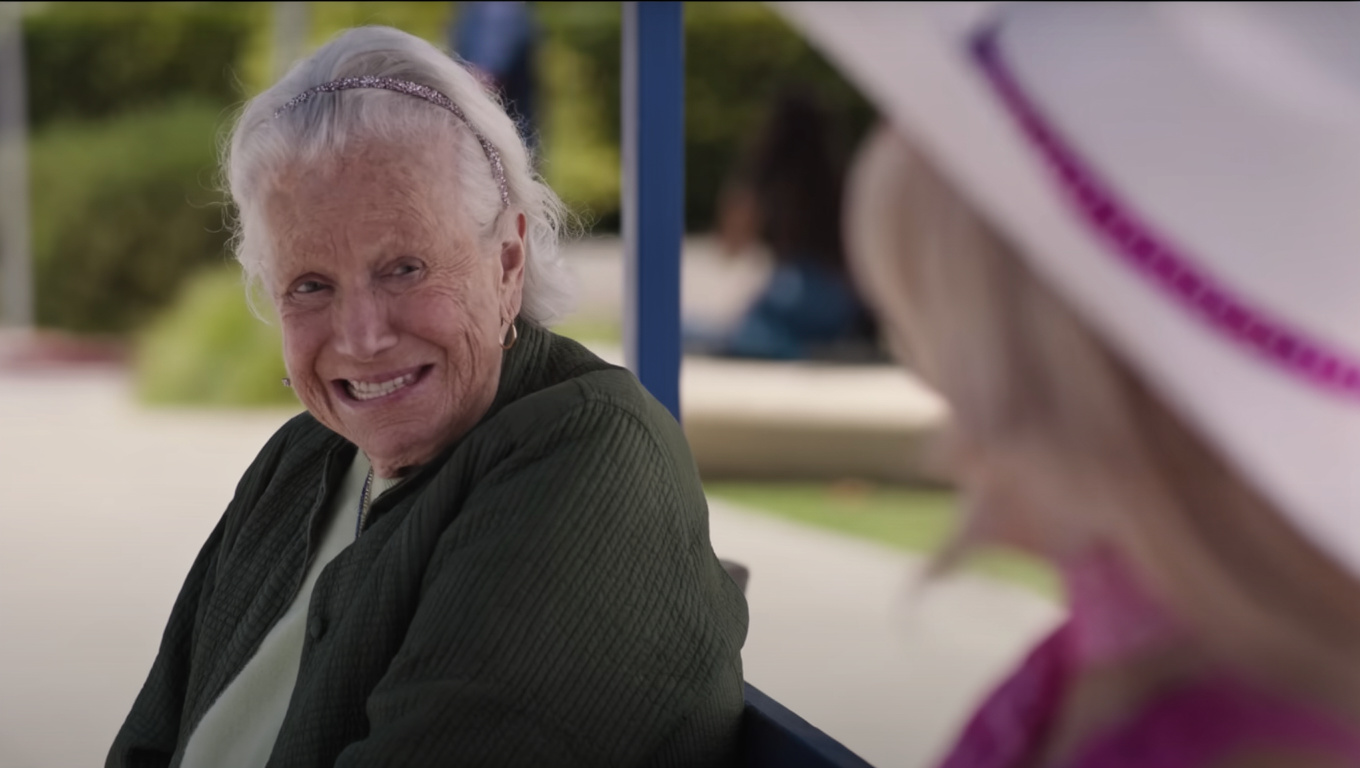}}\endgroup & \begingroup\setlength{\geditmissedcelltall}{0.705882\geditmissedcell}\geditmissedtxtcaptall{\fontsize{5.2}{5.9}\selectfont A cinematic medium shot of a friendly young woman with smooth skin and light brown hair, sitting on a blue bench outdoors and giving a wide, toothy smile towards\,\ldots{}}\endgroup & \begingroup\setlength{\geditmissedcelltall}{0.705882\geditmissedcell}\parbox[c][\geditmissedcelltall][c]{\geditmissedcell}{\includegraphics[width=\geditmissedcell,height=\geditmissedcelltall,trim={136.000000bp 0.000000bp 136.000000bp 0.000000bp},clip]{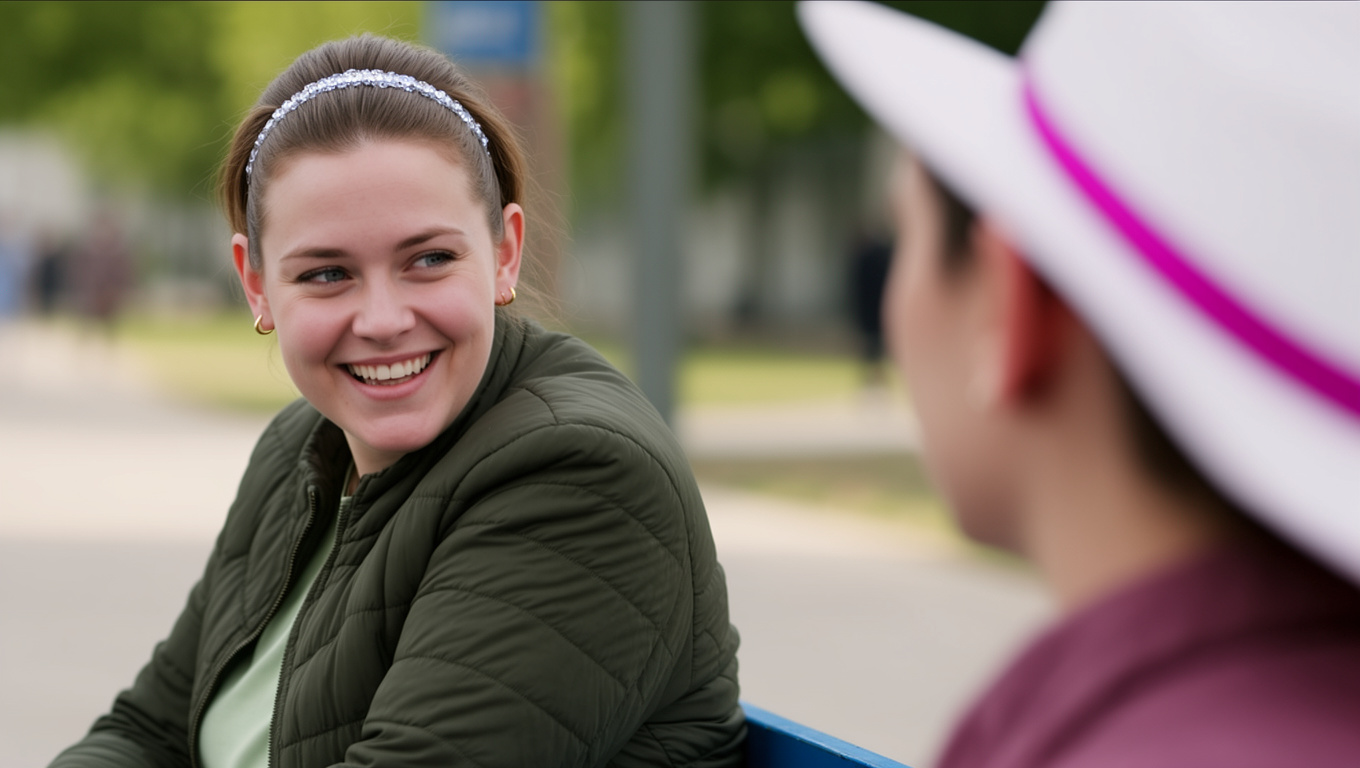}}\endgroup & \begingroup\setlength{\geditmissedcelltall}{0.705882\geditmissedcell}\parbox[c][\geditmissedcelltall][c]{\geditmissedcell}{\includegraphics[width=\geditmissedcell,height=\geditmissedcelltall,trim={136.000000bp 0.000000bp 136.000000bp 0.000000bp},clip]{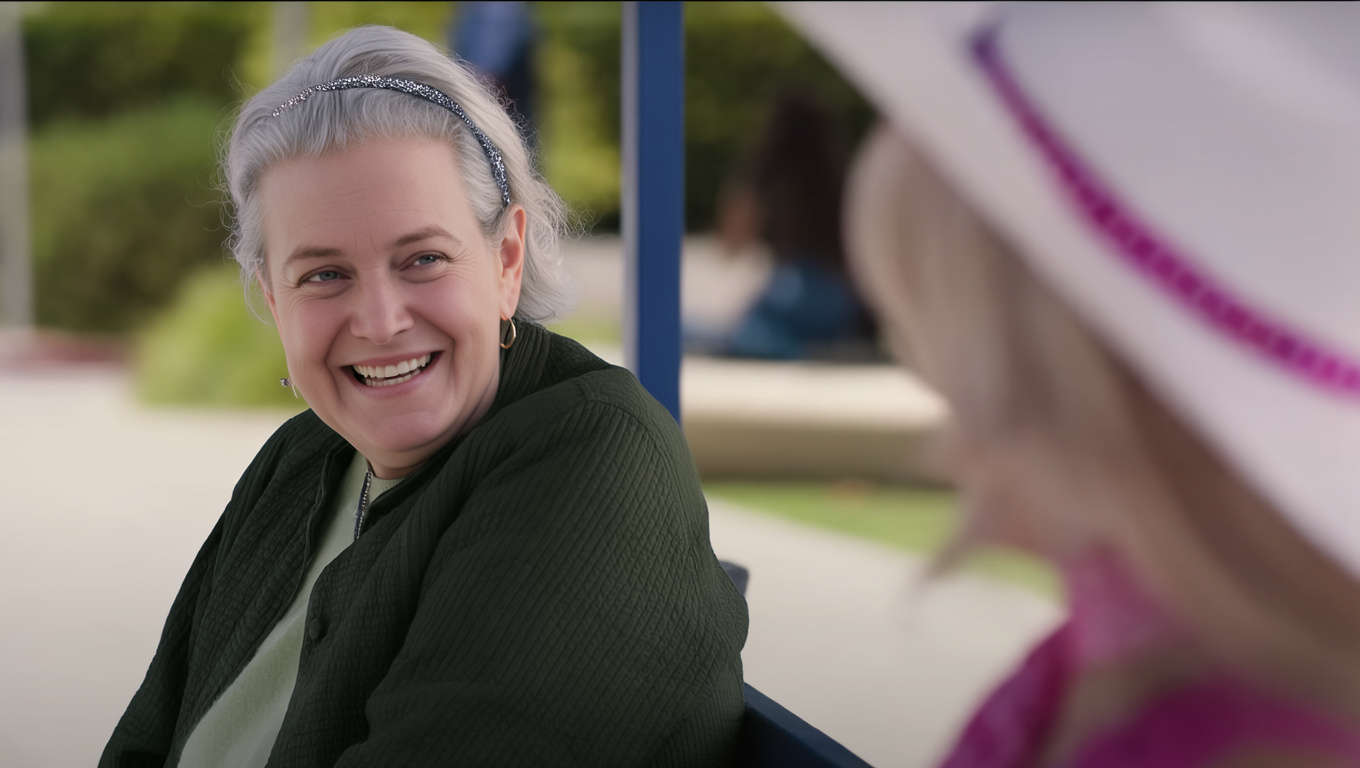}}\endgroup \\
\begingroup\setlength{\geditmissedcelltall}{0.767123\geditmissedcell}\geditmissedtxtinstrtall{\fontsize{6}{7}\selectfont Remove the clutter from the photo.}\endgroup & \begingroup\setlength{\geditmissedcelltall}{0.767123\geditmissedcell}\parbox[c][\geditmissedcelltall][c]{\geditmissedcell}{\includegraphics[width=\geditmissedcell,height=\geditmissedcelltall,trim={0.000000bp 0.000000bp 0.000000bp 0.000000bp},clip]{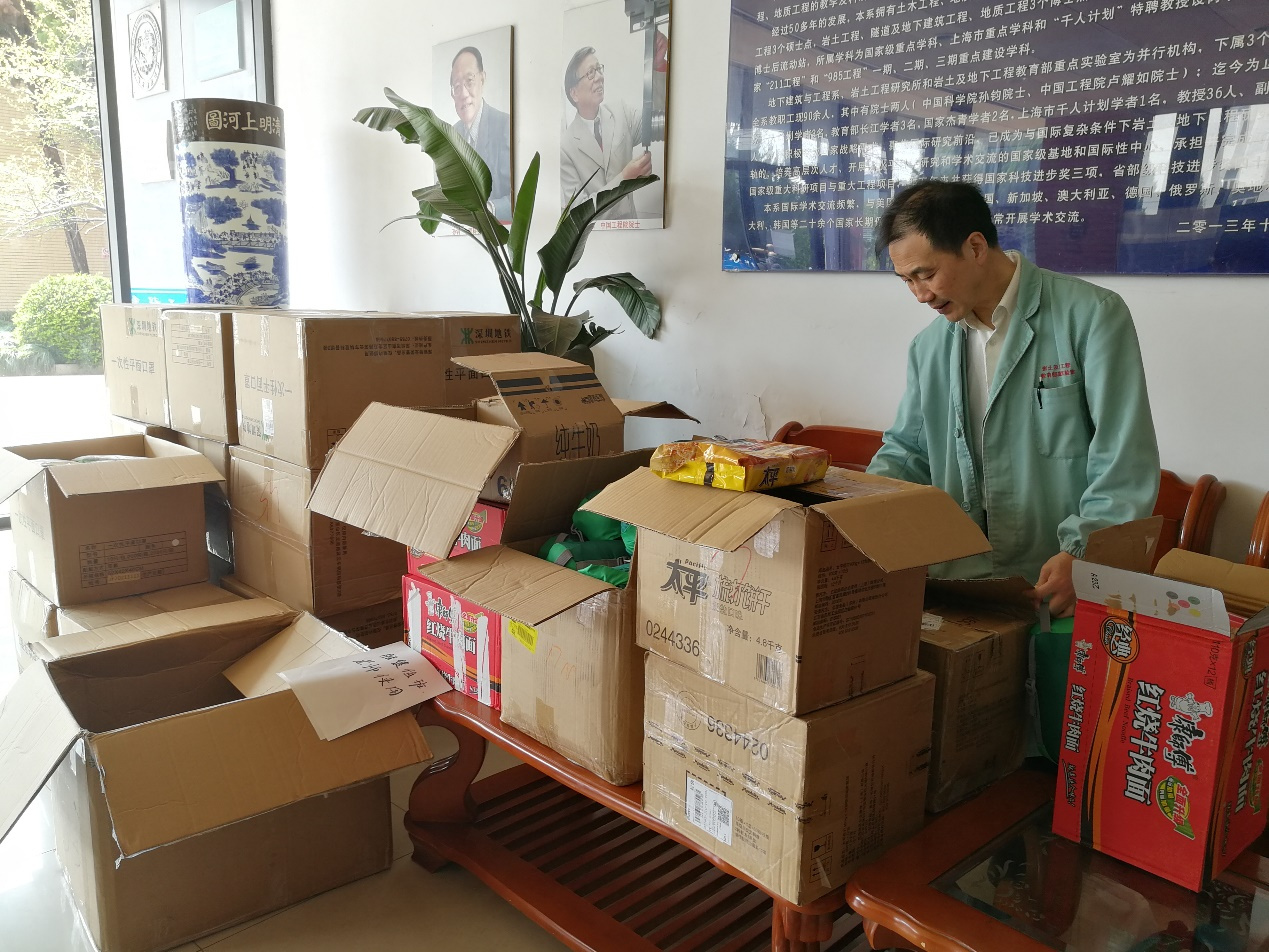}}\endgroup & \begingroup\setlength{\geditmissedcelltall}{0.767123\geditmissedcell}\parbox[c][\geditmissedcelltall][c]{\geditmissedcell}{\includegraphics[width=\geditmissedcell,height=\geditmissedcelltall,trim={0.000000bp 0.000000bp 0.000000bp 0.000000bp},clip]{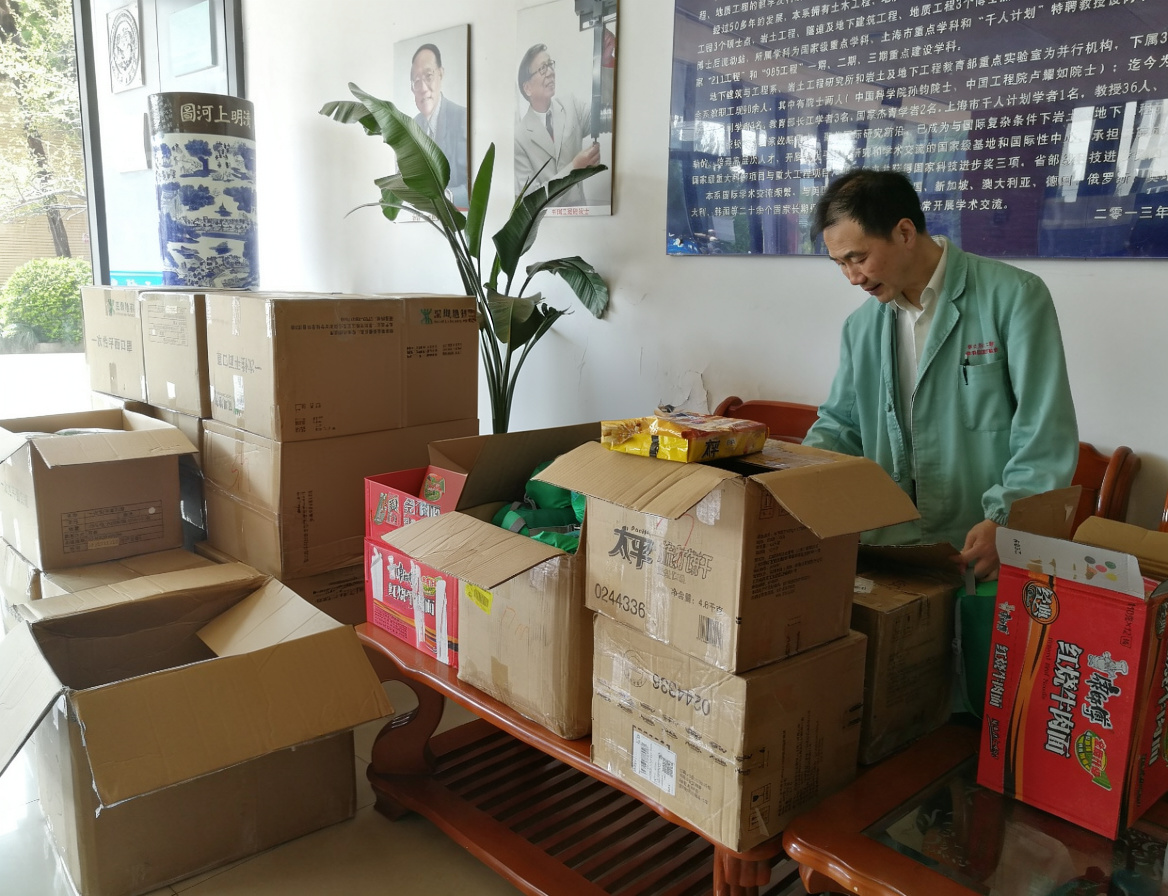}}\endgroup & \begingroup\setlength{\geditmissedcelltall}{0.767123\geditmissedcell}\geditmissedtxtcaptall{\fontsize{5.2}{5.9}\selectfont A realistic photograph of a brightly lit office interior, captured from an eye-level perspective. A middle-aged Asian man with short black hair, wearing a light\,\ldots{}}\endgroup & \begingroup\setlength{\geditmissedcelltall}{0.767123\geditmissedcell}\parbox[c][\geditmissedcelltall][c]{\geditmissedcell}{\includegraphics[width=\geditmissedcell,height=\geditmissedcelltall,trim={0.000000bp 0.000000bp 0.000000bp 0.000000bp},clip]{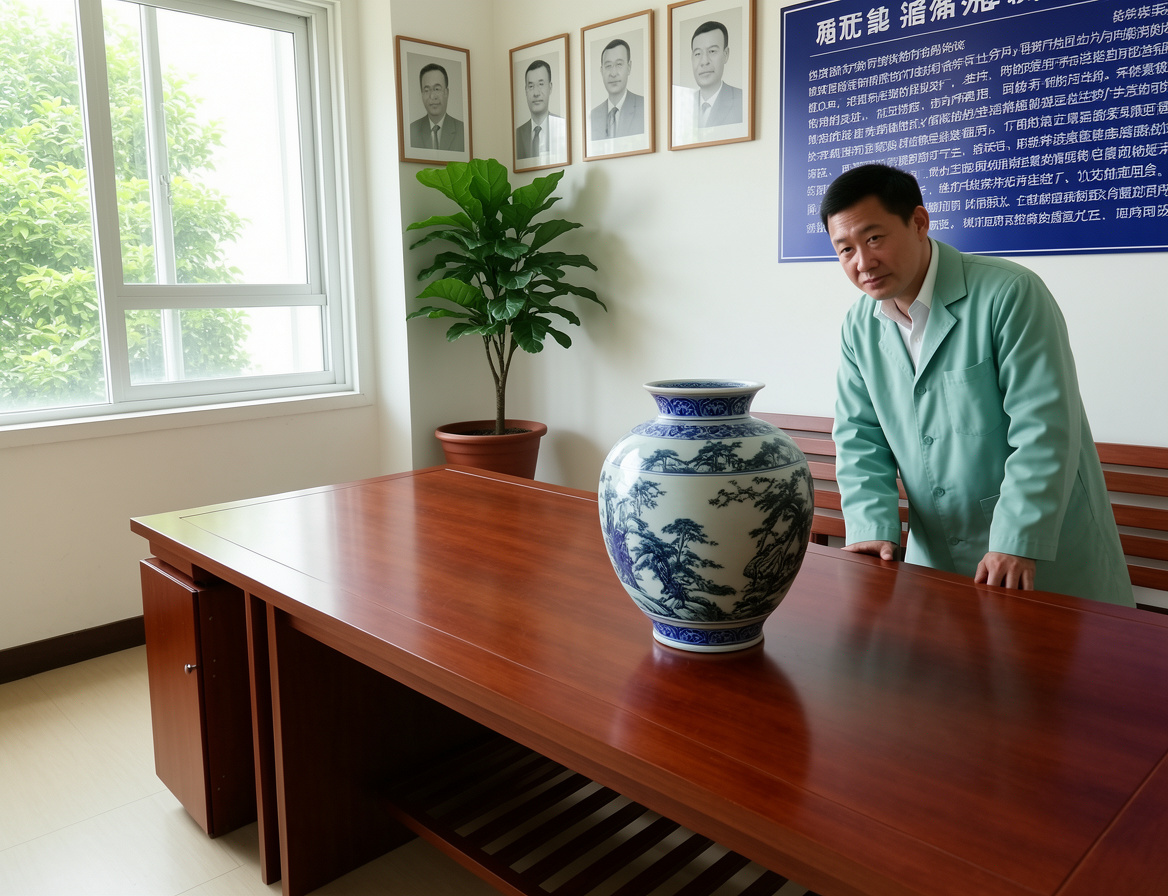}}\endgroup & \begingroup\setlength{\geditmissedcelltall}{0.767123\geditmissedcell}\parbox[c][\geditmissedcelltall][c]{\geditmissedcell}{\includegraphics[width=\geditmissedcell,height=\geditmissedcelltall,trim={0.000000bp 0.000000bp 0.000000bp 0.000000bp},clip]{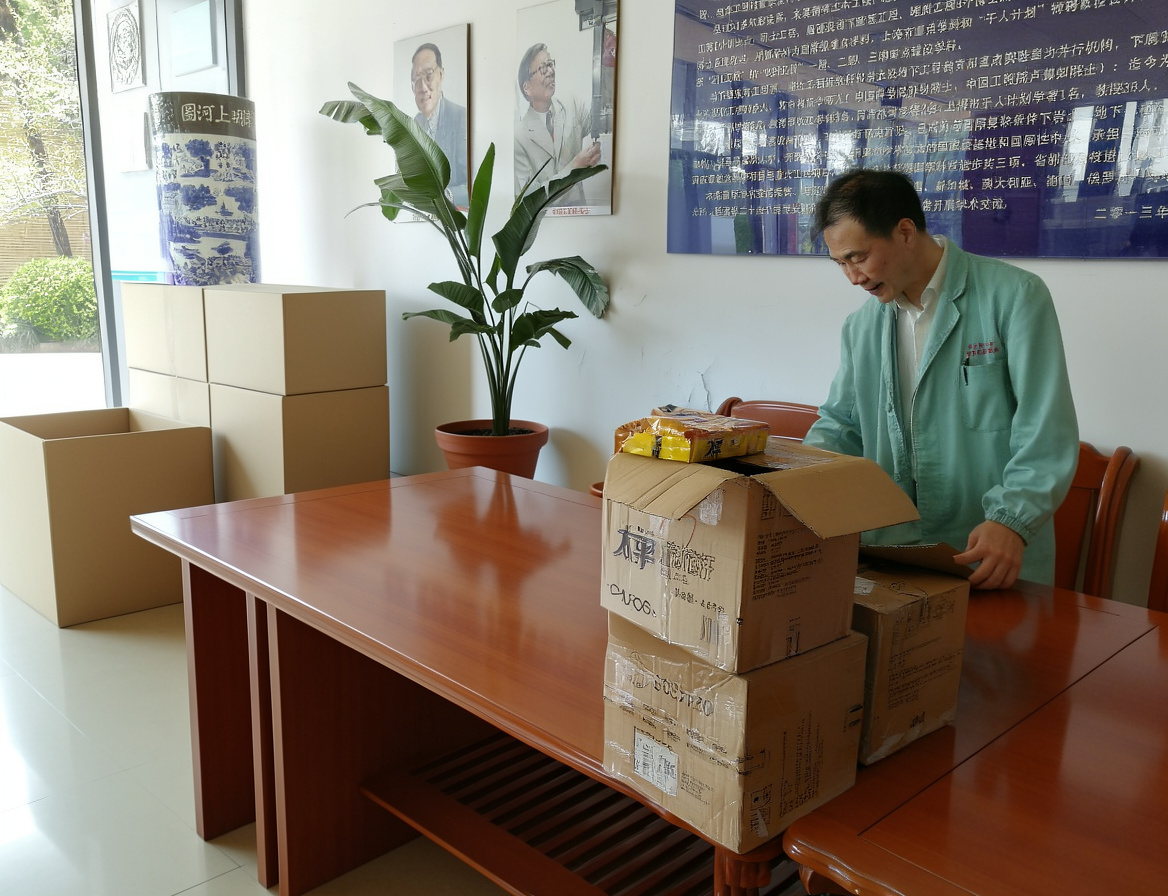}}\endgroup \\
\begingroup\setlength{\geditmissedcelltall}{1.471698\geditmissedcell}\geditmissedtxtinstrtall{\fontsize{6}{7}\selectfont Give her a different outfit.}\endgroup & \begingroup\setlength{\geditmissedcelltall}{1.471698\geditmissedcell}\parbox[c][\geditmissedcelltall][c]{\geditmissedcell}{\includegraphics[width=\geditmissedcell,height=\geditmissedcelltall,trim={0.000000bp 0.000000bp 0.000000bp 0.000000bp},clip]{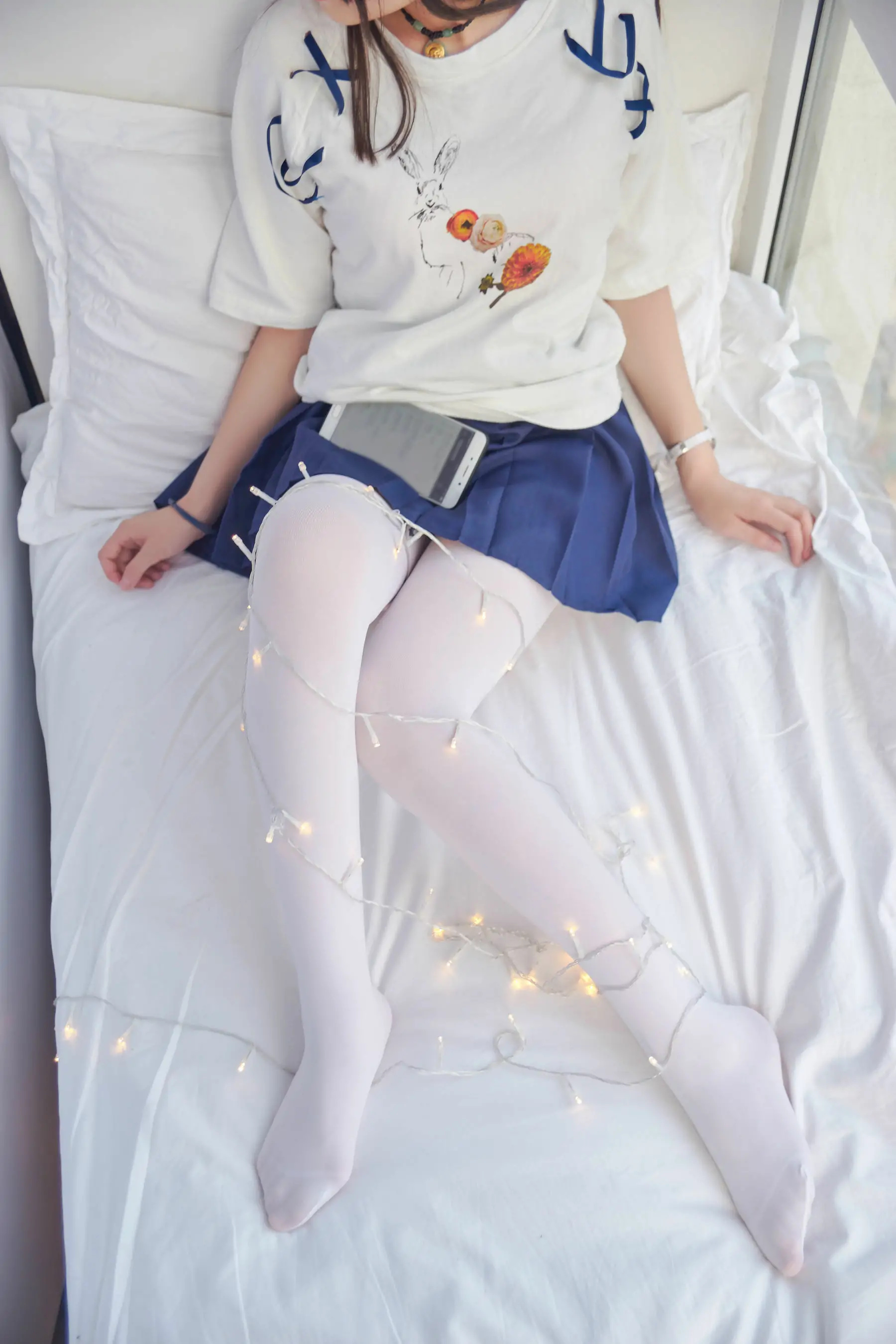}}\endgroup & \begingroup\setlength{\geditmissedcelltall}{1.471698\geditmissedcell}\parbox[c][\geditmissedcelltall][c]{\geditmissedcell}{\includegraphics[width=\geditmissedcell,height=\geditmissedcelltall,trim={0.000000bp 0.000000bp 0.000000bp 0.000000bp},clip]{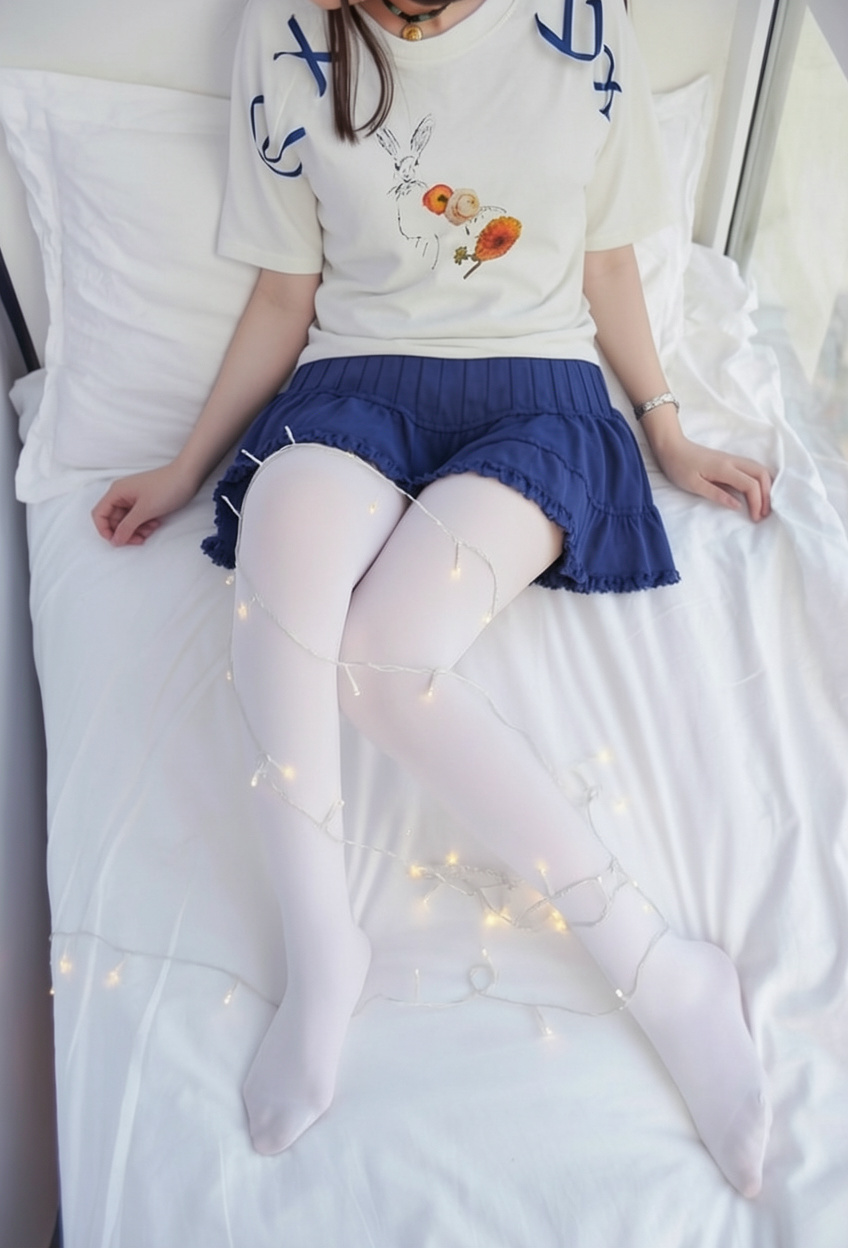}}\endgroup & \begingroup\setlength{\geditmissedcelltall}{1.471698\geditmissedcell}\geditmissedtxtcaptall{\fontsize{5.2}{5.9}\selectfont A high-angle photograph of a young woman with dark hair sitting on a bed covered in rumpled white sheets. She is wearing a cozy, oversized pink chunky-knit sweater\,\ldots{}}\endgroup & \begingroup\setlength{\geditmissedcelltall}{1.471698\geditmissedcell}\parbox[c][\geditmissedcelltall][c]{\geditmissedcell}{\includegraphics[width=\geditmissedcell,height=\geditmissedcelltall,trim={0.000000bp 0.000000bp 0.000000bp 0.000000bp},clip]{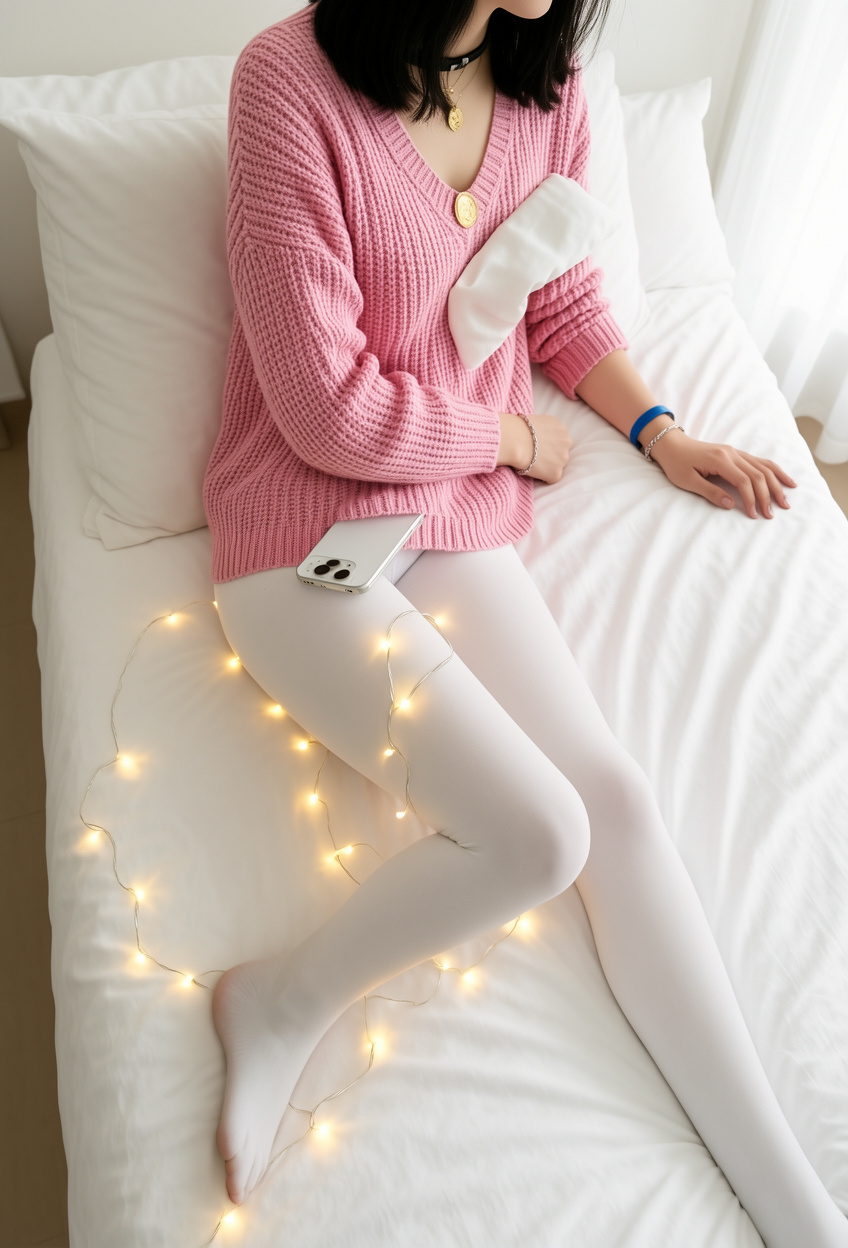}}\endgroup & \begingroup\setlength{\geditmissedcelltall}{1.471698\geditmissedcell}\parbox[c][\geditmissedcelltall][c]{\geditmissedcell}{\includegraphics[width=\geditmissedcell,height=\geditmissedcelltall,trim={0.000000bp 0.000000bp 0.000000bp 0.000000bp},clip]{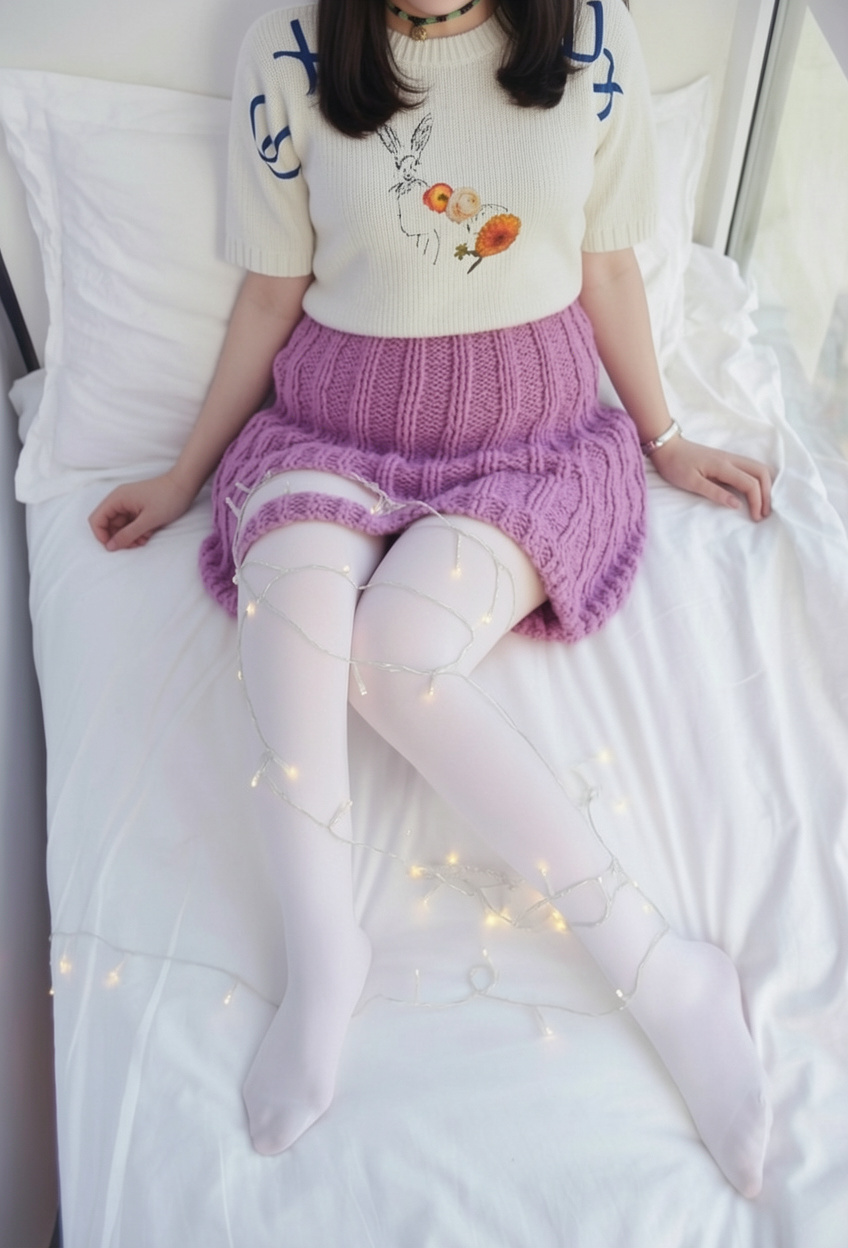}}\endgroup \\
\end{tabular}
\endgroup
\caption{\textbf{Missed edits (FLUX2-Klein~4B, GEdit).} Pure editing leaves the requested change largely unexecuted. A one-way switch uses T2I from step 10 onward; E$\to$T2I$\to$E returns to editing after $[10,16)$.}
\label{fig:appx-gedit-missed}
\end{figure}

\clearpage

\begin{figure}[!htb]
\centering

\begingroup
\sffamily
\setlength{\tabcolsep}{0pt}
\renewcommand{\arraystretch}{0}
\setlength{\fboxsep}{0pt}
\newlength{\geditincorrectinstr}
\newlength{\geditincorrectcell}
\newlength{\geditincorrectcelltall}
\setlength{\geditincorrectinstr}{0.110\linewidth}
\setlength{\geditincorrectcell}{0.178\linewidth}
\setlength{\geditincorrectcelltall}{1.000\geditincorrectcell}
\providecommand{\geditincorrectimg}[1]{%
  \parbox[c][\geditincorrectcell][c]{\geditincorrectcell}{\centering\includegraphics[width=\geditincorrectcell,height=\geditincorrectcell,keepaspectratio]{#1}}%
}
\providecommand{\geditincorrectimgtall}[1]{%
  \parbox[c][\geditincorrectcelltall][c]{\geditincorrectcell}{\centering\includegraphics[width=\geditincorrectcell]{#1}}%
}
\providecommand{\geditincorrectmissing}{%
  \parbox[c][\geditincorrectcell][c]{\geditincorrectcell}{\centering\textcolor{black!10}{\rule{\geditincorrectcell}{\geditincorrectcell}}}%
}
\providecommand{\geditincorrectmissingtall}{%
  \parbox[c][\geditincorrectcelltall][c]{\geditincorrectcell}{\centering\textcolor{black!10}{\rule{\geditincorrectcell}{\geditincorrectcelltall}}}%
}
\providecommand{\geditincorrecttextpad}{\vspace{2pt}\vspace{0pt}}
\providecommand{\geditincorrecttxtinstr}[1]{%
  \parbox[c][\geditincorrectcell][t]{\dimexpr\geditincorrectinstr-2pt\relax}{\geditincorrecttextpad\raggedright #1\par}%
}
\providecommand{\geditincorrecttxtinstrtall}[1]{%
  \parbox[c][\geditincorrectcelltall][t]{\dimexpr\geditincorrectinstr-2pt\relax}{\geditincorrecttextpad\raggedright #1\par}%
}
\providecommand{\geditincorrecttxtcap}[1]{%
  \parbox[c][\geditincorrectcell][t]{\dimexpr\geditincorrectcell-2pt\relax}{\geditincorrecttextpad\raggedright #1\par}%
}
\providecommand{\geditincorrecttxtcaptall}[1]{%
  \parbox[c][\geditincorrectcelltall][t]{\dimexpr\geditincorrectcell-2pt\relax}{\geditincorrecttextpad\raggedright #1\par}%
}
\providecommand{\geditincorrecthdrinstr}[1]{%
  \parbox[t]{\dimexpr\geditincorrectinstr-2pt\relax}{\centering\geditincorrecttextpad #1\par}%
}
\providecommand{\geditincorrecthdr}[1]{%
  \parbox[t]{\dimexpr\geditincorrectcell-2pt\relax}{\centering\geditincorrecttextpad #1\par}%
}

\begin{tabular}{@{}>{\raggedright\arraybackslash}m{\geditincorrectinstr}*{5}{>{\centering\arraybackslash}m{\geditincorrectcell}}@{}}
\geditincorrecthdrinstr{\fontsize{7}{8}\selectfont\bfseries instruction} & \geditincorrecthdr{\fontsize{7}{8}\selectfont\bfseries source} & \geditincorrecthdr{\fontsize{7}{8}\selectfont\bfseries pure-E} & \geditincorrecthdr{\fontsize{7}{8}\selectfont\bfseries T2I caption} & \geditincorrecthdr{\fontsize{7}{8}\selectfont\bfseries E$\to$T2I $k{=}10$} & \geditincorrecthdr{\fontsize{7}{8}\selectfont\bfseries E$\to$T2I$\to$E\\$[10,16)$} \\
\noalign{\vskip 1pt}
\begingroup\setlength{\geditincorrectcelltall}{1.714286\geditincorrectcell}\geditincorrecttxtinstrtall{\fontsize{5.4}{6.2}\selectfont Replace the sword in the image with a diamond sword.}\endgroup & \begingroup\setlength{\geditincorrectcelltall}{1.714286\geditincorrectcell}\parbox[c][\geditincorrectcelltall][c]{\geditincorrectcell}{\includegraphics[width=\geditincorrectcell,height=\geditincorrectcelltall,trim={0.000000bp 0.000000bp 0.000000bp 0.000000bp},clip]{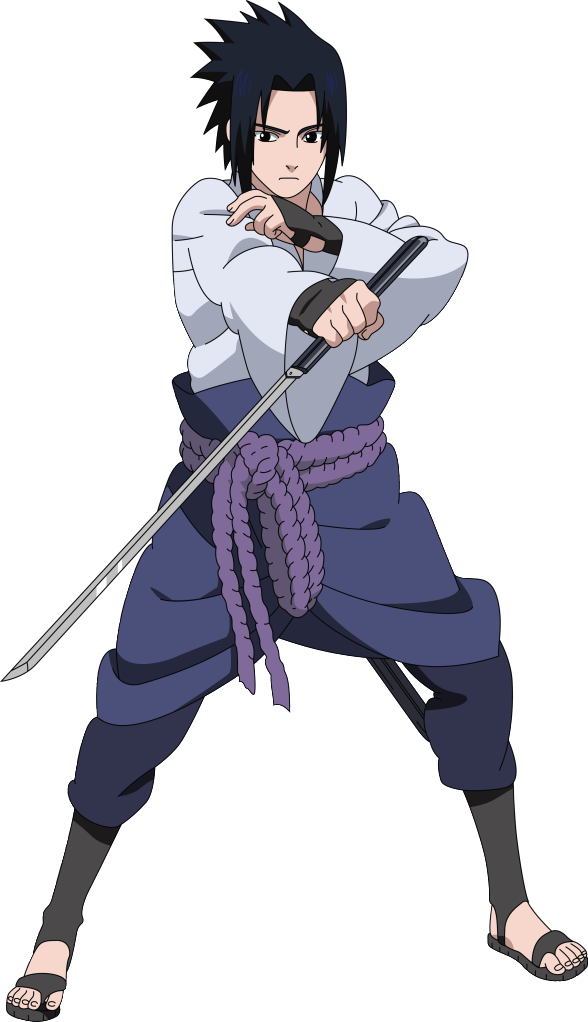}}\endgroup & \begingroup\setlength{\geditincorrectcelltall}{1.714286\geditincorrectcell}\parbox[c][\geditincorrectcelltall][c]{\geditincorrectcell}{\includegraphics[width=\geditincorrectcell,height=\geditincorrectcelltall,trim={0.000000bp 0.000000bp 0.000000bp 0.000000bp},clip]{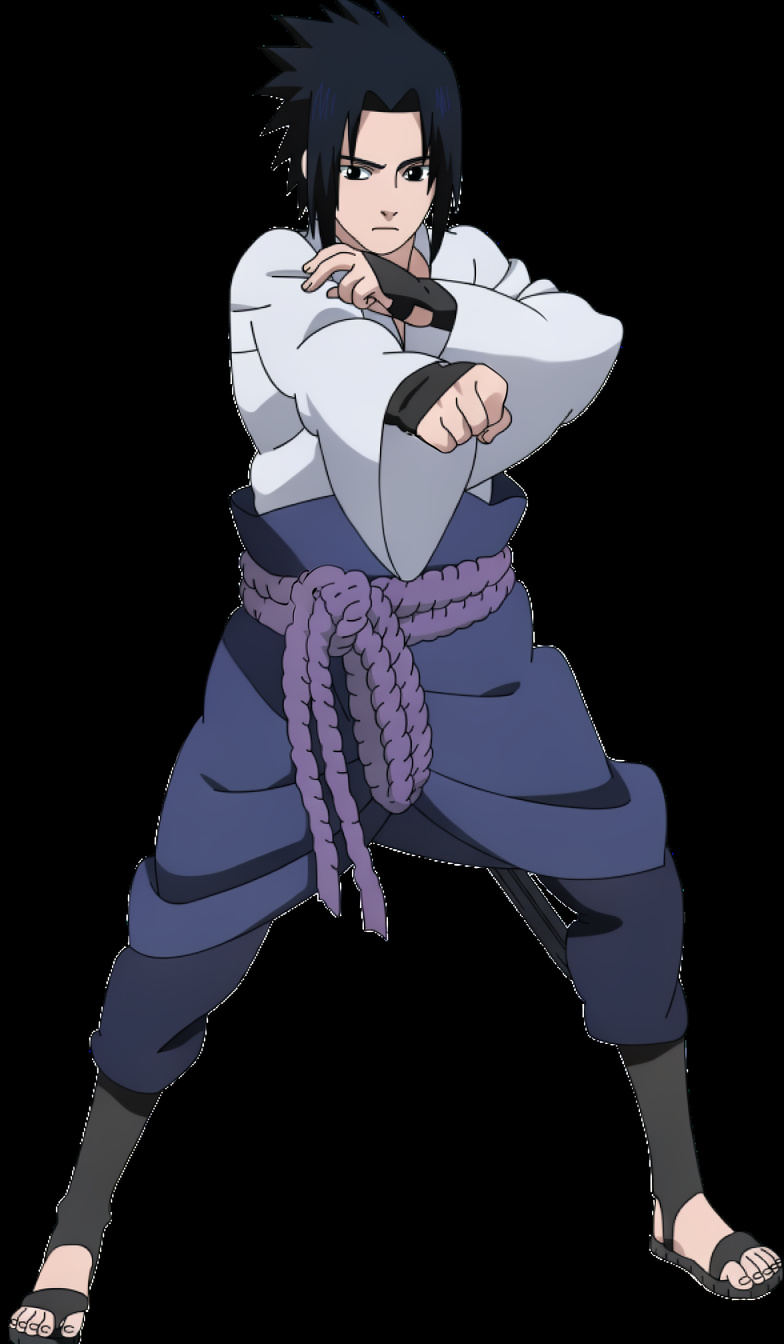}}\endgroup & \begingroup\setlength{\geditincorrectcelltall}{1.714286\geditincorrectcell}\geditincorrecttxtcaptall{\fontsize{5.2}{5.9}\selectfont A full-body, 2D anime illustration of Sasuke Uchiha in a wide, low fighting stance against a plain white background. He has spiky black hair, pale skin, and a\,\ldots{}}\endgroup & \begingroup\setlength{\geditincorrectcelltall}{1.714286\geditincorrectcell}\parbox[c][\geditincorrectcelltall][c]{\geditincorrectcell}{\includegraphics[width=\geditincorrectcell,height=\geditincorrectcelltall,trim={0.000000bp 0.000000bp 0.000000bp 0.000000bp},clip]{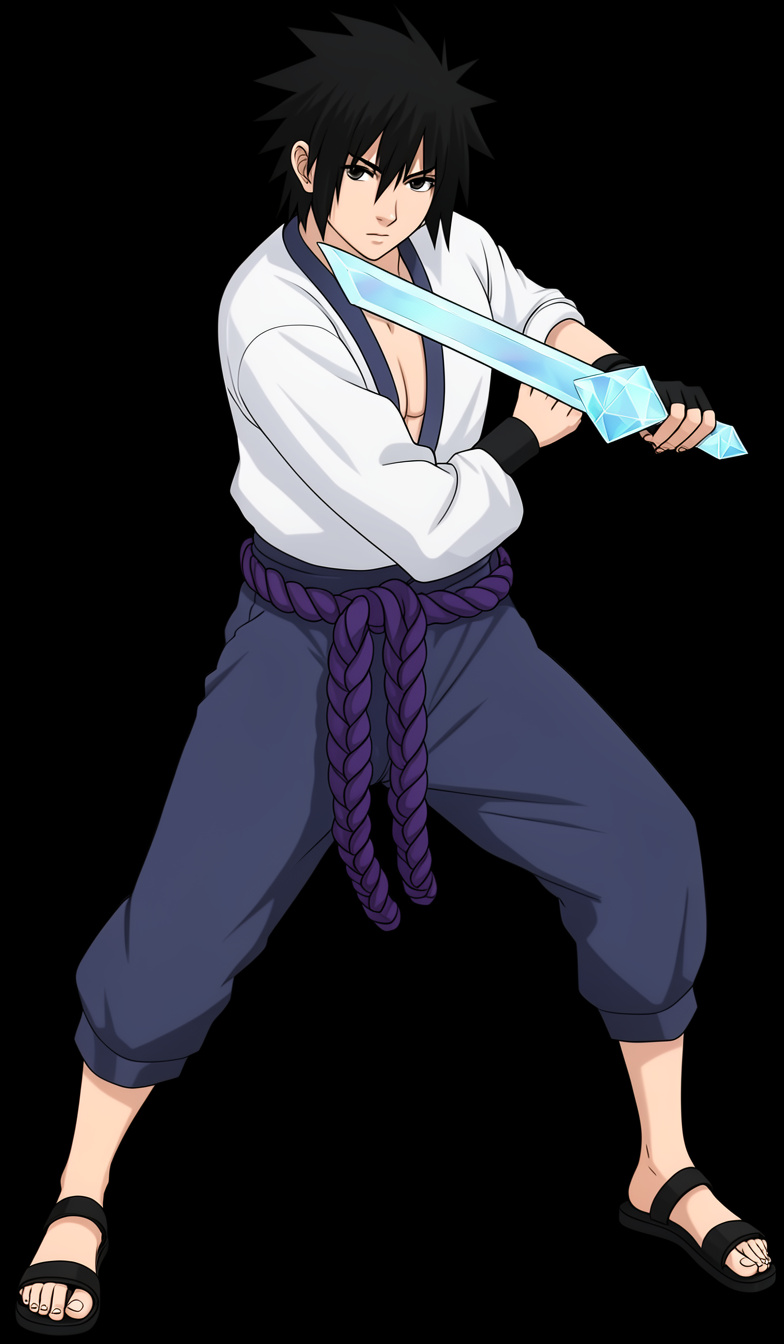}}\endgroup & \begingroup\setlength{\geditincorrectcelltall}{1.714286\geditincorrectcell}\parbox[c][\geditincorrectcelltall][c]{\geditincorrectcell}{\includegraphics[width=\geditincorrectcell,height=\geditincorrectcelltall,trim={0.000000bp 0.000000bp 0.000000bp 0.000000bp},clip]{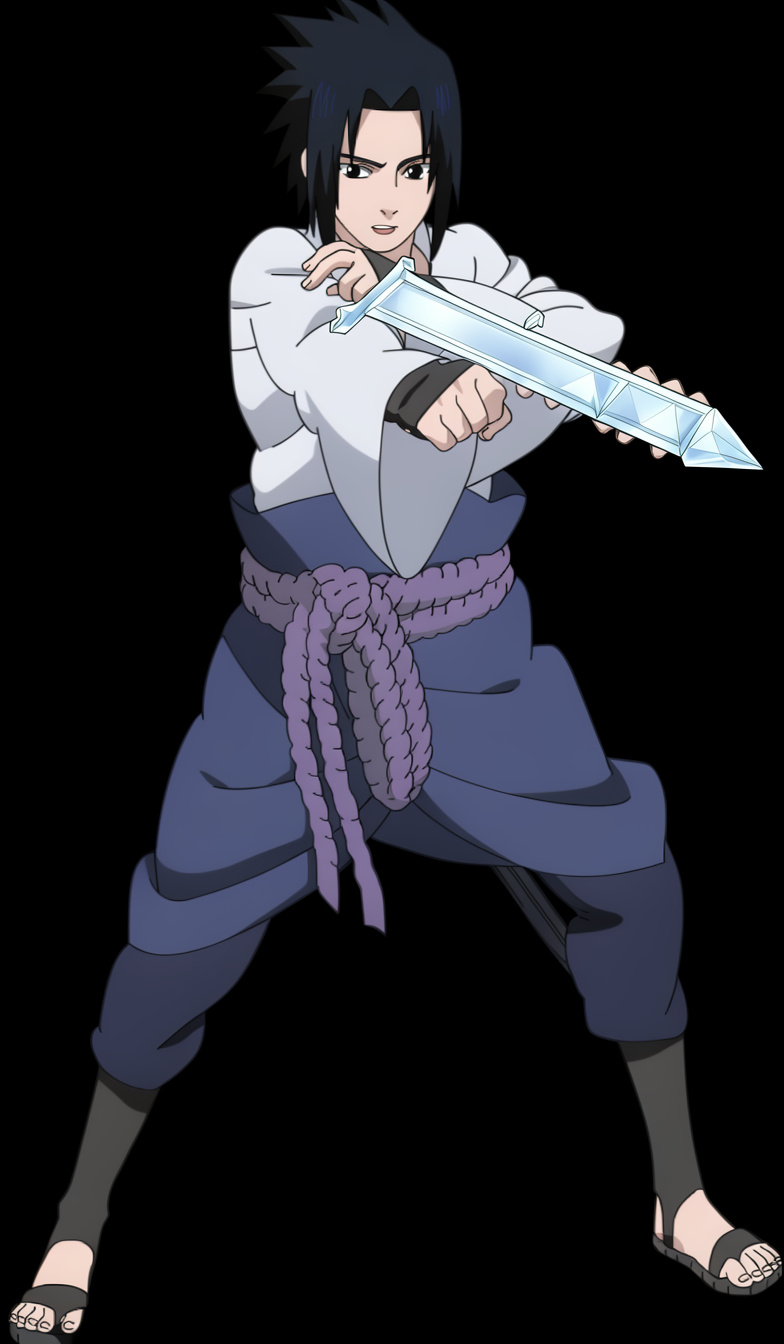}}\endgroup \\
\begingroup\setlength{\geditincorrectcelltall}{1.303571\geditincorrectcell}\geditincorrecttxtinstrtall{\fontsize{4.8}{5.6}\selectfont Can you Photoshop a girlfriend for me? Sitting alone is boring, and I've already left space for her.}\endgroup & \begingroup\setlength{\geditincorrectcelltall}{1.303571\geditincorrectcell}\parbox[c][\geditincorrectcelltall][c]{\geditincorrectcell}{\includegraphics[width=\geditincorrectcell,height=\geditincorrectcelltall,trim={0.000000bp 0.000000bp 0.000000bp 0.000000bp},clip]{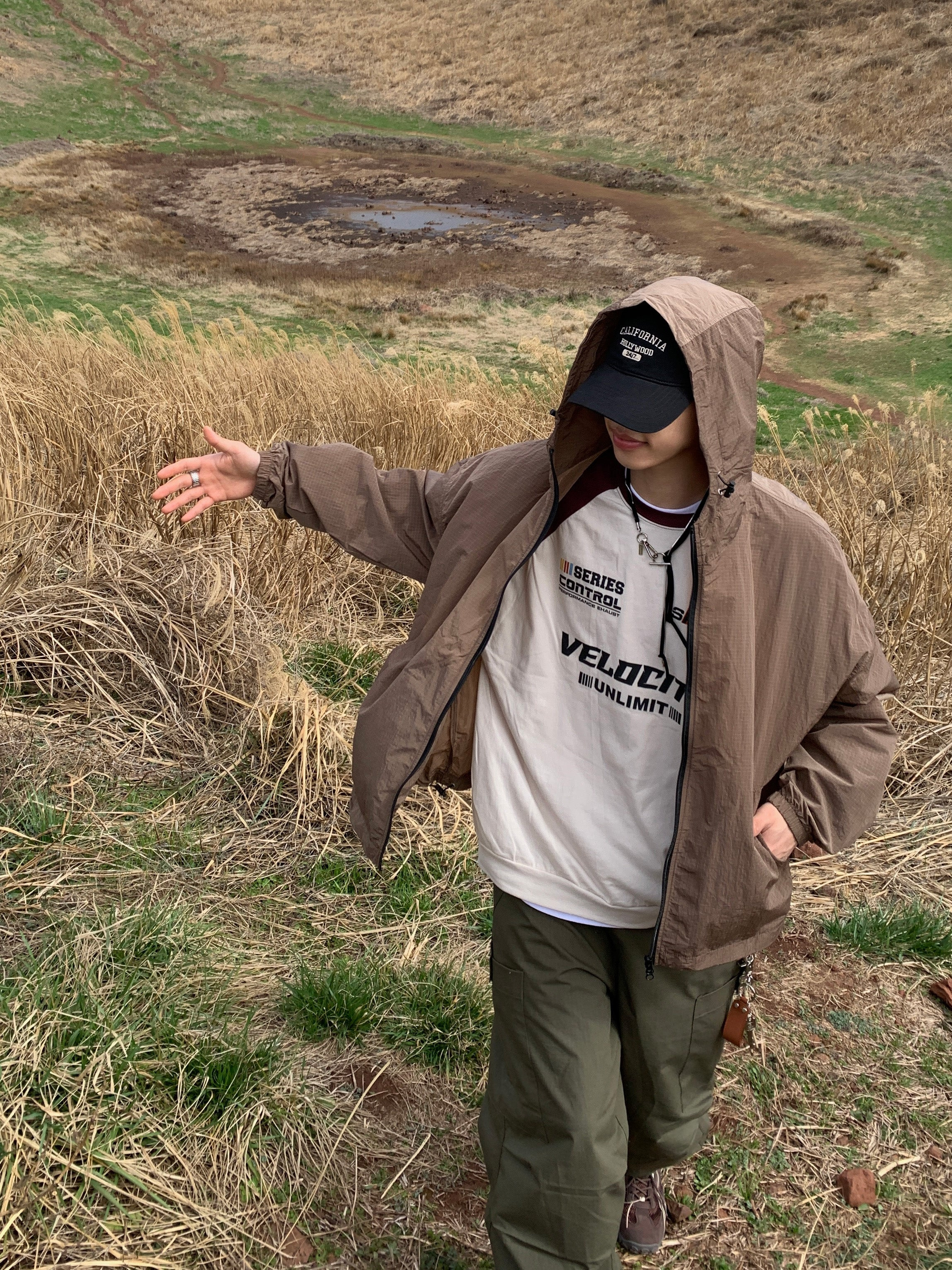}}\endgroup & \begingroup\setlength{\geditincorrectcelltall}{1.303571\geditincorrectcell}\parbox[c][\geditincorrectcelltall][c]{\geditincorrectcell}{\includegraphics[width=\geditincorrectcell,height=\geditincorrectcelltall,trim={0.000000bp 0.000000bp 0.000000bp 0.000000bp},clip]{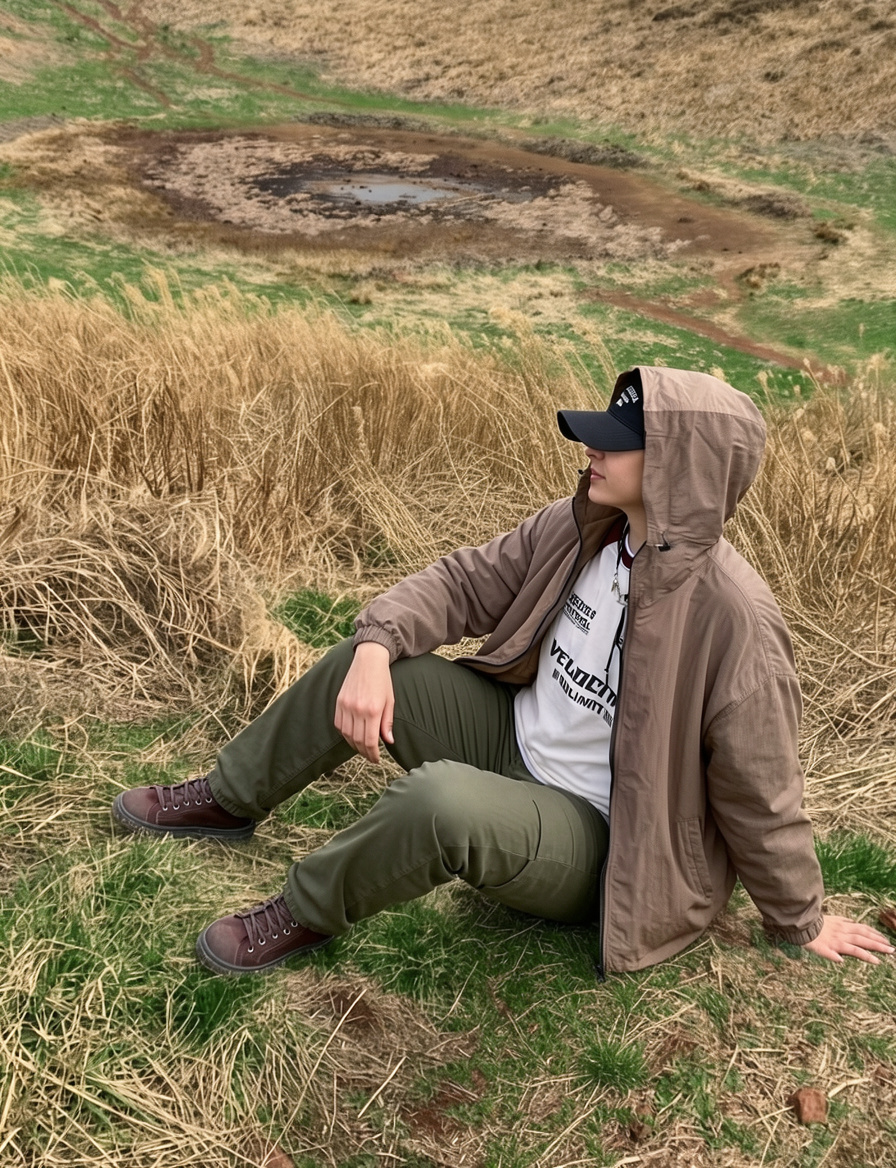}}\endgroup & \begingroup\setlength{\geditincorrectcelltall}{1.303571\geditincorrectcell}\geditincorrecttxtcaptall{\fontsize{5.2}{5.9}\selectfont A high-angle, medium photograph of a young couple standing close together on a grassy, hilly field. On the right, a young man wears an unzipped brown hooded\,\ldots{}}\endgroup & \begingroup\setlength{\geditincorrectcelltall}{1.303571\geditincorrectcell}\parbox[c][\geditincorrectcelltall][c]{\geditincorrectcell}{\includegraphics[width=\geditincorrectcell,height=\geditincorrectcelltall,trim={0.000000bp 0.000000bp 0.000000bp 0.000000bp},clip]{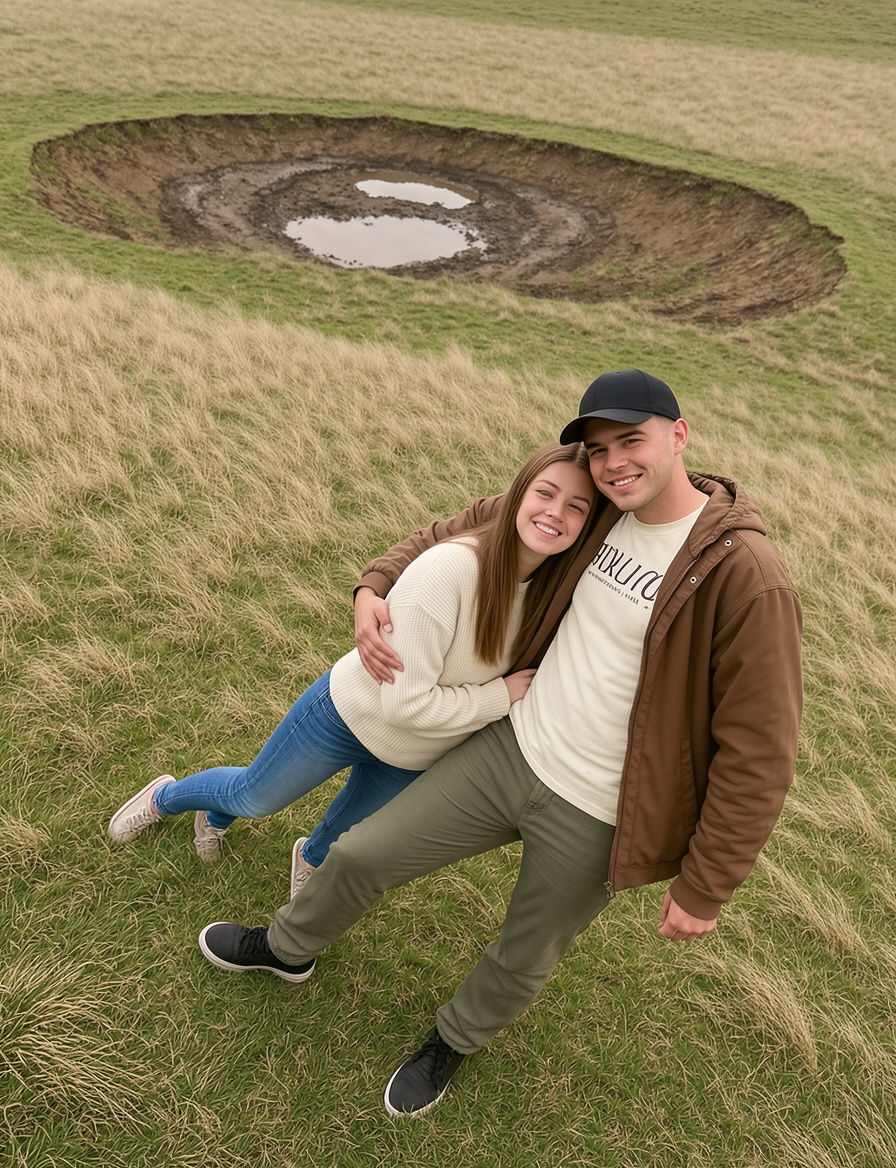}}\endgroup & \begingroup\setlength{\geditincorrectcelltall}{1.303571\geditincorrectcell}\parbox[c][\geditincorrectcelltall][c]{\geditincorrectcell}{\includegraphics[width=\geditincorrectcell,height=\geditincorrectcelltall,trim={0.000000bp 0.000000bp 0.000000bp 0.000000bp},clip]{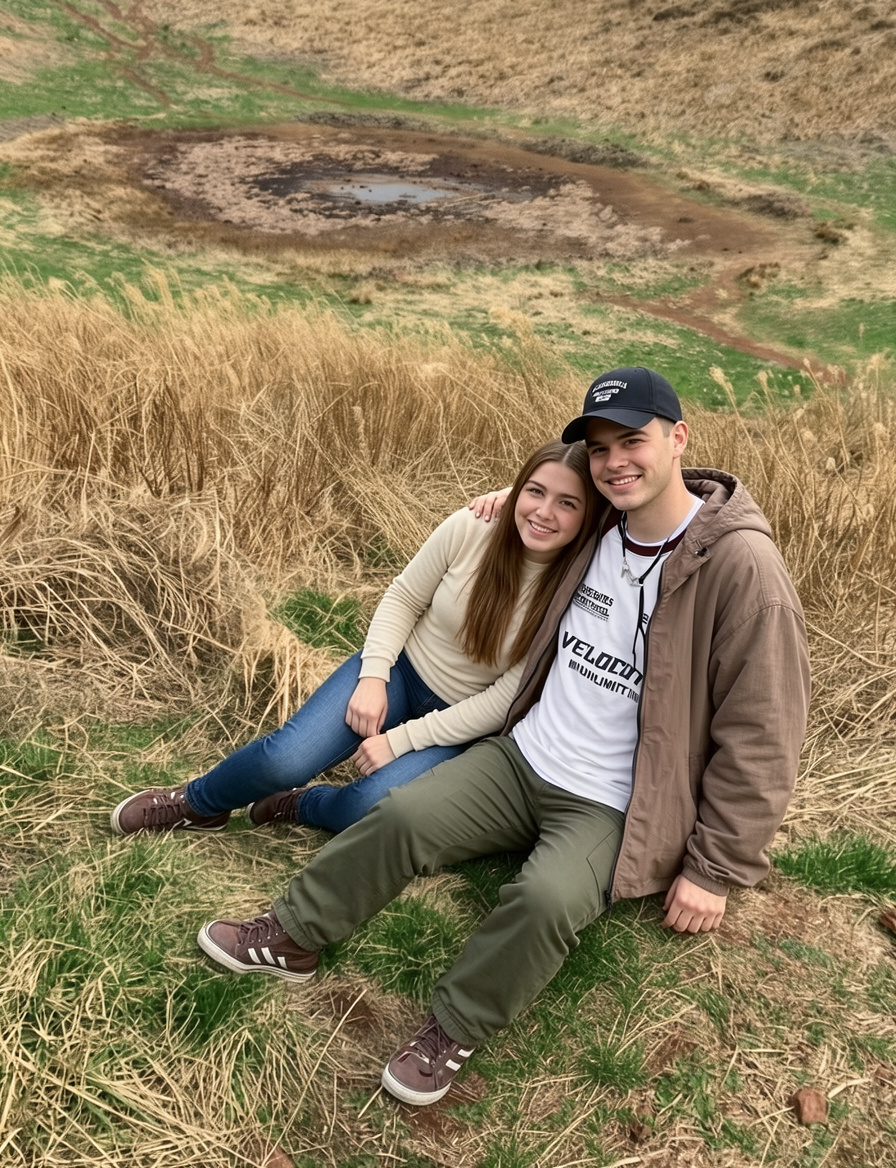}}\endgroup \\
\begingroup\setlength{\geditincorrectcelltall}{1.095000\geditincorrectcell}\geditincorrecttxtinstrtall{\fontsize{5.4}{6.2}\selectfont Add a beautiful woman to accompany the boyfriend in the image.}\endgroup & \begingroup\setlength{\geditincorrectcelltall}{1.095000\geditincorrectcell}\parbox[c][\geditincorrectcelltall][c]{\geditincorrectcell}{\includegraphics[width=\geditincorrectcell,height=\geditincorrectcelltall,trim={0.000000bp 86.400000bp 0.000000bp 144.000000bp},clip]{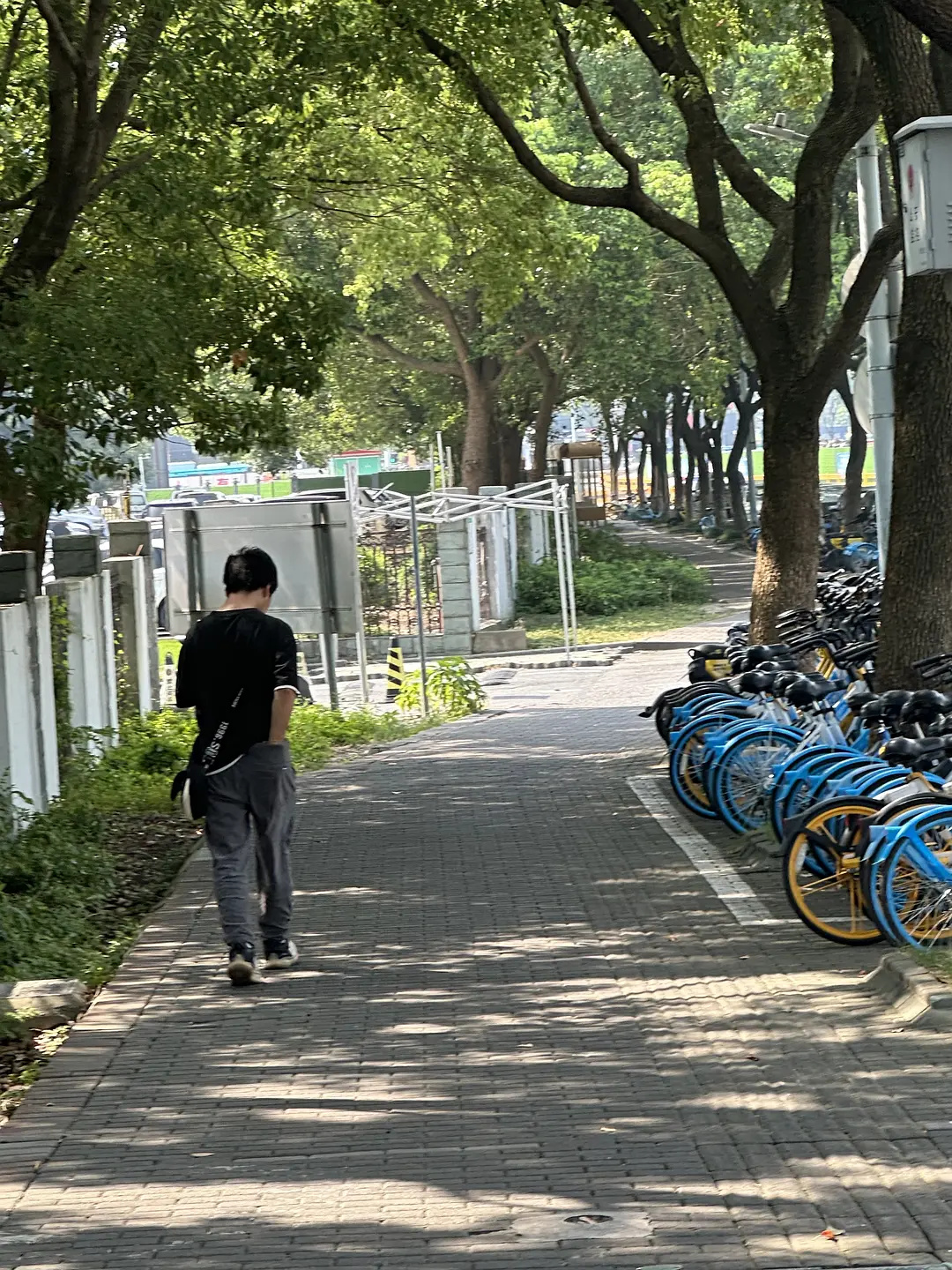}}\endgroup & \begingroup\setlength{\geditincorrectcelltall}{1.095000\geditincorrectcell}\parbox[c][\geditincorrectcelltall][c]{\geditincorrectcell}{\includegraphics[width=\geditincorrectcell,height=\geditincorrectcelltall,trim={0.000000bp 70.080000bp 0.000000bp 116.800000bp},clip]{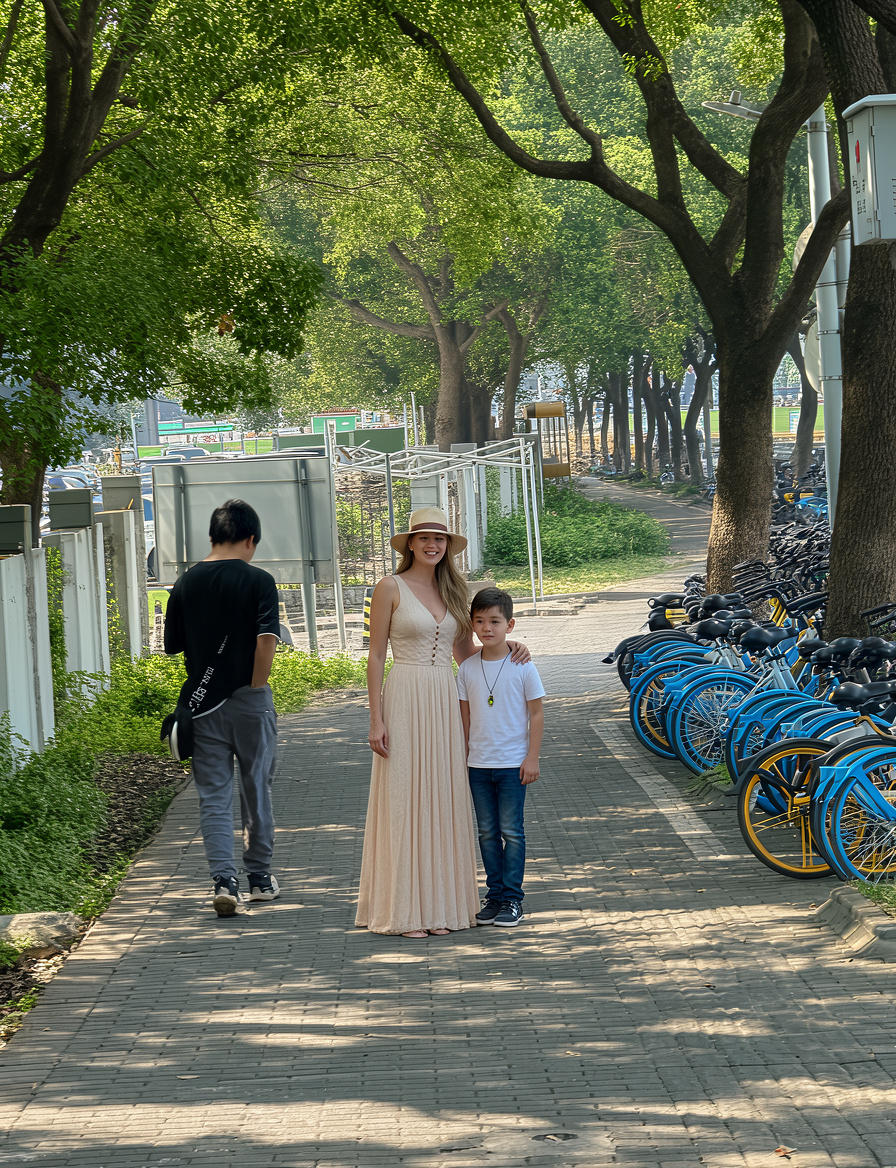}}\endgroup & \begingroup\setlength{\geditincorrectcelltall}{1.095000\geditincorrectcell}\geditincorrecttxtcaptall{\fontsize{5.2}{5.9}\selectfont A candid, eye-level photograph from behind of a young couple walking away from the camera down a tree-lined path on a sunny day. The man on the left has short black\,\ldots{}}\endgroup & \begingroup\setlength{\geditincorrectcelltall}{1.095000\geditincorrectcell}\parbox[c][\geditincorrectcelltall][c]{\geditincorrectcell}{\includegraphics[width=\geditincorrectcell,height=\geditincorrectcelltall,trim={0.000000bp 70.080000bp 0.000000bp 116.800000bp},clip]{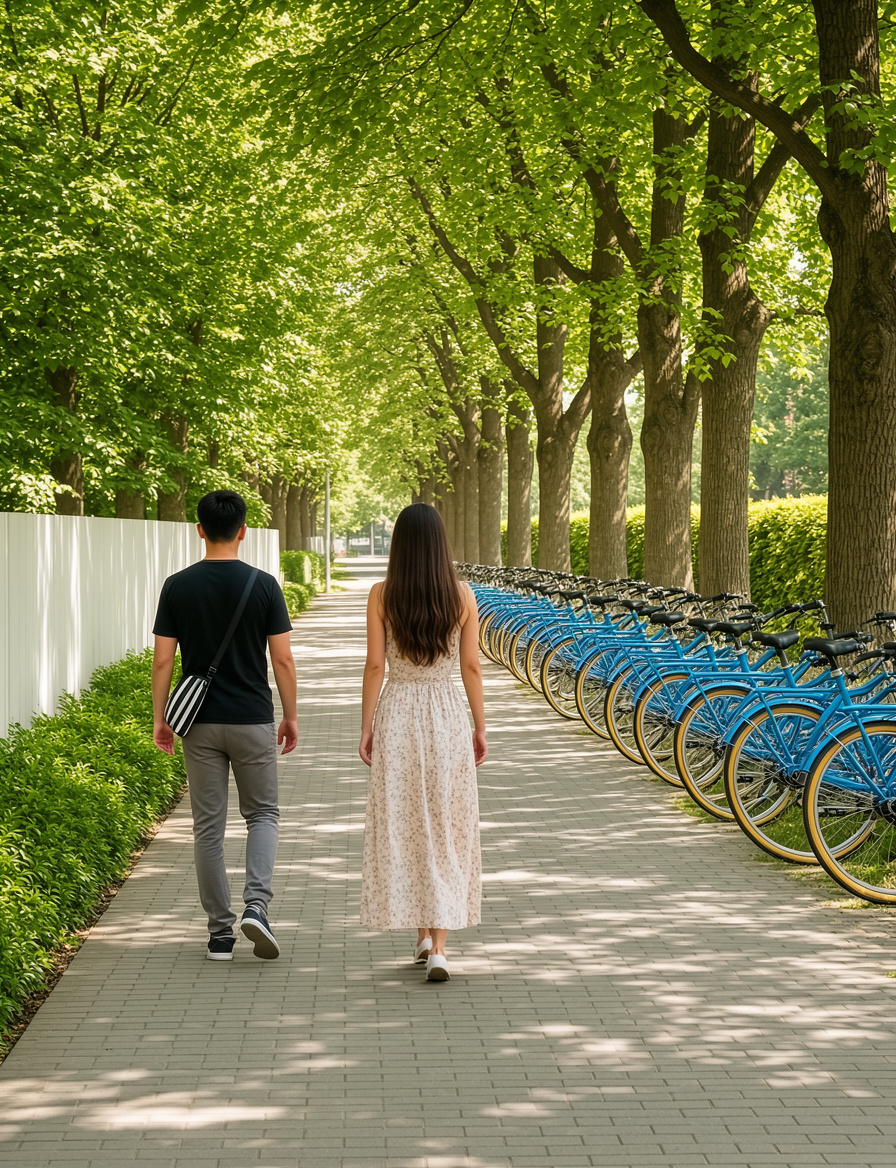}}\endgroup & \begingroup\setlength{\geditincorrectcelltall}{1.095000\geditincorrectcell}\parbox[c][\geditincorrectcelltall][c]{\geditincorrectcell}{\includegraphics[width=\geditincorrectcell,height=\geditincorrectcelltall,trim={0.000000bp 70.080000bp 0.000000bp 116.800000bp},clip]{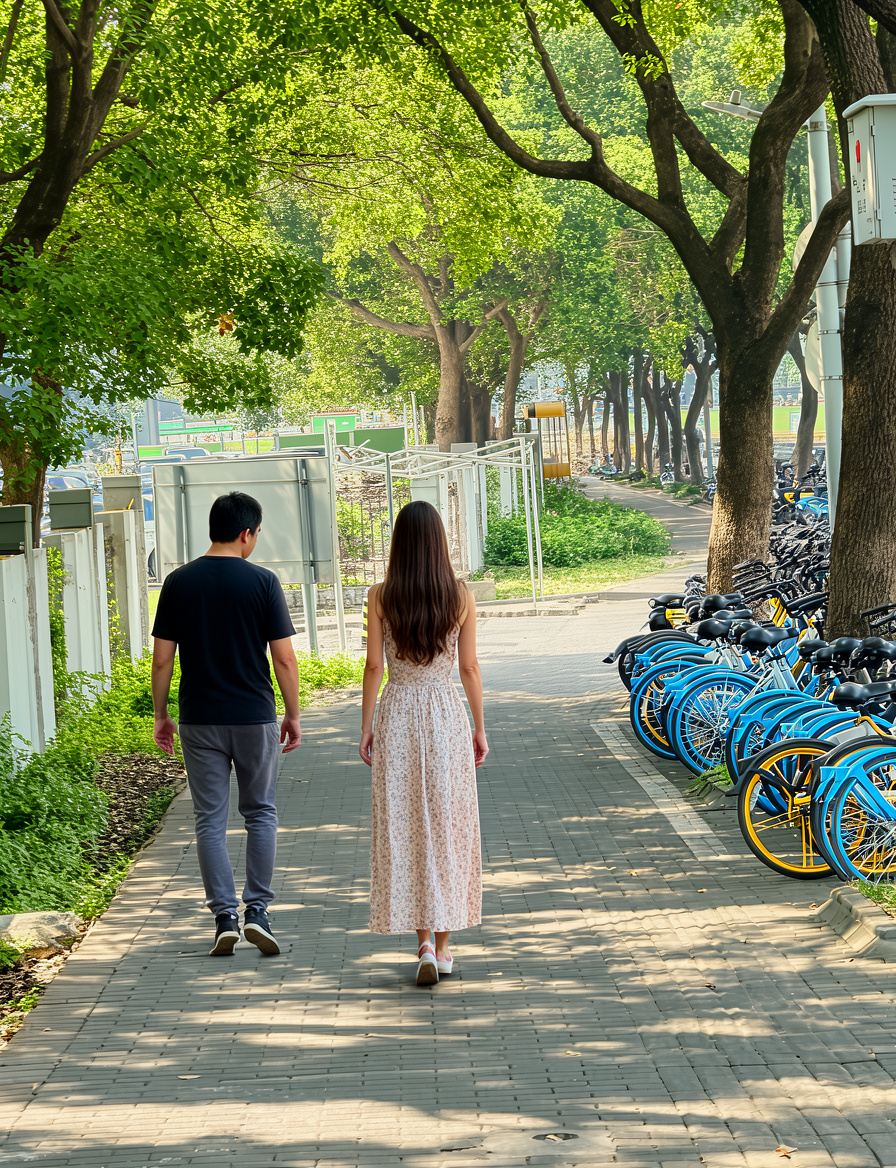}}\endgroup \\
\begingroup\setlength{\geditincorrectcelltall}{0.679487\geditincorrectcell}\geditincorrecttxtinstrtall{\fontsize{6}{7}\selectfont Make all the people in the image laugh.}\endgroup & \begingroup\setlength{\geditincorrectcelltall}{0.679487\geditincorrectcell}\parbox[c][\geditincorrectcelltall][c]{\geditincorrectcell}{\includegraphics[width=\geditincorrectcell,height=\geditincorrectcelltall,trim={0.000000bp 0.000000bp 0.000000bp 0.000000bp},clip]{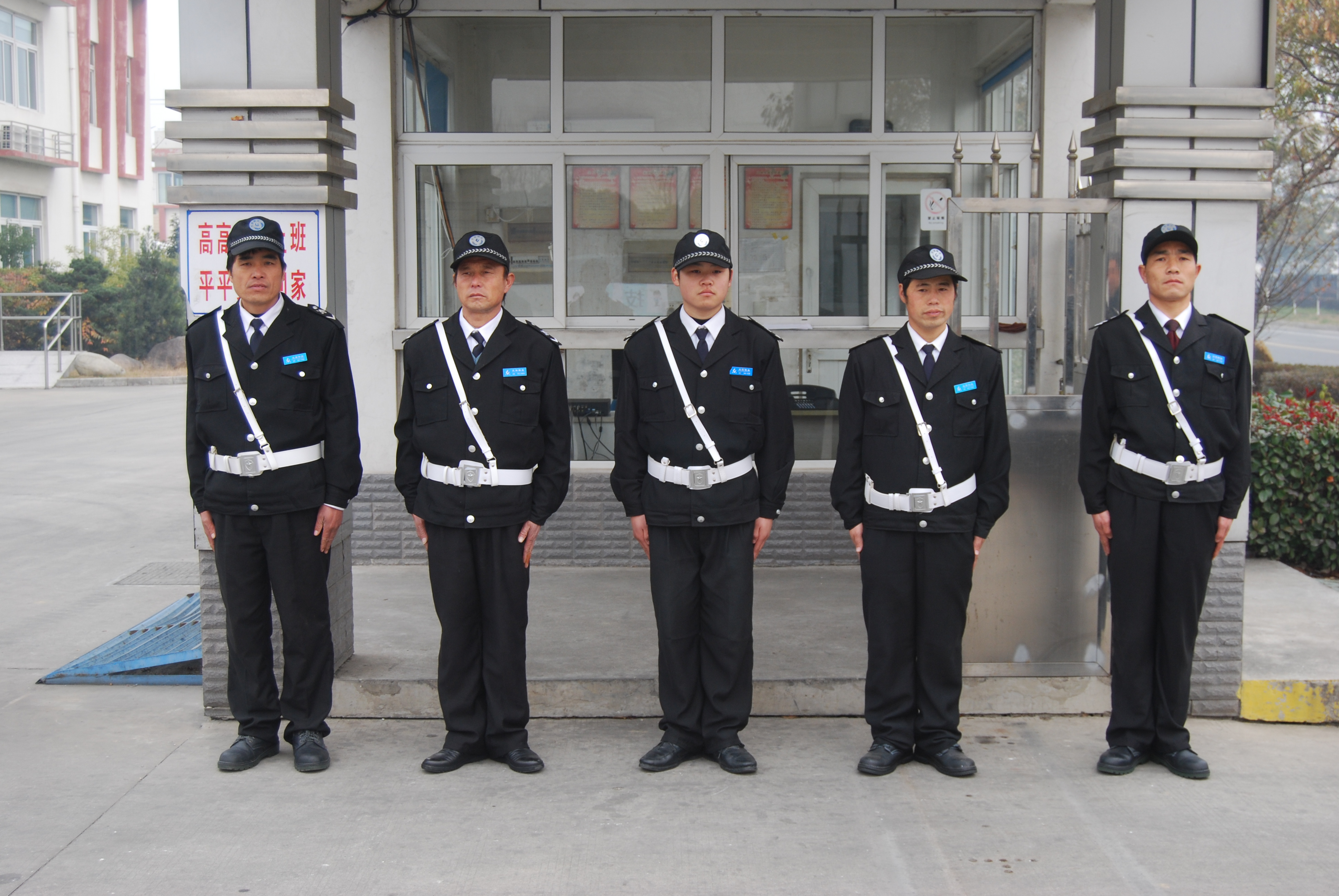}}\endgroup & \begingroup\setlength{\geditincorrectcelltall}{0.679487\geditincorrectcell}\parbox[c][\geditincorrectcelltall][c]{\geditincorrectcell}{\includegraphics[width=\geditincorrectcell,height=\geditincorrectcelltall,trim={0.000000bp 0.000000bp 0.000000bp 0.000000bp},clip]{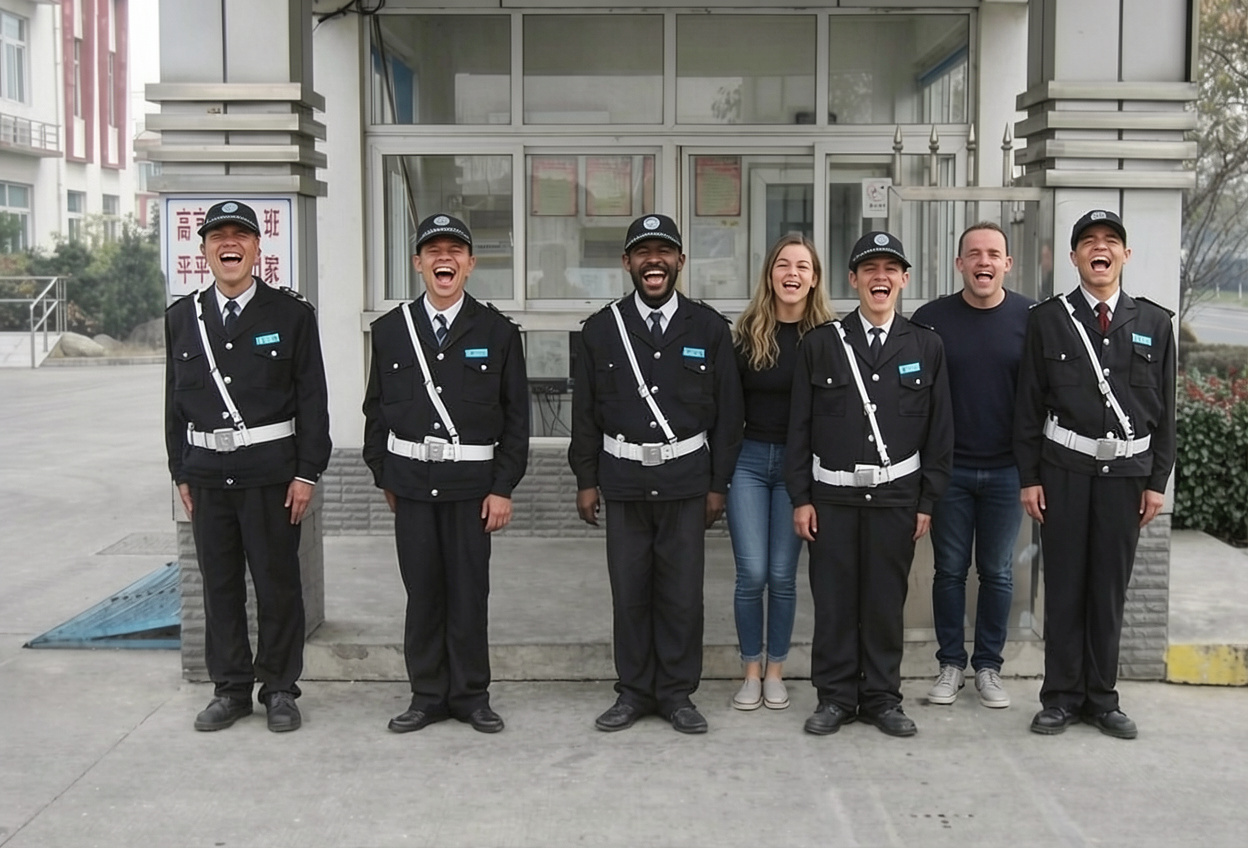}}\endgroup & \begingroup\setlength{\geditincorrectcelltall}{0.679487\geditincorrectcell}\geditincorrecttxtcaptall{\fontsize{5.2}{5.9}\selectfont A full-length, eye-level photograph captures five Asian male security guards standing in a perfectly straight line at attention, all laughing heartily with joyous\,\ldots{}}\endgroup & \begingroup\setlength{\geditincorrectcelltall}{0.679487\geditincorrectcell}\parbox[c][\geditincorrectcelltall][c]{\geditincorrectcell}{\includegraphics[width=\geditincorrectcell,height=\geditincorrectcelltall,trim={0.000000bp 0.000000bp 0.000000bp 0.000000bp},clip]{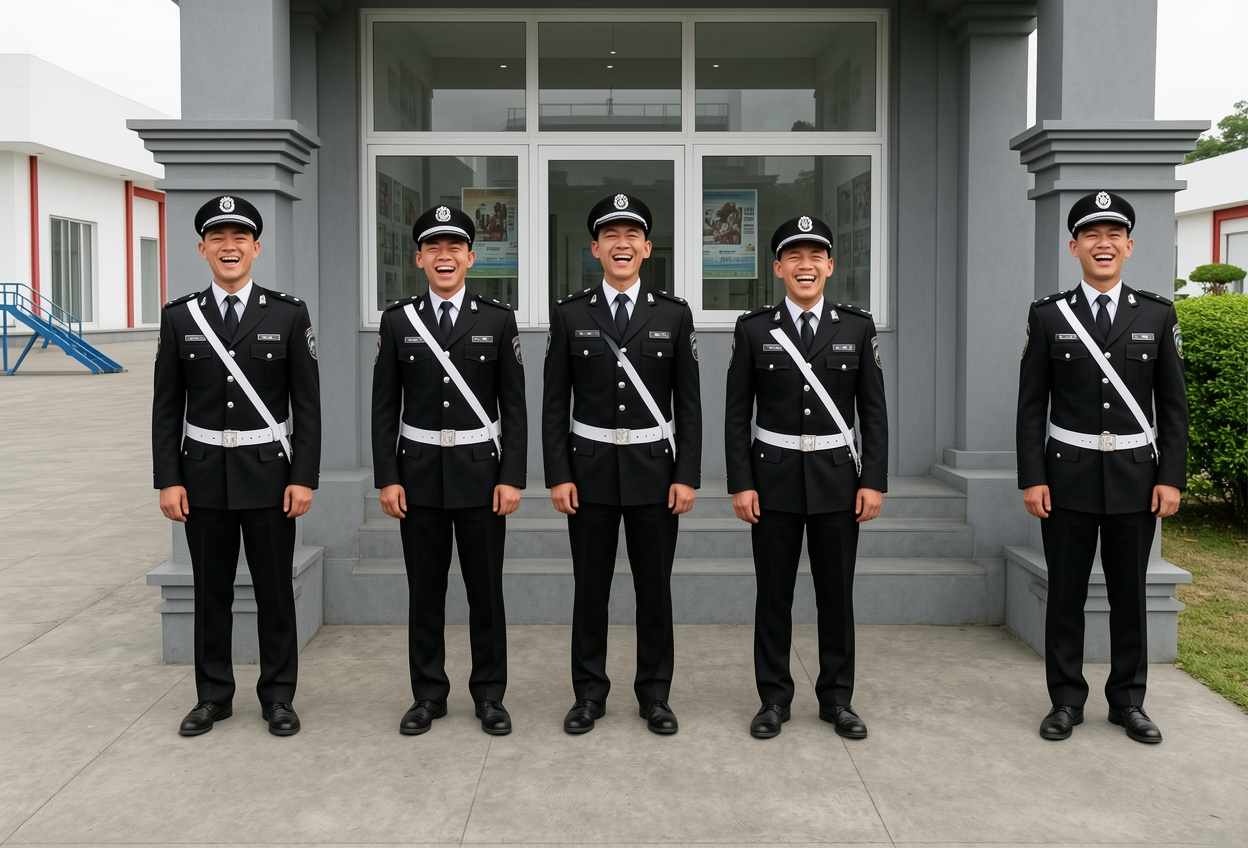}}\endgroup & \begingroup\setlength{\geditincorrectcelltall}{0.679487\geditincorrectcell}\parbox[c][\geditincorrectcelltall][c]{\geditincorrectcell}{\includegraphics[width=\geditincorrectcell,height=\geditincorrectcelltall,trim={0.000000bp 0.000000bp 0.000000bp 0.000000bp},clip]{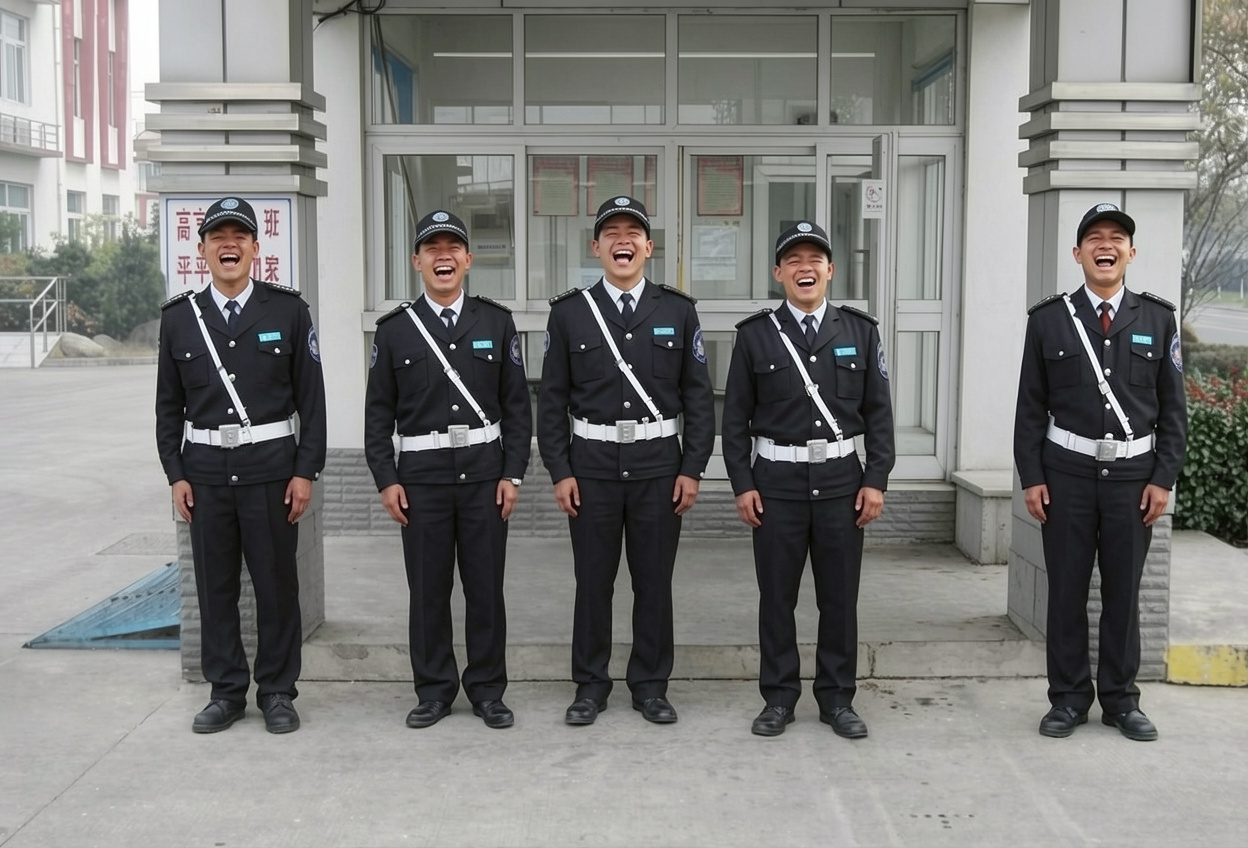}}\endgroup \\
\begingroup\setlength{\geditincorrectcelltall}{1.000000\geditincorrectcell}\geditincorrecttxtinstrtall{\fontsize{6}{7}\selectfont Replace the text 'duolingo' with 'greenowl'}\endgroup & \begingroup\setlength{\geditincorrectcelltall}{1.000000\geditincorrectcell}\parbox[c][\geditincorrectcelltall][c]{\geditincorrectcell}{\includegraphics[width=\geditincorrectcell,height=\geditincorrectcelltall,trim={331.121951bp 0.000000bp 331.121951bp 0.000000bp},clip]{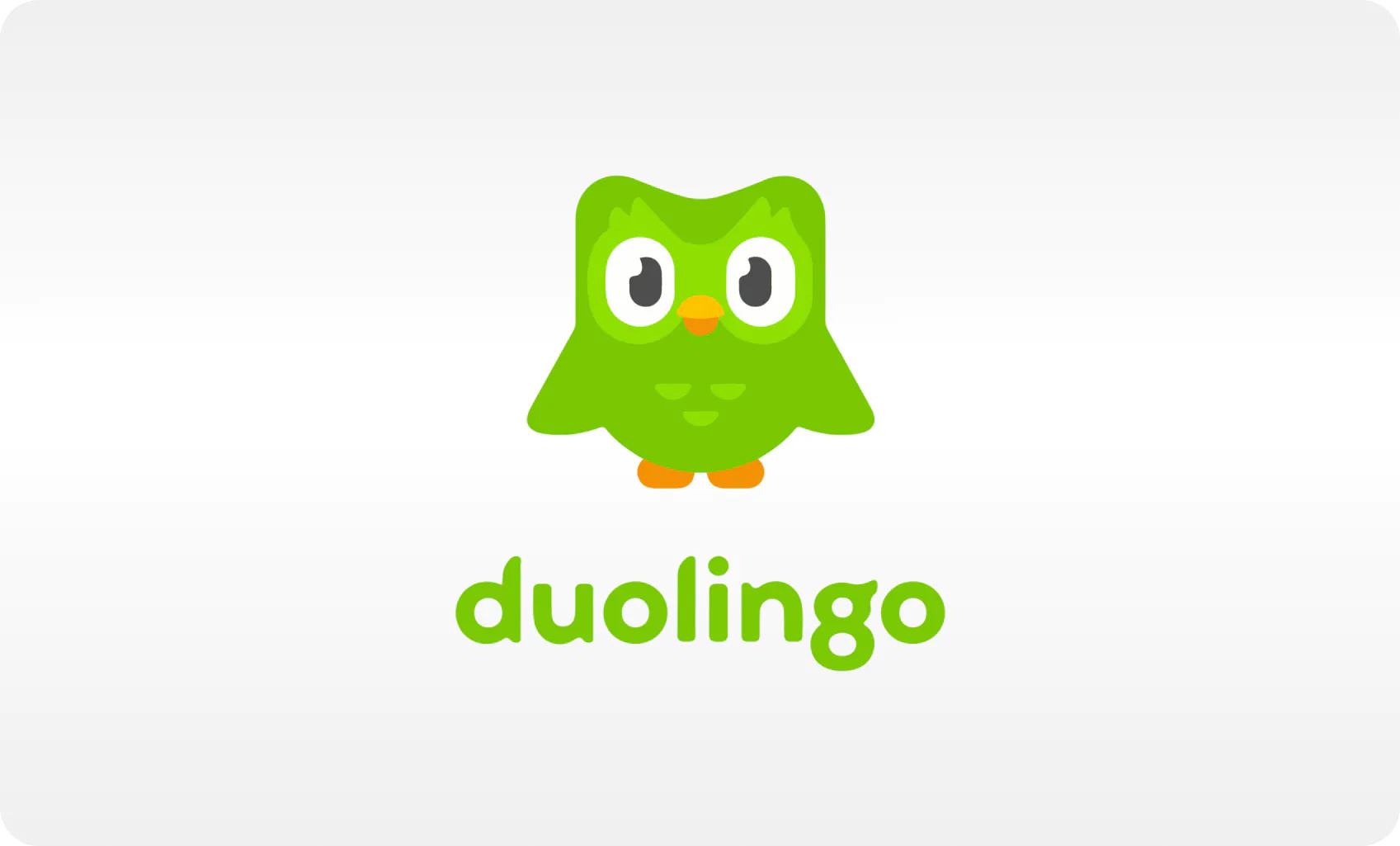}}\endgroup & \begingroup\setlength{\geditincorrectcelltall}{1.000000\geditincorrectcell}\parbox[c][\geditincorrectcelltall][c]{\geditincorrectcell}{\includegraphics[width=\geditincorrectcell,height=\geditincorrectcelltall,trim={256.000000bp 0.000000bp 256.000000bp 0.000000bp},clip]{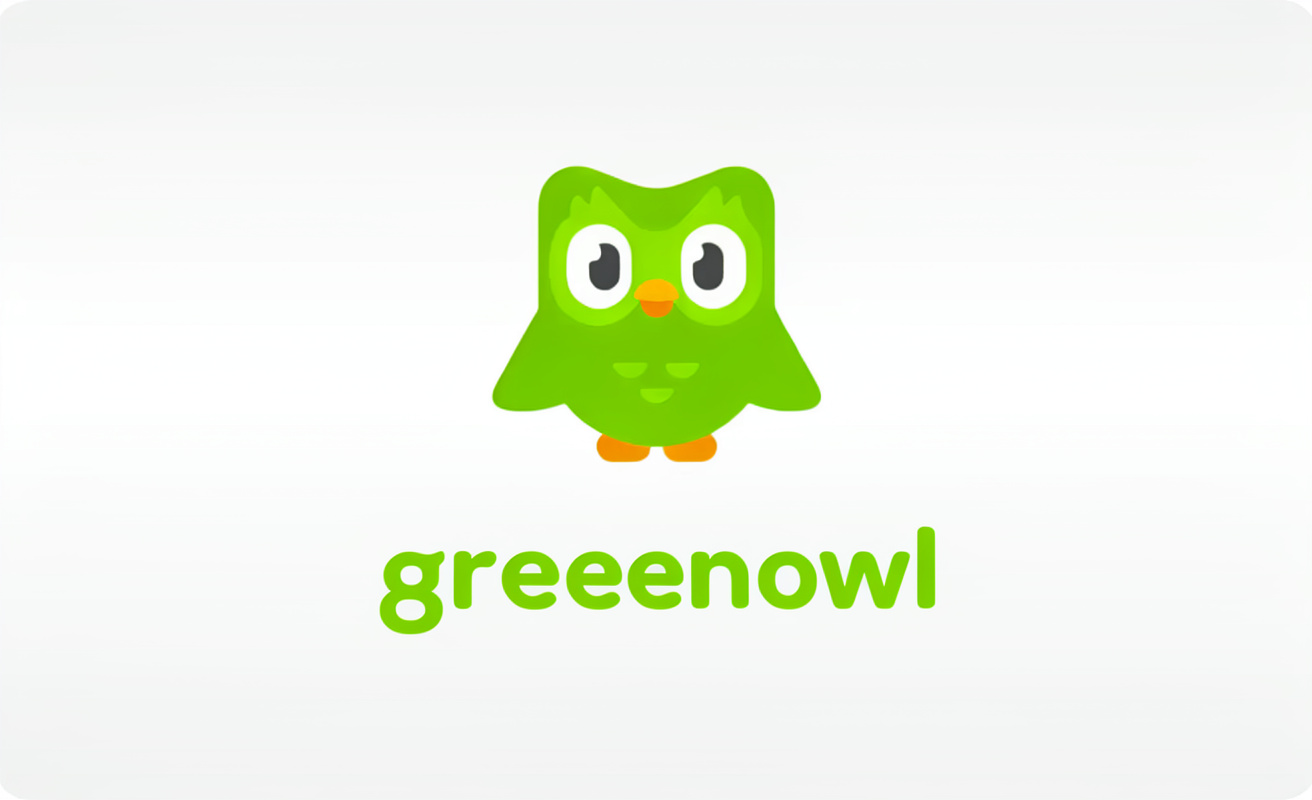}}\endgroup & \begingroup\setlength{\geditincorrectcelltall}{1.000000\geditincorrectcell}\geditincorrecttxtcaptall{\fontsize{5.2}{5.9}\selectfont A flat vector graphic of a stylized cartoon green owl mascot, centered on a minimalist off-white background with a subtle light grey horizontal gradient. The owl is\,\ldots{}}\endgroup & \begingroup\setlength{\geditincorrectcelltall}{1.000000\geditincorrectcell}\parbox[c][\geditincorrectcelltall][c]{\geditincorrectcell}{\includegraphics[width=\geditincorrectcell,height=\geditincorrectcelltall,trim={256.000000bp 0.000000bp 256.000000bp 0.000000bp},clip]{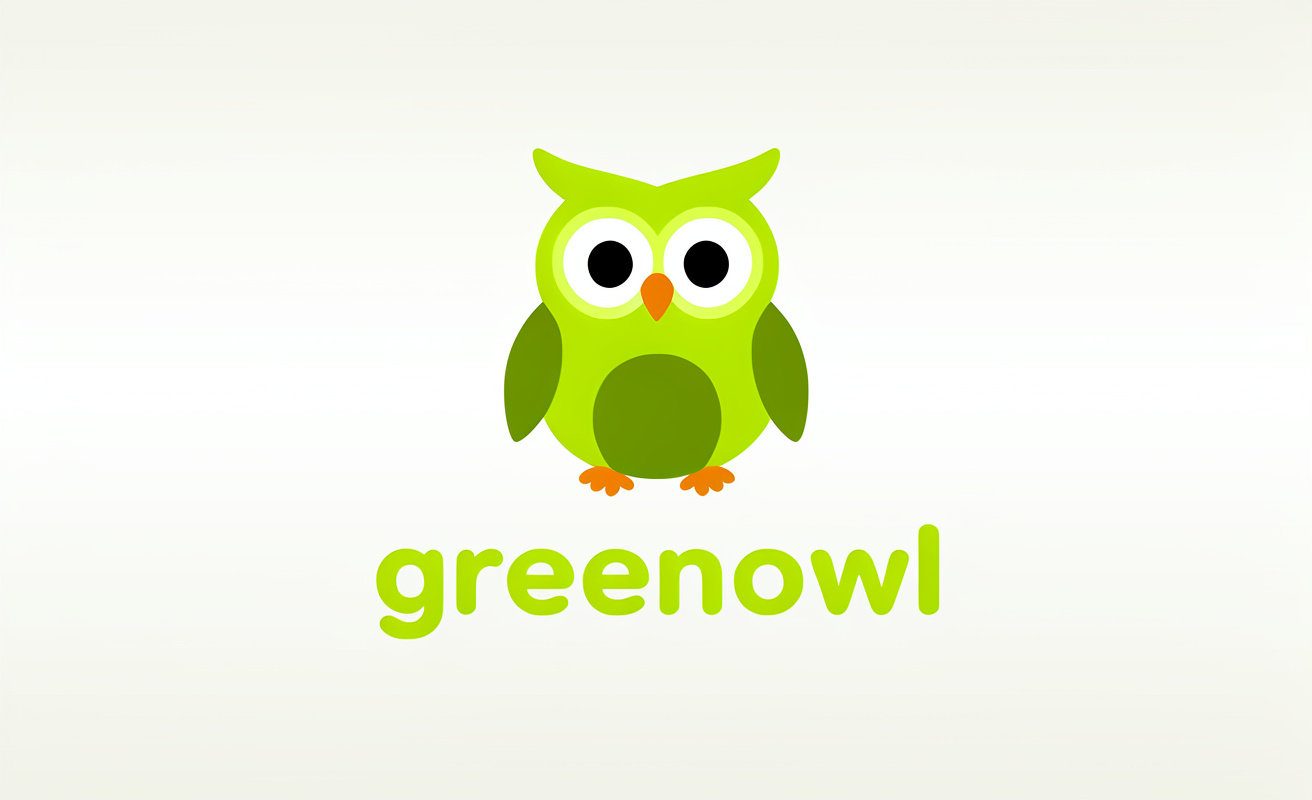}}\endgroup & \begingroup\setlength{\geditincorrectcelltall}{1.000000\geditincorrectcell}\parbox[c][\geditincorrectcelltall][c]{\geditincorrectcell}{\includegraphics[width=\geditincorrectcell,height=\geditincorrectcelltall,trim={256.000000bp 0.000000bp 256.000000bp 0.000000bp},clip]{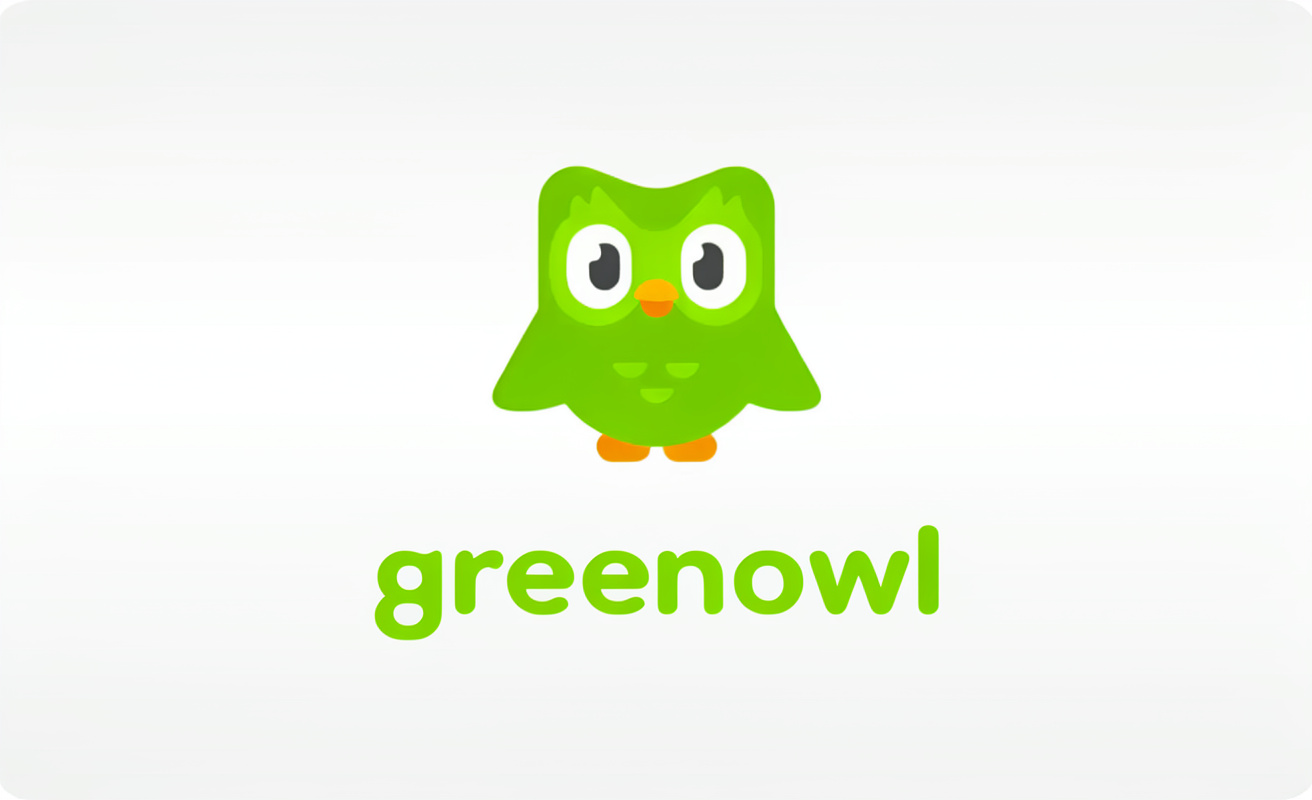}}\endgroup \\
\end{tabular}
\endgroup
\caption{\textbf{Incorrect edits (FLUX2-Klein~4B, GEdit).} Pure editing changes the image but does not produce the requested result. Columns as in \cref{fig:appx-gedit-missed}.}
\label{fig:appx-gedit-incorrect}
\end{figure}

\clearpage

\begin{figure}[!htb]
\centering

\begingroup
\sffamily
\setlength{\tabcolsep}{0pt}
\renewcommand{\arraystretch}{0}
\setlength{\fboxsep}{0pt}
\newlength{\geditunnaturalincompleteinstr}
\newlength{\geditunnaturalincompletecell}
\newlength{\geditunnaturalincompletecelltall}
\setlength{\geditunnaturalincompleteinstr}{0.110\linewidth}
\setlength{\geditunnaturalincompletecell}{0.178\linewidth}
\setlength{\geditunnaturalincompletecelltall}{1.000\geditunnaturalincompletecell}
\providecommand{\geditunnaturalincompleteimg}[1]{%
  \parbox[c][\geditunnaturalincompletecell][c]{\geditunnaturalincompletecell}{\centering\includegraphics[width=\geditunnaturalincompletecell,height=\geditunnaturalincompletecell,keepaspectratio]{#1}}%
}
\providecommand{\geditunnaturalincompleteimgtall}[1]{%
  \parbox[c][\geditunnaturalincompletecelltall][c]{\geditunnaturalincompletecell}{\centering\includegraphics[width=\geditunnaturalincompletecell]{#1}}%
}
\providecommand{\geditunnaturalincompletemissing}{%
  \parbox[c][\geditunnaturalincompletecell][c]{\geditunnaturalincompletecell}{\centering\textcolor{black!10}{\rule{\geditunnaturalincompletecell}{\geditunnaturalincompletecell}}}%
}
\providecommand{\geditunnaturalincompletemissingtall}{%
  \parbox[c][\geditunnaturalincompletecelltall][c]{\geditunnaturalincompletecell}{\centering\textcolor{black!10}{\rule{\geditunnaturalincompletecell}{\geditunnaturalincompletecelltall}}}%
}
\providecommand{\geditunnaturalincompletetextpad}{\vspace{2pt}\vspace{0pt}}
\providecommand{\geditunnaturalincompletetxtinstr}[1]{%
  \parbox[c][\geditunnaturalincompletecell][t]{\dimexpr\geditunnaturalincompleteinstr-2pt\relax}{\geditunnaturalincompletetextpad\raggedright #1\par}%
}
\providecommand{\geditunnaturalincompletetxtinstrtall}[1]{%
  \parbox[c][\geditunnaturalincompletecelltall][t]{\dimexpr\geditunnaturalincompleteinstr-2pt\relax}{\geditunnaturalincompletetextpad\raggedright #1\par}%
}
\providecommand{\geditunnaturalincompletetxtcap}[1]{%
  \parbox[c][\geditunnaturalincompletecell][t]{\dimexpr\geditunnaturalincompletecell-2pt\relax}{\geditunnaturalincompletetextpad\raggedright #1\par}%
}
\providecommand{\geditunnaturalincompletetxtcaptall}[1]{%
  \parbox[c][\geditunnaturalincompletecelltall][t]{\dimexpr\geditunnaturalincompletecell-2pt\relax}{\geditunnaturalincompletetextpad\raggedright #1\par}%
}
\providecommand{\geditunnaturalincompletehdrinstr}[1]{%
  \parbox[t]{\dimexpr\geditunnaturalincompleteinstr-2pt\relax}{\centering\geditunnaturalincompletetextpad #1\par}%
}
\providecommand{\geditunnaturalincompletehdr}[1]{%
  \parbox[t]{\dimexpr\geditunnaturalincompletecell-2pt\relax}{\centering\geditunnaturalincompletetextpad #1\par}%
}

\begin{tabular}{@{}>{\raggedright\arraybackslash}m{\geditunnaturalincompleteinstr}*{5}{>{\centering\arraybackslash}m{\geditunnaturalincompletecell}}@{}}
\geditunnaturalincompletehdrinstr{\fontsize{7}{8}\selectfont\bfseries instruction} & \geditunnaturalincompletehdr{\fontsize{7}{8}\selectfont\bfseries source} & \geditunnaturalincompletehdr{\fontsize{7}{8}\selectfont\bfseries pure-E} & \geditunnaturalincompletehdr{\fontsize{7}{8}\selectfont\bfseries T2I caption} & \geditunnaturalincompletehdr{\fontsize{7}{8}\selectfont\bfseries E$\to$T2I $k{=}10$} & \geditunnaturalincompletehdr{\fontsize{7}{8}\selectfont\bfseries E$\to$T2I$\to$E\\$[10,16)$} \\
\noalign{\vskip 1pt}
\begingroup\setlength{\geditunnaturalincompletecelltall}{1.000000\geditunnaturalincompletecell}\geditunnaturalincompletetxtinstrtall{\fontsize{6}{7}\selectfont Enhance my nose.}\endgroup & \begingroup\setlength{\geditunnaturalincompletecelltall}{1.000000\geditunnaturalincompletecell}\parbox[c][\geditunnaturalincompletecelltall][c]{\geditunnaturalincompletecell}{\includegraphics[width=\geditunnaturalincompletecell,height=\geditunnaturalincompletecelltall,trim={0.000000bp 0.000000bp 0.000000bp 0.000000bp},clip]{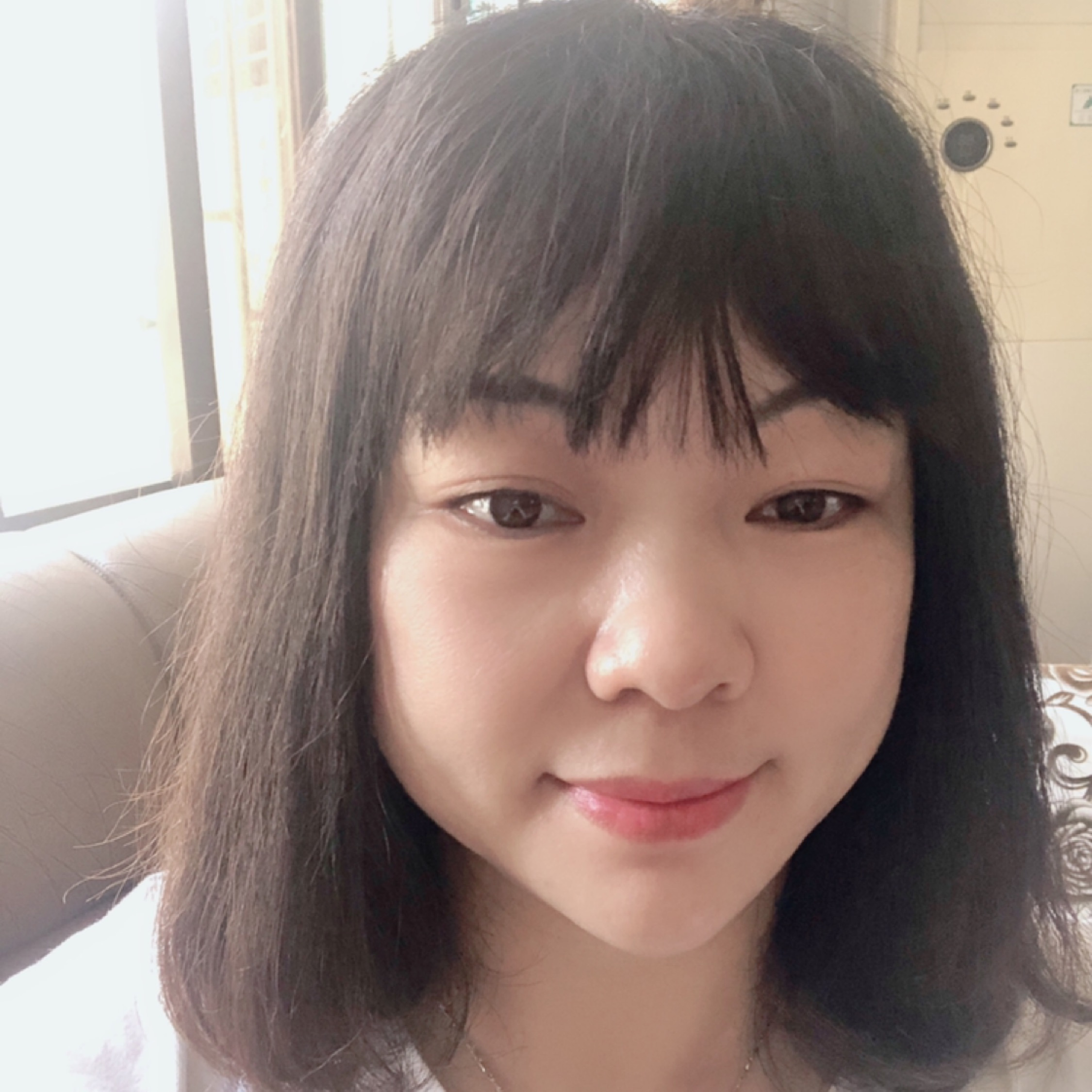}}\endgroup & \begingroup\setlength{\geditunnaturalincompletecelltall}{1.000000\geditunnaturalincompletecell}\parbox[c][\geditunnaturalincompletecelltall][c]{\geditunnaturalincompletecell}{\includegraphics[width=\geditunnaturalincompletecell,height=\geditunnaturalincompletecelltall,trim={0.000000bp 0.000000bp 0.000000bp 0.000000bp},clip]{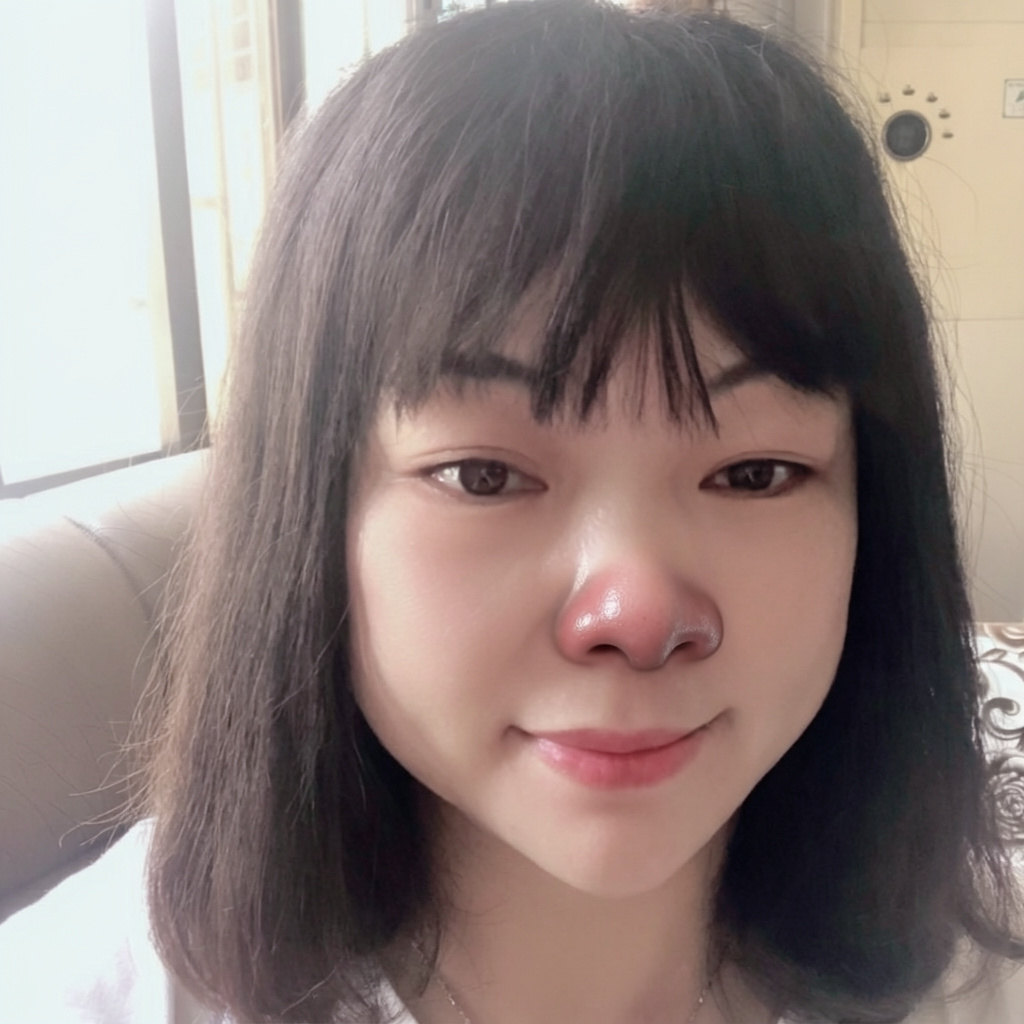}}\endgroup & \begingroup\setlength{\geditunnaturalincompletecelltall}{1.000000\geditunnaturalincompletecell}\geditunnaturalincompletetxtcaptall{\fontsize{5.2}{5.9}\selectfont A close-up, natural-light selfie of a young East Asian woman with fair skin, looking at the camera with a gentle smile. She has shoulder-length, dark brown hair\,\ldots{}}\endgroup & \begingroup\setlength{\geditunnaturalincompletecelltall}{1.000000\geditunnaturalincompletecell}\parbox[c][\geditunnaturalincompletecelltall][c]{\geditunnaturalincompletecell}{\includegraphics[width=\geditunnaturalincompletecell,height=\geditunnaturalincompletecelltall,trim={0.000000bp 0.000000bp 0.000000bp 0.000000bp},clip]{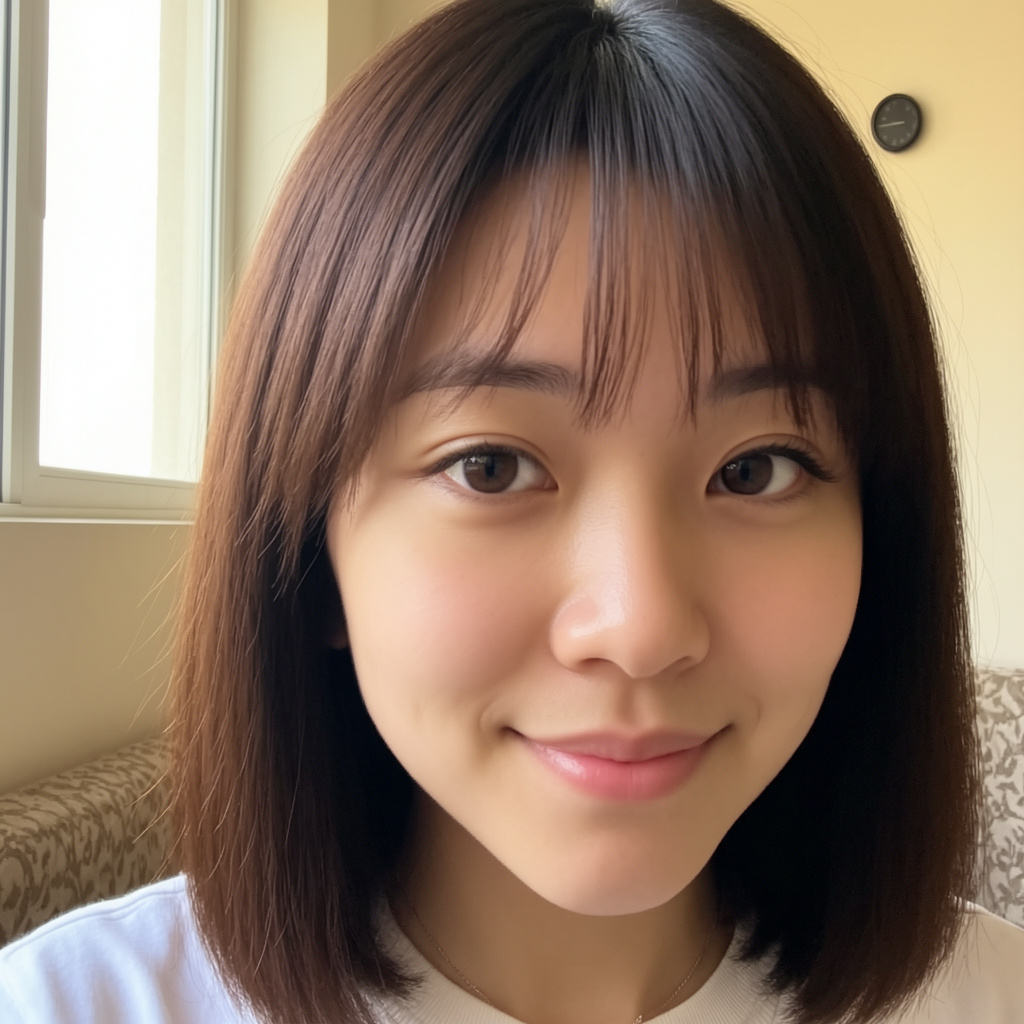}}\endgroup & \begingroup\setlength{\geditunnaturalincompletecelltall}{1.000000\geditunnaturalincompletecell}\parbox[c][\geditunnaturalincompletecelltall][c]{\geditunnaturalincompletecell}{\includegraphics[width=\geditunnaturalincompletecell,height=\geditunnaturalincompletecelltall,trim={0.000000bp 0.000000bp 0.000000bp 0.000000bp},clip]{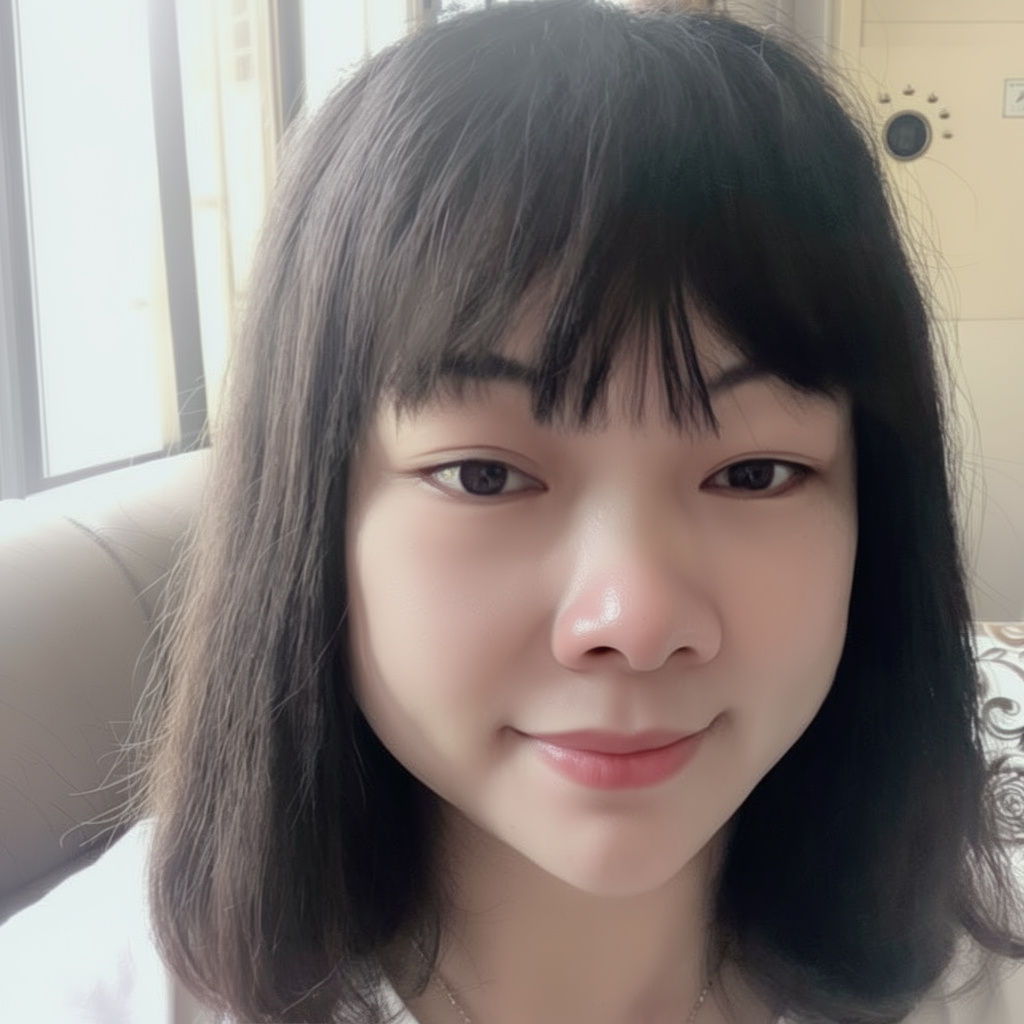}}\endgroup \\
\begingroup\setlength{\geditunnaturalincompletecelltall}{1.471698\geditunnaturalincompletecell}\geditunnaturalincompletetxtinstrtall{\fontsize{6}{7}\selectfont Make his nose higher.}\endgroup & \begingroup\setlength{\geditunnaturalincompletecelltall}{1.471698\geditunnaturalincompletecell}\parbox[c][\geditunnaturalincompletecelltall][c]{\geditunnaturalincompletecell}{\includegraphics[width=\geditunnaturalincompletecell,height=\geditunnaturalincompletecelltall,trim={0.000000bp 0.000000bp 0.000000bp 0.000000bp},clip]{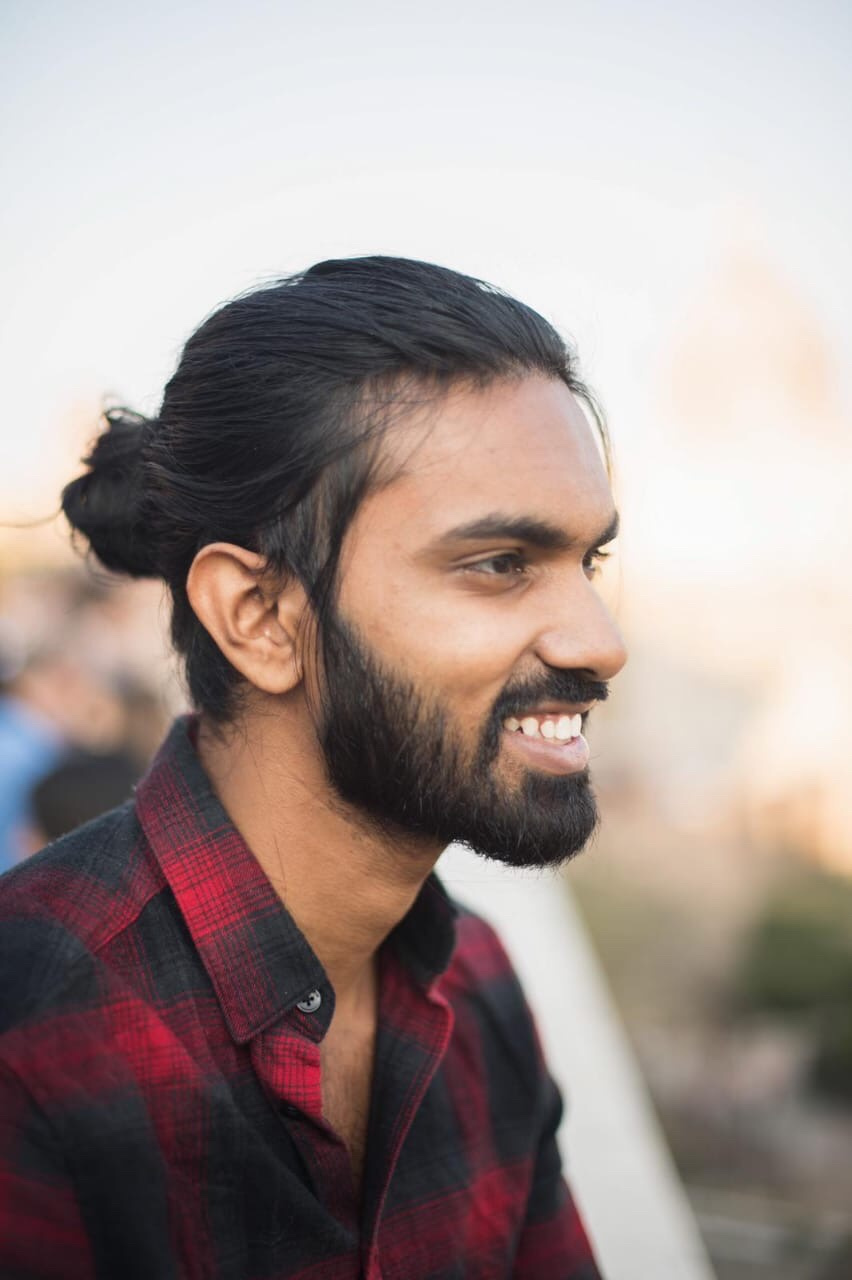}}\endgroup & \begingroup\setlength{\geditunnaturalincompletecelltall}{1.471698\geditunnaturalincompletecell}\parbox[c][\geditunnaturalincompletecelltall][c]{\geditunnaturalincompletecell}{\includegraphics[width=\geditunnaturalincompletecell,height=\geditunnaturalincompletecelltall,trim={0.000000bp 0.000000bp 0.000000bp 0.000000bp},clip]{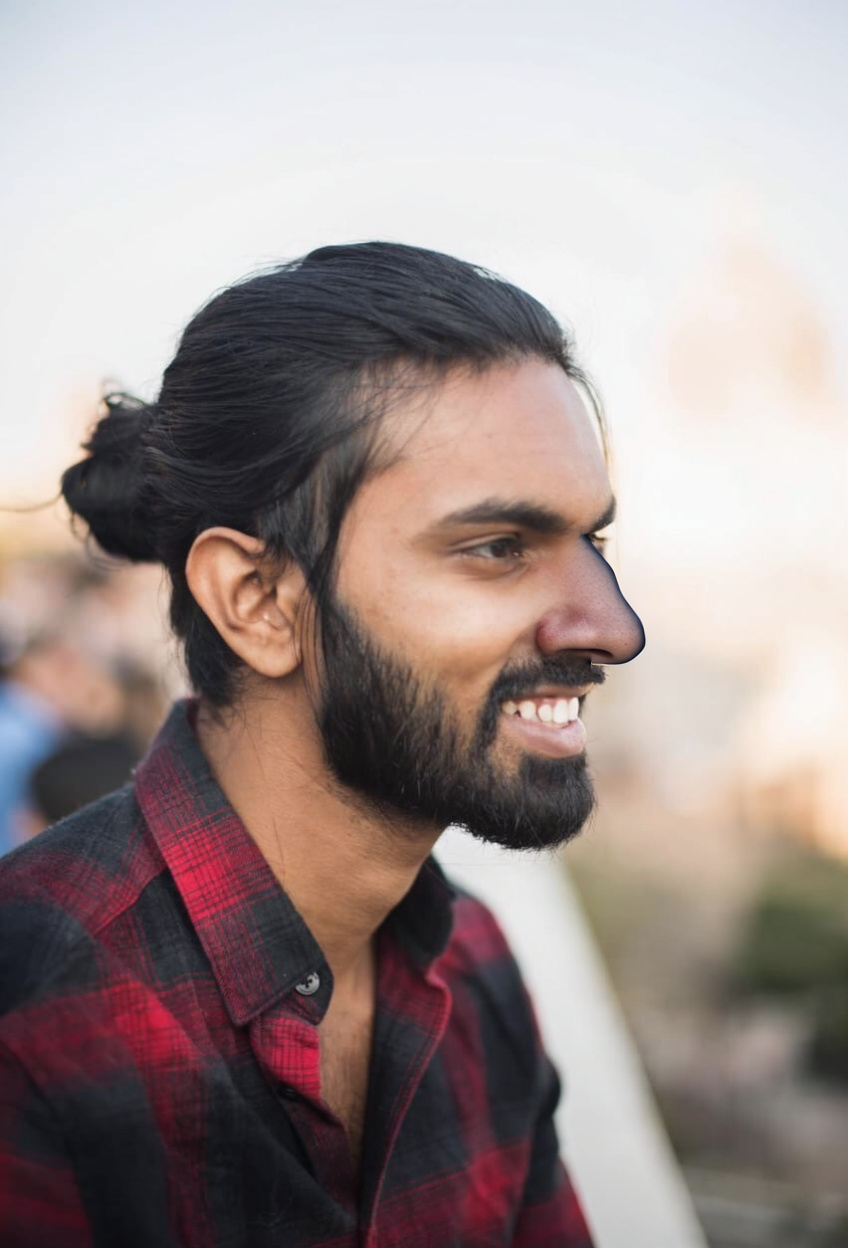}}\endgroup & \begingroup\setlength{\geditunnaturalincompletecelltall}{1.471698\geditunnaturalincompletecell}\geditunnaturalincompletetxtcaptall{\fontsize{5.2}{5.9}\selectfont A photographic close-up portrait of a young South Asian man with dark skin, viewed from a three-quarter profile as he smiles and looks towards the right. He has a\,\ldots{}}\endgroup & \begingroup\setlength{\geditunnaturalincompletecelltall}{1.471698\geditunnaturalincompletecell}\parbox[c][\geditunnaturalincompletecelltall][c]{\geditunnaturalincompletecell}{\includegraphics[width=\geditunnaturalincompletecell,height=\geditunnaturalincompletecelltall,trim={0.000000bp 0.000000bp 0.000000bp 0.000000bp},clip]{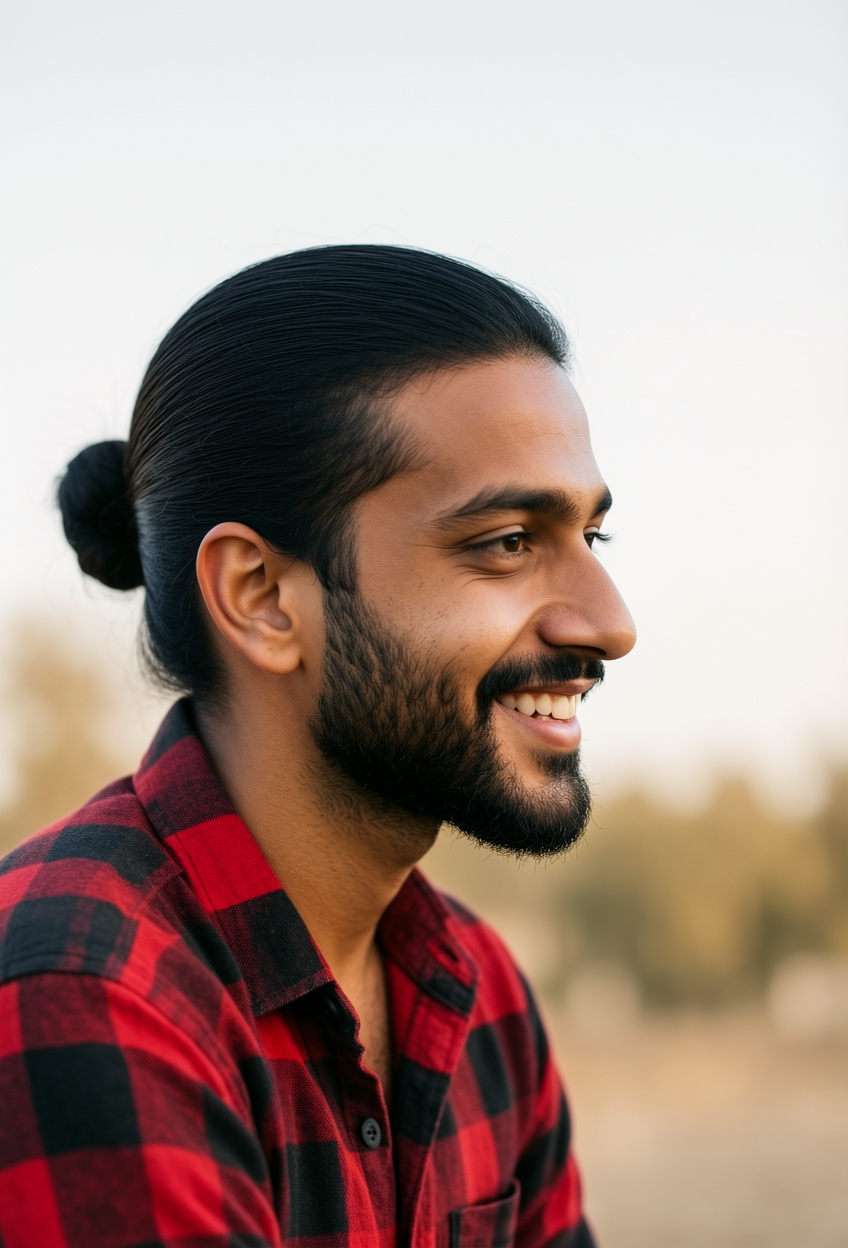}}\endgroup & \begingroup\setlength{\geditunnaturalincompletecelltall}{1.471698\geditunnaturalincompletecell}\parbox[c][\geditunnaturalincompletecelltall][c]{\geditunnaturalincompletecell}{\includegraphics[width=\geditunnaturalincompletecell,height=\geditunnaturalincompletecelltall,trim={0.000000bp 0.000000bp 0.000000bp 0.000000bp},clip]{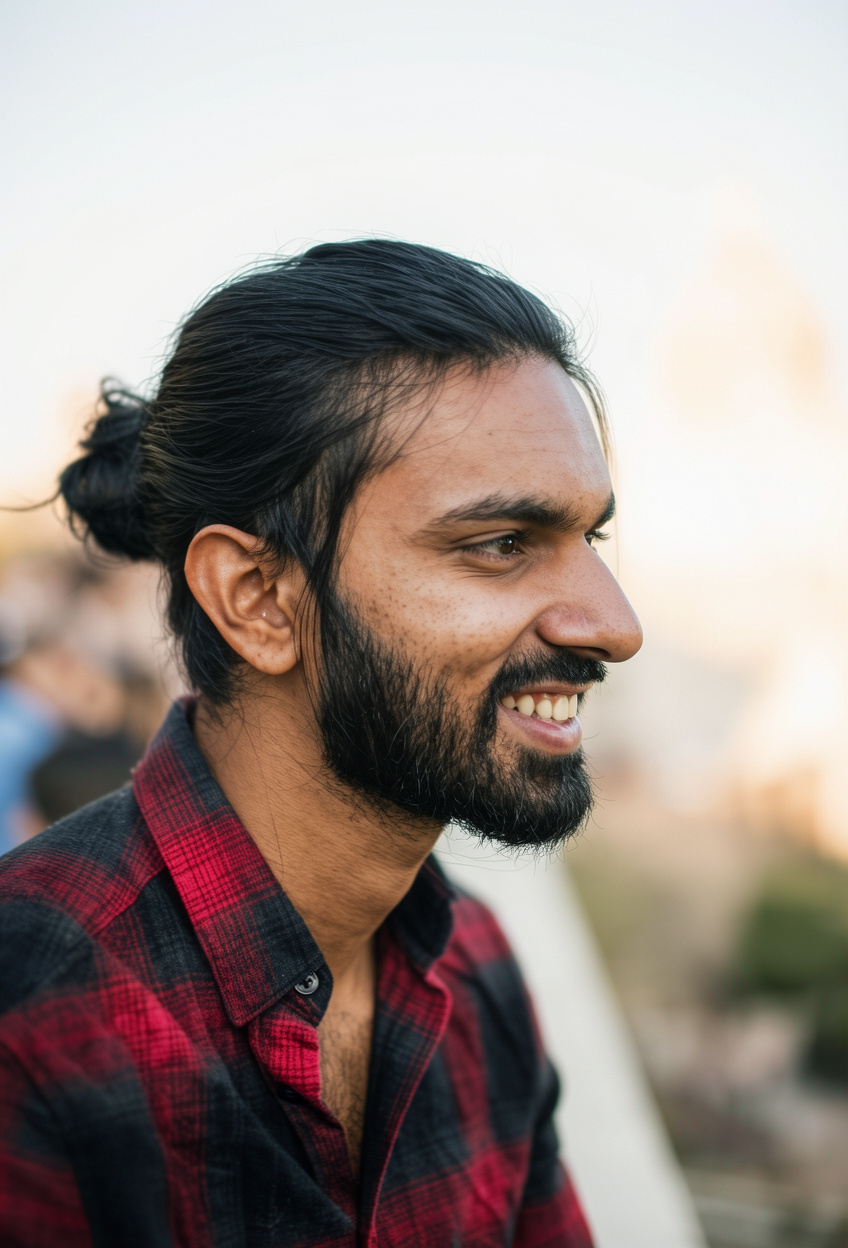}}\endgroup \\
\begingroup\setlength{\geditunnaturalincompletecelltall}{1.303571\geditunnaturalincompletecell}\geditunnaturalincompletetxtinstrtall{\fontsize{6}{7}\selectfont Change the character into a male.}\endgroup & \begingroup\setlength{\geditunnaturalincompletecelltall}{1.303571\geditunnaturalincompletecell}\parbox[c][\geditunnaturalincompletecelltall][c]{\geditunnaturalincompletecell}{\includegraphics[width=\geditunnaturalincompletecell,height=\geditunnaturalincompletecelltall,trim={0.000000bp 0.000000bp 0.000000bp 0.000000bp},clip]{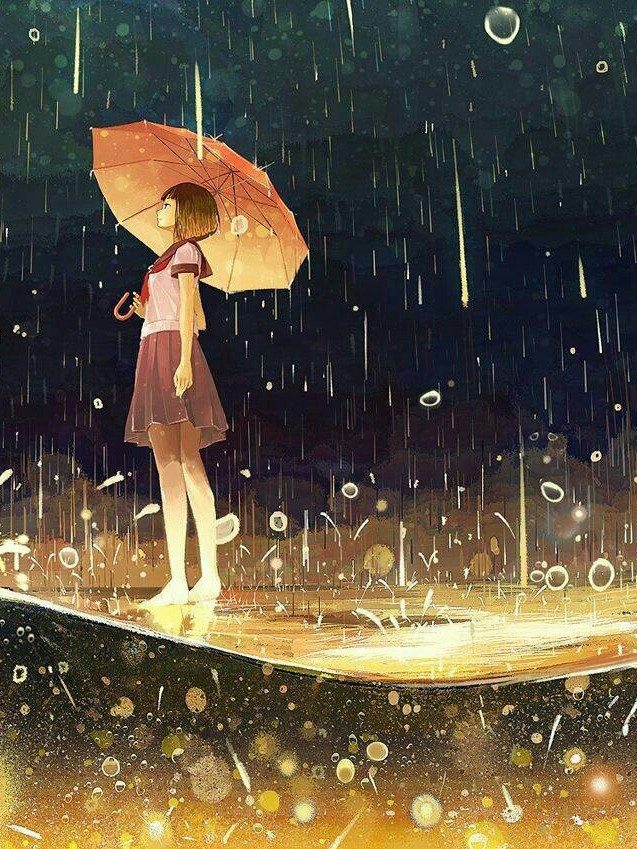}}\endgroup & \begingroup\setlength{\geditunnaturalincompletecelltall}{1.303571\geditunnaturalincompletecell}\parbox[c][\geditunnaturalincompletecelltall][c]{\geditunnaturalincompletecell}{\includegraphics[width=\geditunnaturalincompletecell,height=\geditunnaturalincompletecelltall,trim={0.000000bp 0.000000bp 0.000000bp 0.000000bp},clip]{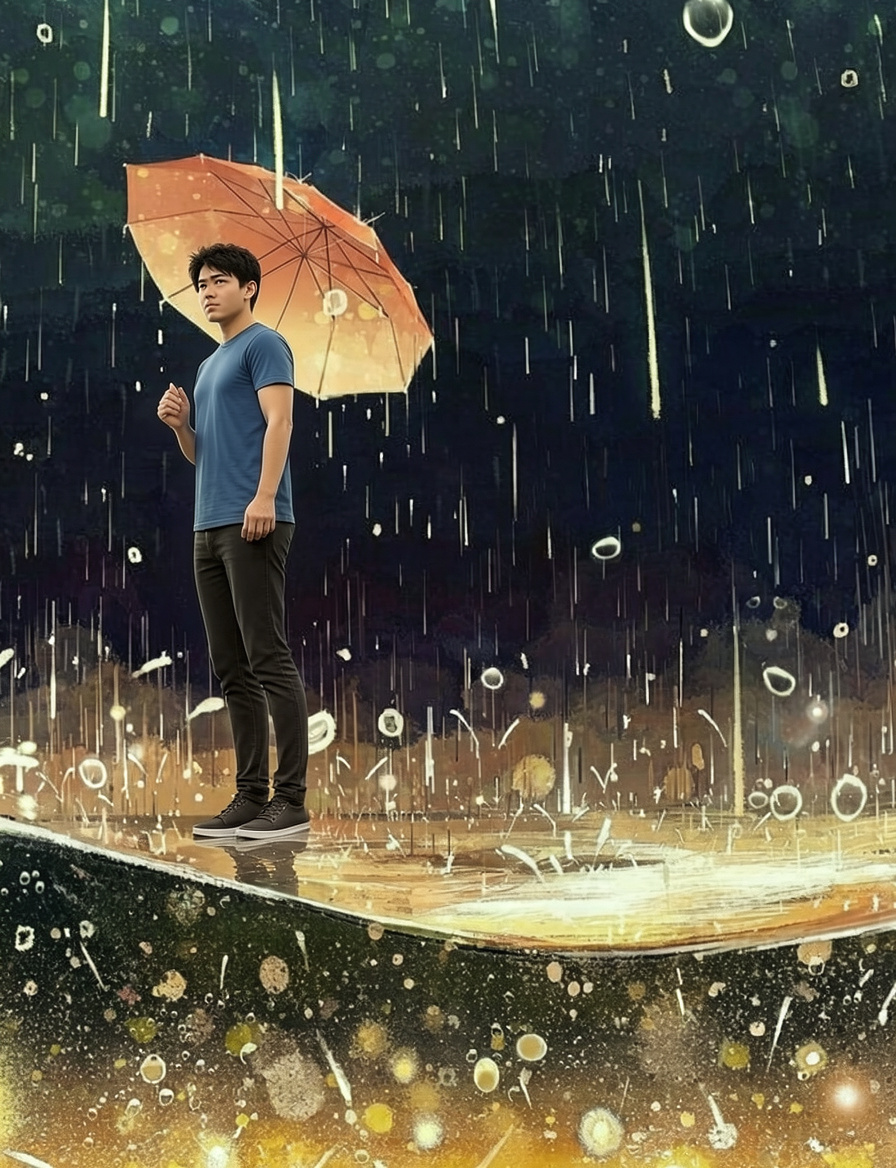}}\endgroup & \begingroup\setlength{\geditunnaturalincompletecelltall}{1.303571\geditunnaturalincompletecell}\geditunnaturalincompletetxtcaptall{\fontsize{5.2}{5.9}\selectfont A young anime-style male stands barefoot on a glowing, reflective surface during a rainstorm, holding a translucent orange umbrella above his head. He has short\,\ldots{}}\endgroup & \begingroup\setlength{\geditunnaturalincompletecelltall}{1.303571\geditunnaturalincompletecell}\parbox[c][\geditunnaturalincompletecelltall][c]{\geditunnaturalincompletecell}{\includegraphics[width=\geditunnaturalincompletecell,height=\geditunnaturalincompletecelltall,trim={0.000000bp 0.000000bp 0.000000bp 0.000000bp},clip]{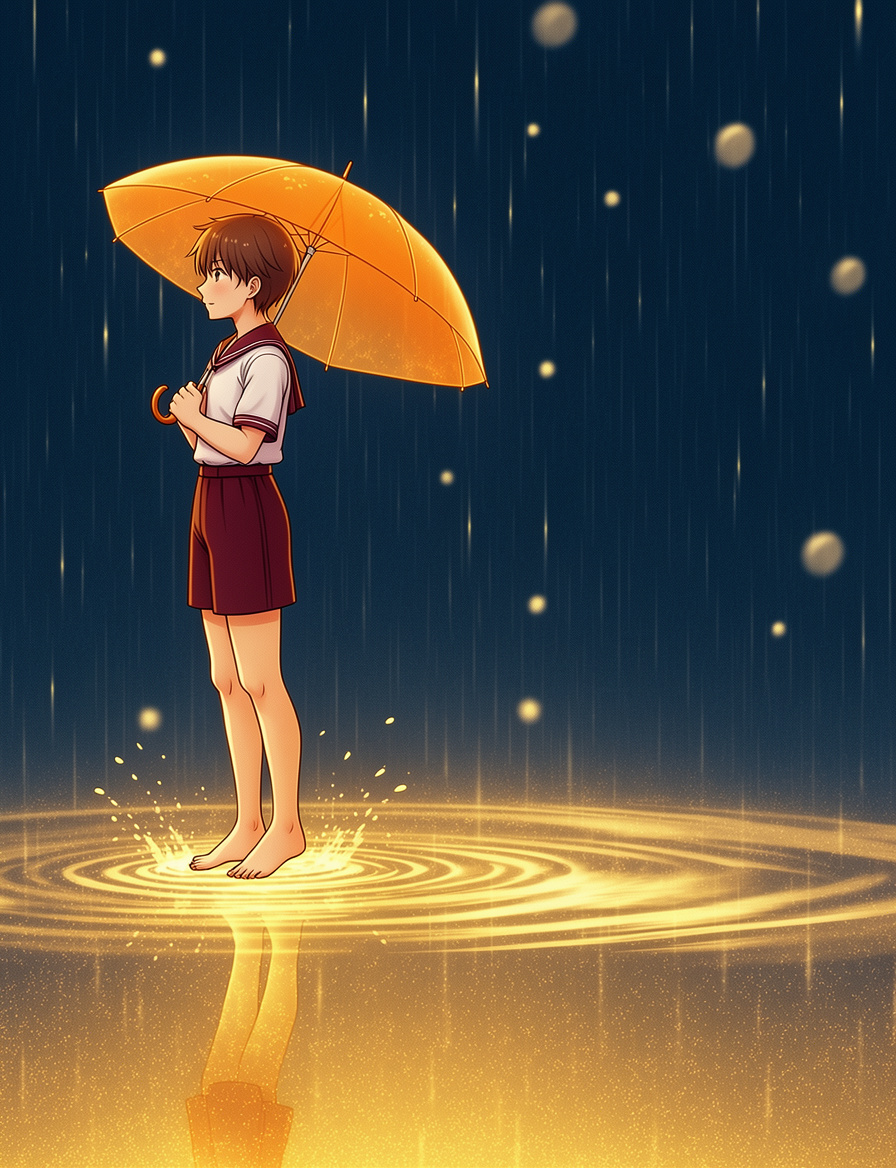}}\endgroup & \begingroup\setlength{\geditunnaturalincompletecelltall}{1.303571\geditunnaturalincompletecell}\parbox[c][\geditunnaturalincompletecelltall][c]{\geditunnaturalincompletecell}{\includegraphics[width=\geditunnaturalincompletecell,height=\geditunnaturalincompletecelltall,trim={0.000000bp 0.000000bp 0.000000bp 0.000000bp},clip]{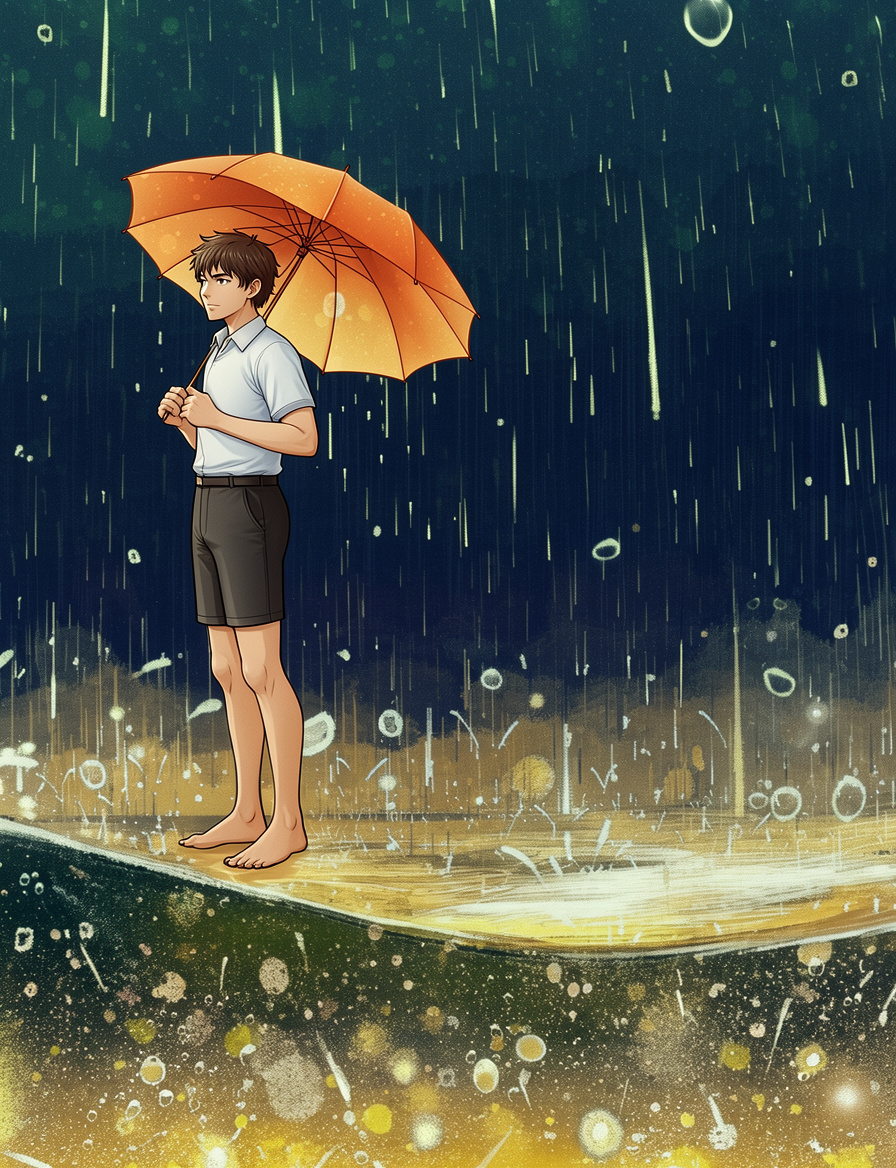}}\endgroup \\
\begingroup\setlength{\geditunnaturalincompletecelltall}{1.303571\geditunnaturalincompletecell}\geditunnaturalincompletetxtinstrtall{\fontsize{5.4}{6.2}\selectfont Change the background of this photo to a traditional Chinese landscape painting.}\endgroup & \begingroup\setlength{\geditunnaturalincompletecelltall}{1.303571\geditunnaturalincompletecell}\parbox[c][\geditunnaturalincompletecelltall][c]{\geditunnaturalincompletecell}{\includegraphics[width=\geditunnaturalincompletecell,height=\geditunnaturalincompletecelltall,trim={0.000000bp 0.000000bp 0.000000bp 0.000000bp},clip]{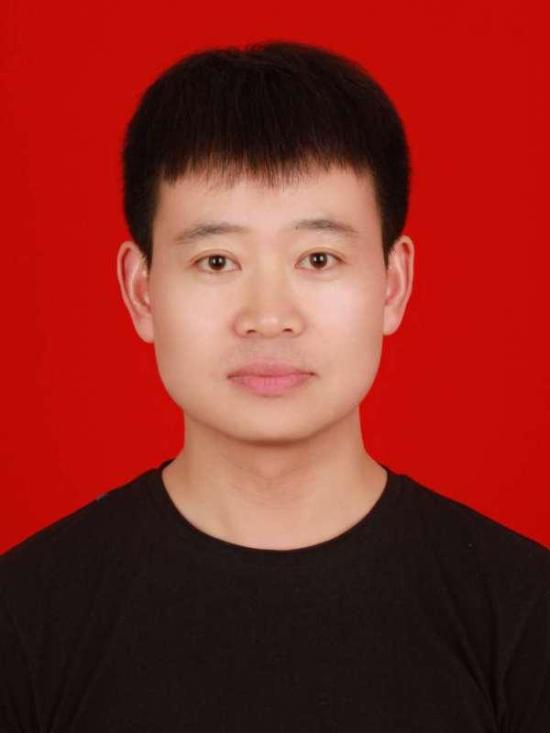}}\endgroup & \begingroup\setlength{\geditunnaturalincompletecelltall}{1.303571\geditunnaturalincompletecell}\parbox[c][\geditunnaturalincompletecelltall][c]{\geditunnaturalincompletecell}{\includegraphics[width=\geditunnaturalincompletecell,height=\geditunnaturalincompletecelltall,trim={0.000000bp 0.000000bp 0.000000bp 0.000000bp},clip]{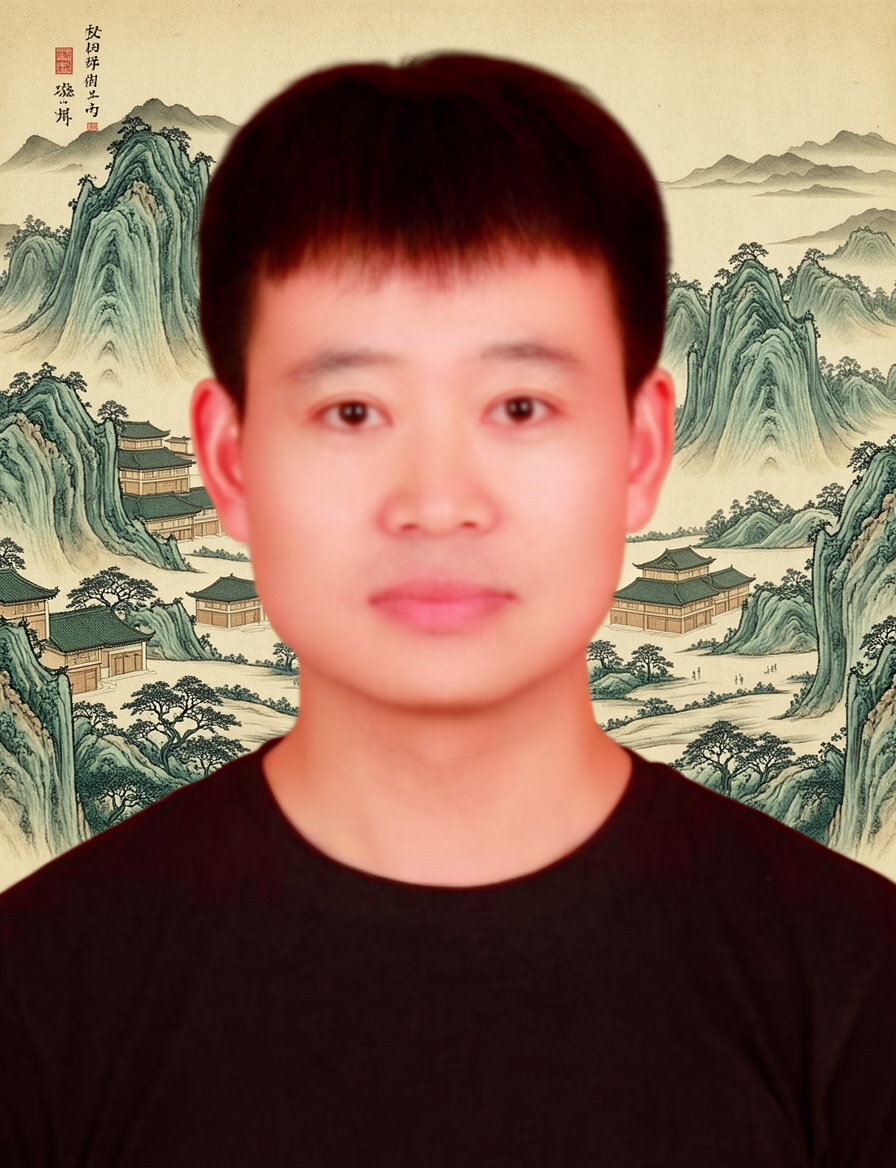}}\endgroup & \begingroup\setlength{\geditunnaturalincompletecelltall}{1.303571\geditunnaturalincompletecell}\geditunnaturalincompletetxtcaptall{\fontsize{5.2}{5.9}\selectfont A centered, eye-level head-and-shoulders portrait of a young East Asian man with short, straight black hair and a neat fringe, looking directly at the camera with a\,\ldots{}}\endgroup & \begingroup\setlength{\geditunnaturalincompletecelltall}{1.303571\geditunnaturalincompletecell}\parbox[c][\geditunnaturalincompletecelltall][c]{\geditunnaturalincompletecell}{\includegraphics[width=\geditunnaturalincompletecell,height=\geditunnaturalincompletecelltall,trim={0.000000bp 0.000000bp 0.000000bp 0.000000bp},clip]{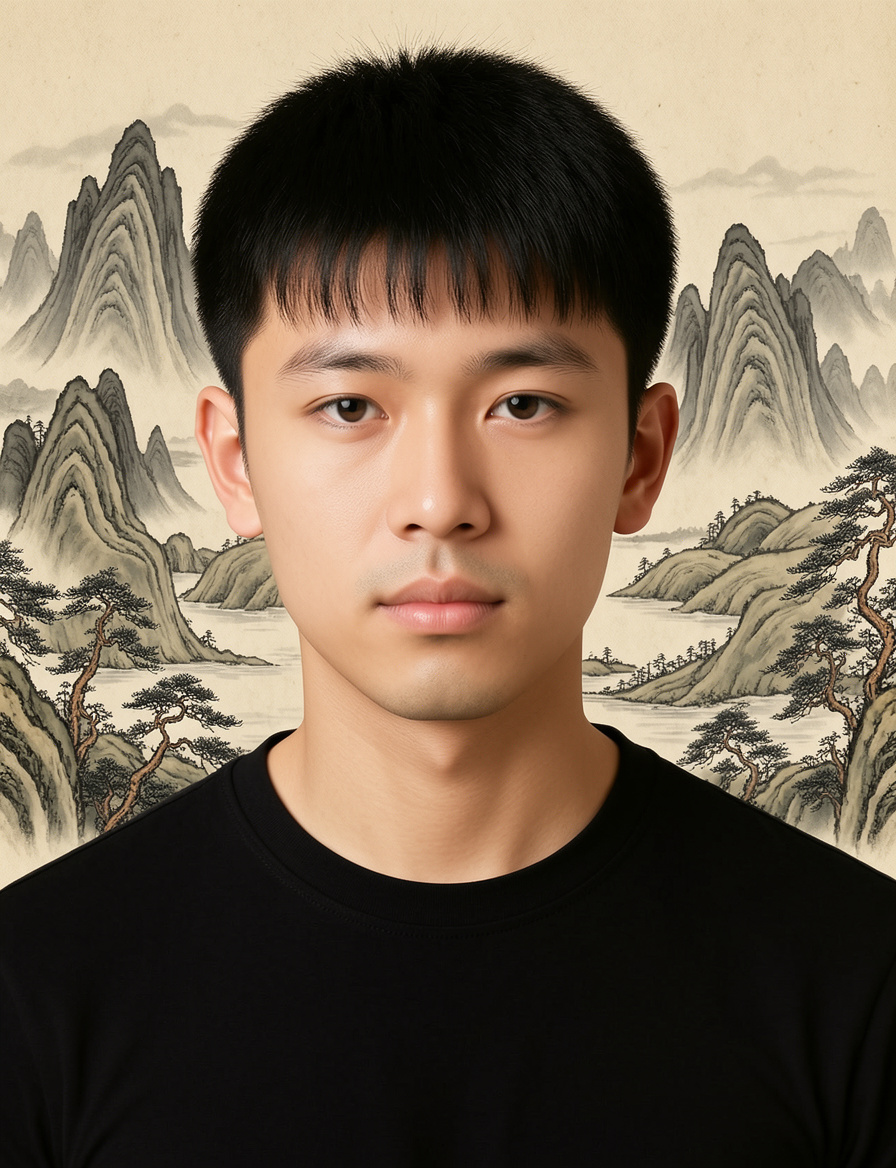}}\endgroup & \begingroup\setlength{\geditunnaturalincompletecelltall}{1.303571\geditunnaturalincompletecell}\parbox[c][\geditunnaturalincompletecelltall][c]{\geditunnaturalincompletecell}{\includegraphics[width=\geditunnaturalincompletecell,height=\geditunnaturalincompletecelltall,trim={0.000000bp 0.000000bp 0.000000bp 0.000000bp},clip]{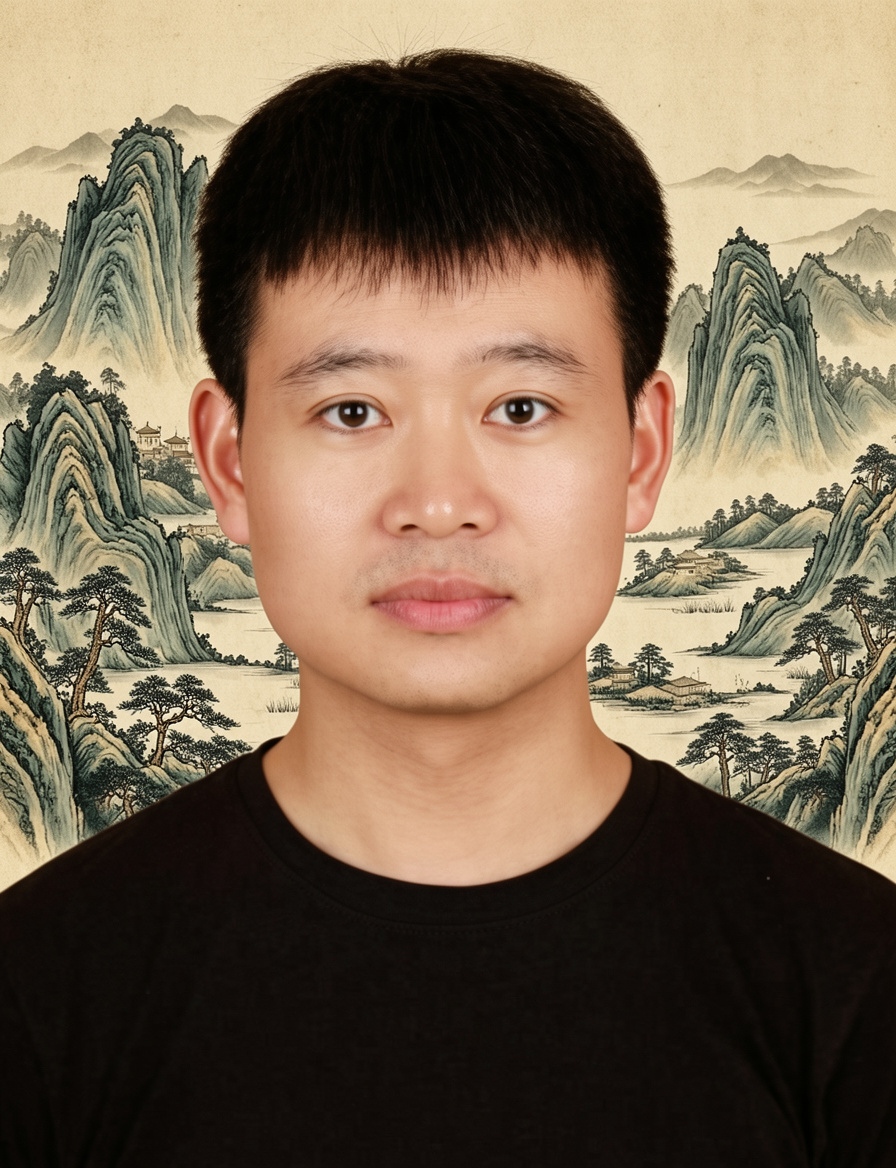}}\endgroup \\
\begingroup\setlength{\geditunnaturalincompletecelltall}{1.150000\geditunnaturalincompletecell}\geditunnaturalincompletetxtinstrtall{\fontsize{6}{7}\selectfont Make the sheep from jade.}\endgroup & \begingroup\setlength{\geditunnaturalincompletecelltall}{1.150000\geditunnaturalincompletecell}\parbox[c][\geditunnaturalincompletecelltall][c]{\geditunnaturalincompletecell}{\includegraphics[width=\geditunnaturalincompletecell,height=\geditunnaturalincompletecelltall,trim={0.000000bp 0.000000bp 0.000000bp 0.000000bp},clip]{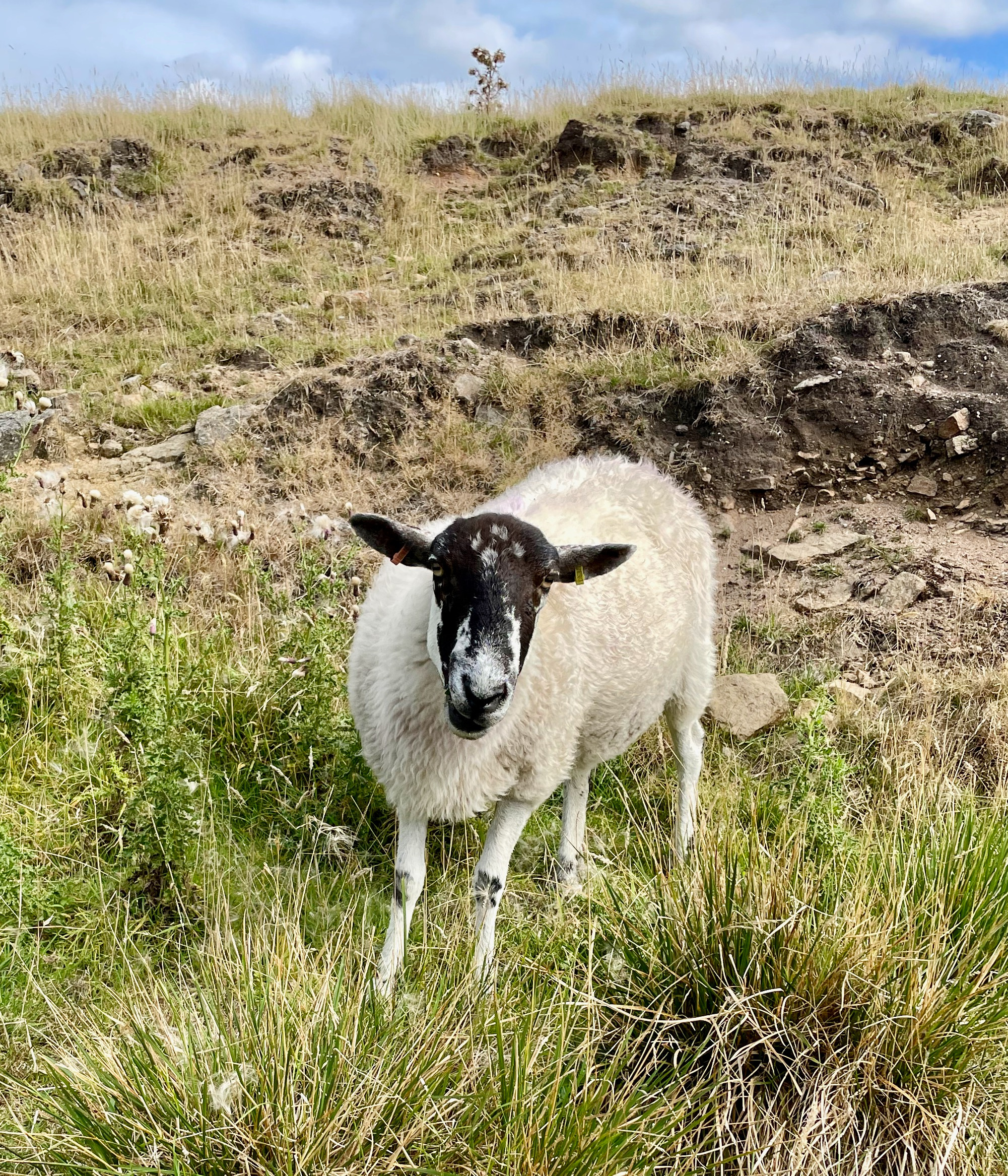}}\endgroup & \begingroup\setlength{\geditunnaturalincompletecelltall}{1.150000\geditunnaturalincompletecell}\parbox[c][\geditunnaturalincompletecelltall][c]{\geditunnaturalincompletecell}{\includegraphics[width=\geditunnaturalincompletecell,height=\geditunnaturalincompletecelltall,trim={0.000000bp 0.000000bp 0.000000bp 0.000000bp},clip]{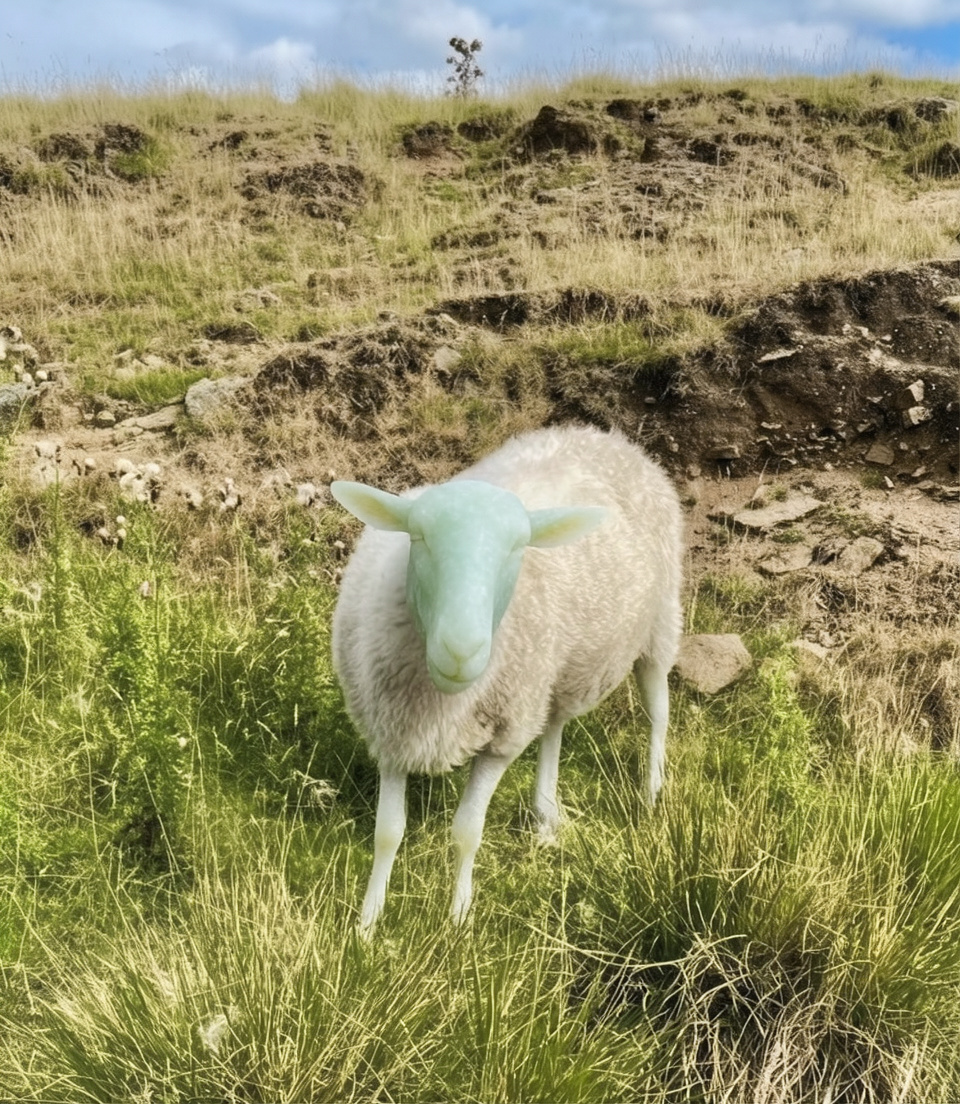}}\endgroup & \begingroup\setlength{\geditunnaturalincompletecelltall}{1.150000\geditunnaturalincompletecell}\geditunnaturalincompletetxtcaptall{\fontsize{5.2}{5.9}\selectfont A highly detailed, realistic photograph of an intricately carved jade sheep sculpture standing on a grassy hillside. The sculpture is the central focus, positioned\,\ldots{}}\endgroup & \begingroup\setlength{\geditunnaturalincompletecelltall}{1.150000\geditunnaturalincompletecell}\parbox[c][\geditunnaturalincompletecelltall][c]{\geditunnaturalincompletecell}{\includegraphics[width=\geditunnaturalincompletecell,height=\geditunnaturalincompletecelltall,trim={0.000000bp 0.000000bp 0.000000bp 0.000000bp},clip]{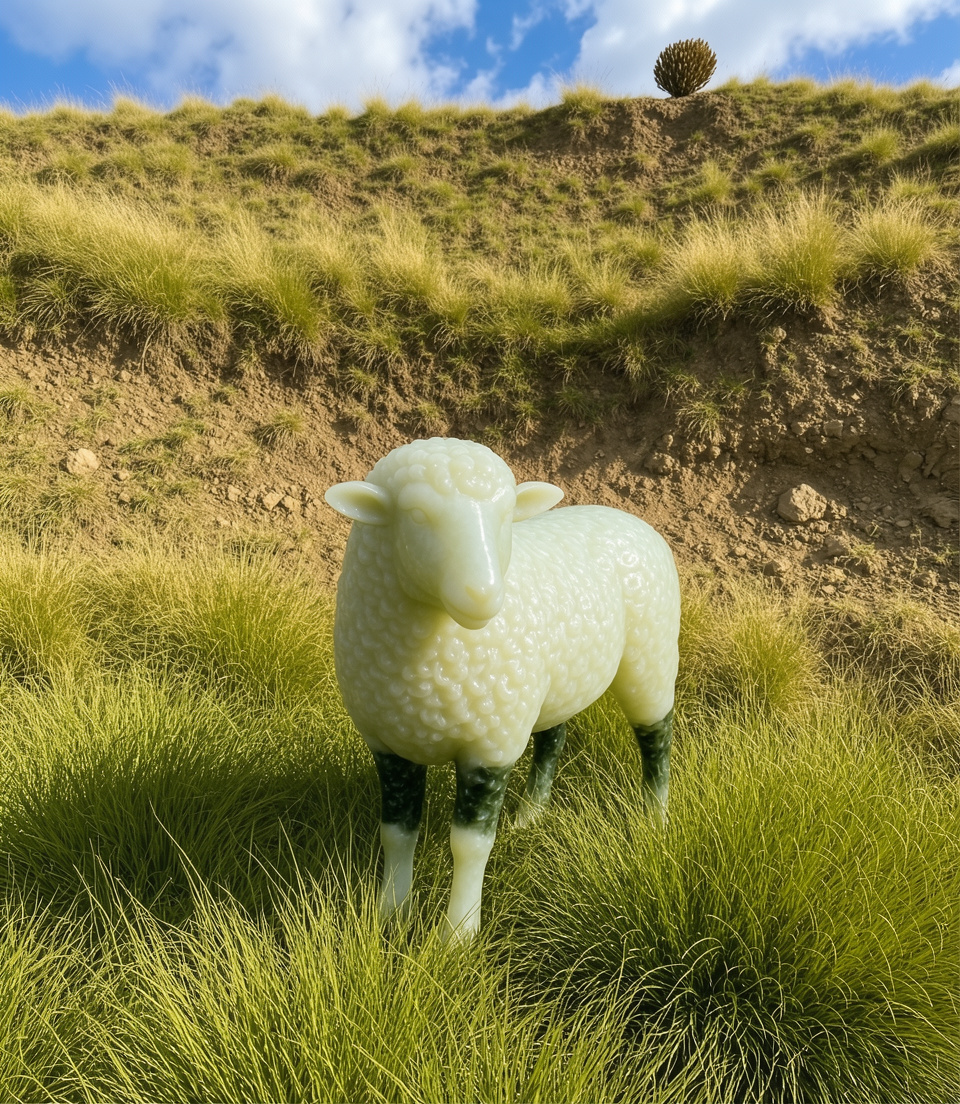}}\endgroup & \begingroup\setlength{\geditunnaturalincompletecelltall}{1.150000\geditunnaturalincompletecell}\parbox[c][\geditunnaturalincompletecelltall][c]{\geditunnaturalincompletecell}{\includegraphics[width=\geditunnaturalincompletecell,height=\geditunnaturalincompletecelltall,trim={0.000000bp 0.000000bp 0.000000bp 0.000000bp},clip]{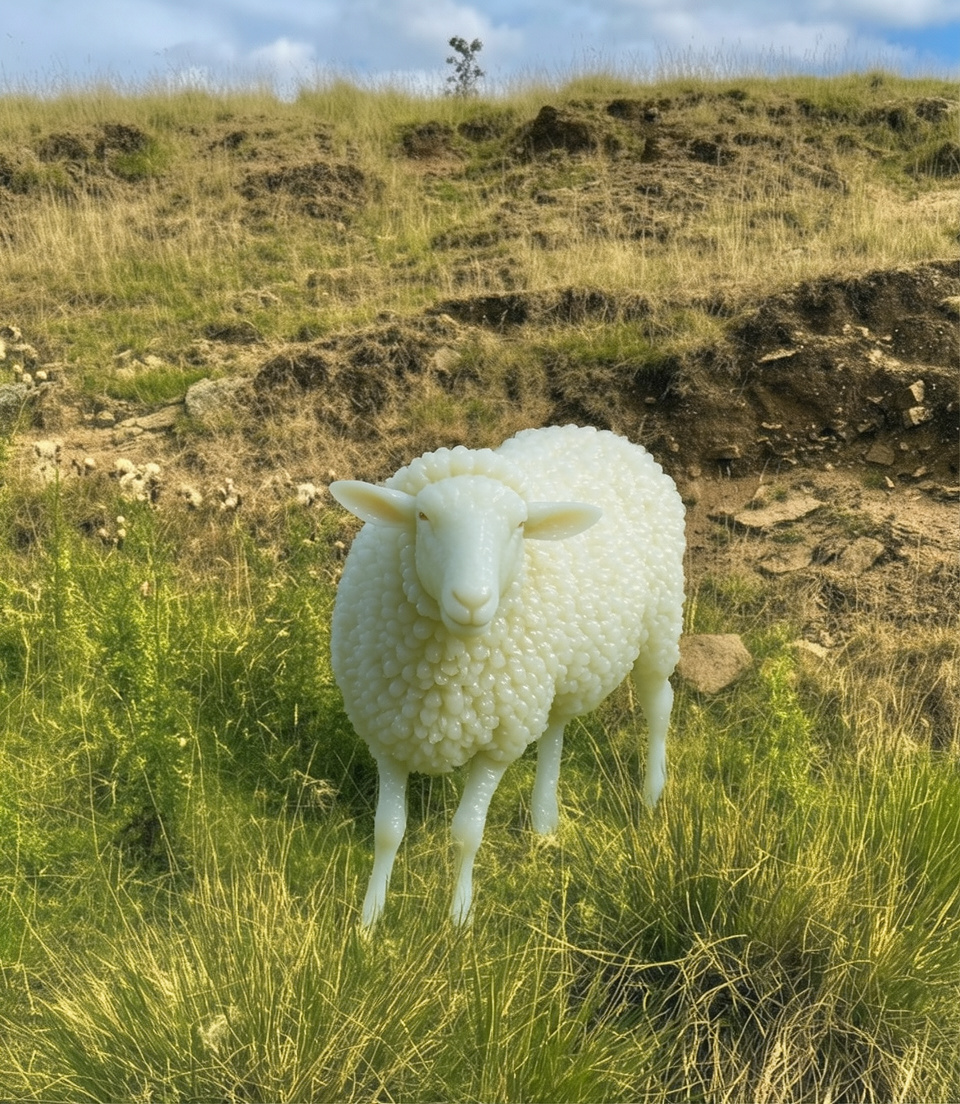}}\endgroup \\
\end{tabular}
\endgroup
\caption{\textbf{Unnatural and incomplete edits (FLUX2-Klein~4B, GEdit).} The first four rows illustrate unnatural appearance or artifacts under pure editing; the final row illustrates an incomplete transformation. Columns as in \cref{fig:appx-gedit-missed}.}
\label{fig:appx-gedit-unnatural-incomplete}
\end{figure}
\FloatBarrier

\clearpage

\section{Extended editing results}
\label{app:results}
The following tables report benchmark and conditioning sweeps for FLUX2-Klein~4B, together with ImgEdit category breakdowns for all three models. A single switch $\mathrm{E}\to\mathrm{T2I}$ retains T2I conditioning from step $k$ onward; $\mathrm{E}\to\mathrm{T2I}\to\mathrm{E}$ uses it only on $[k_1,k_2)$. Selective rows apply $[10,16)$ according to the attention rule, with thresholds in \cref{app:hparams}. Conditioning controls are defined in \cref{sec:fixed-duet}.

\paragraph{Reading the tables.}
Green and red shading indicate improvement and deterioration relative to baseline, scaled separately for each metric within a comparison block. Higher is better except for MSE. GPT-4.1 and Qwen3.5 GEdit results use different judges and are presented separately.

\begin{table}[!ht]
\centering
\caption{\textbf{GIE-Bench results (FLUX2-Klein~4B, GPT-4o correctness judge).} Functional correctness by edit category and source-preservation metrics for fixed schedules, the improved-instruction control, and selective application.}
\label{tab:appx-giebench-fc}
\vspace{0.75em}
{\footnotesize
\setlength{\tabcolsep}{3pt}
\resizebox{\linewidth}{!}{
\begin{tabular}{ll|cccccccccc|ccccc}
\toprule
 & & \multicolumn{10}{c|}{\textbf{Functional correctness}} & \multicolumn{5}{c}{\textbf{Preservation}} \\
Schedule & Interval & \shortstack{Add\\Object} & \shortstack{Attribute\\Change} & \shortstack{Color\\Change} & \shortstack{Layout\\Modif.} & \shortstack{Object\\Replace} & \shortstack{Remove\\Object} & \shortstack{Scene /\\Bkgd\\Change} & \shortstack{Size\\Change} & \shortstack{Textual\\Edit} & Overall & SSIM$\uparrow$ & CLIP$\uparrow$ & \shortstack{Unmasked\\CLIP$\uparrow$} & PSNR$\uparrow$ & MSE$\downarrow$ \\
\midrule
Baseline & --- & 91.67 & 88.33 & 90.76 & 44.17 & 93.33 & 68.33 & 95.83 & 20.83 & 67.50 & 73.40 & 0.805 & 0.97 & 0.913 & 20.88 & 1882.32 \\
\midrule
$\mathrm{E}\to\mathrm{E}^{*}\to\mathrm{E}$ & $[0,10)$ & 91.67 & \cellcolor[RGB]{224,246,228}93.33 & \cellcolor[RGB]{198,239,206}95.80 & \cellcolor[RGB]{255,250,251}43.33 & 93.33 & \cellcolor[RGB]{243,251,244}71.43 & \cellcolor[RGB]{226,247,230}96.67 & \cellcolor[RGB]{233,249,236}27.5 & \cellcolor[RGB]{255,199,206}61.67 & \cellcolor[RGB]{245,252,246}74.95 & \cellcolor[RGB]{255,254,254}0.801 & 0.970 & \cellcolor[RGB]{255,252,253}0.911 & \cellcolor[RGB]{255,252,253}20.32 & \cellcolor[RGB]{255,253,253}2117.10 \\
$\mathrm{E}\to\mathrm{E}^{*}\to\mathrm{E}$ & $[25,35)$ & \cellcolor[RGB]{255,199,206}89.52 & \cellcolor[RGB]{255,199,206}88.00 & \cellcolor[RGB]{231,248,235}92.86 & \cellcolor[RGB]{250,254,251}45.45 & \cellcolor[RGB]{238,250,240}94.29 & \cellcolor[RGB]{255,199,206}66.34 & \cellcolor[RGB]{255,199,206}93.48 & \cellcolor[RGB]{246,252,247}23.58 & \cellcolor[RGB]{254,255,255}67.65 & \cellcolor[RGB]{255,199,206}73.13 & \cellcolor[RGB]{198,239,206}0.81 & 0.97 & \cellcolor[RGB]{198,239,206}0.914 & \cellcolor[RGB]{255,254,254}20.76 & \cellcolor[RGB]{255,255,255}1897.70 \\
$\mathrm{E}\to\mathrm{E}^{*}\to\mathrm{E}$ & $[35,45)$ & \cellcolor[RGB]{255,233,236}90.83 & \cellcolor[RGB]{250,254,251}89.17 & \cellcolor[RGB]{246,252,247}91.60 & \cellcolor[RGB]{239,251,241}48.33 & \cellcolor[RGB]{255,252,252}93.28 & \cellcolor[RGB]{255,232,235}67.50 & 95.83 & \cellcolor[RGB]{241,251,243}25.00 & \cellcolor[RGB]{255,223,227}64.17 & \cellcolor[RGB]{251,254,252}73.93 & \cellcolor[RGB]{198,239,206}0.81 & 0.97 & \cellcolor[RGB]{255,255,255}0.913 & \cellcolor[RGB]{255,255,255}20.84 & \cellcolor[RGB]{198,239,206}1878.95 \\
E$^{*}$ & $[0,50)$ & \cellcolor[RGB]{255,233,236}90.83 & \cellcolor[RGB]{224,246,228}93.33 & \cellcolor[RGB]{217,244,222}94.12 & \cellcolor[RGB]{255,240,242}41.67 & \cellcolor[RGB]{225,246,229}95 & \cellcolor[RGB]{246,252,247}70.59 & \cellcolor[RGB]{226,247,230}96.67 & \cellcolor[RGB]{222,246,227}30.83 & \cellcolor[RGB]{246,252,247}70 & \cellcolor[RGB]{238,250,241}75.88 & \cellcolor[RGB]{255,254,254}0.799 & 0.970 & \cellcolor[RGB]{255,251,252}0.910 & \cellcolor[RGB]{255,252,252}20.25 & \cellcolor[RGB]{255,253,253}2174.08 \\
\midrule
E$\to$T2I & $k=5$ & \cellcolor[RGB]{198,239,206}95.00 & \cellcolor[RGB]{208,242,215}95.83 & \cellcolor[RGB]{236,250,239}92.44 & \cellcolor[RGB]{198,239,206}59.17 & \cellcolor[RGB]{240,251,242}94.17 & \cellcolor[RGB]{198,239,206}82.50 & \cellcolor[RGB]{226,247,230}96.67 & \cellcolor[RGB]{198,239,206}38.33 & \cellcolor[RGB]{198,239,206}82.50 & \cellcolor[RGB]{198,239,206}81.84 & \cellcolor[RGB]{255,199,206}0.54 & \cellcolor[RGB]{255,199,206}0.93 & \cellcolor[RGB]{255,199,206}0.861 & \cellcolor[RGB]{255,199,206}9.21 & \cellcolor[RGB]{255,199,206}8892.47 \\
E$\to$T2I & $k=15$ & \cellcolor[RGB]{227,247,231}93.33 & \cellcolor[RGB]{239,251,242}90.83 & \cellcolor[RGB]{217,244,222}94.12 & \cellcolor[RGB]{255,199,206}35.00 & \cellcolor[RGB]{241,251,243}94.12 & \cellcolor[RGB]{211,243,218}79.17 & 95.83 & 20.83 & \cellcolor[RGB]{217,244,222}77.50 & \cellcolor[RGB]{240,251,242}75.60 & \cellcolor[RGB]{255,220,224}0.64 & \cellcolor[RGB]{255,227,230}0.95 & \cellcolor[RGB]{255,218,223}0.879 & \cellcolor[RGB]{255,202,208}9.78 & \cellcolor[RGB]{255,205,211}8177.44 \\
$\mathrm{E}\to\mathrm{T2I}\to\mathrm{E}$ & $[10,16)$ & \cellcolor[RGB]{212,243,218}94.17 & \cellcolor[RGB]{219,245,224}94.17 & \cellcolor[RGB]{208,242,214}94.96 & \cellcolor[RGB]{252,254,252}45.00 & \cellcolor[RGB]{225,246,229}95.00 & \cellcolor[RGB]{245,252,246}70.83 & \cellcolor[RGB]{198,239,206}97.50 & \cellcolor[RGB]{233,249,236}27.50 & \cellcolor[RGB]{249,253,250}69.17 & \cellcolor[RGB]{234,249,237}76.46 & \cellcolor[RGB]{255,250,250}0.78 & 0.97 & \cellcolor[RGB]{255,247,248}0.906 & \cellcolor[RGB]{255,243,245}18.45 & \cellcolor[RGB]{255,250,250}2569.91 \\
$\mathrm{E}\to\mathrm{T2I}\to\mathrm{E}$ & $[10,20)$ & \cellcolor[RGB]{227,247,231}93.33 & \cellcolor[RGB]{198,239,206}97.50 & \cellcolor[RGB]{226,247,230}93.28 & \cellcolor[RGB]{255,224,228}39.17 & \cellcolor[RGB]{255,199,206}92.50 & \cellcolor[RGB]{205,241,212}80.83 & \cellcolor[RGB]{255,215,220}94.17 & \cellcolor[RGB]{220,245,225}31.67 & \cellcolor[RGB]{252,254,252}68.33 & \cellcolor[RGB]{232,249,236}76.74 & \cellcolor[RGB]{255,229,232}0.68 & \cellcolor[RGB]{255,227,230}0.95 & \cellcolor[RGB]{255,229,232}0.889 & \cellcolor[RGB]{255,215,220}12.54 & \cellcolor[RGB]{255,226,229}5568.90 \\
$\mathrm{E}\to\mathrm{T2I}\to\mathrm{E}$ & $[20,35)$ & \cellcolor[RGB]{227,247,231}93.33 & \cellcolor[RGB]{250,254,251}89.17 & \cellcolor[RGB]{236,250,239}92.44 & \cellcolor[RGB]{255,204,210}35.83 & \cellcolor[RGB]{225,246,229}95.00 & \cellcolor[RGB]{208,242,215}80.00 & 95.83 & \cellcolor[RGB]{247,253,248}23.33 & \cellcolor[RGB]{226,247,230}75.00 & \cellcolor[RGB]{241,251,243}75.53 & \cellcolor[RGB]{255,233,236}0.70 & \cellcolor[RGB]{255,241,243}0.96 & \cellcolor[RGB]{255,231,234}0.891 & \cellcolor[RGB]{255,207,213}10.95 & \cellcolor[RGB]{255,213,219}7090.06 \\
$\mathrm{E}\to\mathrm{T2I}\to\mathrm{E}$ & $[25,35)$ & \cellcolor[RGB]{255,253,253}91.59 & \cellcolor[RGB]{241,251,243}90.65 & \cellcolor[RGB]{235,249,238}92.55 & \cellcolor[RGB]{255,231,234}40.19 & \cellcolor[RGB]{228,248,232}94.79 & \cellcolor[RGB]{225,247,229}75.73 & \cellcolor[RGB]{255,243,245}95.33 & \cellcolor[RGB]{250,254,251}22.22 & \cellcolor[RGB]{237,250,239}72.28 & \cellcolor[RGB]{244,252,246}75.03 & \cellcolor[RGB]{255,243,245}0.75 & \cellcolor[RGB]{255,241,243}0.96 & \cellcolor[RGB]{255,240,242}0.900 & \cellcolor[RGB]{255,217,222}12.93 & \cellcolor[RGB]{255,226,230}5518.98 \\
$\mathrm{E}\to\mathrm{T2I}\to\mathrm{E}$ & $[35,45)$ & \cellcolor[RGB]{238,250,241}92.65 & \cellcolor[RGB]{254,255,254}88.52 & \cellcolor[RGB]{249,253,250}91.30 & \cellcolor[RGB]{255,228,231}39.71 & \cellcolor[RGB]{198,239,206}96.47 & \cellcolor[RGB]{220,245,225}77.11 & \cellcolor[RGB]{255,205,212}93.75 & \cellcolor[RGB]{244,252,246}24.14 & \cellcolor[RGB]{255,224,228}64.29 & \cellcolor[RGB]{255,199,206}73.13 & \cellcolor[RGB]{255,250,250}0.78 & 0.97 & \cellcolor[RGB]{255,246,247}0.905 & \cellcolor[RGB]{255,222,226}13.96 & \cellcolor[RGB]{255,230,233}5005.39 \\
Selective & $[10,16)$ & \cellcolor[RGB]{212,243,218}94.17 & \cellcolor[RGB]{224,246,228}93.33 & \cellcolor[RGB]{208,242,214}94.96 & \cellcolor[RGB]{246,252,247}46.67 & \cellcolor[RGB]{240,251,242}94.17 & \cellcolor[RGB]{232,248,235}74.17 & \cellcolor[RGB]{226,247,230}96.67 & \cellcolor[RGB]{236,250,239}26.67 & \cellcolor[RGB]{233,249,236}73.33 & \cellcolor[RGB]{230,248,233}77.11 & \cellcolor[RGB]{255,252,252}0.79 & 0.97 & \cellcolor[RGB]{255,249,250}0.908 & \cellcolor[RGB]{255,246,247}19.00 & \cellcolor[RGB]{255,251,252}2370.95 \\
\bottomrule
\end{tabular}
}
}
\end{table}

\begin{table}[!ht]
\centering
\noindent
\begin{minipage}[t]{0.38\linewidth}
\centering
\captionof{table}{\textbf{MagicBrush CLIP-T-gen (FLUX2-Klein 4B).} Metric reported in \cref{tab:main}.}
\label{tab:appx-magicbrush}
\vspace{0.75em}
{\footnotesize
\setlength{\tabcolsep}{3pt}
\centering
\sbox0{\begin{tabular}{llc}
\toprule
Schedule & Interval & CLIP-T-gen \\
\midrule
Baseline & --- & 0.317 \\
\midrule
$\mathrm{E}\to\mathrm{E}^{*}\to\mathrm{E}$ & $[0,10)$ & 0.317 \\
E$^{*}$ & $[0,50)$ & 0.317 \\
\midrule
E$\to$T2I & $k=5$ & \cellcolor[RGB]{219,245,224}0.322 \\
E$\to$T2I & $k=10$ & \cellcolor[RGB]{205,241,212}0.324 \\
E$\to$T2I & $k=15$ & \cellcolor[RGB]{198,239,206}0.325 \\
E$\to$T2I & $k=25$ & \cellcolor[RGB]{226,247,230}0.321 \\
E$\to$T2I & $k=35$ & \cellcolor[RGB]{248,253,249}0.318 \\
E$\to$T2I & $k=40$ & \cellcolor[RGB]{248,253,249}0.318 \\
E$\to$T2I & $k=45$ & 0.317 \\
$\mathrm{E}\to\mathrm{T2I}\to\mathrm{E}$ & $[10,16)$ & \cellcolor[RGB]{219,245,224}0.322 \\
$\mathrm{E}\to\mathrm{T2I}\to\mathrm{E}$ & $[10,20)$ & \cellcolor[RGB]{205,241,212}0.324 \\
$\mathrm{E}\to\mathrm{T2I}\to\mathrm{E}$ & $[15,30)$ & \cellcolor[RGB]{198,239,206}0.325 \\
$\mathrm{E}\to\mathrm{T2I}\to\mathrm{E}$ & $[15,35)$ & \cellcolor[RGB]{198,239,206}0.325 \\
$\mathrm{E}\to\mathrm{T2I}\to\mathrm{E}$ & $[20,30)$ & \cellcolor[RGB]{219,245,224}0.322 \\
$\mathrm{E}\to\mathrm{T2I}\to\mathrm{E}$ & $[20,35)$ & \cellcolor[RGB]{212,243,218}0.323 \\
$\mathrm{E}\to\mathrm{T2I}\to\mathrm{E}$ & $[25,35)$ & \cellcolor[RGB]{226,247,230}0.321 \\
$\mathrm{E}\to\mathrm{T2I}\to\mathrm{E}$ & $[25,40)$ & \cellcolor[RGB]{226,247,230}0.321 \\
$\mathrm{E}\to\mathrm{T2I}\to\mathrm{E}$ & $[35,45)$ & \cellcolor[RGB]{248,253,249}0.318 \\
Selective & $[10,16)$ & \cellcolor[RGB]{241,251,243}0.319 \\
\bottomrule
\end{tabular}}
\ifdim\wd0>\linewidth\relax
  \resizebox{\linewidth}{!}{\usebox0}
\else
  \usebox0
\fi
}

\end{minipage}
\hfill
\begin{minipage}[t]{0.60\linewidth}
\centering
{\captionsetup{width=0.78\linewidth, margin={1.2em,0pt}}
\captionof{table}{\textbf{GEdit sweep (FLUX2-Klein 4B, GPT-4.1 judge).} Single-switch and return-to-edit sweeps, including early and full-trajectory improved-instruction controls.}
\label{tab:appx-gedit-gpt}
}
\vspace{0.75em}
{\footnotesize
\setlength{\tabcolsep}{3pt}
\centering
\sbox0{\begin{tabular}{llccc}
\toprule
Schedule & Interval & G\_SC & G\_PQ & G\_O \\
\midrule
Baseline & --- & 7.72 & 7.46 & 7.11 \\
\midrule
$\mathrm{E}\to\mathrm{E}^{*}\to\mathrm{E}$ & $[0,10)$ & \cellcolor[RGB]{199,239,207}8.16 & \cellcolor[RGB]{253,255,254}7.49 & \cellcolor[RGB]{211,243,217}7.54 \\
$\mathrm{E}\to\mathrm{E}^{*}\to\mathrm{E}$ & $[25,35)$ & \cellcolor[RGB]{249,253,250}7.77 & \cellcolor[RGB]{254,255,255}7.47 & \cellcolor[RGB]{250,254,251}7.16 \\
$\mathrm{E}\to\mathrm{E}^{*}\to\mathrm{E}$ & $[35,45)$ & \cellcolor[RGB]{254,255,254}7.73 & 7.46 & \cellcolor[RGB]{255,254,254}7.09 \\
E$^{*}$ & $[0,50)$ & \cellcolor[RGB]{198,239,206}8.17 & \cellcolor[RGB]{253,255,254}7.49 & \cellcolor[RGB]{211,243,217}7.54 \\
\midrule
E$\to$T2I & $k=5$ & \cellcolor[RGB]{255,199,206}5.31 & \cellcolor[RGB]{198,239,206}8.50 & \cellcolor[RGB]{255,199,206}6.15 \\
E$\to$T2I & $k=10$ & \cellcolor[RGB]{255,210,216}5.79 & \cellcolor[RGB]{201,240,208}8.45 & \cellcolor[RGB]{255,218,222}6.47 \\
E$\to$T2I & $k=15$ & \cellcolor[RGB]{255,224,228}6.39 & \cellcolor[RGB]{210,242,216}8.28 & \cellcolor[RGB]{255,232,235}6.72 \\
E$\to$T2I & $k=25$ & \cellcolor[RGB]{255,244,246}7.26 & \cellcolor[RGB]{219,245,224}8.12 & \cellcolor[RGB]{254,255,254}7.12 \\
E$\to$T2I & $k=35$ & \cellcolor[RGB]{255,247,248}7.36 & \cellcolor[RGB]{231,248,235}7.89 & \cellcolor[RGB]{255,250,250}7.02 \\
E$\to$T2I & $k=40$ & \cellcolor[RGB]{255,245,246}7.28 & \cellcolor[RGB]{240,251,242}7.73 & \cellcolor[RGB]{255,240,242}6.86 \\
E$\to$T2I & $k=45$ & \cellcolor[RGB]{255,242,244}7.18 & \cellcolor[RGB]{248,253,249}7.59 & \cellcolor[RGB]{255,232,235}6.71 \\
$\mathrm{E}\to\mathrm{T2I}\to\mathrm{E}$ & $[10,16)$ & \cellcolor[RGB]{209,242,216}8.08 & \cellcolor[RGB]{231,248,234}7.90 & \cellcolor[RGB]{198,239,206}7.67 \\
$\mathrm{E}\to\mathrm{T2I}\to\mathrm{E}$ & $[10,20)$ & \cellcolor[RGB]{255,251,251}7.53 & \cellcolor[RGB]{214,243,220}8.21 & \cellcolor[RGB]{212,243,218}7.53 \\
$\mathrm{E}\to\mathrm{T2I}\to\mathrm{E}$ & $[15,30)$ & \cellcolor[RGB]{255,231,234}6.70 & \cellcolor[RGB]{211,243,217}8.27 & \cellcolor[RGB]{255,242,244}6.89 \\
$\mathrm{E}\to\mathrm{T2I}\to\mathrm{E}$ & $[15,35)$ & \cellcolor[RGB]{255,229,232}6.59 & \cellcolor[RGB]{211,243,217}8.26 & \cellcolor[RGB]{255,244,245}6.92 \\
$\mathrm{E}\to\mathrm{T2I}\to\mathrm{E}$ & $[20,30)$ & \cellcolor[RGB]{255,254,254}7.67 & \cellcolor[RGB]{225,247,229}8.01 & \cellcolor[RGB]{222,246,227}7.43 \\
$\mathrm{E}\to\mathrm{T2I}\to\mathrm{E}$ & $[20,35)$ & \cellcolor[RGB]{255,244,246}7.26 & \cellcolor[RGB]{216,244,221}8.18 & \cellcolor[RGB]{233,249,236}7.33 \\
$\mathrm{E}\to\mathrm{T2I}\to\mathrm{E}$ & $[25,35)$ & \cellcolor[RGB]{240,251,242}7.84 & \cellcolor[RGB]{225,247,229}8.01 & \cellcolor[RGB]{211,243,217}7.54 \\
$\mathrm{E}\to\mathrm{T2I}\to\mathrm{E}$ & $[25,40)$ & \cellcolor[RGB]{255,253,253}7.63 & \cellcolor[RGB]{220,245,225}8.09 & \cellcolor[RGB]{218,245,224}7.47 \\
$\mathrm{E}\to\mathrm{T2I}\to\mathrm{E}$ & $[35,45)$ & \cellcolor[RGB]{242,251,244}7.82 & \cellcolor[RGB]{231,248,235}7.89 & \cellcolor[RGB]{224,246,229}7.41 \\
Selective & $[10,16)$ & \cellcolor[RGB]{218,245,223}8.01 & \cellcolor[RGB]{243,252,245}7.67 & \cellcolor[RGB]{220,245,225}7.45 \\
\bottomrule
\end{tabular}}
\ifdim\wd0>0.74\linewidth\relax
  \resizebox{0.74\linewidth}{!}{\usebox0}
\else
  \usebox0
\fi
}

\end{minipage}
\smallskip
\end{table}

\clearpage
\begin{table}[p]
\begin{minipage}{\linewidth}
\centering
\caption{\textbf{GEdit interval-conditioning matrix (FLUX2-Klein 4B, Qwen3.5).}
Rows specify image conditioning before, during, and after $[k_1,k_2)$; columns specify the text schedule.
Here $p=p_{\mathrm{edit}}$, $p^{*}=p_{\mathrm{edit}}^{*}$, $c=c_{\mathrm{tgt}}$, and $x=x_{\mathrm{src}}$.
All interval variants share the same 12-interval grid.
SC/PQ/O denote semantic consistency, perceptual quality, and overall score.}
\label{tab:appx-gedit-qwen}
\fontsize{8}{9}\selectfont
\setlength{\tabcolsep}{2.5pt}
\renewcommand{\arraystretch}{0.95}
\begin{tabular}{@{}ll|c|c|c|c@{}}
\toprule
Image schedule & & Text $(p,p,p)$ & Text $(p,c,p)$ & Text $(c,c,c)$ & Text $(p,p^{*},p)$ \\
\midrule
$(x,x,x)$ & \begin{tabular}[t]{@{}l@{}}
Interval \\
\midrule
$[0,10)$ \\
$[0,15)$ \\
$[10,20)$ \\
$[15,30)$ \\
$[15,35)$ \\
$[20,30)$ \\
$[20,35)$ \\
$[25,35)$ \\
$[25,40)$ \\
$[35,45)$ \\
$[35,50)$ \\
$[40,50)$ \\
\end{tabular} & \begin{tabular}[t]{@{}c@{}}
\begin{tabular}[t]{*{3}{>{\centering\arraybackslash}p{17pt}}}
SC & PQ & O \\
\midrule
\cellcolor[RGB]{238,238,238}7.91 & \cellcolor[RGB]{238,238,238}7.65 & \cellcolor[RGB]{238,238,238}7.33 \\
\end{tabular} \\
\end{tabular} & \begin{tabular}[t]{@{}c@{}}
\begin{tabular}[t]{*{3}{>{\centering\arraybackslash}p{17pt}}}
SC & PQ & O \\
\midrule
\cellcolor[RGB]{255,254,255}7.83 & \cellcolor[RGB]{223,246,227}8.06 & \cellcolor[RGB]{239,251,241}7.46 \\
\cellcolor[RGB]{255,255,255}7.87 & \cellcolor[RGB]{221,245,226}8.09 & \cellcolor[RGB]{235,249,238}7.49 \\
\cellcolor[RGB]{234,249,237}8.13 & \cellcolor[RGB]{244,252,246}7.79 & \cellcolor[RGB]{231,248,234}7.53 \\
\cellcolor[RGB]{243,252,245}8.03 & \cellcolor[RGB]{247,253,248}7.75 & \cellcolor[RGB]{240,251,242}7.45 \\
\cellcolor[RGB]{234,249,237}8.13 & \cellcolor[RGB]{246,252,247}7.77 & \cellcolor[RGB]{229,248,233}7.54 \\
\cellcolor[RGB]{245,252,247}8.01 & \cellcolor[RGB]{251,254,252}7.70 & \cellcolor[RGB]{245,252,247}7.41 \\
\cellcolor[RGB]{244,252,246}8.02 & \cellcolor[RGB]{253,255,254}7.67 & \cellcolor[RGB]{246,253,248}7.40 \\
\cellcolor[RGB]{244,252,246}8.02 & \cellcolor[RGB]{255,250,251}7.64 & \cellcolor[RGB]{249,253,250}7.38 \\
\cellcolor[RGB]{248,253,249}7.98 & \cellcolor[RGB]{255,241,243}7.62 & \cellcolor[RGB]{250,254,251}7.37 \\
\cellcolor[RGB]{250,254,251}7.96 & \cellcolor[RGB]{252,254,252}7.69 & \cellcolor[RGB]{248,253,249}7.39 \\
\cellcolor[RGB]{250,254,251}7.96 & \cellcolor[RGB]{250,254,251}7.71 & \cellcolor[RGB]{248,253,249}7.39 \\
\cellcolor[RGB]{247,253,248}7.99 & \cellcolor[RGB]{255,255,255}7.65 & \cellcolor[RGB]{248,253,249}7.39 \\
\end{tabular} \\
\end{tabular} & \begin{tabular}[t]{@{}c@{}}
\begin{tabular}[t]{*{3}{>{\centering\arraybackslash}p{17pt}}}
SC & PQ & O \\
\midrule
\cellcolor[RGB]{255,253,253}7.57 & \cellcolor[RGB]{214,243,219}8.18 & \cellcolor[RGB]{255,255,255}7.28 \\
\end{tabular} \\
\end{tabular} & \begin{tabular}[t]{@{}c@{}}
\begin{tabular}[t]{*{3}{>{\centering\arraybackslash}p{17pt}}}
SC & PQ & O \\
\midrule
\cellcolor[RGB]{198,239,206}8.50 & \cellcolor[RGB]{252,254,253}7.69 & \cellcolor[RGB]{198,239,206}7.80 \\
\cellcolor[RGB]{200,239,207}8.48 & \cellcolor[RGB]{255,252,253}7.64 & \cellcolor[RGB]{202,240,209}7.76 \\
\cellcolor[RGB]{230,248,233}8.17 & \cellcolor[RGB]{253,255,254}7.67 & \cellcolor[RGB]{234,249,237}7.50 \\
\cellcolor[RGB]{241,251,243}8.05 & \cellcolor[RGB]{255,255,255}7.65 & \cellcolor[RGB]{243,252,244}7.43 \\
\cellcolor[RGB]{233,249,236}8.14 & \cellcolor[RGB]{255,246,247}7.63 & \cellcolor[RGB]{238,250,240}7.47 \\
\cellcolor[RGB]{240,251,242}8.07 & \cellcolor[RGB]{255,227,230}7.59 & \cellcolor[RGB]{249,253,250}7.38 \\
\cellcolor[RGB]{240,251,242}8.07 & \cellcolor[RGB]{255,227,230}7.59 & \cellcolor[RGB]{249,253,250}7.38 \\
\cellcolor[RGB]{243,252,245}8.03 & \cellcolor[RGB]{255,241,243}7.62 & \cellcolor[RGB]{250,254,251}7.37 \\
\cellcolor[RGB]{244,252,246}8.02 & \cellcolor[RGB]{255,246,247}7.63 & \cellcolor[RGB]{251,254,252}7.36 \\
\cellcolor[RGB]{242,251,244}8.04 & \cellcolor[RGB]{255,232,235}7.60 & \cellcolor[RGB]{250,254,251}7.37 \\
\cellcolor[RGB]{247,253,248}7.99 & \cellcolor[RGB]{255,246,247}7.63 & \cellcolor[RGB]{254,255,254}7.34 \\
\cellcolor[RGB]{246,253,248}8.00 & \cellcolor[RGB]{255,246,247}7.63 & \cellcolor[RGB]{250,254,251}7.37 \\
\end{tabular} \\
\end{tabular} \\
\midrule
$(x,\varnothing,x)$ & \begin{tabular}[t]{@{}l@{}}
Interval \\
\midrule
$[0,10)$ \\
$[0,15)$ \\
$[10,20)$ \\
$[15,30)$ \\
$[15,35)$ \\
$[20,30)$ \\
$[20,35)$ \\
$[25,35)$ \\
$[25,40)$ \\
$[35,45)$ \\
$[35,50)$ \\
$[40,50)$ \\
\end{tabular} & \begin{tabular}[t]{@{}c@{}}
\begin{tabular}[t]{*{3}{>{\centering\arraybackslash}p{17pt}}}
SC & PQ & O \\
\midrule
\cellcolor[RGB]{255,206,213}1.15 & \cellcolor[RGB]{255,224,228}7.58 & \cellcolor[RGB]{255,208,214}1.36 \\
\cellcolor[RGB]{255,201,208}0.39 & \cellcolor[RGB]{248,253,249}7.74 & \cellcolor[RGB]{255,202,208}0.51 \\
\cellcolor[RGB]{255,236,238}5.21 & \cellcolor[RGB]{243,251,244}7.81 & \cellcolor[RGB]{255,241,242}5.47 \\
\cellcolor[RGB]{255,234,236}4.94 & \cellcolor[RGB]{234,249,237}7.92 & \cellcolor[RGB]{255,240,242}5.46 \\
\cellcolor[RGB]{255,230,233}4.46 & \cellcolor[RGB]{230,248,234}7.97 & \cellcolor[RGB]{255,237,239}4.99 \\
\cellcolor[RGB]{255,251,252}7.36 & \cellcolor[RGB]{239,251,242}7.85 & \cellcolor[RGB]{255,254,254}7.19 \\
\cellcolor[RGB]{255,244,245}6.31 & \cellcolor[RGB]{239,251,242}7.85 & \cellcolor[RGB]{255,249,249}6.51 \\
\cellcolor[RGB]{255,253,253}7.61 & \cellcolor[RGB]{247,253,248}7.75 & \cellcolor[RGB]{255,255,255}7.32 \\
\cellcolor[RGB]{255,249,250}7.13 & \cellcolor[RGB]{245,252,246}7.78 & \cellcolor[RGB]{255,253,253}7.02 \\
\cellcolor[RGB]{255,254,254}7.78 & \cellcolor[RGB]{252,254,252}7.69 & \cellcolor[RGB]{255,255,255}7.33 \\
\cellcolor[RGB]{255,254,254}7.74 & \cellcolor[RGB]{251,254,252}7.70 & \cellcolor[RGB]{255,255,255}7.32 \\
\cellcolor[RGB]{255,255,255}7.86 & \cellcolor[RGB]{255,246,247}7.63 & \cellcolor[RGB]{254,255,254}7.34 \\
\end{tabular} \\
\end{tabular} & \begin{tabular}[t]{@{}c@{}}
\begin{tabular}[t]{*{3}{>{\centering\arraybackslash}p{17pt}}}
SC & PQ & O \\
\midrule
\cellcolor[RGB]{255,250,250}7.17 & \cellcolor[RGB]{209,242,215}8.24 & \cellcolor[RGB]{255,254,255}7.26 \\
\cellcolor[RGB]{255,247,248}6.74 & \cellcolor[RGB]{207,241,213}8.27 & \cellcolor[RGB]{255,253,253}7.04 \\
\cellcolor[RGB]{255,255,255}7.88 & \cellcolor[RGB]{224,246,228}8.05 & \cellcolor[RGB]{213,243,219}7.67 \\
\cellcolor[RGB]{255,252,252}7.48 & \cellcolor[RGB]{223,246,227}8.06 & \cellcolor[RGB]{240,251,242}7.45 \\
\cellcolor[RGB]{255,251,251}7.32 & \cellcolor[RGB]{219,245,224}8.11 & \cellcolor[RGB]{250,254,251}7.37 \\
\cellcolor[RGB]{242,251,244}8.04 & \cellcolor[RGB]{233,249,236}7.93 & \cellcolor[RGB]{211,243,217}7.69 \\
\cellcolor[RGB]{255,254,254}7.79 & \cellcolor[RGB]{228,247,232}8.00 & \cellcolor[RGB]{226,247,230}7.57 \\
\cellcolor[RGB]{240,251,242}8.07 & \cellcolor[RGB]{238,250,240}7.87 & \cellcolor[RGB]{218,245,223}7.63 \\
\cellcolor[RGB]{252,254,253}7.94 & \cellcolor[RGB]{232,248,235}7.95 & \cellcolor[RGB]{221,245,226}7.61 \\
\cellcolor[RGB]{247,253,248}7.99 & \cellcolor[RGB]{246,253,248}7.76 & \cellcolor[RGB]{234,249,237}7.50 \\
\cellcolor[RGB]{248,253,249}7.98 & \cellcolor[RGB]{244,252,246}7.79 & \cellcolor[RGB]{237,250,239}7.48 \\
\cellcolor[RGB]{245,252,247}8.01 & \cellcolor[RGB]{249,253,250}7.73 & \cellcolor[RGB]{237,250,239}7.48 \\
\end{tabular} \\
\end{tabular} & \begin{tabular}[t]{@{}c@{}}
\begin{tabular}[t]{*{3}{>{\centering\arraybackslash}p{17pt}}}
SC & PQ & O \\
\midrule
\cellcolor[RGB]{255,248,249}6.95 & \cellcolor[RGB]{202,240,209}8.33 & \cellcolor[RGB]{255,254,254}7.20 \\
\cellcolor[RGB]{255,246,247}6.59 & \cellcolor[RGB]{203,240,211}8.31 & \cellcolor[RGB]{255,252,252}6.94 \\
\cellcolor[RGB]{255,251,252}7.42 & \cellcolor[RGB]{202,240,209}8.33 & \cellcolor[RGB]{230,248,234}7.53 \\
\cellcolor[RGB]{255,251,251}7.29 & \cellcolor[RGB]{207,242,214}8.26 & \cellcolor[RGB]{247,253,248}7.39 \\
\cellcolor[RGB]{255,250,250}7.16 & \cellcolor[RGB]{207,242,214}8.26 & \cellcolor[RGB]{255,255,255}7.33 \\
\cellcolor[RGB]{255,252,253}7.51 & \cellcolor[RGB]{207,242,214}8.26 & \cellcolor[RGB]{239,250,241}7.46 \\
\cellcolor[RGB]{255,251,251}7.34 & \cellcolor[RGB]{209,242,216}8.24 & \cellcolor[RGB]{252,254,252}7.36 \\
\cellcolor[RGB]{255,252,253}7.54 & \cellcolor[RGB]{211,243,217}8.22 & \cellcolor[RGB]{247,253,248}7.40 \\
\cellcolor[RGB]{255,251,252}7.38 & \cellcolor[RGB]{210,242,216}8.22 & \cellcolor[RGB]{253,255,254}7.34 \\
\cellcolor[RGB]{255,252,253}7.52 & \cellcolor[RGB]{214,243,219}8.18 & \cellcolor[RGB]{255,255,255}7.32 \\
\cellcolor[RGB]{255,252,253}7.52 & \cellcolor[RGB]{210,242,217}8.22 & \cellcolor[RGB]{251,254,251}7.36 \\
\cellcolor[RGB]{255,253,253}7.58 & \cellcolor[RGB]{216,244,221}8.15 & \cellcolor[RGB]{254,255,254}7.34 \\
\end{tabular} \\
\end{tabular} & \begin{tabular}[t]{c}---\end{tabular} \\
\bottomrule
\end{tabular}
\par\smallskip
\raggedright
Green/red denotes improvement/degradation relative to pure E (gray), with shared per-metric scales in \cref{tab:appx-gedit-qwen,tab:appx-gedit-qwen-single,tab:appx-gedit-qwen-summary}. Constant-conditioning results are independent of placement; --- marks an unavailable or untested combination.
\end{minipage}
\par\medskip
\begin{minipage}{\linewidth}
\centering
\caption{\textbf{GEdit single-switch conditioning matrix (FLUX2-Klein 4B, Qwen3.5).}
Rows specify image conditioning before/after $k$; columns specify the text schedule.
The reverse switch T2I$\to$E is shown on the right.
Notation and colors follow \cref{tab:appx-gedit-qwen}.}
\label{tab:appx-gedit-qwen-single}
\fontsize{8}{9}\selectfont
\setlength{\tabcolsep}{1.5pt}
\renewcommand{\arraystretch}{0.95}
\raggedright
\begin{minipage}[c]{0.79\linewidth}
\raggedright
\begin{tabular*}{\linewidth}{@{\extracolsep{\fill}}ll|c|c|c|c@{}}
\toprule
Image schedule & & Text $(p,p)$ & Text $(p,c)$ & Text $(c,c)$ & Text $(p,p^{*})$ \\
\midrule
$(x,x)$ & \begin{tabular}[t]{@{}l@{}}
$k$ \\
\midrule
$0$ \\
$5$ \\
$10$ \\
$15$ \\
$20$ \\
$25$ \\
$30$ \\
$35$ \\
$40$ \\
$45$ \\
\end{tabular} & \begin{tabular}[t]{@{}c@{}}
\begin{tabular}[t]{*{3}{>{\centering\arraybackslash}p{16pt}}}
SC & PQ & O \\
\midrule
\cellcolor[RGB]{238,238,238}7.91 & \cellcolor[RGB]{238,238,238}7.65 & \cellcolor[RGB]{238,238,238}7.33 \\
\end{tabular} \\
\end{tabular} & \begin{tabular}[t]{@{}c@{}}
\begin{tabular}[t]{*{3}{>{\centering\arraybackslash}p{16pt}}}
SC & PQ & O \\
\midrule
\cellcolor[RGB]{255,253,253}7.57 & \cellcolor[RGB]{214,243,219}8.18 & \cellcolor[RGB]{255,255,255}7.28 \\
\cellcolor[RGB]{245,252,247}8.01 & \cellcolor[RGB]{230,248,234}7.97 & \cellcolor[RGB]{234,249,237}7.50 \\
\cellcolor[RGB]{239,250,241}8.08 & \cellcolor[RGB]{234,249,237}7.92 & \cellcolor[RGB]{224,246,229}7.58 \\
\cellcolor[RGB]{233,249,236}8.14 & \cellcolor[RGB]{239,250,241}7.86 & \cellcolor[RGB]{228,247,232}7.55 \\
\cellcolor[RGB]{238,250,241}8.08 & \cellcolor[RGB]{249,253,250}7.72 & \cellcolor[RGB]{240,251,242}7.45 \\
\cellcolor[RGB]{241,251,243}8.05 & \cellcolor[RGB]{253,255,254}7.67 & \cellcolor[RGB]{245,252,247}7.41 \\
\cellcolor[RGB]{242,251,244}8.04 & \cellcolor[RGB]{253,254,253}7.68 & \cellcolor[RGB]{247,253,248}7.39 \\
\cellcolor[RGB]{250,254,251}7.96 & \cellcolor[RGB]{250,254,251}7.71 & \cellcolor[RGB]{248,253,249}7.39 \\
\cellcolor[RGB]{247,253,248}7.99 & \cellcolor[RGB]{255,255,255}7.65 & \cellcolor[RGB]{248,253,249}7.39 \\
\cellcolor[RGB]{255,255,255}7.87 & \cellcolor[RGB]{255,246,247}7.63 & \cellcolor[RGB]{255,255,255}7.29 \\
\end{tabular} \\
\end{tabular} & \begin{tabular}[t]{@{}c@{}}
\begin{tabular}[t]{*{3}{>{\centering\arraybackslash}p{16pt}}}
SC & PQ & O \\
\midrule
\cellcolor[RGB]{255,253,253}7.57 & \cellcolor[RGB]{214,243,219}8.18 & \cellcolor[RGB]{255,255,255}7.28 \\
\end{tabular} \\
\end{tabular} & \begin{tabular}[t]{@{}c@{}}
\begin{tabular}[t]{*{3}{>{\centering\arraybackslash}p{16pt}}}
SC & PQ & O \\
\midrule
\cellcolor[RGB]{204,241,211}8.44 & \cellcolor[RGB]{255,199,206}7.53 & \cellcolor[RGB]{207,242,214}7.72 \\
\cellcolor[RGB]{219,245,224}8.28 & \cellcolor[RGB]{254,255,254}7.66 & \cellcolor[RGB]{221,245,226}7.61 \\
\cellcolor[RGB]{236,250,238}8.11 & \cellcolor[RGB]{254,255,254}7.66 & \cellcolor[RGB]{235,250,238}7.49 \\
\cellcolor[RGB]{237,250,239}8.10 & \cellcolor[RGB]{255,241,243}7.62 & \cellcolor[RGB]{242,251,243}7.44 \\
\cellcolor[RGB]{244,252,246}8.02 & \cellcolor[RGB]{255,240,242}7.62 & \cellcolor[RGB]{252,254,253}7.35 \\
\cellcolor[RGB]{236,250,238}8.11 & \cellcolor[RGB]{255,232,235}7.60 & \cellcolor[RGB]{244,252,246}7.42 \\
\cellcolor[RGB]{243,252,245}8.03 & \cellcolor[RGB]{255,247,248}7.63 & \cellcolor[RGB]{250,254,251}7.37 \\
\cellcolor[RGB]{247,253,248}7.99 & \cellcolor[RGB]{255,246,247}7.63 & \cellcolor[RGB]{254,255,254}7.34 \\
\cellcolor[RGB]{246,253,248}8.00 & \cellcolor[RGB]{255,246,247}7.63 & \cellcolor[RGB]{250,254,251}7.37 \\
\cellcolor[RGB]{252,254,253}7.94 & \cellcolor[RGB]{255,232,235}7.60 & \cellcolor[RGB]{255,255,255}7.32 \\
\end{tabular} \\
\end{tabular} \\
\midrule
$(x,\varnothing)$ & \begin{tabular}[t]{@{}l@{}}
$k$ \\
\midrule
$0$ \\
$5$ \\
$10$ \\
$15$ \\
$20$ \\
$25$ \\
$30$ \\
$35$ \\
$40$ \\
$45$ \\
\end{tabular} & \begin{tabular}[t]{@{}c@{}}
\begin{tabular}[t]{*{3}{>{\centering\arraybackslash}p{16pt}}}
SC & PQ & O \\
\midrule
\cellcolor[RGB]{255,199,206}0.11 & \cellcolor[RGB]{246,252,247}7.77 & \cellcolor[RGB]{255,199,206}0.15 \\
\cellcolor[RGB]{255,203,209}0.61 & \cellcolor[RGB]{198,239,206}8.38 & \cellcolor[RGB]{255,204,211}0.82 \\
\cellcolor[RGB]{255,213,218}2.03 & \cellcolor[RGB]{214,243,219}8.18 & \cellcolor[RGB]{255,218,222}2.55 \\
\cellcolor[RGB]{255,228,231}4.12 & \cellcolor[RGB]{230,248,234}7.97 & \cellcolor[RGB]{255,235,237}4.70 \\
\cellcolor[RGB]{255,240,242}5.81 & \cellcolor[RGB]{235,250,238}7.90 & \cellcolor[RGB]{255,246,247}6.17 \\
\cellcolor[RGB]{255,248,249}6.88 & \cellcolor[RGB]{243,252,245}7.80 & \cellcolor[RGB]{255,251,252}6.88 \\
\cellcolor[RGB]{255,252,252}7.51 & \cellcolor[RGB]{251,254,252}7.70 & \cellcolor[RGB]{255,254,254}7.20 \\
\cellcolor[RGB]{255,254,254}7.74 & \cellcolor[RGB]{251,254,252}7.70 & \cellcolor[RGB]{255,255,255}7.32 \\
\cellcolor[RGB]{255,255,255}7.86 & \cellcolor[RGB]{255,246,247}7.63 & \cellcolor[RGB]{254,255,254}7.34 \\
\cellcolor[RGB]{251,254,252}7.95 & \cellcolor[RGB]{253,254,253}7.68 & \cellcolor[RGB]{248,253,249}7.39 \\
\end{tabular} \\
\end{tabular} & \begin{tabular}[t]{@{}c@{}}
\begin{tabular}[t]{*{3}{>{\centering\arraybackslash}p{16pt}}}
SC & PQ & O \\
\midrule
\cellcolor[RGB]{255,238,240}5.53 & \cellcolor[RGB]{200,239,207}8.36 & \cellcolor[RGB]{255,246,247}6.16 \\
\cellcolor[RGB]{255,242,244}6.09 & \cellcolor[RGB]{207,241,213}8.27 & \cellcolor[RGB]{255,249,250}6.53 \\
\cellcolor[RGB]{255,247,248}6.73 & \cellcolor[RGB]{210,242,216}8.23 & \cellcolor[RGB]{255,253,253}7.04 \\
\cellcolor[RGB]{255,249,250}7.12 & \cellcolor[RGB]{217,244,222}8.14 & \cellcolor[RGB]{255,254,254}7.23 \\
\cellcolor[RGB]{255,253,253}7.62 & \cellcolor[RGB]{226,247,230}8.02 & \cellcolor[RGB]{233,249,236}7.51 \\
\cellcolor[RGB]{255,255,255}7.85 & \cellcolor[RGB]{231,248,234}7.96 & \cellcolor[RGB]{228,247,232}7.55 \\
\cellcolor[RGB]{249,253,250}7.97 & \cellcolor[RGB]{239,250,241}7.86 & \cellcolor[RGB]{227,247,231}7.56 \\
\cellcolor[RGB]{248,253,249}7.98 & \cellcolor[RGB]{244,252,246}7.79 & \cellcolor[RGB]{237,250,239}7.48 \\
\cellcolor[RGB]{245,252,247}8.01 & \cellcolor[RGB]{249,253,250}7.73 & \cellcolor[RGB]{237,250,239}7.48 \\
\cellcolor[RGB]{244,252,246}8.02 & \cellcolor[RGB]{254,255,254}7.66 & \cellcolor[RGB]{245,252,247}7.41 \\
\end{tabular} \\
\end{tabular} & \begin{tabular}[t]{@{}c@{}}
\begin{tabular}[t]{*{3}{>{\centering\arraybackslash}p{16pt}}}
SC & PQ & O \\
\midrule
\cellcolor[RGB]{255,238,240}5.53 & \cellcolor[RGB]{200,239,207}8.36 & \cellcolor[RGB]{255,246,247}6.16 \\
\cellcolor[RGB]{255,241,243}5.93 & \cellcolor[RGB]{201,240,208}8.34 & \cellcolor[RGB]{255,249,249}6.52 \\
\cellcolor[RGB]{255,245,246}6.46 & \cellcolor[RGB]{204,241,211}8.30 & \cellcolor[RGB]{255,251,252}6.85 \\
\cellcolor[RGB]{255,248,249}6.88 & \cellcolor[RGB]{208,242,214}8.26 & \cellcolor[RGB]{255,254,254}7.14 \\
\cellcolor[RGB]{255,249,249}7.02 & \cellcolor[RGB]{209,242,216}8.24 & \cellcolor[RGB]{255,254,254}7.17 \\
\cellcolor[RGB]{255,250,251}7.24 & \cellcolor[RGB]{211,243,217}8.21 & \cellcolor[RGB]{255,254,254}7.25 \\
\cellcolor[RGB]{255,252,252}7.44 & \cellcolor[RGB]{213,243,219}8.19 & \cellcolor[RGB]{255,255,255}7.33 \\
\cellcolor[RGB]{255,252,253}7.52 & \cellcolor[RGB]{210,242,217}8.22 & \cellcolor[RGB]{251,254,251}7.36 \\
\cellcolor[RGB]{255,253,253}7.58 & \cellcolor[RGB]{216,244,221}8.15 & \cellcolor[RGB]{254,255,254}7.34 \\
\cellcolor[RGB]{255,253,253}7.58 & \cellcolor[RGB]{215,244,221}8.16 & \cellcolor[RGB]{255,255,255}7.32 \\
\end{tabular} \\
\end{tabular} & \begin{tabular}[t]{c}---\end{tabular} \\
\bottomrule
\end{tabular*}
\end{minipage}%
\hspace{8pt}%
\begin{minipage}[c]{\dimexpr0.21\linewidth-8pt\relax}
\raggedright
\begin{tabular}{@{}c@{}}
\textbf{Reverse switch} \\
$\mathrm{T2I}\to\mathrm{E}$ \\
Image $(\varnothing,x)$ \\
Text $(c,p)$ \\
\begin{tabular*}{\linewidth}{@{\extracolsep{\fill}}lccc@{}}
\toprule
$k$ & SC & PQ & O \\
\midrule
$0$ & \cellcolor[RGB]{238,238,238}7.91 & \cellcolor[RGB]{238,238,238}7.65 & \cellcolor[RGB]{238,238,238}7.33 \\
$5$ & \cellcolor[RGB]{255,254,254}7.79 & \cellcolor[RGB]{222,246,227}8.07 & \cellcolor[RGB]{232,248,235}7.52 \\
$10$ & \cellcolor[RGB]{255,250,250}7.17 & \cellcolor[RGB]{209,242,215}8.24 & \cellcolor[RGB]{255,254,255}7.26 \\
$15$ & \cellcolor[RGB]{255,247,248}6.74 & \cellcolor[RGB]{207,241,213}8.27 & \cellcolor[RGB]{255,253,253}7.04 \\
$20$ & \cellcolor[RGB]{255,244,245}6.33 & \cellcolor[RGB]{207,242,214}8.26 & \cellcolor[RGB]{255,250,251}6.73 \\
$25$ & \cellcolor[RGB]{255,242,243}6.04 & \cellcolor[RGB]{205,241,212}8.29 & \cellcolor[RGB]{255,249,250}6.54 \\
$30$ & \cellcolor[RGB]{255,241,243}5.98 & \cellcolor[RGB]{206,241,213}8.28 & \cellcolor[RGB]{255,248,249}6.48 \\
$35$ & \cellcolor[RGB]{255,241,242}5.90 & \cellcolor[RGB]{202,240,209}8.33 & \cellcolor[RGB]{255,248,249}6.45 \\
$40$ & \cellcolor[RGB]{255,239,241}5.66 & \cellcolor[RGB]{200,240,208}8.35 & \cellcolor[RGB]{255,247,248}6.27 \\
$45$ & \cellcolor[RGB]{255,239,241}5.73 & \cellcolor[RGB]{200,239,207}8.36 & \cellcolor[RGB]{255,247,248}6.35 \\
\bottomrule
\end{tabular*} \\
\end{tabular}
\end{minipage}
\end{minipage}
\end{table}
\clearpage

The per-interval and single-switch results show that:
\begin{itemize}
\item Editing with the improved instruction ($\mathrm{E}^{*}$) raises semantic consistency, with essentially no gain in perceptual quality.
\item Removing image conditioning gives the highest perceptual quality in both the interval and single-switch sweeps.
\item The text should follow the task: use the target caption while the image is absent, and the editing instruction at the source-conditioned start and end. Keeping the original instruction during the image-free interval, or extending the caption outside it, sacrifices semantic consistency.
\end{itemize}

\begin{table}[!htbp]
\centering
\caption{\textbf{GEdit best scores over the displayed placements (FLUX2-Klein 4B, Qwen3.5).}
Each row pools interval and single-switch results for one conditioning schedule. Metrics are optimized separately; parentheses give a maximizing interval, so entries in a row need not describe one configuration. Image/text triples specify conditioning before, during, and after the interval. Forward and reverse switches are written as $[k,50)$ and $[0,k)$, respectively. Ties use the earliest start, then the earliest end.}
\label{tab:appx-gedit-qwen-summary}
\small
\setlength{\tabcolsep}{3pt}
\begin{tabular}{@{}lllccc@{}}
\toprule
Schedule & Image & Text & Best SC & Best PQ & Best O \\
\midrule
\cellcolor[gray]{0.93}\parbox[c][24pt][c]{93pt}{$\mathrm{E}$} & \cellcolor[gray]{0.93}$(x,x,x)$ & \cellcolor[gray]{0.93}$(p,p,p)$ & \cellcolor[RGB]{238,238,238}7.91 & \cellcolor[RGB]{238,238,238}7.65 & \cellcolor[RGB]{238,238,238}7.33 \\
\midrule[0.25pt]
\parbox[c][24pt][c]{93pt}{$\mathrm{E}\to\mathrm{E}^{*}\to\mathrm{E}$} & $(x,x,x)$ & $(p,p^{*},p)$ & \cellcolor[RGB]{198,239,206}8.50 {\scriptsize $([0,10))$} & \cellcolor[RGB]{252,254,253}7.69 {\scriptsize $([0,10))$} & \cellcolor[RGB]{198,239,206}7.80 {\scriptsize $([0,10))$} \\
\midrule[0.25pt]
\parbox[c][24pt][c]{93pt}{\shortstack[l]{Caption in interval;\\image retained}} & $(x,x,x)$ & $(p,c,p)$ & \cellcolor[RGB]{233,249,236}8.14 {\scriptsize $([15,50))$} & \cellcolor[RGB]{214,243,219}8.18 {\scriptsize $([0,50))$} & \cellcolor[RGB]{224,246,229}7.58 {\scriptsize $([10,50))$} \\
\midrule[0.25pt]
\parbox[c][24pt][c]{93pt}{\shortstack[l]{Image off in interval;\\original instruction}} & $(x,\varnothing,x)$ & $(p,p,p)$ & \cellcolor[RGB]{251,254,252}7.95 {\scriptsize $([45,50))$} & \cellcolor[RGB]{198,239,206}8.38 {\scriptsize $([5,50))$} & \cellcolor[RGB]{248,253,249}7.39 {\scriptsize $([45,50))$} \\
\midrule[0.25pt]
\parbox[c][24pt][c]{93pt}{\shortstack[l]{Caption throughout;\\image off in interval}} & $(x,\varnothing,x)$ & $(c,c,c)$ & \cellcolor[RGB]{255,253,253}7.58 {\scriptsize $([40,50))$} & \cellcolor[RGB]{200,239,207}8.36 {\scriptsize $([0,50))$} & \cellcolor[RGB]{230,248,234}7.53 {\scriptsize $([10,20))$} \\
\midrule[0.25pt]
\parbox[c][24pt][c]{93pt}{$\mathrm{E}\to\mathrm{T2I}\to\mathrm{E}$} & $(x,\varnothing,x)$ & $(p,c,p)$ & \cellcolor[RGB]{240,251,242}8.07 {\scriptsize $([25,35))$} & \cellcolor[RGB]{200,239,207}8.36 {\scriptsize $([0,45))$} & \cellcolor[RGB]{211,243,217}7.69 {\scriptsize $([20,30))$} \\
\bottomrule
\end{tabular}
\end{table}
\FloatBarrier

\begin{table}[!htbp]
\centering
\caption{\textbf{ImgEdit per-category scores (GPT-4o judge).} Category breakdown for FLUX2-Klein 4B, 9B, and BAGEL: baseline, improved-instruction controls, T2I schedules, and selective application where reported.}
\label{tab:appx-imgedit}
\vspace{0.75em}
{\footnotesize
\setlength{\tabcolsep}{3pt}
\resizebox{\linewidth}{!}{
\begin{tabular}{llcccccccccc}
\toprule
Schedule & Interval & background & adjust & style & extract & remove & add & replace & compose & action & overall \\
\midrule
\multicolumn{12}{l}{\textbf{FLUX2-Klein 4B}} \\
\midrule

Baseline & --- & 4.21 & 3.85 & 4.80 & 1.47 & 2.99 & 4.40 & 4.18 & 3.26 & 4.70 & 3.74 \\
\midrule
$\mathrm{E}\to\mathrm{E}^{*}\to\mathrm{E}$ & $[0,10)$ & \cellcolor[RGB]{255,249,250}4.11 & \cellcolor[RGB]{255,240,242}3.63 & 4.80 & \cellcolor[RGB]{202,240,209}3.74 & \cellcolor[RGB]{198,239,206}4.33 & \cellcolor[RGB]{238,250,241}4.49 & \cellcolor[RGB]{255,253,253}4.16 & \cellcolor[RGB]{208,242,215}3.68 & \cellcolor[RGB]{247,253,248}4.73 & \cellcolor[RGB]{198,239,206}4.18 \\
$\mathrm{E}\to\mathrm{E}^{*}\to\mathrm{E}$ & $[25,35)$ & \cellcolor[RGB]{255,247,248}4.08 & \cellcolor[RGB]{255,254,254}3.83 & \cellcolor[RGB]{255,244,246}4.74 & \cellcolor[RGB]{255,199,206}1.38 & \cellcolor[RGB]{255,199,206}2.98 & \cellcolor[RGB]{255,239,241}4.36 & \cellcolor[RGB]{255,251,252}4.14 & \cellcolor[RGB]{255,252,252}3.19 & \cellcolor[RGB]{198,239,206}4.91 & \cellcolor[RGB]{255,244,246}3.69 \\
$\mathrm{E}\to\mathrm{E}^{*}\to\mathrm{E}$ & $[35,45)$ & \cellcolor[RGB]{255,251,252}4.15 & \cellcolor[RGB]{255,252,253}3.81 & 4.80 & \cellcolor[RGB]{253,254,253}1.55 & \cellcolor[RGB]{254,255,254}3.01 & 4.40 & \cellcolor[RGB]{251,254,251}4.20 & 3.26 & \cellcolor[RGB]{228,247,232}4.80 & \cellcolor[RGB]{255,253,253}3.73 \\
E$^{*}$ & $[0,50)$ & \cellcolor[RGB]{255,247,248}4.08 & \cellcolor[RGB]{255,247,248}3.73 & \cellcolor[RGB]{255,246,247}4.75 & \cellcolor[RGB]{204,241,211}3.64 & \cellcolor[RGB]{200,240,208}4.28 & \cellcolor[RGB]{251,254,252}4.42 & \cellcolor[RGB]{253,254,253}4.19 & \cellcolor[RGB]{223,246,227}3.55 & \cellcolor[RGB]{247,253,248}4.73 & \cellcolor[RGB]{202,240,209}4.15 \\
\midrule
T2I$\to$E & $k=25$ & \cellcolor[RGB]{225,247,229}4.40 & \cellcolor[RGB]{255,199,206}3.04 & \cellcolor[RGB]{255,237,239}4.70 & \cellcolor[RGB]{198,239,206}3.90 & \cellcolor[RGB]{221,246,226}3.78 & \cellcolor[RGB]{246,252,247}4.45 & \cellcolor[RGB]{255,199,206}3.55 & \cellcolor[RGB]{255,200,207}2.01 & \cellcolor[RGB]{255,237,239}4.36 & \cellcolor[RGB]{229,248,233}3.94 \\
E$\to$T2I & $k=5$ & \cellcolor[RGB]{220,245,225}4.43 & \cellcolor[RGB]{255,209,214}3.18 & \cellcolor[RGB]{255,248,249}4.76 & \cellcolor[RGB]{206,241,212}3.58 & \cellcolor[RGB]{203,240,210}4.21 & \cellcolor[RGB]{242,251,244}4.47 & \cellcolor[RGB]{255,212,218}3.70 & \cellcolor[RGB]{255,206,212}2.14 & \cellcolor[RGB]{255,199,206}3.65 & \cellcolor[RGB]{226,247,230}3.96 \\
E$\to$T2I & $k=10$ & \cellcolor[RGB]{215,244,221}4.46 & \cellcolor[RGB]{255,242,244}3.66 & \cellcolor[RGB]{244,252,245}4.83 & \cellcolor[RGB]{229,248,233}2.58 & \cellcolor[RGB]{230,248,233}3.58 & \cellcolor[RGB]{222,246,227}4.58 & \cellcolor[RGB]{255,221,225}3.80 & \cellcolor[RGB]{255,199,206}1.99 & \cellcolor[RGB]{255,214,219}3.93 & \cellcolor[RGB]{238,250,241}3.87 \\
E$\to$T2I & $k=15$ & \cellcolor[RGB]{211,243,217}4.49 & \cellcolor[RGB]{241,251,243}4.06 & \cellcolor[RGB]{225,246,229}4.88 & \cellcolor[RGB]{243,252,244}2.00 & \cellcolor[RGB]{238,250,240}3.40 & \cellcolor[RGB]{211,243,217}4.64 & \cellcolor[RGB]{255,243,244}4.04 & \cellcolor[RGB]{255,240,241}2.91 & \cellcolor[RGB]{255,230,233}4.23 & \cellcolor[RGB]{230,248,234}3.93 \\
E$\to$T2I & $k=25$ & \cellcolor[RGB]{215,244,221}4.46 & \cellcolor[RGB]{202,240,210}4.62 & \cellcolor[RGB]{206,241,213}4.93 & \cellcolor[RGB]{244,252,246}1.94 & \cellcolor[RGB]{242,251,244}3.30 & \cellcolor[RGB]{209,242,215}4.65 & \cellcolor[RGB]{231,248,234}4.30 & \cellcolor[RGB]{206,241,213}3.70 & \cellcolor[RGB]{255,254,255}4.69 & \cellcolor[RGB]{211,243,217}4.08 \\
E$\to$T2I & $k=35$ & \cellcolor[RGB]{244,252,245}4.28 & \cellcolor[RGB]{215,244,221}4.43 & \cellcolor[RGB]{209,242,216}4.92 & \cellcolor[RGB]{251,254,252}1.63 & \cellcolor[RGB]{246,252,247}3.20 & \cellcolor[RGB]{231,248,234}4.53 & \cellcolor[RGB]{235,249,237}4.28 & \cellcolor[RGB]{208,242,215}3.68 & \cellcolor[RGB]{241,251,243}4.75 & \cellcolor[RGB]{230,248,234}3.93 \\
E$\to$T2I & $k=40$ & \cellcolor[RGB]{239,251,241}4.31 & \cellcolor[RGB]{232,248,235}4.19 & \cellcolor[RGB]{213,243,219}4.91 & \cellcolor[RGB]{252,254,252}1.61 & \cellcolor[RGB]{248,253,249}3.15 & \cellcolor[RGB]{235,249,238}4.51 & \cellcolor[RGB]{237,250,239}4.27 & \cellcolor[RGB]{227,247,231}3.51 & \cellcolor[RGB]{220,245,225}4.83 & \cellcolor[RGB]{237,250,239}3.88 \\
E$\to$T2I & $k=45$ & \cellcolor[RGB]{244,252,245}4.28 & \cellcolor[RGB]{241,251,243}4.06 & \cellcolor[RGB]{228,248,232}4.87 & \cellcolor[RGB]{255,255,255}1.48 & \cellcolor[RGB]{248,253,249}3.15 & \cellcolor[RGB]{248,253,249}4.44 & \cellcolor[RGB]{249,253,250}4.21 & \cellcolor[RGB]{244,252,245}3.36 & \cellcolor[RGB]{203,241,211}4.89 & \cellcolor[RGB]{245,252,246}3.82 \\
$\mathrm{E}\to\mathrm{T2I}\to\mathrm{E}$ & $[10,16)$ & \cellcolor[RGB]{208,242,214}4.51 & \cellcolor[RGB]{236,250,239}4.12 & \cellcolor[RGB]{221,245,226}4.89 & \cellcolor[RGB]{245,252,247}1.88 & \cellcolor[RGB]{237,250,240}3.41 & \cellcolor[RGB]{224,246,228}4.57 & \cellcolor[RGB]{243,252,244}4.24 & \cellcolor[RGB]{201,240,209}3.74 & \cellcolor[RGB]{212,243,218}4.86 & \cellcolor[RGB]{223,246,227}3.99 \\
$\mathrm{E}\to\mathrm{T2I}\to\mathrm{E}$ & $[10,20)$ & \cellcolor[RGB]{209,242,216}4.50 & \cellcolor[RGB]{223,246,228}4.31 & \cellcolor[RGB]{198,239,206}4.95 & \cellcolor[RGB]{234,249,237}2.37 & \cellcolor[RGB]{221,245,226}3.79 & \cellcolor[RGB]{198,239,206}4.71 & \cellcolor[RGB]{198,239,206}4.46 & \cellcolor[RGB]{230,248,234}3.48 & \cellcolor[RGB]{236,250,239}4.77 & \cellcolor[RGB]{201,240,208}4.16 \\
$\mathrm{E}\to\mathrm{T2I}\to\mathrm{E}$ & $[15,30)$ & \cellcolor[RGB]{198,239,206}4.57 & \cellcolor[RGB]{202,240,210}4.62 & \cellcolor[RGB]{206,241,213}4.93 & \cellcolor[RGB]{243,252,244}2.00 & \cellcolor[RGB]{229,248,233}3.59 & \cellcolor[RGB]{200,240,208}4.70 & \cellcolor[RGB]{255,251,252}4.14 & \cellcolor[RGB]{230,248,234}3.48 & \cellcolor[RGB]{247,253,248}4.73 & \cellcolor[RGB]{206,241,213}4.12 \\
$\mathrm{E}\to\mathrm{T2I}\to\mathrm{E}$ & $[15,35)$ & \cellcolor[RGB]{200,239,207}4.56 & \cellcolor[RGB]{217,244,223}4.40 & \cellcolor[RGB]{213,243,219}4.91 & \cellcolor[RGB]{240,251,242}2.13 & \cellcolor[RGB]{237,250,239}3.42 & \cellcolor[RGB]{209,242,215}4.65 & \cellcolor[RGB]{255,249,250}4.11 & \cellcolor[RGB]{245,252,246}3.35 & \cellcolor[RGB]{255,254,255}4.69 & \cellcolor[RGB]{214,243,219}4.06 \\
$\mathrm{E}\to\mathrm{T2I}\to\mathrm{E}$ & $[20,30)$ & \cellcolor[RGB]{219,245,224}4.44 & \cellcolor[RGB]{206,241,212}4.57 & \cellcolor[RGB]{217,244,222}4.90 & \cellcolor[RGB]{245,252,246}1.91 & \cellcolor[RGB]{240,251,242}3.34 & \cellcolor[RGB]{215,244,220}4.62 & \cellcolor[RGB]{231,248,234}4.30 & \cellcolor[RGB]{201,240,209}3.74 & \cellcolor[RGB]{217,244,222}4.84 & \cellcolor[RGB]{212,243,218}4.07 \\
$\mathrm{E}\to\mathrm{T2I}\to\mathrm{E}$ & $[20,35)$ & \cellcolor[RGB]{214,243,220}4.47 & \cellcolor[RGB]{211,243,217}4.49 & \cellcolor[RGB]{206,241,213}4.93 & \cellcolor[RGB]{245,252,246}1.90 & \cellcolor[RGB]{243,252,245}3.27 & \cellcolor[RGB]{207,242,214}4.66 & \cellcolor[RGB]{247,253,248}4.22 & \cellcolor[RGB]{198,239,206}3.77 & \cellcolor[RGB]{241,251,243}4.75 & \cellcolor[RGB]{216,244,222}4.04 \\
$\mathrm{E}\to\mathrm{T2I}\to\mathrm{E}$ & $[25,35)$ & \cellcolor[RGB]{220,245,225}4.43 & \cellcolor[RGB]{208,242,214}4.54 & \cellcolor[RGB]{209,242,216}4.92 & \cellcolor[RGB]{246,252,247}1.86 & \cellcolor[RGB]{241,251,243}3.32 & \cellcolor[RGB]{211,243,217}4.64 & \cellcolor[RGB]{237,250,239}4.27 & \cellcolor[RGB]{200,240,208}3.75 & \cellcolor[RGB]{225,247,229}4.81 & \cellcolor[RGB]{216,244,222}4.04 \\
$\mathrm{E}\to\mathrm{T2I}\to\mathrm{E}$ & $[25,40)$ & \cellcolor[RGB]{222,246,226}4.42 & \cellcolor[RGB]{198,239,206}4.68 & \cellcolor[RGB]{202,240,209}4.94 & \cellcolor[RGB]{249,253,250}1.74 & \cellcolor[RGB]{239,251,241}3.36 & \cellcolor[RGB]{202,240,209}4.69 & \cellcolor[RGB]{235,249,237}4.28 & \cellcolor[RGB]{211,243,218}3.65 & \cellcolor[RGB]{247,253,248}4.73 & \cellcolor[RGB]{212,243,218}4.07 \\
$\mathrm{E}\to\mathrm{T2I}\to\mathrm{E}$ & $[35,45)$ & \cellcolor[RGB]{236,250,239}4.33 & \cellcolor[RGB]{221,245,225}4.35 & \cellcolor[RGB]{206,241,213}4.93 & \cellcolor[RGB]{251,254,251}1.66 & \cellcolor[RGB]{246,252,247}3.21 & \cellcolor[RGB]{237,250,239}4.50 & \cellcolor[RGB]{231,248,234}4.30 & \cellcolor[RGB]{236,250,239}3.43 & \cellcolor[RGB]{214,244,220}4.85 & \cellcolor[RGB]{232,248,235}3.92 \\
Selective & $[10,16)$ & \cellcolor[RGB]{231,248,235}4.36 & \cellcolor[RGB]{245,252,247}3.99 & \cellcolor[RGB]{221,245,226}4.89 & \cellcolor[RGB]{249,253,250}1.73 & \cellcolor[RGB]{244,252,245}3.26 & 4.40 & \cellcolor[RGB]{239,250,241}4.26 & \cellcolor[RGB]{208,242,215}3.68 & \cellcolor[RGB]{225,247,229}4.81 & \cellcolor[RGB]{237,250,239}3.88 \\
\midrule
\multicolumn{12}{l}{\textbf{FLUX2-Klein 9B}} \\
\midrule

Baseline & --- & 4.22 & 4.17 & 4.92 & 1.94 & 4.14 & 4.61 & 4.47 & 3.29 & 4.81 & 4.02 \\
\midrule
$\mathrm{E}\to\mathrm{E}^{*}\to\mathrm{E}$ & $[25,35)$ & \cellcolor[RGB]{241,251,243}4.29 & \cellcolor[RGB]{249,253,250}4.21 & \cellcolor[RGB]{255,233,235}4.90 & \cellcolor[RGB]{251,254,252}2.05 & \cellcolor[RGB]{255,243,244}4.12 & \cellcolor[RGB]{255,199,206}4.55 & \cellcolor[RGB]{234,249,237}4.51 & \cellcolor[RGB]{229,248,233}3.58 & 4.81 & \cellcolor[RGB]{244,252,246}4.08 \\
$\mathrm{E}\to\mathrm{E}^{*}\to\mathrm{E}$ & $[35,45)$ & \cellcolor[RGB]{253,254,253}4.23 & \cellcolor[RGB]{245,252,246}4.24 & \cellcolor[RGB]{255,233,235}4.90 & \cellcolor[RGB]{250,254,251}2.09 & \cellcolor[RGB]{255,199,206}4.05 & \cellcolor[RGB]{255,218,222}4.57 & \cellcolor[RGB]{250,254,251}4.48 & \cellcolor[RGB]{252,254,253}3.32 & \cellcolor[RGB]{255,199,206}4.74 & \cellcolor[RGB]{246,252,247}4.07 \\
E$^{*}$ & $[0,50)$ & \cellcolor[RGB]{255,199,206}4.09 & \cellcolor[RGB]{240,251,242}4.27 & \cellcolor[RGB]{255,210,216}4.88 & \cellcolor[RGB]{198,239,206}3.63 & \cellcolor[RGB]{198,239,206}4.55 & \cellcolor[RGB]{255,218,222}4.57 & \cellcolor[RGB]{255,250,251}4.46 & \cellcolor[RGB]{217,244,222}3.72 & 4.81 & \cellcolor[RGB]{198,239,206}4.34 \\
\midrule
$\mathrm{E}\to\mathrm{T2I}\to\mathrm{E}$ & $[10,16)$ & \cellcolor[RGB]{231,248,234}4.34 & \cellcolor[RGB]{237,250,240}4.29 & \cellcolor[RGB]{241,251,243}4.93 & \cellcolor[RGB]{236,250,239}2.50 & \cellcolor[RGB]{224,246,229}4.36 & \cellcolor[RGB]{244,252,245}4.62 & \cellcolor[RGB]{198,239,206}4.58 & \cellcolor[RGB]{227,247,231}3.61 & \cellcolor[RGB]{198,239,206}4.92 & \cellcolor[RGB]{218,244,223}4.23 \\
$\mathrm{E}\to\mathrm{T2I}\to\mathrm{E}$ & $[10,20)$ & \cellcolor[RGB]{198,239,206}4.50 & \cellcolor[RGB]{223,246,227}4.39 & \cellcolor[RGB]{255,199,206}4.87 & \cellcolor[RGB]{220,245,225}2.98 & \cellcolor[RGB]{205,241,212}4.50 & \cellcolor[RGB]{221,245,226}4.64 & \cellcolor[RGB]{255,199,206}4.36 & \cellcolor[RGB]{240,251,242}3.46 & \cellcolor[RGB]{250,254,251}4.82 & \cellcolor[RGB]{200,240,208}4.33 \\
$\mathrm{E}\to\mathrm{T2I}\to\mathrm{E}$ & $[20,30)$ & \cellcolor[RGB]{220,245,225}4.39 & \cellcolor[RGB]{198,239,206}4.56 & \cellcolor[RGB]{198,239,206}4.96 & \cellcolor[RGB]{241,251,243}2.36 & \cellcolor[RGB]{238,250,241}4.26 & \cellcolor[RGB]{198,239,206}4.66 & \cellcolor[RGB]{245,252,246}4.49 & \cellcolor[RGB]{215,244,221}3.74 & \cellcolor[RGB]{219,245,224}4.88 & \cellcolor[RGB]{212,243,218}4.26 \\
$\mathrm{E}\to\mathrm{T2I}\to\mathrm{E}$ & $[25,35)$ & \cellcolor[RGB]{214,244,220}4.42 & \cellcolor[RGB]{204,241,211}4.52 & \cellcolor[RGB]{212,243,218}4.95 & \cellcolor[RGB]{242,251,244}2.33 & \cellcolor[RGB]{237,250,239}4.27 & \cellcolor[RGB]{209,242,216}4.65 & \cellcolor[RGB]{255,235,237}4.43 & \cellcolor[RGB]{210,242,216}3.80 & \cellcolor[RGB]{224,246,228}4.87 & \cellcolor[RGB]{216,244,221}4.24 \\
$\mathrm{E}\to\mathrm{T2I}\to\mathrm{E}$ & $[35,45)$ & \cellcolor[RGB]{231,248,234}4.34 & \cellcolor[RGB]{218,245,224}4.42 & \cellcolor[RGB]{226,247,230}4.94 & \cellcolor[RGB]{245,252,247}2.23 & \cellcolor[RGB]{251,254,251}4.17 & \cellcolor[RGB]{221,245,226}4.64 & \cellcolor[RGB]{255,219,224}4.40 & \cellcolor[RGB]{198,239,206}3.93 & \cellcolor[RGB]{255,231,234}4.78 & \cellcolor[RGB]{226,247,230}4.18 \\
\midrule
\multicolumn{12}{l}{\textbf{BAGEL}} \\
\midrule

Baseline & --- & 3.38 & 3.51 & 4.43 & 1.38 & 3.00 & 3.93 & 3.64 & 2.53 & 4.22 & 3.32 \\
\midrule
$\mathrm{E}\to\mathrm{E}^{*}\to\mathrm{E}$ & $[25,35)$ & \cellcolor[RGB]{234,249,237}3.47 & \cellcolor[RGB]{247,253,248}3.54 & 4.43 & \cellcolor[RGB]{255,199,206}1.35 & 3.00 & \cellcolor[RGB]{255,248,249}3.92 & \cellcolor[RGB]{255,236,239}3.63 & \cellcolor[RGB]{200,240,208}2.77 & \cellcolor[RGB]{251,254,252}4.24 & \cellcolor[RGB]{253,254,253}3.33 \\
$\mathrm{E}\to\mathrm{E}^{*}\to\mathrm{E}$ & $[35,45)$ & \cellcolor[RGB]{243,252,245}3.43 & \cellcolor[RGB]{255,249,250}3.43 & \cellcolor[RGB]{242,251,244}4.49 & \cellcolor[RGB]{252,254,252}1.42 & \cellcolor[RGB]{254,255,254}3.01 & \cellcolor[RGB]{255,199,206}3.85 & \cellcolor[RGB]{255,199,206}3.61 & \cellcolor[RGB]{241,251,243}2.59 & \cellcolor[RGB]{255,199,206}4.19 & \cellcolor[RGB]{255,199,206}3.30 \\
E$^{*}$ & $[0,50)$ & \cellcolor[RGB]{241,251,243}3.44 & \cellcolor[RGB]{255,245,246}3.37 & \cellcolor[RGB]{255,199,206}4.41 & \cellcolor[RGB]{198,239,206}2.10 & \cellcolor[RGB]{235,249,238}3.25 & \cellcolor[RGB]{255,227,230}3.89 & \cellcolor[RGB]{246,252,247}3.69 & \cellcolor[RGB]{223,246,228}2.67 & \cellcolor[RGB]{230,248,234}4.36 & \cellcolor[RGB]{228,247,231}3.44 \\
\midrule
$\mathrm{E}\to\mathrm{T2I}\to\mathrm{E}$ & $[10,16)$ & \cellcolor[RGB]{219,245,224}3.53 & \cellcolor[RGB]{255,251,251}3.45 & \cellcolor[RGB]{209,242,215}4.65 & \cellcolor[RGB]{242,251,244}1.54 & \cellcolor[RGB]{238,250,240}3.21 & \cellcolor[RGB]{229,248,232}4.05 & \cellcolor[RGB]{231,248,234}3.77 & \cellcolor[RGB]{205,241,212}2.75 & \cellcolor[RGB]{244,252,246}4.28 & \cellcolor[RGB]{228,247,231}3.44 \\
$\mathrm{E}\to\mathrm{T2I}\to\mathrm{E}$ & $[10,20)$ & \cellcolor[RGB]{215,244,220}3.55 & \cellcolor[RGB]{255,199,206}2.75 & \cellcolor[RGB]{204,241,211}4.67 & \cellcolor[RGB]{235,249,238}1.63 & \cellcolor[RGB]{231,248,235}3.29 & \cellcolor[RGB]{246,253,247}3.97 & \cellcolor[RGB]{246,252,247}3.69 & \cellcolor[RGB]{255,199,206}2.35 & \cellcolor[RGB]{255,218,222}4.20 & \cellcolor[RGB]{248,253,249}3.35 \\
$\mathrm{E}\to\mathrm{T2I}\to\mathrm{E}$ & $[20,30)$ & \cellcolor[RGB]{224,246,228}3.51 & \cellcolor[RGB]{255,244,245}3.36 & \cellcolor[RGB]{202,240,210}4.68 & \cellcolor[RGB]{245,252,246}1.51 & \cellcolor[RGB]{231,248,235}3.29 & \cellcolor[RGB]{218,245,223}4.10 & \cellcolor[RGB]{218,245,223}3.84 & \cellcolor[RGB]{232,249,235}2.63 & \cellcolor[RGB]{239,250,241}4.31 & \cellcolor[RGB]{223,246,228}3.46 \\
$\mathrm{E}\to\mathrm{T2I}\to\mathrm{E}$ & $[25,35)$ & \cellcolor[RGB]{205,241,212}3.59 & \cellcolor[RGB]{255,251,252}3.46 & \cellcolor[RGB]{198,239,206}4.70 & \cellcolor[RGB]{246,252,247}1.50 & \cellcolor[RGB]{224,246,228}3.38 & \cellcolor[RGB]{198,239,206}4.19 & \cellcolor[RGB]{207,242,214}3.90 & \cellcolor[RGB]{255,243,244}2.49 & \cellcolor[RGB]{221,246,226}4.41 & \cellcolor[RGB]{212,243,218}3.51 \\
$\mathrm{E}\to\mathrm{T2I}\to\mathrm{E}$ & $[35,45)$ & \cellcolor[RGB]{198,239,206}3.62 & \cellcolor[RGB]{198,239,206}3.73 & \cellcolor[RGB]{219,245,224}4.60 & \cellcolor[RGB]{254,255,254}1.39 & \cellcolor[RGB]{198,239,206}3.70 & \cellcolor[RGB]{200,240,208}4.18 & \cellcolor[RGB]{198,239,206}3.95 & \cellcolor[RGB]{198,239,206}2.78 & \cellcolor[RGB]{198,239,206}4.54 & \cellcolor[RGB]{198,239,206}3.57 \\
\bottomrule
\end{tabular}
}
}
\end{table}

\begin{table}[!htbp]
\centering
\caption{\textbf{Improved-instruction controls and best observed T2I intervals across three models.}
The evaluation protocols match \cref{tab:main}: GPT-4.1 for GEdit and GPT-4o for ImgEdit and GIE correctness.
The best E$\to$T2I$\to$E score is selected separately for each column from the available interval sweep; the row below reports one winning interval. These maxima need not share one configuration, and sweep coverage varies by model and benchmark.
Higher is better; green/red intensity shows changes from each model's pure-editing baseline, using the per-metric scales of \cref{tab:main}. The $\pm$ rows are three-seed baseline standard deviations, as in \cref{tab:main}.}
\label{tab:appx-estar-comparison}
{\small
\setlength{\tabcolsep}{3pt}
\begin{tabular}{lcccccccc}
\toprule
 & \multicolumn{3}{c}{GEdit} & ImgEdit & MagicBrush & \multicolumn{3}{c}{GIE-Bench} \\
\cmidrule(lr){2-4}\cmidrule(lr){5-5}\cmidrule(lr){6-6}\cmidrule(lr){7-9}
Sampling & SC & PQ & O & overall & CLIP-T & FC & SSIM & U-CLIP \\
\midrule
\multicolumn{9}{l}{\textbf{FLUX2-Klein 4B}} \\
\stdrow{0.05}{0.03}{0.04}{0.04}{0.001}{0.35}{0.001}{0.002} \\
Pure E & 7.72 & 7.46 & 7.11 & 3.74 & 0.317 & 73.40 & 0.805 & 0.913 \\
E$^{*}\!\to$E $(10)$ & \cellcolor[RGB]{199,239,207}8.16 & \cellcolor[RGB]{253,254,253}7.49 & \cellcolor[RGB]{211,243,217}7.54 & \cellcolor[RGB]{198,239,206}4.18 & 0.317 & \cellcolor[RGB]{238,250,240}74.95 & \cellcolor[RGB]{255,249,250}0.801 & \cellcolor[RGB]{255,239,241}0.911 \\
E$^{*}$ & \cellcolor[RGB]{198,239,206}8.17 & \cellcolor[RGB]{253,254,253}7.49 & \cellcolor[RGB]{211,243,217}7.54 & \cellcolor[RGB]{202,240,209}4.15 & 0.317 & \cellcolor[RGB]{228,247,232}75.88 & \cellcolor[RGB]{255,247,248}0.799 & \cellcolor[RGB]{255,231,234}0.910 \\
E$\to$T2I$\to$E (best) & \cellcolor[RGB]{209,242,216}8.08 & \cellcolor[RGB]{198,239,206}8.27 & \cellcolor[RGB]{198,239,206}7.67 & \cellcolor[RGB]{201,240,208}4.16 & \cellcolor[RGB]{198,239,206}0.325 & \cellcolor[RGB]{219,245,224}76.74 & \cellcolor[RGB]{255,221,226}0.781 & \cellcolor[RGB]{255,199,206}0.906 \\
{\scriptsize Winning interval} & {\scriptsize $[10,16)$} & {\scriptsize $[15,30)$} & {\scriptsize $[10,16)$} & {\scriptsize $[10,20)$} & {\scriptsize $[15,30)$} & {\scriptsize $[10,20)$} & {\scriptsize $[10,16)$} & {\scriptsize $[10,16)$} \\
\midrule
\multicolumn{9}{l}{\textbf{FLUX2-Klein 9B}} \\
\stdrow{0.01}{0.01}{0.01}{0.02}{0.001}{0.72}{0.001}{0.0003} \\
Pure E & 8.38 & 7.79 & 7.74 & 4.02 & 0.319 & 77.94 & 0.939 & 0.928 \\
E$^{*}$ & \cellcolor[RGB]{203,240,210}8.79 & \cellcolor[RGB]{254,255,254}7.80 & \cellcolor[RGB]{218,245,224}8.10 & \cellcolor[RGB]{214,243,219}4.34 & 0.319 & \cellcolor[RGB]{228,247,232}80.43 & \cellcolor[RGB]{255,252,253}0.937 & \cellcolor[RGB]{255,239,241}0.926 \\
E$\to$T2I$\to$E (best) & \cellcolor[RGB]{233,249,236}8.55 & \cellcolor[RGB]{225,247,229}8.22 & \cellcolor[RGB]{233,249,236}7.96 & \cellcolor[RGB]{215,244,220}4.33 & \cellcolor[RGB]{205,241,212}0.326 & \cellcolor[RGB]{219,245,224}81.28 & \cellcolor[RGB]{255,199,206}0.899 & \cellcolor[RGB]{255,207,213}0.922 \\
{\scriptsize Winning interval} & {\scriptsize $[10,16)$} & {\scriptsize $[10,20)$} & {\scriptsize $[10,16)$} & {\scriptsize $[10,20)$} & {\scriptsize $[10,20)$} & {\scriptsize $[10,16)$} & {\scriptsize $[10,16)$} & {\scriptsize $[10,16)$} \\
\midrule
\multicolumn{9}{l}{\textbf{BAGEL}} \\
\stdrow{0.06}{0.09}{0.07}{0.03}{0.001}{1.25}{0.001}{0.001} \\
Pure E & 7.68 & 6.38 & 6.64 & 3.32 & 0.320 & 67.16 & 0.960 & 0.928 \\
E$^{*}$ & \cellcolor[RGB]{255,236,239}7.53 & \cellcolor[RGB]{249,253,250}6.47 & \cellcolor[RGB]{255,253,253}6.62 & \cellcolor[RGB]{239,251,242}3.44 & \cellcolor[RGB]{248,253,249}0.321 & \cellcolor[RGB]{255,247,248}66.42 & \cellcolor[RGB]{255,254,254}0.959 & \cellcolor[RGB]{255,207,213}0.922 \\
E$\to$T2I$\to$E (best) & \cellcolor[RGB]{222,246,227}7.94 & \cellcolor[RGB]{230,248,233}6.74 & \cellcolor[RGB]{219,245,224}6.99 & \cellcolor[RGB]{223,246,227}3.57 & \cellcolor[RGB]{226,247,230}0.324 & \cellcolor[RGB]{198,239,206}72.38 & \cellcolor[RGB]{255,244,245}0.952 & \cellcolor[RGB]{255,223,227}0.924 \\
{\scriptsize Winning interval} & {\scriptsize $[35,45)$} & {\scriptsize $[35,45)$} & {\scriptsize $[35,45)$} & {\scriptsize $[35,45)$} & {\scriptsize $[10,16)$} & {\scriptsize $[10,16)$} & {\scriptsize $[10,16)$} & {\scriptsize $[10,16)$} \\
\bottomrule
\end{tabular}}
\end{table}

\FloatBarrier

\section{Continuous guidance comparisons}
\label{app:guidance}
On FLUX2-Klein~4B, we test whether lowering image guidance reproduces the effect of a T2I interval, and introduce a multimodal guidance formulation that continuously combines editing with target-caption-conditioned generation. Let $U$, $T$, $I$, and $E$ denote denoiser predictions conditioned on neither image nor text, the target caption alone, the source image alone, and the source image with the edit instruction, respectively.

\paragraph{Image guidance.}
Following InstructPix2Pix~\citep{brooks2023instructpix2pix}, we use
\begin{equation}
 U+s_i(I-U)+s_t(E-I),
 \label{eq:image-guidance}
\end{equation}
where $s_i$ and $s_t$ control image and text guidance. We hold $s_t$ fixed and reduce $s_i=\alpha$ from the editing baseline $\alpha=1$, either on a limited interval or throughout sampling. Even at $\alpha=0$ the image-conditioned predictions $E$ and $I$ still enter the update, so image guidance and caption-only T2I stay distinct controls.

\paragraph{Multimodal guidance.}
We extend multimodal CFG~\citep{goel2024pair} to combine source-conditioned editing with a separate target-caption branch:
\begin{equation}
 U+w_t(T-U)+w_i(I-U)+w_{ti}(E-T-I+U).
 \label{eq:mm-guidance}
\end{equation}
Here $E$ uses the edit instruction, whereas $T$ uses the target caption. Setting $w_t=s_t$, $w_i=\alpha$, and $w_{ti}=\alpha s_t$ gives the interpolation $\alpha v_{\mathrm{E}}+(1-\alpha)v_{\mathrm{T2I}}$, where $v_{\mathrm{E}}=s_tE+(1-s_t)I$ and $v_{\mathrm{T2I}}=s_tT+(1-s_t)U$. We apply multimodal guidance on a bounded interval and retain edit conditioning elsewhere: $\alpha=0$ recovers $\mathrm{E}\to\mathrm{T2I}\to\mathrm{E}$ and $\alpha=1$ recovers pure editing.

\begin{figure}[t]
 \centering
 \includegraphics[width=\linewidth]{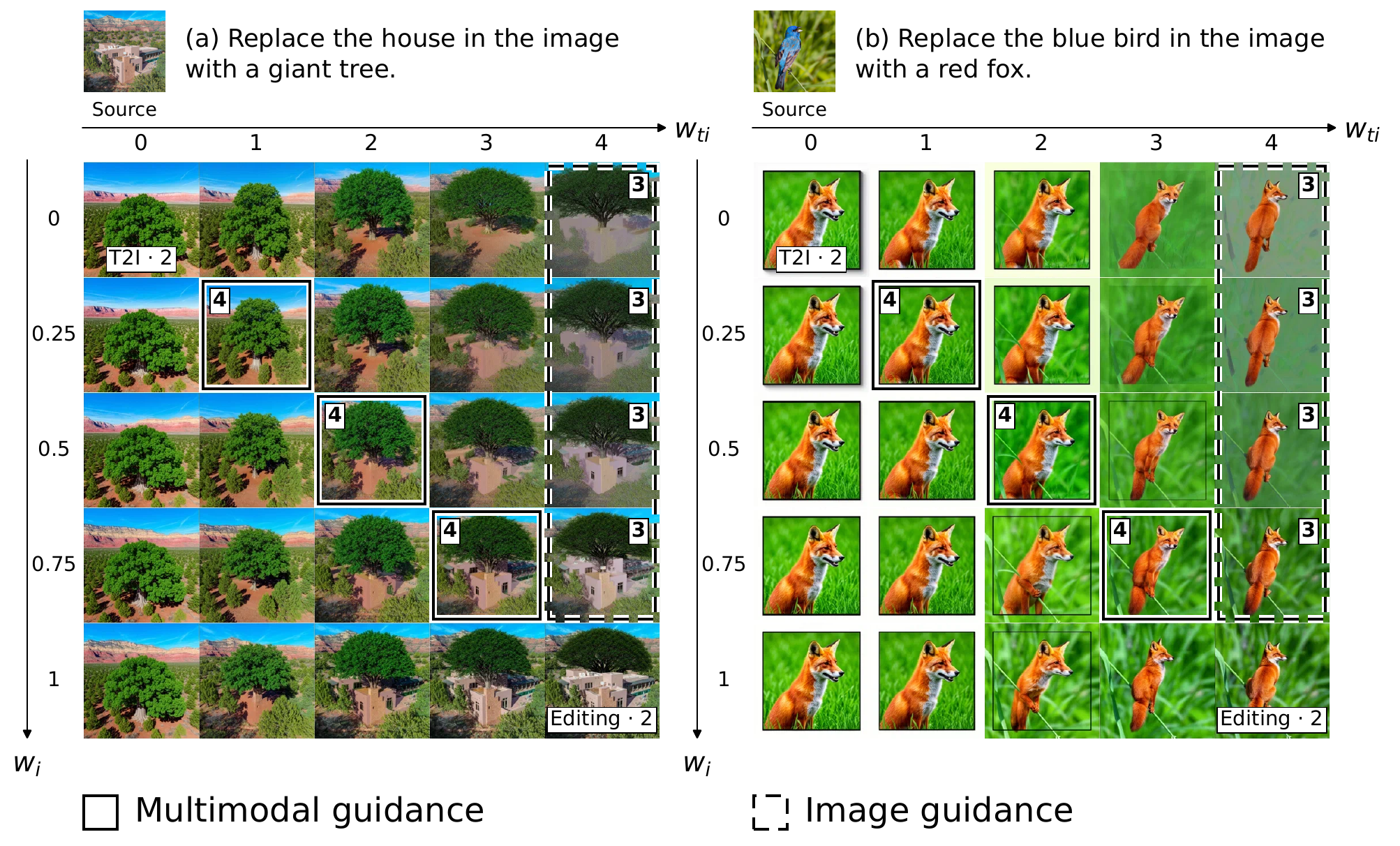}
 \caption{\textbf{Continuous guidance in the factorial weight plane (FLUX2-Klein~4B).} We fix $w_t=4$ and apply each weight pair throughout 50 denoising steps. Multimodal guidance follows the diagonal $(w_i,w_{ti})=(\alpha,4\alpha)$, interpolating from caption-only T2I at $\alpha=0$ to editing at $\alpha=1$; solid outlines mark $0<\alpha<1$. Image guidance follows the rightmost column, $(w_i,w_{ti})=(\alpha,4)$; dashed outlines mark $0\leq\alpha<1$. Even at $\alpha=0$, image guidance uses the image-conditioned predictions $E$ and $I$. Numbers indicate forward passes per denoising step: four for intermediate multimodal guidance, three for image guidance with $\alpha<1$, and two for the T2I and editing endpoints.}
	\label{fig:guidance-grid}
\end{figure}

\Cref{fig:guidance-grid} illustrates the two guidance paths in the same weight plane.

\Cref{tab:appx-mmdynamic} compares these controls on $[10,20)$ and $[25,35)$. At $[10,20)$, multimodal guidance with $\alpha=0.25$ improves GEdit semantic consistency and overall score, GIE-Bench functional correctness, and preservation over the pure T2I interval, while reducing perceptual quality, ImgEdit, and CLIP-T scores. At $[25,35)$, the pure T2I endpoint has the highest GEdit overall, ImgEdit, CLIP-T, and functional-correctness scores among the multimodal-guidance settings, while intermediate weights retain more source structure. Lowering image guidance on either interval changes GEdit overall by at most $+0.06$ and reduces ImgEdit. Applied over the full trajectory, it lowers both further.

For FLUX2-Klein~4B/9B, the denoiser forward counts in \cref{tab:appx-fwd} follow directly from the required predictions: two at either endpoint, three for image guidance with $\alpha<1$, and four for multimodal guidance with $0<\alpha<1$.

\begin{table}[!ht]
\centering
\caption{\textbf{Forward passes within the denoising trajectory (FLUX2-Klein~4B/9B).} Counts are per sampling step, counting batched conditioning branches separately. Pure editing and the pure T2I endpoint each require two passes.}
\label{tab:appx-fwd}
\vspace{0.5em}
{\footnotesize
\begin{tabular}{lcl}
\toprule
Conditioning & Forwards / step & Required predictions \\
\midrule
Pure E (baseline) & \textbf{2} & $E,\,I$ \\
Pure T2I & \textbf{2} & $T,\,U$ \\
Image guidance ($\alpha<1$)                & 3          & $E,\,I,\,U$ \\
Multimodal guidance ($0<\alpha<1$)       & 4          & $E,\,T,\,I,\,U$ \\
\bottomrule
\end{tabular}
}
\end{table}
\begin{table}[!ht]
\centering
\caption{\textbf{Image guidance and multimodal guidance (FLUX2-Klein~4B).} GEdit uses GPT-4.1; ImgEdit and GIE correctness use GPT-4o. $\alpha=1$ is editing. For multimodal guidance, $\alpha=0$ is the T2I endpoint; for image guidance, it sets the image guidance scale to zero. Shading uses one baseline-centered scale per metric across all blocks.}
\label{tab:appx-mmdynamic}
\vspace{0.5em}
{\small
\setlength{\tabcolsep}{4pt}
\begin{tabular}{lc|ccc|c|c|ccc}
\toprule
 & & \multicolumn{3}{c|}{\textbf{GEdit}} & \textbf{ImgEdit} & \textbf{MagicBrush} & \multicolumn{3}{c}{\textbf{GIE-Bench}} \\
$\alpha$ & \shortstack{Fwd/\\step} & SC & PQ & O & overall & CLIP-T & FC & SSIM & U-CLIP \\
\midrule
\multicolumn{10}{l}{\textbf{Image guidance}} \\
\cmidrule(lr){1-10}
\multicolumn{10}{l}{\textit{Interval} $[10,20)$} \\
$0.0$ & 3 & \cellcolor[RGB]{255,253,254}7.69 & \cellcolor[RGB]{255,249,250}7.27 & \cellcolor[RGB]{255,254,254}7.10 & \cellcolor[RGB]{255,248,249}3.66 & \cellcolor[RGB]{235,249,237}0.319 & \cellcolor[RGB]{234,249,237}74.84 & \cellcolor[RGB]{255,244,245}0.782 & \cellcolor[RGB]{255,243,245}0.891 \\
$0.25$ & 3 & \cellcolor[RGB]{253,254,253}7.74 & \cellcolor[RGB]{255,252,252}7.36 & \cellcolor[RGB]{251,253,251}7.15 & \cellcolor[RGB]{255,252,253}3.71 & \cellcolor[RGB]{240,250,242}0.319 & \cellcolor[RGB]{232,248,236}74.98 & \cellcolor[RGB]{255,247,248}0.790 & \cellcolor[RGB]{255,247,248}0.898 \\
$0.5$ & 3 & \cellcolor[RGB]{255,254,254}7.70 & \cellcolor[RGB]{255,254,254}7.43 & \cellcolor[RGB]{251,253,251}7.15 & \cellcolor[RGB]{255,254,254}3.73 & \cellcolor[RGB]{240,250,242}0.319 & \cellcolor[RGB]{255,253,253}73.35 & \cellcolor[RGB]{255,251,251}0.797 & \cellcolor[RGB]{255,250,251}0.904 \\
$0.75$ & 3 & \cellcolor[RGB]{245,252,246}7.79 & \cellcolor[RGB]{253,254,253}7.48 & \cellcolor[RGB]{249,253,250}7.17 & \cellcolor[RGB]{255,253,253}3.72 & \cellcolor[RGB]{246,252,247}0.318 & \cellcolor[RGB]{248,253,249}73.88 & \cellcolor[RGB]{255,253,253}0.802 & \cellcolor[RGB]{255,253,253}0.910 \\
$1.0$ (Editing) & 2 & 7.72 & 7.46 & 7.11 & 3.74 & 0.317 & 73.40 & 0.805 & 0.913 \\
\cmidrule(lr){1-10}
\multicolumn{10}{l}{\textit{Interval} $[25,35)$} \\
$0.0$ & 3 & \cellcolor[RGB]{255,253,253}7.69 & \cellcolor[RGB]{255,252,253}7.38 & \cellcolor[RGB]{252,254,252}7.13 & \cellcolor[RGB]{255,254,254}3.73 & \cellcolor[RGB]{255,254,254}0.317 & 73.40 & \cellcolor[RGB]{255,237,240}0.768 & \cellcolor[RGB]{255,243,244}0.890 \\
$0.25$ & 3 & \cellcolor[RGB]{255,254,254}7.72 & \cellcolor[RGB]{255,251,251}7.32 & \cellcolor[RGB]{255,254,254}7.10 & \cellcolor[RGB]{255,254,254}3.73 & \cellcolor[RGB]{253,254,253}0.317 & \cellcolor[RGB]{252,254,252}73.57 & \cellcolor[RGB]{255,245,246}0.785 & \cellcolor[RGB]{255,246,247}0.896 \\
$0.5$ & 3 & \cellcolor[RGB]{247,252,248}7.77 & \cellcolor[RGB]{255,254,254}7.44 & \cellcolor[RGB]{249,253,250}7.17 & \cellcolor[RGB]{255,254,254}3.73 & \cellcolor[RGB]{250,253,250}0.318 & \cellcolor[RGB]{244,252,246}74.14 & \cellcolor[RGB]{255,251,252}0.799 & \cellcolor[RGB]{255,250,251}0.905 \\
$0.75$ & 3 & \cellcolor[RGB]{246,252,247}7.78 & \cellcolor[RGB]{255,254,254}7.43 & \cellcolor[RGB]{252,254,252}7.13 & \cellcolor[RGB]{255,254,254}3.73 & \cellcolor[RGB]{253,254,253}0.317 & \cellcolor[RGB]{255,199,206}71.90 & \cellcolor[RGB]{255,253,254}0.803 & \cellcolor[RGB]{255,253,253}0.911 \\
$1.0$ (Editing) & 2 & 7.72 & 7.46 & 7.11 & 3.74 & 0.317 & 73.40 & 0.805 & 0.913 \\
\cmidrule(lr){1-10}
\multicolumn{10}{l}{\textit{Full trajectory} $[0,50)$} \\
$0.5$ & 3 & \cellcolor[RGB]{255,241,243}7.35 & \cellcolor[RGB]{255,236,239}6.80 & \cellcolor[RGB]{255,241,243}6.71 & \cellcolor[RGB]{255,239,241}3.55 & \cellcolor[RGB]{238,250,240}0.319 & \cellcolor[RGB]{229,247,233}75.21 & \cellcolor[RGB]{255,231,234}0.755 & \cellcolor[RGB]{255,228,231}0.860 \\
$0.0$ & 3 & \cellcolor[RGB]{255,199,206}6.12 & \cellcolor[RGB]{255,199,206}5.42 & \cellcolor[RGB]{255,199,206}5.45 & \cellcolor[RGB]{255,199,206}3.07 & \cellcolor[RGB]{255,199,206}0.311 & \cellcolor[RGB]{255,231,234}72.76 & \cellcolor[RGB]{255,206,212}0.700 & \cellcolor[RGB]{255,199,206}0.803 \\
\midrule
\multicolumn{10}{l}{\textbf{Multimodal guidance}} \\
\cmidrule(lr){1-10}
\multicolumn{10}{l}{\textit{Interval} $[10,20)$} \\
$0.0$ (T2I) & 2 & \cellcolor[RGB]{255,248,248}7.53 & \cellcolor[RGB]{198,239,206}8.21 & \cellcolor[RGB]{215,244,221}7.53 & \cellcolor[RGB]{198,239,206}4.16 & \cellcolor[RGB]{198,239,206}0.324 & \cellcolor[RGB]{208,241,214}76.74 & \cellcolor[RGB]{255,199,206}0.684 & \cellcolor[RGB]{255,242,244}0.889 \\
$0.25$ & 4 & \cellcolor[RGB]{198,239,206}8.10 & \cellcolor[RGB]{220,245,225}7.92 & \cellcolor[RGB]{198,239,206}7.73 & \cellcolor[RGB]{223,246,228}3.97 & \cellcolor[RGB]{206,241,212}0.323 & \cellcolor[RGB]{198,239,206}77.48 & \cellcolor[RGB]{255,236,239}0.766 & \cellcolor[RGB]{255,249,250}0.903 \\
$0.5$ & 4 & \cellcolor[RGB]{214,243,220}7.99 & \cellcolor[RGB]{230,248,233}7.79 & \cellcolor[RGB]{218,244,223}7.51 & \cellcolor[RGB]{238,250,241}3.86 & \cellcolor[RGB]{223,246,228}0.321 & \cellcolor[RGB]{239,250,241}74.49 & \cellcolor[RGB]{255,251,251}0.798 & \cellcolor[RGB]{255,253,253}0.910 \\
$0.75$ & 4 & \cellcolor[RGB]{230,248,234}7.88 & \cellcolor[RGB]{243,251,245}7.61 & \cellcolor[RGB]{235,249,238}7.32 & \cellcolor[RGB]{249,253,250}3.78 & \cellcolor[RGB]{247,252,248}0.318 & 73.40 & \cellcolor[RGB]{255,254,254}0.804 & 0.913 \\
$1.0$ (Editing) & 2 & 7.72 & 7.46 & 7.11 & 3.74 & 0.317 & 73.40 & 0.805 & 0.913 \\
\cmidrule(lr){1-10}
\multicolumn{10}{l}{\textit{Interval} $[25,35)$} \\
$0.0$ (T2I) & 2 & \cellcolor[RGB]{237,249,239}7.84 & \cellcolor[RGB]{213,243,219}8.01 & \cellcolor[RGB]{215,244,221}7.54 & \cellcolor[RGB]{214,243,220}4.04 & \cellcolor[RGB]{224,246,229}0.321 & \cellcolor[RGB]{232,248,235}75.03 & \cellcolor[RGB]{255,228,232}0.749 & \cellcolor[RGB]{255,248,248}0.900 \\
$0.25$ & 4 & \cellcolor[RGB]{230,247,233}7.89 & \cellcolor[RGB]{224,246,228}7.87 & \cellcolor[RGB]{222,245,226}7.46 & \cellcolor[RGB]{225,246,229}3.96 & \cellcolor[RGB]{240,250,242}0.319 & \cellcolor[RGB]{243,251,244}74.24 & \cellcolor[RGB]{255,240,241}0.773 & \cellcolor[RGB]{255,249,250}0.903 \\
$0.5$ & 4 & \cellcolor[RGB]{228,247,232}7.90 & \cellcolor[RGB]{238,250,240}7.68 & \cellcolor[RGB]{232,248,235}7.36 & \cellcolor[RGB]{242,251,244}3.83 & \cellcolor[RGB]{244,252,246}0.318 & \cellcolor[RGB]{248,253,249}73.86 & \cellcolor[RGB]{255,252,252}0.800 & \cellcolor[RGB]{255,253,253}0.910 \\
$0.75$ & 4 & \cellcolor[RGB]{255,254,254}7.71 & \cellcolor[RGB]{246,252,247}7.57 & \cellcolor[RGB]{251,254,252}7.14 & \cellcolor[RGB]{251,253,251}3.76 & \cellcolor[RGB]{252,254,253}0.317 & \cellcolor[RGB]{247,252,248}73.96 & \cellcolor[RGB]{255,254,254}0.804 & \cellcolor[RGB]{255,254,254}0.913 \\
$1.0$ (Editing) & 2 & 7.72 & 7.46 & 7.11 & 3.74 & 0.317 & 73.40 & 0.805 & 0.913 \\
\bottomrule
\end{tabular}
}
\end{table}

\FloatBarrier

\section{End-to-end timing and captioner choice}
\label{app:timing}
Switching reduces image-generation time because T2I omits source-image tokens while keeping the same denoiser pass count as editing on the FLUX models (\cref{tab:appx-fwd,tab:appx-caption-timing}). End-to-end latency depends on the captioner: with the Qwen captioners~\citep{qwen2026qwen35}, total time stays close to pure editing, whereas the Gemini API call dominates the measured runtime.

The quality gains do not depend on Gemini captions (\cref{tab:appx-caption-quality}). At the same switching interval, the evaluated Qwen alternatives retain improvements over pure editing in GEdit overall, ImgEdit, MagicBrush alignment, and GIE-Bench correctness, with a similar preservation cost. A faster and cheaper captioner can therefore retain the benefit of task switching while avoiding the large captioning overhead.
\begin{table}[!htbp]
\centering
\caption{\textbf{End-to-end time per edit, including caption generation.} The $\mathrm{E}\to\mathrm{T2I}\to\mathrm{E}$ schedule uses the interval $[10,20)$. Timings use FLUX2-Klein~4B on $1{,}000$ samples; all times are means in seconds. For Gemini, the VLM-call time is the API-call latency. In addition to image generation and the VLM call, total time includes VAE encoding and decoding and image resizing.}
\label{tab:appx-caption-timing}
{\small
\setlength{\tabcolsep}{5pt}
\begin{tabular*}{\linewidth}{@{\extracolsep{\fill}}llrrr@{}}
\toprule
Sampling & Caption model & Image generation & VLM call & Total \\
\midrule
Pure editing & --- & 5.325 & --- & 5.475 \\
$\mathrm{E}\to\mathrm{T2I}\to\mathrm{E}$ & Gemini-2.5-pro & 4.900 & 26.760 & 31.810 \\
$\mathrm{E}\to\mathrm{T2I}\to\mathrm{E}$ & Qwen3.5-9B & 4.900 & 0.460 & 5.510 \\
$\mathrm{E}\to\mathrm{T2I}\to\mathrm{E}$ & Qwen3.5-27B & 4.900 & 1.143 & 6.193 \\
$\mathrm{E}\to\mathrm{T2I}\to\mathrm{E}$ & Qwen3.5-35B-A3B & 4.900 & 0.239 & 5.289 \\
\bottomrule
\end{tabular*}}
\end{table}

\begin{table}[!htbp]
\centering
\caption{\textbf{Captioner choice for $\mathrm{E}\to\mathrm{T2I}\to\mathrm{E}$ on $[10,20)$ (FLUX2-Klein~4B).} The interval matches the timing measurements in \cref{tab:appx-caption-timing}. Pure editing reproduces \cref{tab:main}; Gemini-caption results use the same $[10,20)$ interval as the Qwen results. Caption models supply target captions; the benchmark metrics follow \cref{sec:exps}. Higher is better. Green/red intensity shows gains/losses from pure editing, normalized by the largest absolute change per metric within this table. The $\pm$ row gives the three-seed standard deviation of the pure-editing baseline.}
\label{tab:appx-caption-quality}
{\small
\setlength{\tabcolsep}{4pt}
\begin{tabular}{@{}lrrrrr@{}}
\toprule
 & GEdit & ImgEdit & MagicBrush & \multicolumn{2}{c}{GIE-Bench} \\
\cmidrule(lr){2-2}\cmidrule(lr){3-3}\cmidrule(lr){4-4}\cmidrule(lr){5-6}
Caption model & O & Overall & CLIP-T & FC (\%) & SSIM \\
\midrule
{\footnotesize\textcolor{black!55}{$\pm$ std (3 seeds)}} & {\footnotesize\textcolor{black!55}{0.04}} & {\footnotesize\textcolor{black!55}{0.04}} & {\footnotesize\textcolor{black!55}{0.001}} & {\footnotesize\textcolor{black!55}{0.35}} & {\footnotesize\textcolor{black!55}{0.001}} \\
Pure editing (no caption) & 7.11 & 3.74 & 0.317 & 73.40 & 0.805 \\
\midrule
Gemini-2.5-pro & \cellcolor[RGB]{198,239,206}7.53 & \cellcolor[RGB]{198,239,206}4.16 & \cellcolor[RGB]{198,239,206}0.324 & \cellcolor[RGB]{207,242,214}76.74 & \cellcolor[RGB]{255,199,206}0.684 \\
Qwen3.5-27B & \cellcolor[RGB]{201,240,208}7.51 & \cellcolor[RGB]{206,241,213}4.10 & \cellcolor[RGB]{198,239,206}0.324 & \cellcolor[RGB]{203,240,210}77.02 & \cellcolor[RGB]{255,199,206}0.684 \\
Qwen3.5-35B-A3B & \cellcolor[RGB]{205,241,212}7.48 & \cellcolor[RGB]{205,241,212}4.11 & \cellcolor[RGB]{206,241,213}0.323 & \cellcolor[RGB]{198,239,206}77.37 & \cellcolor[RGB]{255,199,206}0.683 \\
\bottomrule
\end{tabular}}
\end{table}

\FloatBarrier

\section{Selective-rule settings}
\label{app:hparams}
All models use the rule in \cref{sec:selective-duet}, with the thresholds below shared across benchmarks. The probe step ($k=9$), T2I interval ($[10,16)$), and statistic definitions are unchanged: source attention is averaged across heads and all transformer layers before computing spatial CV and the $75$th percentile $\pi$. The probe sits at the last step before the interval because attention patterns grow more differentiated over steps $0$--$9$.

\begin{center}
\begin{tabular}{lcc}
\toprule
Model & $\tau_{\mathrm{cv}}$ & $\tau_{\pi}$ \\
\midrule
FLUX2-Klein 4B & 0.50 & 0.22 \\
FLUX2-Klein 9B & 0.50 & 0.209 \\
BAGEL & 0.166 & 0.469 \\
\bottomrule
\end{tabular}
\end{center}

For 4B, we replayed pure-editing and fixed-interval outcomes on the GEdit dataset, seeking a task-averaged fidelity gain of at least $0.25$ with a preservation loss of at most $0.05$. Of $111$ candidate threshold rules, $8$ met both criteria, and $(0.50,0.22)$ was selected as one feasible operating point.

For 9B and BAGEL, we transferred the 4B operating point using probe distributions on the GEdit dataset. On 9B, we retained the CV threshold and adjusted $\pi$ to approximately match 4B's gate and overall switching rates. On BAGEL, both thresholds were obtained by matching each gate's exceedance rate separately to 4B ($13.5\%$ for CV and $18.5\%$ for $\pi$), then rounding. ImgEdit, MagicBrush, and GIE-Bench were not used to select these thresholds.

\paragraph{Comparison with random routing.}
To test whether the attention rule outperforms choosing edits at random, we replay per-sample GEdit scores (seed $1337$) under pure editing and under the fixed $[10,16)$ interval, and route each edit to one or the other. Random routing with probability $p$ has expected macro-averaged scores equal to the $p$-weighted mix of the two; we set $p$ so that its mean preservation equals that of the attention rule, and compare quality at that matched preservation. \Cref{tab:appx-coin} reports the results for FLUX2-Klein~4B and 9B. On both models the rule exceeds the preservation-matched random baseline on overall, perceptual-quality, and fidelity scores, with the larger margin on 4B, where the thresholds were selected, and a smaller but same-signed margin on 9B, where they were transferred by distribution matching alone. BAGEL was not included in this replay.

\begin{table}[!ht]
\centering
\caption{\textbf{Attention rule versus preservation-matched random routing (GEdit, GPT-4.1, macro averaging).} Random routing sends each edit to the $[10,16)$ interval with a probability chosen to match the rule's mean preservation. Differences are rule minus random; positive favors the rule.}
\label{tab:appx-coin}
\vspace{0.5em}
{\footnotesize
\setlength{\tabcolsep}{3pt}
\begin{tabular}{llcccc}
\toprule
Model & Routing & Preservation & Fidelity & PQ & O \\
\midrule
\multirow{4}{*}{FLUX2-Klein 4B}
 & Pure editing & 8.902 & 8.395 & 7.421 & 7.112 \\
 & Attention rule ($68\%$ switched) & 8.860 & 8.696 & 7.698 & 7.459 \\
 & Random, preservation-matched ($p=0.33$) & 8.860 & 8.546 & 7.578 & 7.295 \\
 & Difference & $0$ & $+0.149$ & $+0.120$ & $+0.164$ \\
\midrule
\multirow{4}{*}{FLUX2-Klein 9B}
 & Pure editing & 9.127 & 9.033 & 7.746 & 7.778 \\
 & Attention rule ($69\%$ switched) & 9.085 & 9.201 & 7.869 & 7.949 \\
 & Random, preservation-matched ($p=0.39$) & 9.085 & 9.110 & 7.813 & 7.856 \\
 & Difference & $0$ & $+0.091$ & $+0.056$ & $+0.093$ \\
\bottomrule
\end{tabular}
}
\end{table}

\section{Preservation analysis}
\label{app:preservation}
\paragraph{Judge protocol.}
We use GPT-4.1, the same judge as in our main GEdit evaluation, providing the source image, edited image, and editing instruction. The prompt and scoring scale are adapted from GEdit's semantic-consistency evaluation by retaining the preservation assessment and removing the fidelity assessment. The preservation-only prompt asks the judge to score unnecessary changes to unrelated content from $0$ to $10$, without penalizing changes needed to carry out the instruction. A score of $0$ indicates substantial unrelated changes; $10$ indicates that all unrelated content is preserved.
\paragraph{Paired comparisons.}
We evaluate preservation on GIE-Bench using these GPT scores and SSIM outside the edit mask. For each comparison, we match edits by identifier and average paired score differences with equal weight per edit. Because multiple edits can share a source image, confidence intervals and significance tests use standard errors clustered by source image.

\Cref{tab:appx-preservation} reports paired mean differences with confidence intervals and significance tests adjusted for multiple comparisons.

\begin{table}[!ht]
\centering
\caption{\textbf{Paired preservation effects on GIE-Bench.} Positive differences indicate better preservation. Shading marks $p<0.05$ (green: gain; red: loss). Fixed and selective application use $[10,16)$. For each metric, Bonferroni $95\%$ confidence intervals and Holm-adjusted $p$ values account for multiple comparisons within the 4B baseline sweep and, separately, the pooled 9B/BAGEL comparisons. $^{\dagger}$The exploratory 4B selective--fixed comparison uses an ordinary $95\%$ interval and an unadjusted $p$ value.}
\label{tab:appx-preservation}
\vspace{0.5em}
{\footnotesize
\setlength{\tabcolsep}{3pt}
\resizebox{\linewidth}{!}{
\begin{tabular}{lrrr|rrr}
\toprule
 & \multicolumn{3}{c|}{GPT preservation} & \multicolumn{3}{c}{SSIM} \\
Comparison & $\Delta$ & 95\% CI & $p$ & $\Delta$ & 95\% CI & $p$ \\
\midrule
\multicolumn{7}{l}{\textbf{FLUX2-Klein 4B: comparisons against pure editing}} \\
$\mathrm{E}\to\mathrm{T2I}$, $k=35$ & $-0.059$ & $[-0.207,+0.088]$ & $0.946$ & \cellcolor[RGB]{255,199,206}$-0.0371$ & \cellcolor[RGB]{255,199,206}$[-0.0404,-0.0338]$ & $<0.001$ \\
$\mathrm{E}\to\mathrm{T2I}$, $k=40$ & $+0.099$ & $[-0.046,+0.244]$ & $0.391$ & \cellcolor[RGB]{255,199,206}$-0.0187$ & \cellcolor[RGB]{255,199,206}$[-0.0207,-0.0168]$ & $<0.001$ \\
$\mathrm{E}\to\mathrm{T2I}$, $k=45$ & $+0.093$ & $[-0.043,+0.228]$ & $0.391$ & \cellcolor[RGB]{255,199,206}$-0.0053$ & \cellcolor[RGB]{255,199,206}$[-0.0061,-0.0044]$ & $<0.001$ \\
$\mathrm{E}\to\mathrm{T2I}\to\mathrm{E}$, $[25,35)$ & \cellcolor[RGB]{255,199,206}$-0.341$ & \cellcolor[RGB]{255,199,206}$[-0.521,-0.161]$ & $<0.001$ & \cellcolor[RGB]{255,199,206}$-0.0567$ & \cellcolor[RGB]{255,199,206}$[-0.0613,-0.0521]$ & $<0.001$ \\
$\mathrm{E}\to\mathrm{T2I}\to\mathrm{E}$, $[35,45)$ & $+0.030$ & $[-0.125,+0.184]$ & $0.993$ & \cellcolor[RGB]{255,199,206}$-0.0270$ & \cellcolor[RGB]{255,199,206}$[-0.0297,-0.0244]$ & $<0.001$ \\
Fixed $[10,16)$ & $-0.075$ & $[-0.246,+0.096]$ & $0.946$ & \cellcolor[RGB]{255,199,206}$-0.0246$ & \cellcolor[RGB]{255,199,206}$[-0.0283,-0.0209]$ & $<0.001$ \\
Selective $[10,16)$ & $-0.041$ & $[-0.202,+0.121]$ & $0.993$ & \cellcolor[RGB]{255,199,206}$-0.0191$ & \cellcolor[RGB]{255,199,206}$[-0.0225,-0.0157]$ & $<0.001$ \\
\midrule
\multicolumn{7}{l}{\textbf{FLUX2-Klein 4B: follow-up comparison}} \\
Selective $-$ fixed$^{\dagger}$ & $+0.034$ & $[-0.047,+0.116]$ & $0.410$ & \cellcolor[RGB]{198,239,206}$+0.0055$ & \cellcolor[RGB]{198,239,206}$[+0.0041,+0.0069]$ & $<0.001$ \\
\midrule
\multicolumn{7}{l}{\textbf{FLUX2-Klein 9B}} \\
Fixed $-$ pure editing & $-0.158$ & $[-0.338,+0.023]$ & $0.127$ & \cellcolor[RGB]{255,199,206}$-0.0391$ & \cellcolor[RGB]{255,199,206}$[-0.0442,-0.0340]$ & $<0.001$ \\
Selective $-$ pure editing & $-0.084$ & $[-0.235,+0.067]$ & $0.561$ & \cellcolor[RGB]{255,199,206}$-0.0267$ & \cellcolor[RGB]{255,199,206}$[-0.0312,-0.0223]$ & $<0.001$ \\
Selective $-$ fixed & $+0.073$ & $[-0.049,+0.195]$ & $0.561$ & \cellcolor[RGB]{198,239,206}$+0.0124$ & \cellcolor[RGB]{198,239,206}$[+0.0093,+0.0155]$ & $<0.001$ \\
\midrule
\multicolumn{7}{l}{\textbf{BAGEL}} \\
Fixed $-$ pure editing & $-0.044$ & $[-0.226,+0.139]$ & $1.000$ & \cellcolor[RGB]{255,199,206}$-0.0082$ & \cellcolor[RGB]{255,199,206}$[-0.0096,-0.0068]$ & $<0.001$ \\
Selective $-$ pure editing & $+0.005$ & $[-0.165,+0.174]$ & $1.000$ & \cellcolor[RGB]{255,199,206}$-0.0055$ & \cellcolor[RGB]{255,199,206}$[-0.0067,-0.0042]$ & $<0.001$ \\
Selective $-$ fixed & $+0.048$ & $[-0.076,+0.173]$ & $0.920$ & \cellcolor[RGB]{198,239,206}$+0.0027$ & \cellcolor[RGB]{198,239,206}$[+0.0019,+0.0035]$ & $<0.001$ \\
\bottomrule
\end{tabular}
}
}
\end{table}

\FloatBarrier

Among the paired comparisons in \cref{tab:appx-preservation}, only the 4B $[25,35)$ interval shows a significant mean GPT-preservation reduction after correction. Every fixed and selective configuration lowers SSIM significantly relative to pure editing. Selective application recovers part of that loss, with a significant SSIM gain over fixed $[10,16)$ on all three models and no significant GPT-preservation reduction relative to pure editing.

\paragraph{Figure measurements.}
Generations use fixed seeds, and preservation and correctness are assessed in independent judge runs. \Cref{fig:frontier} combines GPT-4.1 preservation and Gemini-3-Flash correctness by configuration. Its horizontal error bars are ordinary $95\%$ confidence intervals for individual preservation means, using source-clustered standard errors and a $t$ critical value with degrees of freedom equal to the number of source groups minus one. Configurations without per-edit preservation records are shown as means without error bars. The paired differences and simultaneous confidence intervals in \cref{tab:appx-preservation} provide the comparisons against pure editing and between fixed and selective switching.

\FloatBarrier

\end{document}